%% file: main.tex
\documentclass{article} % For LaTeX2e
\usepackage{iclr2021_conference,times}
\usepackage{booktabs}
\usepackage{amsmath,amsfonts,bm}
\input{math_commands.tex}
\usepackage{hyperref}
\usepackage{multirow}

\usepackage{url}
\usepackage{xcolor}
\usepackage{tikz}
\usepackage{array}
\usepackage{xcolor}
\usepackage{calc}
\def\basiceval#1{\the\numexpr#1\relax}
\usepackage{float}

\usepackage[font=small,labelfont=bf]{caption}
\usepackage{tabularx} % in the preamble
\usepackage{setspace}

\usepackage{titletoc}
\usepackage{appendix}

\usepackage{etoc}

\newcommand\numberthis{\addtocounter{equation}{1}\tag{\theequation}}

\newlength\MAX  

\newcommand*\VChart[3]{ #1\%  ~\rlap{\textcolor{black!2}{\rule{\MAX}{1.8ex}}}\rlap{\textcolor{red!17}{\rule{#3\MAX}{1.8ex}}}\textcolor{orange}{\rule{#2\MAX}{1.8ex}}}

\usepackage[colorinlistoftodos,prependcaption,textsize=tiny]{todonotes}
\usepackage{xargs}
\newcommandx{\gk}[2][1=]{\todo[backgroundcolor=orange!25,#1]{GK: #2}}
\newcommandx{\md}[2][1=]{\todo[backgroundcolor=blue!25,#1]{MD: #2}}
\newcommandx{\he}[2][1=]{\todo[backgroundcolor=olive!25,#1]{HE: #2}}
\usepackage{wrapfig}

\usepackage{color,soul}
\definecolor{g0}{HTML}{30b52d}
\definecolor{y1}{HSB}{38,10,255}
\definecolor{y2}{HSB}{38,20,255}
\definecolor{y3}{HSB}{38,25,255}
\definecolor{y4}{HSB}{38,50,255}
\definecolor{y5}{HSB}{38,80,255}
\definecolor{y6}{HSB}{38,110,255}
\definecolor{y7}{HSB}{38,150,255}
\definecolor{y8}{HSB}{38,200,255}
\definecolor{y9}{HSB}{38,250,255}
\definecolor{y10}{HSB}{38,255,255}
\definecolor{r0}{RGB}{255,181,195}

\DeclareRobustCommand{\ye}[1]{{\sethlcolor{y5}\hl{#1}}}
\DeclareRobustCommand{\yf}[1]{{\sethlcolor{y6}\hl{#1}}}
\DeclareRobustCommand{\yg}[1]{{\sethlcolor{y7}\hl{#1}}}
\DeclareRobustCommand{\yh}[1]{{\sethlcolor{y8}\hl{#1}}}

\DeclareRobustCommand{\yj}[1]{{\sethlcolor{y10}\hl{#1}}}
\DeclareRobustCommand{\g}[1]{{\sethlcolor{g0}\hl{#1}}}
\DeclareRobustCommand{\r}[1]{{\sethlcolor{r0}\hl{#1}}}

\title{A Distributional Approach to\\ Controlled Text Generation}

\author{
Muhammad Khalifa
\footnotemark[1]\thanks{Equal Contributions.}
\; 
\footnotemark[2]\thanks{Work done during an internship at NAVER Labs Europe.}
\\ \ Cairo University
\And
Hady Elsahar\footnotemark[1]\\
Naver Labs Europe
\And
Marc Dymetman\footnotemark[1] \\
Naver Labs Europe
\AND
\vspace*{-2.5em}
\normalfont{\texttt{\{hady.elsahar,marc.dymetman\}@naverlabs.com}} \\
\hspace{0.2cm}
\texttt{m.khalifa@grad.fci-cu.edu.eg}
\vspace*{0em}
}

\newcommand{\MK}[1]{\textcolor{brown}{ #1}}

\newcommand{\HEFN}[1]{}
\newcommand{\MKFN}[1]{}
\newcommand{\MDFN}[1]{}
\newcommand{\CC}{\mathcal{C}} % Constraint manifold of distributions

\newcommand{\blambda}{{\bm{\lambda}}}
\newcommand{\bphi}{{\bm{\phi}}}
\newcommand{\bmu}{{\bm{\mu}}}
\newcommand{\tmu}{{\bar{\mu}}}
\newcommand{\tbmu}{{\bar{\bmu}}}

\newcommand{\pit}{{\pi_\theta}}

\newcommand{\sprod}[2]{{\left<#1,#2\right>}}

\newtheorem{theorem}{Theorem}

\newcommand{\precite}[1]{[[#1]]}
\usepackage{array}

\usepackage{enumitem}
\setitemize{noitemsep,topsep=0pt,parsep=0pt,partopsep=0pt}

\usepackage{xspace}

\usepackage{comment}
\usepackage{enumitem}

\usepackage[noend]{algpseudocode}
\usepackage{algorithm}
\usepackage{textcomp}

\renewcommand{\textuparrow}{$\uparrow$}
\renewcommand{\textdownarrow}{$\downarrow$}
\usepackage{wrapfig}
\usepackage{enumitem}
\usepackage{bbm}

\iclrfinalcopy % Uncomment for camera-ready version, but NOT for submission.
\begin{document}
\maketitle

\input{abstract}

\input{sections/intro_relatedwork}

\input{sections/formalization}

\input{sections/experiments}
\input{sections/conclusion}

%
\subsubsection*{Acknowledgments}
We would like to thank the anonymous reviewers for their insightful feedback that helped enhancing the final version of this manuscript. We also thank Germán Kruszewski, Laurent Besacier, Matthias Gallé and Christopher Dance for providing technical feedback on this work and proof-reading the manuscript, as well as Tetiana Parshakova and Jean-Marc Andreoli for their work on the original versions of the SNIS and DPG algorithms.

% \clearpage
%\input{Notes/notes_on_draft}
%
\clearpage
\bibliography{ref}
\bibliographystyle{iclr2020_conference}
\newpage
\appendix

% Comment next line for faster compilation
\input{sections/appendix}

\end{document}

%% file: math_commands.tex
\def\eqref#1{equation~\ref{#1}}
\def\1{\bm{1}}

\DeclareMathAlphabet{\mathsfit}{\encodingdefault}{\sfdefault}{m}{sl}
\SetMathAlphabet{\mathsfit}{bold}{\encodingdefault}{\sfdefault}{bx}{n}

\newcommand{\E}{\mathbb{E}}

\newcommand{\KL}{D_{\mathrm{KL}}}
\newcommand{\KLhat}{\hat{D}_{\mathrm{KL}}}
\newcommand{\TVD}{\mathrm{TVD}}

\DeclareMathOperator*{\argmax}{arg\,max}
\DeclareMathOperator*{\argmin}{arg\,min}

\newcommand{\GDC}{GDC\xspace}
\newcommand{\REINFORCE}{REINFORCE }
\newcommand{\REINFORCEP}{REINFORCE\textsubscript{P(x)} }
\newcommand{\ZIEGLER}{ZIEGLER }

%% file: abstract.tex
\begin{abstract}
We propose a Distributional Approach for addressing Controlled Text Generation from pre-trained Language Models (LMs). This approach permits to specify, in a single formal framework, both ``pointwise'' and ``distributional'' constraints over the target LM --- to our knowledge, the first model with such generality --- while minimizing KL divergence from the initial LM distribution. The optimal target distribution is then uniquely determined as an explicit EBM (Energy-Based Model) representation.  From that optimal representation we then train a target controlled Autoregressive LM through an adaptive distributional variant of Policy Gradient. We conduct a first set of experiments over pointwise constraints showing the advantages of our approach over a set of baselines, in terms of obtaining a controlled LM balancing constraint satisfaction with divergence from the initial LM.  We then perform experiments over distributional constraints, a unique feature of our approach, demonstrating its potential as a remedy to the problem of Bias in Language Models. Through an ablation study, we show the effectiveness of our adaptive technique for obtaining faster convergence.\footnote{Code available at \url{https://github.com/naver/gdc}}
\end{abstract}

%% file: sections/intro_relatedwork.tex
\section{Introduction}
Neural language models, such as GPT-2/3 \citep{radford2019language,Brown2020LanguageMA}, pretrained on huge amounts of text, have become pre-eminent in NLP, producing texts of unprecedented quality. 
% %% todo mention they fail to be steered 
In this paper, we are concerned with the problem of controlling a generic pretrained LM in order to satisfy certain desiderata. For instance, we may want to avoid toxic content; prevent certain demographic biases; or steer generations towards a certain topic or style. 
Prior work, taking inspiration from Reinforcement Learning (RL), has aimed at inducing autoregressive models to optimize global objectives using task specific rewards such as BLEU and ROUGE for Machine Translation and Summarization~\citep{seq_lvl_train_RanzatoCAZ15, BahdanauBXGLPCB17}, or hand crafted rewards~\citep{RL_dialogue_LiMRJGG16,RL_TambwekarDMMHR19} to improve certain a priori desirable features. 
% todo rewrite
% Such work rely on techniques such as policy gradients as in~\cite{seq_lvl_train_RanzatoCAZ15}, actor critics~\citep{BahdanauBXGLPCB17} or continuous approximations~\citep{ShettyRHFS17}. And often done in a fine-tuning step after a standard MLE training. 
%

%
%%% degeneration
However, such an optimization process is not infallible; \cite{LiuLSNCP16} noted that it often leads to ``degeneration'', producing poor examples that improve the average reward but forgo coherence and fluency. 
%% todo decide to remove
% This has inspired previous work to rely on hand-crafted rewards to ensure the quality of generated text~\citep{Wu_googleMT16,PaulusXS18,gumbel_textgen_yang_NIPS2018}, which is mostly non-trivial, and often the its results cannot be fully trusted.
%
%% KL penalty and hacks
This degeneration is often diagnosed as an effect of deviating too much from the original pretrained LM during optimization. Consequently, prior work has regarded proximity to the pretrained model as a prescription for sample quality. This view is most prominent in open-domain generation where no gold references are available for fine-tuning, making the pretrained LM itself the yardstick for fluency. 
\cite{KL_Jaques17,Ziegler19} propose a conservative fine-tuning approach moderated by a KL penalty between the trained policy and the original LM, discouraging large deviations. 
% This penalty acts as a regularizer that prevents the trained policy from deviating too much from the original LM.
%
A KL penalty was also used by \cite{plug_and_play_20}, 
%for controlling unconditional LMs,
this time in a plug-and-play rather than a fine-tuning context. However, the authors show that balancing policy deviations from the original LM while also satisfying the 
control conditions is delicate. To combat degeneration they had to combine the KL penalty with post-norm fusion, reranking, and early-stopping procedures. 

%% most of controlled NLG models has considered what we call pointwise view.  
Most of the existing work on Controlled Generation has taken what we refer to as a ``{pointwise}'' view, namely focusing on the quality of each \emph{individual} output, a view that is encouraged by the standard RL goal of maximizing rewards computed at the individual level.
Such techniques are incapable of enforcing ``{distributional}'' conditions, where some collective statistical properties are desired over the set of \emph{all} generations.
%i.e. when the desired control condition is defined in form of feature expectations over the set of whole generations (e.g. balancing gender bias).
% todo: find better e.g. than (balancin gender bias). or remove

Distributional control is key to solving the problem of social biases in LMs trained on large, uncurated Web corpora.   
Those LMs - dubbed \emph{``Stochastic Parrots''} in ~\citep{stochasticParrots} - tend to encode hegemonic biases that are harmful to marginalized populations. There has been a large body of work analysing these distributional biases~\citep{blodgett-bias-survey, bias_mt_stanovsky-etal-2019,bias_mt_PratesAL20,ShengCNP_LM_bias19,gpt3}.
%
% One problem that is currently causing a lot of concern, and which could much benefit from distributional control, is that of \emph{social biases} conspicuous in pretrained language models
% \citep{blodgett-bias-survey, bias_mt_stanovsky-etal-2019,bias_mt_PratesAL20,ShengCNP_LM_bias19,gpt3}. 
%
%%% todo: maybe rewrite this part below it reads like literature review 
However, applying distributional control on pretrained models is still an understudied problem.
\cite{babysitter2_Sheng2020} introduce a method relying on adversarial triggers~\citep{WallaceFKGS19}; this method does not de-bias the whole distribution but only obtains non-biased continuations of given prompts. \cite{BordiaB19} introduce a regularization term for reducing gender bias when training a language model from scratch (as opposed to de-biasing a pretrained model).%
\footnote{Additional Related Work is provided in \S\ref{sec:appendix:relatedwork}. We use \S A, \S B ... to refer to sections in the Appendix.}
%

%%%%%%%%%%%%%%%
%%%%% OUR Work %%%%% 
% We introduce GDC ... 
% In this work we formalize the problem of controlled text generation as a constraint satisfaction problem over the distribution $p$. By definition we constrain the expectations (\textit{Moments}) of certain features, relative to $p$, to have certain values;  However, moment constraints alone underspecify the distribution $p$ -- in general, a great many distributions, especially on high-dimensional data such as sequences, can fit a finite number of moment constraints; Moreover they do not explicitly refer to the pretrained model $a(x)$, meaning $p$ cannot exploit $a$ regarding fluency or semantic adequacy. \\
%
In this work, we present our \emph{Generation with Distributional Control} (\GDC) approach, in which we 
formalize the problem of controlled text generation as a \emph{constraint satisfaction} problem over the \emph{probability distribution} $p$ representing the desired target LM. Namely, we require the expectations  (``moments'') relative to $p$ of certain output features to have specific values; this permits for instance to condition all outputs to speak about sports (a \textit{pointwise constraint}), and 50\% of them to mention female characters (a \textit{distributional constraint}). Additionally, we require $p$ to have a \emph{minimal KL divergence} $\KL(p,a)$ from the original pretrained LM $a$. This has the effect that $p$ now inherits favorable linguistic qualities from $a$. As we will explain, this formulation is a generalization of the \emph{Maximum Entropy Principle}
%inside Information Geometry \citep{Nielsen-intro-information-geometry}, 
and leads to a unique solution $P(x)$. $P(x)$ is an unnormalized distribution, aka an \emph{Energy-Based Model} (EBM)
\citep{Hinton02,lecun_tutorial_2006,Bakhtin2020EnergyBasedMF}, %\citep{Hinton02,RanzatoBCL07},
of which $p(x) =1/Z\ P(x)$ is the normalized version, where $Z \doteq \sum_x P(x)$ is the partition function of $P$. 

Computing the EBM representation $P$ is a crucial step, as it fully determines the \emph{optimal} distribution $p$ we are looking for. 
%However, it is not the end of the story. There are two issues. First, in order to compute $p(x)$, we must compute $Z$, which can be challenging. Second, and even more seriously, even if we know $Z$, we are not able to directly \emph{sample} from $p$.%
However, it is not the end of the story, because the representation thus obtained does not enable us to directly \emph{sample} from $p$, an essential property of any LM.%
\footnote{One possible sampling approach here would be to employ MCMC techniques, such as Metropolis-Hastings~\citep{Robert:2005:MCS:1051451}. These come with theoretical convergence guarantees in the limit but in practice convergence can be very difficult to assess, and furthermore, obtaining samples can be extremely slow.}
%
% By contrast, our objective is to produce an autoregressive ``policy'' (aka  LM) $\pit$ that can be used at inference time as an efficient sampler. 
%
To this end, we introduce \emph{KL-adaptive DPG (Distributional Policy Gradient)}, a variant of an algorithm recently proposed in \citep{opt-rl-arxiv-2019}. We train the policy $\pit$ to approximate $p$ in an adaptive way, by speeding up the next round of approximations based on approximations previously obtained. 
At the end of this process, we obtain a final $\pit$, our target LM, on which we can estimate diverse metrics, including $\KL(p,\pit)$, measuring the approximation quality of $\pit$ relative to the optimal $p$,
and $\KL(\pit,a)$, measuring the divergence of $\pit$ relative to the original LM $a$.

This two-step approach differs from much research in NLP-oriented work with EBMs, which tends to use EBM representations \emph{inside} the training loops of neural networks, blurring different dimensions of the problem. By contrast --- similarly to \cite{A-parshakova-etal-2019-global,opt-rl-arxiv-2019} in a different context --- we clearly \emph{decouple} the relatively simple problem of determining a ``pivot'' optimal EBM from the more difficult problem of exploiting this EBM at inference time, 
Such decoupling is valuable, because it permits to better diagnose the important challenges to focus on.

Overall, 
% the merits of this work 
our contributions
can be summarized as follows:
\begin{enumerate}%[leftmargin=*]
%\begin{itemize}[noitemsep,topsep=0pt,parsep=5pt,partopsep=0pt,leftmargin=*]
% this next one looks good too
%\begin{itemize}[noitemsep,topsep=0pt,parsep=5pt,partopsep=0pt,leftmargin=5mm]
%\begin{itemize}[noitemsep,topsep=0pt,parsep=5pt,partopsep=0pt]
\item We introduce a Distributional View for controlled text generation formalized as a constraint satisfaction problem combined with a divergence minimization objective, providing a single framework both for ``distributional'' constraints (collective statistical requirements) and for ``pointwise'' constraints (hard requirements on each individual) \textbf{(\S\ref{sec:constraints-info-geo})}. 
To our knowledge, this is the first framework with such generality for controlled text generation.%
\item  We show how these constraints lead to an optimal EBM for the target model \textbf{(\S\ref{sec:constraints2ebm})}, propose the KL-Adaptive DPG algorithm 
%--- a variant of
% \citep{opt-rl-arxiv-2019} 
% tailored for this problem --- 
%
for approximating the optimal EBM distribution by an autoregressive policy \textbf{(\S\ref{sec:ebm-2-policy})}, and show the effectiveness of this adaptive technique for obtaining faster convergence  \textbf{(\S\ref{sec:appendix:exp-ablation})}.%\footnote{In the sequel, \S A, \S B ... refer to sections in the Appendix.}
\item We conduct experiments in a number of pointwise and distributional conditions, assessing results in terms of divergence from GPT-2, fluency and diversity, with better performance than 
% a number of
% a few
strong baselines. The distributional experiments show the potential of our approach as a remedy to the current and important problem of bias in pretrained language models, providing a novel direction for addressing it \textbf{(\S\ref{sec:EXPERIMENTS})}.
\end{enumerate}

%8-pages-constraint

%% file: sections/formalization.tex
\section{Formalization}
\label{section:formalization}

We denote by $X$ the set of all sequences $x$ of bounded length $L_{max}$, by $a$ the initial pretrained model and  
%\todo{flow is broken a bit}, 
by $p$ the desired target model. The probabilities of $x$ according to each model are $a(x)$ and $p(x)$.
%
%\footnote{To avoid certain technical subtleties of limited relevance to our main discussion,we consider only bounded sequences, in order to keep $X$ finite. By taking $L_{max}$ sufficiently large, we can assume that the probability mass of $a$ over $X$ is negligibly different from $1.0$.}
%
Our approach consists in expressing our desiderata through constraints on the desired values 
$\tmu_i$ 
of the \emph{expectations} (aka \emph{moments}) $\mu_i \doteq \E_{x\sim p}\: \phi_i(x)$ of certain predefined real-valued feature functions $\phi_i(x)$, for $i \in \{1,\ldots,k\}$. 

To illustrate, the previous example can be expressed by using two binary features, $\phi_1(x) = 1$ iff $x$ is classified as speaking about sports, $\phi_2(x) = 1$  iff $x$ mentions a female character. Then our ``moment constraints'' take the following form:
%\begin{align} \label{eq:constraints1}
 $ \mu_1 = \E_{x\sim p}\: \phi_1(x) = 1.0, \quad \mu_2 = \E_{x\sim p}\: \phi_2(x) = 0.5$.   
%\end{align}  
\\
The first (pointwise) constraint implies that each individual $x$ has to speak about sports (otherwise $\mu_1$ could not reach its maximum value $1.0$), the second (distributional) constraint that 50\% of the $x$'s have to mention a female character.\footnote{This example uses only binary features, but real-valued features can also be used, for instance scores returned by a soft classifier.}
%
%\footnote{Note that as stated here, the two conditions are not exclusive, but this could easily be enforced by adding a fourth feature forbidding the simultaneous presence of both mentions.}

Let $\CC$ be the set of all distributions $c$ over $X$ that satisfy the moment constraints. % (\ref{eq:constraints1}). This set can be huge. For instance, any set of two sequences, one mentioning a female character and both talking about sports, would already determine a particular distribution $c$ in $\CC$,  with tiny coverage and no quality guarantees.
We then propose to specify $p$ as a distribution respecting the constraints, but also minimizing KL divergence from $a$:
\begin{align} \label{eq:KL_argmin}
  p \doteq \argmin_{c\in \CC} \KL(c,a),
\end{align}  

Equation (\ref{eq:KL_argmin}) is a generalization of the \emph{Maximum Entropy Principle} of \citet{jaynes57}, which corresponds to the limit case where $a$ is the uniform $u$ distribution over $X$, noting that minimizing $\KL(c,u)$ is equivalent to maximizing the entropy of $c$ under the constraints --- in other words, trying to find the least ``specific'' distribution satisfying the constraints.
%For a general $a$, (\ref{eq:KL_argmin}) represents a Generalized MaxEnt  model 
% (aka Minimum Discrimination Information) 
%\citep{Nielsen-dual-geometry-2016}.
%
%%%%%%%%%%%%%%%%%%%%%%%%%%%%%%%%%%%%%%%%%%%%%%%%%%%%%%%%%%%%%%%%%
\begin{figure}[!t]
\centering
\vspace{-1cm}
\includegraphics[trim=0 20 0 0, width=0.85\textwidth, clip]{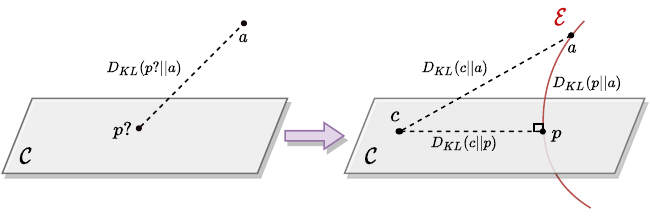}
\caption{
From MaxEnt to EBM through Information Geometry. The Generalized MaxEnt specification (left panel) is looking for a distribution $p$ that lies on the moment constraints manifold $\mathcal{C}$ and that minimizes the forward KL $\KL(p,a)$. The solution is provided by Information Geometry: (1) build the exponential family $\mathcal{E}$ determined by $a$ and $\bphi$, (2) $p$ lies at the intersection between $\mathcal{C}$ and $\mathcal{E}$, (3) for any distribution $c$ satisfying the constraints, the ``Pythagorean identity" holds: $\KL(c||a) = \KL(c||p) + \KL(p||a)$; in particular $p$ is unique.
}
\label{fig:Information-Geometry-1}
\end{figure}
%%%%%%%%%%%%%%%%%%%%%%%%%%%%%%%%%%%%%%%%%%%%%%%%%%%%%%%%%%%%%%%%%
\subsection{Constraints, Information Geometry, Exponential Families}
\label{sec:constraints-info-geo}
To recap our formal approach, we have a finite set $X$,
% \footnote{Generalizations to countable and continuous spaces exist, e.g. \citep{Csiszar1975}.}
a distribution $a$ over $X$ s.t. $a(x)>0, \forall x\in X$, and real functions $\phi_1, ...,\phi_k$ over $X$. 
We specify moment constraints $\mu_i = \tmu_i$ on distributions $c$ over $X$, where $\mu_i \doteq \E_{x\sim c}\ \phi_i(x)$ and the $\tmu_i$'s are given targets; the set of distributions satisfying these constraints is denoted by $\CC$. Our Problem is to find a $p$ such that $p = \argmin_{c \in \CC} \KL(c,a)$. 

We follow \citet{csizarShields2004} on this question, a problem that is at the core of the field of Information Geometry \citep{Nielsen-intro-information-geometry,amari-nagaoka-information-geometry}. Under the assumption that $\CC \neq \emptyset$, they prove the following result (also see \S\ref{appendix-csizar}):

\begin{theorem} \label{theorem:main}
  \textbf{\emph{(A)}} There exists a unique solution $p$ to the problem above, obtained as $p(x) \propto P(x)$ where $P$ is in  \emph{exponential family} 
  form:  
  \begin{align}
      P(x) &= a(x) \ \mathbbm{1}[x\in X_\CC]\ \; e^{\sum_i \lambda_i \phi_i(x)}. \label{eq:exponential_with_X_C}
  \end{align}

In other words $p(x) = 1/Z\ P(x)$, with $Z = \sum_{x\in X} P(x)$; $P$ is an unnormalized distribution, i.e. an \emph{EBM}.
Here $X_\CC = \{x\in X |\ \exists c\in \CC \ s.t.\ c(x) >0\}$ is the ``support set'' associated with $\CC$. The $\lambda_i$'s are real numbers called the \emph{natural parameters} associated with the moments $\mu_i$. \\
%that is, the set of all $x$'s that are in the support of at least one distribution in $\CC$ (by the assumption $\CC \neq \emptyset$, this set is not empty).

  \textbf{\emph{(B)}}  $p$ can be approximated to arbitrary precision by distributions $p_\epsilon$ of the form: 
  \begin{align}
  p_\epsilon(x) \propto a(x) \; e^{\sum_i \lambda_{\epsilon,i} \phi_i(x)}
  \end{align}
  for appropriate real values of the $\lambda_{\epsilon,i}$.\\
  
  \textbf{\emph{(C)}}  $p$ satisfies the \emph{Pythagorean Identity}: $\KL(c,a) = \KL(c,p) + \KL(p,a), \forall c\in\CC$ \emph{(see Fig~\ref{fig:Information-Geometry-1})}.
\end{theorem}

\medskip

The advantage of this version of the connection between Generalized Maximum Entropy and Exponential Families is its generality, which distinguishes it from other presentations, and which makes it ideal for unified application to pointwise, distributional or hybrid constraints.

In the special case of only pointwise constraints, of the form $\E_{x\sim c} \phi_i(x) = 1.0, i\in[1,k]$, with $\phi_i(x) \in \{0,1\}$, let's define the predicate $b(x)$ to be 1 iff $x$ satisfies all the constraints.
Then, using the (A) form of the result, it is an easy exercise (see \S\ref{appendix:details_pointwise}) to prove that $X_\CC = \{x\in X |\ b(x)=1\}$ and that one has $p(x) \propto a(x) b(x)$. In this case $P(x)=a(x)b(x)$ is a very simple EBM that does not involve an exponential part; this is the EBM form that we use for experiments involving only pointwise constraints.

In the general case where some constraints are distributional, the determination of $X_\CC$ is not as direct, and we prefer to use the approximation provided by (B), which permits a generic implementation. 
With only distributional constraints, an exact solution is typically obtained with finite $\lambda$'s. 
With hybrid constraints, some of the $\lambda$'s may tend to infinite (positive or negative) values but thresholding them suffices to get a good approximation.

% Marc 24.02.2021
%\input{sections/snis_algo.tex}
%%%%%%%%%
\begin{wrapfigure}{r}{0.4\textwidth}
\vspace{-45pt}
\begin{minipage}{0.4\textwidth}
\begin{algorithm}[H]
\caption{\ Computing $\blambda$} \label{al:computing_lambdas}
\begin{small}
\begin{algorithmic}[1]
\Require $a$, features $\bphi$, imposed moments $\tbmu$
\State sample a batch ${x_1,\ldots,x_N}$ from $a$
\State for each $j\in[1,N]$: $w_j(\blambda) \gets e^{\blambda\cdot \bphi(x_j)}$
\State $\hat{\bmu}(\blambda) \gets \frac{\sum_{j=1}^N w_j(\blambda)\ \bphi(x_j)}{\sum_{j=1}^N w_j(\blambda)}$
\State solve by SGD: $\argmin_\blambda ||\tbmu - \hat{\bmu}(\blambda)||_2^2$
\Ensure parameter vector $\blambda$
\end{algorithmic}
\end{small}
\end{algorithm}
\end{minipage}
%\vspace{10pt}
\end{wrapfigure}
%%%%%%%%%%%%%%%
%This property means that, even if the support $X_\CC$ is a proper subset of $X$, we can still approximate $p$ by a strict exponential family expression, but where, possibly, some of the $\lambda's$ may tend to arbitrary large positive or negative values. From an application standpoint, we can use this property to avoid explicitation of $X_\CC$ and just clip some of the $\lambda$'s if they cross a large positive or negative threshold.
%
%The connection between expectation constraints and exponential families is a core starting point for the field of ``Information Geometry'' \citep{Nielsen-intro-information-geometry,amari-nagaoka-information-geometry}, which studies general information divergences in relation with differential geometry. An illustration of this viewpoint applied to our problem is provided in Fig.~\ref{fig:Information-Geometry-1}.
%
% \subsubsection{From Moments $\mu$ to Parameters $\lambda$}
%%%%%%%%%%%%%%%%%%%%%%%%%%%%%%%%%%%%%%%%%%%%%%%%%%
%%%%%%%%%%%%% Constraints -> EBM %%%%%%%%%%%%%%%%%
%%%%%%%%%%%%%%%%%%%%%%%%%%%%%%%%%%%%%%%%%%%%%%%%%%
\subsection{From Moment Constraints to EBM}
\label{sec:constraints2ebm}
Let's now consider a set of desired moment constraints $\tbmu$.\footnote{Boldface $\bphi$ and $\bmu$ represents vectors of real values (features and moments).}
In the general case (i.e., when some constraints are distributional), we use Theorem 1.(B), which says that the desired energy-based model $P$ can be approximated arbitrarily closely in the following form: 
\begin{align}
    P(x) \doteq a(x) e^{\blambda \cdot \bphi(x)} .
\end{align}

This EBM defines the desired normalized distribution $p(x) \doteq \frac{P(x)}{Z}$, where $Z \doteq \sum_x P(x)$. What is left is to learn appropriate values for the parameter vector $\blambda$ s.t.: 
\begin{align}
\E_{x\sim p} \bphi(x) \simeq \tbmu.
\end{align}

We address this problem through Algorithm \ref{al:computing_lambdas}. First, we sample a large number $N$ of sequences $x_1 \ldots x_j \ldots x_N$ from $a$. On line 2, we define ``importance weights'' $w_j(\blambda) \doteq \frac{P(x_j)}{a(x_j)} = \exp\sprod{\blambda}{\bphi(x_j)}$. On line 3, we then use {SNIS} (Self Normalized Importance Sampling) %\citep{A-parshakova-etal-2019-global,KimBengio2016,owen_chapter_importance_sampling_2013}:
\citep{KimBengio2016,A-parshakova-etal-2019-global} to estimate  $\bmu(\blambda)\doteq \E_{x\sim p} \bphi(x) $.
% \begin{align}
%     \bmu(\blambda) 
%     \simeq \hat{\bmu}(\blambda) = \frac{\sum_i w_i(\blambda)\ \bphi(x_i)}{\sum_i w_i(\blambda)}.
% \end{align}
SNIS consists in computing:
\begin{align}
\hat{\bmu}(\blambda) = \frac{\sum_{j=1}^N w_j(\blambda)\ \bphi(x_j)}{\sum_{j=1}^N w_j(\blambda)},
\end{align}
and it can be shown that $\hat{\bmu}(\blambda) \simeq \bmu(\blambda)$, with convergence 
%with $N\rightarrow \infty$
in the limit
\citep{owen_chapter_importance_sampling_2013}.
% \footnote{In our experiments, we take $N \simeq 20K$.} 

%$\bmu(\blambda) \simeq \hat{\bmu}(\blambda) = \frac{\sum_i w_i(\blambda)\ \bphi(x_i)}{\sum_i w_i(\blambda)}$.  

Note that the estimate $\hat{\bmu}(\blambda)$ is obtained not as a single number, but as a parametric function of the variable $\blambda$. We want to find $\blambda$ such that $\hat{\bmu}(\blambda) = \tbmu$, a question that we handle on line 4 by performing an SGD optimization 
% \MDFN{I wonder if the rhs term is convex in lambda, in which case the optimization would have nice properties.}
over the objective $\min ||\tbmu - \hat{\bmu}(\blambda)||_2^2$.%
\footnote{$\bmu(\blambda)$ can approximate $\tbmu$ arbitrarily closely, and we know from SNIS theory that with increasing $N$, $\hat{\bmu}(\blambda)$ will become arbitrarily close to $\bmu(\blambda)$. In our experiments we stop the SGD optimization when $||\tbmu - \hat{\bmu}(\blambda)||_2^2$ becomes smaller than $0.01$.}
%\OK, {We are finessing some issues here: quality of the snis estimate, its convexity (?), the fact that the min can reach 0, the problem of infinite lambdas...}
%

At the end of this process, we obtain an estimated value for the parameter vector $\blambda$, and a representation $P(x) = a(x) \exp \sprod{\blambda}{\bphi(x)}$. While $a(x)$ is a normalized distribution by construction, the introduction of the second factor loses this normalization property, making $P(x)$ an EBM.%
\footnote{The class of Energy-Based Models (EBMs) \citep{lecun_tutorial_2006} is much larger than the exponential family models we are considering in this paper. An EBM $P(x)$ is just any unnormalized distribution over an input space $X$, in other words a mapping $P$ from $X$ to the non-negative reals. The terminology comes from physics, and corresponds to writing $P(x)$ in the form $P(x) = e^{-E(x)}$, $E$ being called the ``energy'' associated with $x$.}
\footnote{A question was raised by an anonymous reviewer about the viability of adding new constraints incrementally. The answer is yes, more details provided in the Appendix,  \S\ref{sec:appendix-incrementally}.}
%
% From now on, we will assume that the value obtained for $\blambda$ is fixed and left out of the notation, and simply use $P$ and $p$ respectively for $P_\blambda$ and $p_\blambda$; we also use the notation $Z = \sum_x P(x)$ for the partition function of $P$.
% \HEFN{I believe this part might be clearer with a simple pseudo code in the appendix.}
%
%
% We finally note that in the special case of strictly pointwise constraints, we can exploit the simplification derived in the previous section
% \gk{it wasn't clear to me which one}
% , and obtain an EBM representation of the form $p(x) \propto P(x) = a(x) b(x)$, a shortcut we will use in the strictly pointwise experiments.%
%
\subsection{From EBM to Autoregressive Policy}
\label{sec:ebm-2-policy}
%
%%%%%%%%% KLDPG ALGORITHM %%%%%%%%%%%%%%%%
\begin{wrapfigure}{r}{0.4\textwidth}
\vspace{-30pt}
\begin{minipage}{0.4\textwidth}
\begin{algorithm}[H]
\caption{\ KL-Adaptive DPG \label{al:KL-adaptive-DPG}}
\begin{small}
\begin{algorithmic}[1]
\Require $P$, initial policy $q$
\State $\pi_\theta \gets q$
\For{each iteration}
\For{each episode}
    \State sample $x$ from $q(\cdot)$
    \State $\theta \gets \theta + \alpha^{(\theta)} \frac{P(x)}{q(x)}\ \nabla_\theta \log \pi_\theta(x)$ 
\EndFor
% \If{}
\If{ $\KL(p||\pi_\theta) <  \KL(p||q)$} 
    \State $q \gets \pi_\theta$
\EndIf
\EndFor
\Ensure $\pi_\theta$
\end{algorithmic}
\end{small}
\end{algorithm}
\end{minipage}
%\vspace{10pt}
\end{wrapfigure}
%%%%%%%%%%%%%%%
%

The EBM representation just obtained for $P$ defines the optimal $p = Z^{-1} P$ unambiguously, a crucial intermediate step in the solution of our problem. 
% $$HE$$
% Under an \textbf{\textit{optimization}} intent as in the standard RL objective, one can obtain a policy of which samples maximize the energy P(x). %% "maximize" here because our energy P(x) formalization
% However such policy concentrate the probability mass on only on one or few sequences. 
% %
% Conversely, 
% TBCTBCTBC%%%
%
From it we can immediately compute ratios of the form $p(x)/p(x')$ for two sequences $x,x'$, %
but without knowing $Z$, we cannot compute $p(x)$ and, \emph{even} with such a knowledge, we cannot produce samples from $p$. 

This problem is typical of EBMs at large: they provide a rich and flexible mechanism for specifying models, but they leave a gap between representation and exploitation. 
A range of techniques, from sophisticated MCMC approaches (especially for continuous models in vision) to contrastive learning techniques, have been developed for bridging this gap. 

One technique that is suitable for our objective here, namely sampling from a sequential EBM that includes an autoregressive component $a(x)$, is the DPG (``Distributional Policy Gradient'') algorithm~\citep{opt-rl-arxiv-2019}.

The objective of DPG is to obtain an autoregressive policy $\pit$ that approximates $p$, where approximation is formalized in terms of making the cross-entropy $CE(p,\pit) = -\sum_x p(x) \log \pit(x)$ as small as possible.\footnote{This is equivalent to minimizing $\KL(p,\pit) = CE(p,\pit) - H(p)$.} 
%
% The algorithm (Algorithm \ref{al:KL-adaptive-DPG}}) 
DPG exploits the fact that, for any ``proposal'' distribution $q$ whose support contains the support of $p$, we have  
$$\nabla_\theta CE(p,\pit) = - \nabla_\theta \E_{x\sim p} \log \pit(x) =
-  \E_{x\sim p} \nabla_\theta \log \pit(x) =
- \E_{x\sim q} \frac{p(x)}{q(x)}\ \nabla_\theta \log \pit(x)$$
%
% \gk{Shouldn't the first equation be $\nabla_\theta \E_{x\sim p} \log \pit(x) $},
where the last equality is an instance of importance sampling.\\
Our ``KL-adaptive'' version of DPG is shown in (Algorithm \ref{al:KL-adaptive-DPG}).
We start from an input EBM $P$, along with an initial policy $q$ which is a proxy to $p$; in our case we take $q=a$. During an iteration (think minibatch or set of minibatches), we sample a number of sequences from $q$, do an SGD update of $\theta$ (line 5), where $P$ is used instead of $p$ (noting that they only differ by a multiplicative constant), and where $\alpha^{(\theta)}$ is a learning rate. 
The efficiency of the algorithm is related to how close the proposal $q$ is to the target $p$,%
\footnote{In the limit where $q$ were equal to $p$, the algorithm would be identical to standard supervised training, except that samples would be obtained directly from the underlying process $p$ rather than a training set of samples.}
The algorithm is \emph{adaptive} in the sense that it modifies $q$ periodically to take advantage of the evolving approximations $\pit$. 
%On line 6, we test whether the current $\pit$ is ``superior'' to $q$, and if the test is positive we update $q$ to $\pit$ on line 7. 
On line 6, we we test whether the current $\pit$ is closer than $q$ to $p$ in terms of KL-divergence, and if so  we update $q$ to $\pit$ on line 7.%
\footnote{In the original DPG, the superiority test is done on the basis of the log-likelihood on a validation set. Here we are in the more demanding situation where no validation set is available. To directly estimate the KL divergence from $p$ (line 6), we exploit the identity $\KL(p\|\pi) = -\log Z + 1/Z\ \mathbb{E}_{x\sim q(x)} \frac{P(x)}{q(x)} \log \frac{P(x)}{\pi(x)}$.
See \S\ref{sec:appendix:kld} for derivations and a comparison with using Total Variation Distance (TVD) for assessing divergence.}
\S\ref{sec:appendix:exp-ablation} provides an ablation study showing the effectiveness of this adaptive step for obtaining faster convergence.

%% file: sections/experiments.tex
\section{EXPERIMENTS, RESULTS, AND EVALUATION}
\label{sec:EXPERIMENTS}
In this section we describe our evaluation methodology and perform experiments on pointwise constraints (\S\ref{sec:exp-pointwise}) and on distributional and hybrid constraints (\S\ref{sec:exp-distributional}). The Appendix contains a detailed view of evaluation~(\S\ref{sec:appendix:experiments}), comparison with extra baselines~(\S\ref{sec:appendix:extra-baselines}), and an ablation study (\S\ref{sec:appendix:exp-ablation}). 

\subsection{Evaluation Metrics} \label{sec:metrics}
The main metrics we report are: 
(1) $\E_{x\sim\pi_\theta} \phi_i(x)$, assessing the ability of $\pit$ to reach the expectation goal on the $i$-th constraint, 
(2) $\KL(p || \pi_\theta)$, the forward KL divergence from the optimal distribution (which should be as close to 0 as possible),
(3) $\KL(\pi_\theta || a)$, the reverse KL divergence from the original GPT-2; for details on the estimation of these metrics see \S\ref{sec:appendix:kld}.\\ 
Previous work has mostly focused on the diversity of each individual output using Dist-1,2,3 scores~\citep{li-etal-2016-diversity} to measure repetitions within a \emph{single} generated sequence. However, the shortcomings in terms of \emph{sample} diversity, of optimization techniques when training generative models for text, has recently been
documented in \citep{GAN_short}. So additionally, we report Self-BLEU-3,4,5 \citep{texygen-ZhuLZGZWY18} to measure repetitions at a distributional level across the whole set of generated samples, and also provide a token/type frequency analysis (see Fig.~\ref{fig:main-zipf} and \S\ref{sec:appendix:zipf}). \\
%% maybe not so important:
Note that KL divergence from the original GPT-2 also implicitly captures sample diversity: a distribution that focuses all its probability mass on a few sequences typically displays high divergence from GPT-2. Implementation details and hyper-parameters are available in the Appendix~(\S~\ref{sec:appendix:hyperparams}).

%%%%%%%POINTWISE%%%%%%%%%
\subsection{Pointwise Constraints Experiments}
\label{sec:exp-pointwise}

Pointwise constraints are of the form $\E_p \phi_i(x) = 1$, with $\phi_i$ a binary feature. Contrarily to distributional constraints, they can be directly associated with a ``reward'', namely $\phi_i$ itself. RL-inspired baselines can then be introduced naturally, and this is what we do here.

% To study how each training objective can fine-tune the original policy to satisfy the imposed constraints while keeping minimum deviation from the original pretrained LM. 

%\paragraph{Experiments}
\textbf{Single-Word constraints:}
Here we constrain the presence of a specific word $w$ in the generated text i.e. $\phi(x) = 1$ iff $w$ appears in the sequence $x$. We use 9 single-word constraints of different rarity levels: 
``US" (original frequency: $7{\cdot}10^{-3}$), 
``China" ($4{\cdot}10^{-3}$),
``Canada" ($2{\cdot}10^{-3}$), 
``amazing" ($1{\cdot}10^{-3}$),
``Paris" ($5{\cdot}10^{-4}$), 
``restaurant" ($6{\cdot}10^{-4}$),
``amusing" ($6{\cdot}10^{-5}$),
``Vampire" ($9{\cdot}10^{-5}$),
``Wikileaks" ($8{\cdot}10^{-5}$).
\\
% In order to assess the impact of the adaptivity of DPG, we conduct as set of single-word experiments where $w$ is chosen with different rarity levels. More precisely,  It makes sense for more rare words to be harder to control for since we have a lower probability to obtain samples that contain it.
\textbf{Word-list constraints:}
We use $4$ different word lists among those proposed in \citep{plug_and_play_20}, covering the following topics: ``kitchen", ``fantasy", ``politics", and ``computers". We set $\phi_l(x)=1$ if $x$ contains at least one one word from the word list $l$. \\
\textbf{Classifier-based constraints:}
We use pre-trained classifiers from \citep{plug_and_play_20}, which consist of a linear head on top of GPT-2. We select 4 classes and define corresponding pointwise constraints: ``very positive", ``positive", ``very negative" and ``Clickbait".\MDFN{Is clickbait clear to all readers?} 
See \S\ref{sec:appendix:hyperparams} for details on constraint computations. \\
\textbf{Baselines:} \label{sec:baselines}
We compare our method \textit{\GDC} to three baselines: (1) \textit{\REINFORCE}~\citep{Williams92Reinforce}, using the reward $\phi(x)$, i.e. trying to maximize $\E_{\pit} \phi(x)$;
(2) \textit{\REINFORCEP}: Reinforce again, but now using the reward $P(x)$ based on our energy model $P$, i.e. maximizing $\E_{\pit} P(x)$;
this baseline starts from the same optimal EBM $P$ representation as \GDC but with a standard optimization objective rather than a distributional one; in other words, while \GDC tries to get a similar \emph{sampling} distribution to $p$, this baseline tries to get sequences of \emph{maximal} probability $p(x)$.  
%by this comparison we show the suitability of our KL-Adaptive DPG algorithm in learning an autoregressive policy that approximates the target distribution $p$;
%
(3) \textit{\ZIEGLER}~\citep{Ziegler19}: an approach relying on the RL Proximal Policy Optimization (PPO) algorithm \citep{PPO} and which tries to maximize the objective $\E_\pit \phi(x) - \beta \KL(\pit,a)$, which \emph{interpolates} the reward $\phi(x)$ with a KL-divergence penalty from the pretrained model, but where the goal is not explicitly to satisfy a constraint; for a geometric illustration of the differences with \GDC see \S\ref{sec:illustration-ziegler}. \S\ref{sec:appendix:extra-baselines} provides a comparison of \GDC with two additional baselines.\\
%
%to maximize a reward $\E \phi(x) + \beta \KL(\pi_\theta,a)$which is a combination of $\E \phi(x)$ and a $\KL$ penalty on the reward to prevent the fine-tuned model from drifting too far from the pretrained model re-weighted by an adaptive hyperparameter $\beta$.% 
%\HEFN{add note to the diagram in the appendix}
%
%%%%%%%%%%%%%%%%%%%%%%
%% Figures pointwise %
%%%%%%%%%%%%%%%%%%%%%%
\begin{figure}
\includegraphics[width=\linewidth]{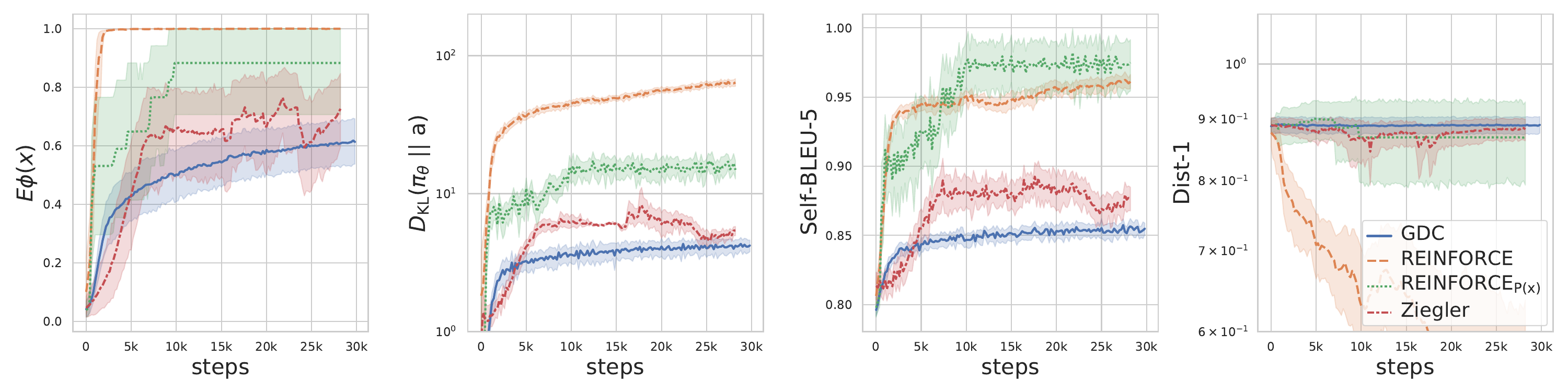}	
\caption{Eval. metrics $\E \phi(s)$, $\KL(\pit\|a)$ ($\downarrow$ better), Self-BLEU-5 ($\downarrow$ better), and Distinct-1 ($\uparrow$ better), aggregated across $17$ point-wise experiments (single words, wordlists, discriminators), performed at each 10 gradient updates, for policies obtained from \GDC against three training baselines \REINFORCE, \REINFORCEP and \ZIEGLER. See Appendix~\ref{sec:appendix:experiments} for a detailed view for each experiment and more evaluation metrics.
\label{fig:exp123-main}}
\end{figure}
%%%%%%
\begin{wrapfigure}{H}{0.4\linewidth}
\center
\includegraphics[width=0.4\textwidth]{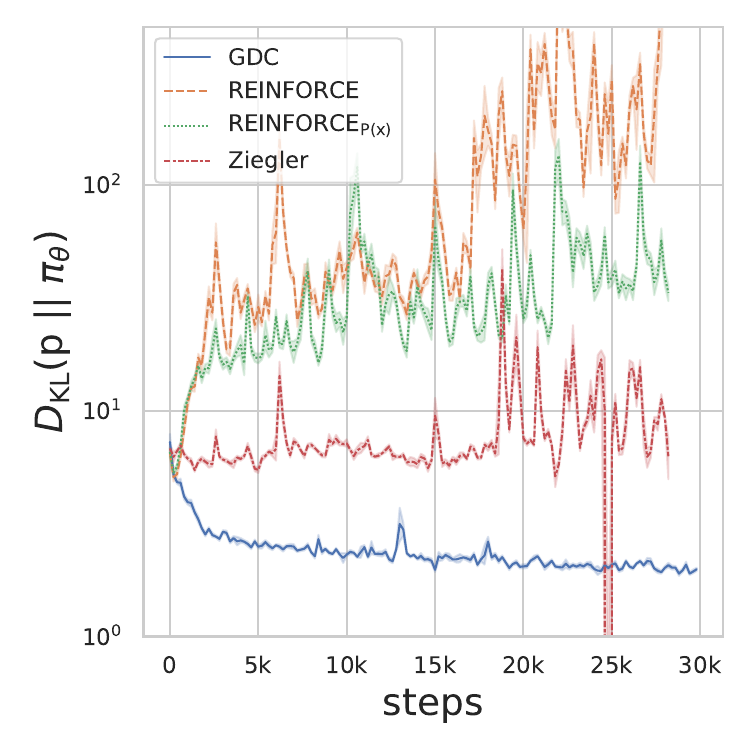}
\caption{
\GDC steadily decreases the KL deviation between the trained policy $\pi_\theta$ and the target distribution $p$.
The Figure is aggregated across $17$ point-wise constraints experiments, see Appendix~\ref{sec:appendix:experiments} for a separate view of each experiment.
\label{fig:klp}
}
\end{wrapfigure}
%

%%%%%%%%%%%%%%%%%%%%%%%%%%%%%%%%%%%%%%%%%%%%%%%%%%%%%%%%%%%%%%%
%%%%%%% comments on pointwise experiments results %%%%%%%%%%%%%
%%%%%%%%%%%%%%%%%%%%%%%%%%%%%%%%%%%%%%%%%%%%%%%%%%%%%%%%%%%%%%%
\textbf{Results:} Figure ~\ref{fig:exp123-main} shows the evolution of the metrics over training steps, aggregated across the $9+4+4=17$ experiments. We observe the following:
the baseline \REINFORCE, which does not have any explicit link in its objective to the pretrained GPT-2, converges very early in the training, reaching a maximum value of $\E_{\pit} \phi(x)$ at the expense of a very large deviation from the original GPT-2. High values of $\KL(\pit|a)$, are translated into low Dist-1 and very high Self-BLEU-5 indicating degeneration and lack of diversity.
\REINFORCEP maximizes the energy model 
% \MDFN{Energy $\rightarrow$ Energy Model (the Energy is $-\log P(x)$}
$P$ by peaking on a few sequences only; this can yield high values of $\E_{\pit} P(x)$, at the expense of low sample diversity as demonstrated in the highest values of SELF-BLEU-5 scores among baselines.%
\footnote{The difference with \REINFORCE makes sense if one observes that $\phi(x)$ can be maximized on many sequences, while $P(x)$ tries to maximize $a(x) \cdot \phi(x)$, which is typically maximized on only one sequence.}
%

\begin{comment}
In the case of \ZIEGLER,  we can see a positive effect of the 
adaptive\MDFN{adaptive?}
interpolation factor $\beta$ that re-weights the KL penalty in the objective function. 
With low values%
\MDFN{Do you mean low \emph{initial} values ?} 
of $\beta$ it deviates away 
and shows in-stability in training (see appendix~\ref{sec:appendix:experiments} for more severe cases).%
\MDFN{I am confused here: I thought that $\beta$ was evolving towards a point defined by a KL-divergence fixed heuristically in Ziegler's implementation. I think we should avoid giving the impression that Ziegler is trying to satisfy the constraint (it does not) or that it has a well defined quantitative objective. It does not, I think ??? }
\end{comment}

\begin{figure}
	\centering
	\begin{minipage}[b]{0.4\textwidth}
	\includegraphics[width=\textwidth]{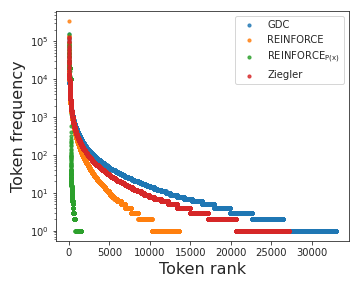}
	\footnotesize{\caption{``Zipf-like'' token frequency analysis on sets of 68000 generated samples from each method (only samples strictly satisfying the constraints are kept, for fair comparison). Longer tails mean a lower concentration of mass on the high frequency tokens, and therefore indicate more vocabulary richness. See Appendix~\ref{sec:appendix:zipf} for details.
	\label{fig:main-zipf}}}
	\end{minipage}
	\hfill
	\begin{minipage}[b]{0.58\textwidth}
		\begin{table}[H]
			\tiny
% 			\hspace{cm}
			\begin{tabular}{p{0.3cm}|p{0.3cm}p{6cm}}
				\toprule
				Reps & \textbf{$\phi(x)$} &  \\ 
				\midrule
				\multicolumn{3}{c}{\textbf{GDC}} \\ 
				1&1& ``Thank you all for the service this site gives me , ” he \ye{said}. ...\\ 
				1&1& This  \ye{book}  is incredibly rich , entertaining , and extremely enjoyable...\\ 
				\hline
				\multicolumn{3}{c}{\textbf{REINFORCE}} \\
				1&1&Featuring the \yh{highest} \yh{quality} \yg{performance}  \yg{performance} \yg{performance}...\\
				1&1&This  \ye{beautiful}  \ye{beautiful}  \yh{quality}   \ye{production}   \yh{quality}   \yg{high}   \yh{quality}...\\ 
                1&1&High  \yh{quality}  \yg{performance}   \yg{high}   \yh{quality}   \yg{performance}  product ...\\ 
				\hline
				\multicolumn{3}{c}{\textbf{REINFORCE\_P(x)}} \\ 
				10k&1&Thank you for  \yg{supporting}  the  \yg{journalism} that our \yg{community} \yg{needs!} ...\\ 
				\hline
				\multicolumn{3}{c}{\textbf{ZIEGLER}} \\ 
				4418&1&Thank you for  \yg{supporting}  the  \yg{journalism}  that our  \yg{community} \yg{needs!} ...\\ 
				3560&1&Be the  \yg{first}  to know. No  \yg{one} \yg{covers} what is  \yg{happening} in our...\\ 
				\bottomrule 
			\end{tabular} 
			
			\footnotesize{\caption{Examples of generations controlled by a discriminator on the class label ``\textit{very positive}''. 
			%\textbf{$\phi(x)$} indicates whether the constraint is met. 
			\texttt{Reps} is the frequency of the whole sequence in a corpus of 10k samples. Tokens highlighted in \ye{y}\yf{e}\yf{l}\yg{l}\yg{o}\yj{w} with different intensities indicates their overall frequencies in the generated corpus. Generations are trimmed to $15$ tokens for display purposes. See \S\ref{sec:appendix:generations} a full list of generations .
			 \label{table:generation-main}}}
			\end{table}
	\end{minipage}
\end{figure}

%%%%%%%%%%%%
%\textit{Alternative: 
In the case of \ZIEGLER  we can see a positive effect of the interpolation factor $\beta$ between the reward and the KL penalty in the objective function. In the aggregated experiments reported here, the reward is slightly better than with \GDC, but with inferior diversity scores (see also Fig.~\ref{fig:main-zipf}, showing that \GDC produces richer vocabulary), and the stability is much worse (a detailed view of each experiment is provided in \S\ref{sec:appendix:experiments}, showing more clearly the instability of this baseline). %\todo{Discuss the previous paragraph}
A complementary evaluation is provided by Figure \ref{fig:klp}, focusing on the ability of $\pit$ to converge to the optimal distribution $p$. We see that \GDC is superior to all baselines in terms of $\KL(p\|\pit)$ and also much more stable.

In summary, in these experiments, we see that with \GDC the constraint expectation $\E_\pit\phi(x)$ smoothly increases while $\pit$ maintains the lowest divergence from GPT-2, becomes closest to the optimal $p$, and has the best diversity scores overall. On the other hand, we also note that at the point where we stop training (30K steps), the average over experiments of $\E_\pit\phi(x)$, while still increasing,  does not reach $100\%$, an issue that we discuss at the end of the paper (\S\ref{sec:discussion}).

\subsection{Distributional and Hybrid Constraints Experiments}
\label{sec:exp-distributional}
%We first start by describing the setting for our distributional constraints. 
%
% Ziegler and reinforce they cannot deal with distributional constraints, as they tend to maximize b(x)
% our EBM view to distributional constraints makes it so natural to control such biases exactly. 
As formalized in \S\ref{section:formalization}, \GDC permits to define pointwise and distributional constraints as well as any mix between them.
This unique feature makes it very suitable to remedy biases that the text generation model may have, a problem identified in several previous works~\citep{babysitter_ShengCNP19}.
%
% Most existing works on Controlled Generation have taken what we have called a \emph{pointwise} view, making them unsuitable for handling distributional requirements.\\
%% Experiments brief
 We employ \GDC to balance gender and profession distributions across biographies generated by a GPT-2 model fine-tuned on Wikipedia Biographies~\citep{DBLP:conf/emnlp/LebretGA16} \textit{(henceforth GPT-2\textsuperscript{bio})} (\S\ref{sec:dist-exp-details} gives additional details).
%
%% Initial bias
The bias in GPT-2\textsuperscript{bio} is significant: we calculated that this model generates only around 7\% female biographies. It also displays a large imbalance between professions related to ``Science" ($1.5\%$), ``Art" ($10.0\%$), ``Business" ($10.9\%$) and ``Sports" ($19.5\%$).
\textbf{Experiment 1: Single Distributional Constraint\em} We use the distributional constraint $\E_{x \sim p} \phi_{female}(x) = 0.5$; \GDC is able to reduce the bias of GPT-2\textsuperscript{bio} to obtain $35.6\%$ female biographies rather than only $7.4\%$ (see Fig.~\ref{table:distributional} for this experiment and the next ones).\\
\textbf{Experiment 2: Multiple Distributional Constraints\em} We then test our framework with several distributional constraints of different values and control directions. We specify four distributional constraints all at once with the goal of \textit{increasing} the expectations of ``science" and ``art" to $40\%$ and \textit{decreasing} those of ``sports" and ``business" to $10\%$. \GDC is able to increase the expectations of the first two professions respectively from $1.5\%$ to $20.3\%$ and from $10$ to $31.6\%$  and to decrease those of ``business" and ``sports" respectively from $10.9\%$ to $10.2\%$ and from $19.5\%$ to $11.9\%$, reaching expectations 
% very 
close to the desired ones for all features using a single training method.\\
\textbf{Experiments 3,4,5,6: Hybrid Constraints\em} Here we want to de-bias the model as in the previous case but we single out biographies of scientists, artists, etc. Formally, our requirements become 
$\E_{x \sim p} \phi_{profession}(x) = 1.0$, a pointwise constraint,
and
$\E_{x \sim p} \phi_{female}(x) = 0.5$, a distributional constraint.
In those 4 hybrid experiments we can clearly see that \GDC can address both pointwise and distributional constraints increasing each simultaneously with just the right amount to reach the desired expectations. Appendix \S\ref{sec:dist-exp-details} further elaborates Fig.~\ref{table:distributional} (convergence curves).
\begin{table}[t]
\scriptsize
\centering
\setlength\extrarowheight{1pt} % or whatever amount is appropriate
\begin{tabular}{p{0.05cm}p{1.7cm}|p{1.66cm}|p{2.4cm}|p{2.4cm}}
\toprule
  & \textbf{Aspect} &  \textbf{Desired} & \textbf{Before} &  \textbf{After}\\
 \hline
 \multicolumn{5}{c}{\textbf{Single Distributional constraint}}\\ 
 \hline
 1 &  Female&     50\% &  \VChart{07.4}{0.074}{0.5} &  \VChart{36.7}{0.367}{0.5}  \\
 \hline
 \multicolumn{5}{c}{\textbf{Multiple distributional constraints}}\\ 
 \hline
  2  &       Art &      40\% $\uparrow$ &   \VChart{10.9}{0.109}{0.40} & $\uparrow$\VChart{31.6}{0.3164}{0.40}  \\
  &  Science &       40\% $\uparrow$   &   \VChart{01.5}{0.015}{0.40} &  $\uparrow$\VChart{20.1}{0.2031}{0.40} \\
  &   Business &     10\% $\downarrow$ &   \VChart{10.9}{0.109}{0.10} &  $\downarrow$\VChart{10.2}{0.102}{0.10} \\
  &   Sports &       10\% $\downarrow$ &   \VChart{19.5}{0.195}{0.10} &  $\downarrow$\VChart{11.9}{0.119}{0.10} \\
  \hline
  \multicolumn{5}{c}{\textbf{Hybrid constraints}}\\ 
  \hline
    3 &    Female&     50\% &   \VChart{07.4}{0.074}{0.5} &  \VChart{31.9}{0.319}{0.5}  \\
      &    Sports &     100\% &  \VChart{17.5}{0.175}{1} &  \VChart{92.9}{0.929}{1} \\
    \hline
    4 &    Female &     50\%&   \VChart{07.4}{0.074}{0.5} &  \VChart{36.6}{0.366}{0.5} \\
      &       Art &     100\%&  \VChart{11.4}{0.114}{1} &  \VChart{88.6}{0.886}{1} \\
    \hline
    5 &    Female &     50\%&   \VChart{07.4}{0.074}{0.5} &  \VChart{37.7}{0.377}{0.5}\\
      &  Business &     100\%&  \VChart{10.1}{0.101}{1} &  \VChart{82.4}{0.824}{1} \\
    \hline
    6 &    Female&     50\% &   \VChart{07.4}{0.074}{0.5} &  \VChart{28.8}{0.288}{0.5} \\
      &   Science&     100\% &   \VChart{01.2}{0.012}{1} &  \VChart{74.7}{0.747}{1} \\
      \bottomrule
\end{tabular}
    \caption{Distributional and hybrid constraints experiments demonstrating the generality of \GDC in dealing with this mixed type of constraints. $\uparrow$/$\downarrow$ indicates which direction (increasing/decreasing) improves the target expectation. See Appendix \S\ref{sec:dist-exp-details} for convergence curves.}
    \label{table:distributional}
\end{table}

%% file: sections/conclusion.tex
\section{Discussion}
\label{sec:discussion}

Our approach to controlled text generation is distinguished by its breadth
--- the first one to handle distributional along with pointwise constraints, with applications to the important problem of Bias in pretrained LMs --- and by the transparency of the supporting formalism. It decouples the training objective along two different dimensions. The first consists in solving the initial constraints specification, and leads through a direct algorithm to an optimal solution in EBM format. The second, where the real computational difficulty lies, consists in approximating this EBM with an autoregressive policy for use at inference time. \\
Sampling from an EBM is an important, hard, and well-identified challenge in the literature. Our approach there consists in proposing a KL-adaptive version of the DPG algorithm, which exploits ascertained improvements of the trained policy to speed up convergence. \\
This is an effective method for rare events, as we show in an ablation study (\S\ref{sec:appendix:exp-ablation}). In the case of pointwise constraints,  where comparisons with baselines can be done, our experiments show the method's superiority in satisfying the constraints while avoiding degeneration. Reaching close to 100\% samples meeting the constraints, can sometimes be obtained in these baselines, but only at a severe cost in terms of quality and sample diversity. Of course, if we do not care about such aspects, obtaining 100\% constraint satisfaction is trivial: just generate \emph{one} sentence satisfying the pointwise constraint!

Our method 
\HEFN{Could mention that Ziegler, CTRL and Plug and play reach 100\% constraint satisfaction. We could formulate it to be
Controlled NLG reaching imposed constraints without degeneration.
}
does not suffer from degeneration, but our end policies still generate a number of samples not satisfying the constraints. A possibility, left for future work, might consist in filling the moderate residual gap with MCMC techniques, which would be guaranteed to reach our optimal $p$ in the limit. We do not go this route here, but conduct an experiment (see \S\ref{Appendix:supervised-experiment}) to better understand the nature of the problem. In the simple case of a single-word  constraint ($x$ includes \textit{``amazing"}), we sample directly 1M samples from GPT-2 and keep the roughly 5K samples containing \textit{amazing} (a variant of rejection sampling, taking two processing days). 
We then do a standard supervised fine-tuning of GPT-2 with these samples, stopping training when the CE validation loss starts to increase, and observe that this model exhibits a worse constraint satisfaction rate than ours.
This experiment does not mean that a much larger fine-tuning dataset, obtained in this slow, non-adaptive way, would not reach better statistics, but it raises doubts about the ability of the GPT-2 architecture to fine-tune over such a non-standard constraint as containing a given word \emph{somewhere} in its output.%
%\footnote{Note how difficult its job would be if the constraint was based on a hash-based predicate filtering one sentence out of two. However the case of a single-word constraint does appear to be simpler.} 

Overall, 
%even if we do not solve our objective completely, 
we believe that 
the proposed decomposition into two sub-problems is a methodological advantage compared to most other works, which directly aim at training a policy with the goal of improving certain evaluation metrics, but without clearly defining what qualifies as an optimal solution. The computational challenge of fully bridging the gap between the optimal EBM and an efficient sampling engine remains, and we hope that the formalism we propose, along with initial applications and experimental validations, will motivate further research along these lines.

%We believe that the real challenge lies in bridging the gap between the optimal EBM and an efficient inference engine, a challenge for which we propose a promising partial solution.

%\subsubsection{What makes a constraint hard to approximate using a policy?}
% % add a analysis here it is not about how rare a condition is but it is about how complex the constraint is
% % combine several words and test DPG against
% % give an extreme example (run experiment?) with random-deterministic scorer

% Reaching 100%:
% - previous related work (plug and play) doesn't reach 100%
% - Ziegler doesn't reach 100%
% - rarity of the constraint not the reason. maybe add comments in the appendix. 
% - discriminator built on top gpt2 will not be able approx the constraint you will not be able to fine tune ur model to do it. 
% - supervised experiment Muhammad did 

%% file: sections/appendix.tex
\renewcommand \thepart{}
\renewcommand \partname{}
\part{Appendix}

\section{Details on Formalization (\S\ref{section:formalization})}
\subsection{Comments on Theorem 1}
\label{appendix-csizar}

Our statement of Theorem 1 is actually a reformulation of two results in section 3 of \citet{csizarShields2004}. Our property (A) is a simple notational transposition of their Remark 3.1 (p.~444). Property (C) is the Pythagorean Identity in their Theorem 3.2 (p.~442). Property (B) reformulates  the last part of the same Theorem 
``... and in general $\mathcal{L} \cap \text{cl}(\mathcal{E}_Q) = {\{P^{*}\}}$'' in terms of a limit of a sequence of distributions.

Note: \citet{csizarShields2004} assume a finite $X$ here, but generalizations to infinite (countable and/or continuous) $X$ spaces are possible, see \citep{Csiszar1975}.

\subsection{The case of Pointwise constraints in \S\ref{sec:constraints2ebm}}
\label{appendix:details_pointwise}
In the case of purely pointwise constraints, if $b(x) = 1$, then the distribution $c = \delta_{x}$ is in $\CC$, hence $x\in X_\CC$. Conversely, if $x\in X_\CC$ then there is some $c\in \CC$ such that $c(x) > 0$, implying that $b(x)=1$. Hence $X_\CC = \{x\in X |\ b(x)=1\}$. Thus, in equation (\ref{eq:exponential_with_X_C}), $P(x) = a(x) b(x) \exp \sum_i \lambda_i \phi_i(x)$; but for $b(x) \neq 0$, $\phi_i(x)=1$, so the exponential factor is a constant, which proves that $P'(x) = a(x) b(x)$ is proportional to $P(x)$, and therefore $p(x) \propto P'(x)$.

\subsection{Incrementally adding new constraints}
\label{sec:appendix-incrementally}
An interesting question\footnote{raised by an anonymous reviewer of our ICLR submission.} is whether the process explained in \S\ref{section:formalization} can be made incremental: if one has already computed a $p$ and a $\pit$ relative to a certain number of constraints, can one add a new constraint without restarting the whole process from scratch? The answer is yes, and here we provide some formal elements to understand why.

\subsubsection{Transitivity Property of Generalized MaxEnt}

According to \citep{csiszar96}, the Generalized MaxEnt of sections \S\ref{sec:constraints-info-geo} and \S\ref{sec:constraints2ebm} has the ``Transitivity property''. In our notation, this says that if we have $k' > k$ constraints, with $C$ the manifold of distributions respecting only the first $k$ constraints, $C'$ the manifold respecting all $k'$ constraints (hence $C' \subset C$), then the maxent projection $p'$ of $a$ onto $C'$ can be obtained by first projecting $a$ onto $C$, obtaining $p$, and then projecting $p$ onto $C'$, obtaining $p'$. In particular, the $k$ lambdas associated with $p$ can be directly reused as the first lambdas of the $k'$ lambda’s associated with $p'$. 

\begin{figure}[H]
\centering
\includegraphics[trim=0 30 50 0,width=0.5\columnwidth]{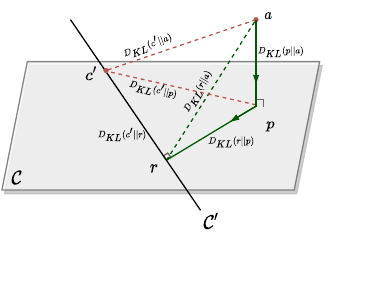}
\caption{Transitivity of Information Projection (aka Generalized MaxEnt).} \label{fig:appendix-Transitivity}
\end{figure}

\citep{csiszar96} gives only a minimal proof sketch, but it is instructive to provide the details, as we do now, because the proof is a neat illustration of the power of information geometry for problems of the kind we consider. The proof, illustrated in Figure \ref{fig:appendix-Transitivity},
is very similar to one of the proofs for the transitivity of the orthogonal projection in Euclidean geometry.

\textbf{Proof}. In the Figure, $p$ is the information projection (Csiszar's terminology for the Generalized Maxent) of $a$ onto $\mathcal C$, as before. Let's define $r$ to be the projection of $p$ onto $C'$. We need to prove that $r$ is identical to the projection $p'$ of $a$ onto $C'$. 
We consider an arbitrary distribution $c'$ in $C'$, and apply the Pythagorean Identity of Theorem \ref{theorem:main} three times. Because $p$ is the projection of $a$ onto $C$, we have $\KL(r,a) = \KL(r,p) + \KL(p,a)$ and also $\KL(c',a) = \KL(c',p) + \KL(p,a)$. Because $r$ is the projection of $p$ onto $C'$, we have $\KL(c',p) = \KL(c',r) + \KL(r,p)$, hence $\KL(c',p) \geq \KL(r,p)$. Putting these three facts together, we find that $\KL(c',a) \geq \KL(r,a)$. As $c'$ is an arbitrary point of $C'$, this proves that $r$ is the projection of $a$ onto $C'$, in other words, $r = p'$.

\subsubsection{Transitivity and Autoregressive Policy}
Due to the Transitivity property, when calculating the EBM representation, it is possible to start from $p$ without re-fitting $p’$ from scratch. 
However the move from EBM to autoregressive policy of \S\ref{sec:ebm-2-policy} remains to be discussed. The question now is the following. We have already obtained a policy $\pi_{\theta}$ approximating $p$, and we are interested in obtaining a policy $\pi_{\theta’}$ approximating $p’$: is it advantageous to start Algorithm 1 with $q=\pi_{\theta}$, rather than starting “from scratch” and taking $q=a$\ ? Intuition says “yes, very probably”, because $\pi_{\theta}$ is by construction an approximation to $p$, which is  closer than $a$ to $p’$ (formally, $\KL(p’,p) \leq \KL(p’,a)$, see Fig.~\ref{fig:appendix-Transitivity}, where $p'=r$). Due to the approximation, we only have $\KL(p',\pit) \simeq \KL(p',p)$ , so a formal proof that $\pit$ is superior to $a$ as a starting point is impossible, but we expect that further experiments would confirm the improvement.

% \begin{figure}[H]
% \centering
% \includegraphics[width=0.35\columnwidth,trim=0 30 50 0]{figures/drawings/adding_new_constraints_2.pdf}
% %
% % \caption{Transitivy and Autoregressive Policy.} \label{fig:appendix-autoregressive_to_p_prime}

% \end{figure}

%%%% EXTRA INFORATOIN ON ADAPTIVITY %%%%%
\section{More on Adaptivity}
\subsection{Details on KL-Adaptivity}
\label{sec:appendix:kld}
\input{sections/adaptivity_extra_appendix}
\newpage
%%%%%%%%%%% ABLATION %%%%%%%%%%%%%%%
\subsection{Ablation on Adaptivity}
\label{sec:appendix:exp-ablation}
\input{sections/ablation}
\newpage
%% Supervised learning exp
\section{Can Standard Supervision Fully Satisfy The Constraints?}
\label{Appendix:supervised-experiment}
\input{sections/supervised_experiments}
\newpage
\section{More Comparisons}

\subsection{Illustration comparing \GDC, REINFORCE, and ZIEGLER}
\label{sec:illustration-ziegler}

The figure below illustrates the difference between \GDC, the RL-based REINFORCE and ZIEGLER baselines for a pointwise constraint. The main points to note are: (1) REINFORCE is trying to find a  distribution $p_R$  maximizing $r(x)$ (meaning that $p_R$ lies on the $\mathcal{C}$ manifold), but this $p_R$ is free to land anywhere on this manifold, and (2) ZIEGLER is trying to find a distribution $p_Z$ that interpolates (with a weight $\beta$) between a high average $r(x)$ and  the KL divergence from $a$; unless $\beta = 0$, in which case we are back to REINFORCE, $p_Z$ does not satisfy the constraint and falls outside of the manifold.

\begin{figure}[H]
\label{fig:appendix-ziegler-geometry}
\centering
% \fbox{\includegraphics[trim=470 220 220 185, clip, width=8cm]{figures/drawings/p_and_reinforce_and-ziegler.pdf}}
\includegraphics[trim=0 35 0 125]{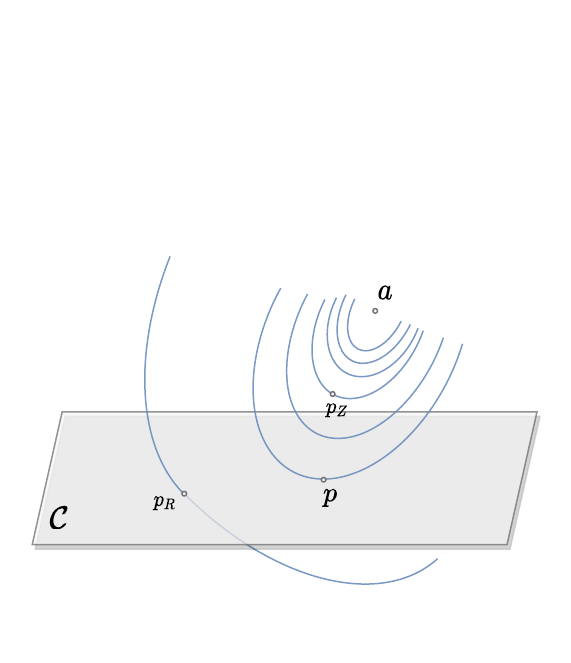}
\caption{Case of a pointwise binary requirement $r(x)=1$: comparison with Reinforce and Ziegler. The curves correspond to different $\KL(\cdot,a)$ levels. The manifold $\mathcal{C}$ is the set of distributions $c$ s.t. $c(x) > 0 \rightarrow r(x)=1$, or, equivalently s.t. $\E_{x\sim c} r(x) = 1$. The curved lines represent increasing levels of the KL divergence $\KL(q,a)$.
According to Reinforce, any distribution $p_R$ s.t. $\E_{x\sim p_R} r(x) =1$, that is, any distribution on $\mathcal{C}$, is optimal. According to Ziegler, to each temperature $\beta > 0$ is associated an optimal distribution $p_Z = \argmin_q \beta \KL(q,a) - \E_{x\sim q} r(x)$, which does not directly lie on $\mathcal{C}$ --- this is because, as indicated in \citep{Ziegler19}, this distribution is of the form $p_Z(x)\propto a(x) e^{r(x)/\beta}$, giving positive probability to all $x$'s in the support of $a$, including to points not lying on $\mathcal{C}$.
Our own optimal $p$ does lie on $\mathcal{C}$ by definition, while minimizing the KL divergence from $a$.
}
\label{fig:Geometry-Ziegler-Reinforce}
\end{figure}

\subsection{Comparison against further baselines}
\label{sec:appendix:extra-baselines}
\label{sec:otherbaselines}
\input{sections/otherbaselines}

\section{Related Work Extended}
\label{sec:appendix:relatedwork}
\input{sections/relatedwork-appendix}

\newpage
\section{Hyperparameters and Training Details}
\label{sec:appendix:hyperparams}
\input{sections/hyperparameters}

\newpage
\section{Distributional and Hybrid Control Experiments For Debiasing Language Models}
\label{sec:dist-exp-details}
\input{sections/exp-extra-distributional}

\section{Extra Details on Pointwise Experiments}
\label{sec:appendix:experiments}
\input{sections/exp-extra-pointwise-klp}

\input{sections/experiments_appendix}
\newpage
\subsection{Token frequency analysis}
\label{sec:appendix:zipf}
\input{sections/tokenfreq-appendix}
\newpage
\subsection{Generation Examples}
\label{sec:appendix:generations}
\input{figures/generations/generations}

%% file: sections/adaptivity_extra_appendix.tex
In this section we provide details on the comparison step in our KL-Adaptive version of the DPG Algorithm, introduced in section~\ref{section:formalization}. We want to assess whether the current $\pit$ is closer than $q$ to $p$, and if the test is positive, we set $\pit$ as the new proposal, hoping to make the proposal more effective for importance sampling.

There are several ways to compute similarity between distributions, two of the most popular ones being on the one hand  KL-divergence and on the other hand Total Variation Distance (TVD) --- where $\TVD(p || p')\doteq 1/2 \sum_x |p(x)-p'(x)|$ --- which is often used in probability and MCMC theory.\footnote{Both metrics are equal to 0 only if the distributions are equal everywhere (in the case of discrete distributions, which are our focus here, otherwise almost everywhere). To our knowledge, there is no obvious best metrics to use when assessing a proposal in importance sampling, leading us to conduct an ablation experiments with both metrics (Appendix~\ref{al:KL-adaptive-DPG})} 
Calculation of these metrics relative to $p$ is not straightforward since the distribution $p \propto P$ is only implicitly represented by the unnormalized EBM $P$, and we cannot easily obtain direct samples from $p$. In this section we describe a workaround.

Given $P$ and a proposal distribution $q$ that we can sample from, using importance sampling 
\citep{owen_chapter_importance_sampling_2013},\MDFN{Find better reference.}
one can calculate the partition function $Z$ as follows: 
\begin{align*} 
%\footnotesize{
            Z &= \sum_x P(x) = \sum_x q(x)\ P(x)/q(x)\\
                    &= \mathbb{E}_{x\sim q(x)}\ P(x)/q(x)\\ \numberthis \label{eqn:z}
%}
\end{align*}
We can then compute $\KL(p || \pi)$ as:  
\begin{align*}
\KL(p || \pi)       &= \sum_x p(x) \log \frac{p(x)}{\pi(x)} = \sum_x p(x) \log \frac{P(x)}{Z \pi(x)} \\
                                &= -\log Z + \sum_x p(x) \log \frac{P(x)}{\pi(x)} =  -\log Z + \sum_x q(x) \frac{p(x)}{q(x)} \log \frac{P(x)}{\pi(x)} \\
 %                               &=  -\log Z + 1/Z \sum_x q(x) \frac{P(x)}{q(x)} \log \frac{P(x)}{\pi(x)} \\
                                &=  -\log Z + 1/Z\ \mathbb{E}_{x\sim q(x)} \frac{P(x)}{q(x)} \log \frac{P(x)}{\pi(x)}
                                \numberthis \label{eqn:kld}\\
\end{align*}
Similarly, for $\TVD(p || \pi)$:
\begin{align*}
            \TVD(p || \pi) &= 1/2 \sum_x |p(x)-\pi(x)| \\
                            &= 1/2 \sum_x q(x)\ \left|\frac{\pi(x)}{q(x)} - \frac{p(x)}{q(x)}\right|
                            = 1/2 \sum_x q(x)\ \left|\frac{\pi(x)}{q(x)} - \frac{P(x)}{Z\ q(x)}\right|\\
                            &= 1/2\ \mathbb{E}_{x\sim q(x)}\ \left|\frac{\pi(x)}{q(x)} - \frac{P(x)}{Z\ q(x)}\right|
                            \numberthis \label{eqn:tvd}\\
\end{align*}
In \S\ref{sec:appendix:exp-ablation} we run an ablation study to compare the use of $\KL$ on line 6 of Algorithm~\ref{al:KL-adaptive-DPG}) or its replacement by TVD.

For both metrics, we need an estimate of $Z$. The precision of this estimate depends on the sample size and the quality of the proposal distribution $q$.  We calculate a moving average estimate $Z_\text{MA}$ of $Z$ is used inside the estimations of $\KL(p\|\pi_\theta)$ and $\KL(p\|q)$ (Algorithm~\ref{appendix:al:DPG}, lines 7 and 8). 
$Z_\text{MA}$ is updated at each iteration of the training, and the moving average estimate is valid due to the fact that $\hat{Z}_i$, based on $K$ samples, is an unbiased estimate of $Z$, and therefore so is $Z_\text{MA}$. % but now taking advantage of \emph{all} the samples obtained in the course of the training.
%
% \MDFN{\textbf{IMPORTANT, Tuesday 22: When looking back at the detailed computation on lines 10 and 11 of the pseudocode below, I noticed something important. Apparently, from this code, the comparison on line 11 does \emph{not} require an estimate of $Z$ !! I think that with TVD, the comparison does require such an estimate still (should be checked). If this is true, this is a rather nice property, but this requires an adaptation of our narration. }}
%
In this way, the estimate benefits from \emph{all} the samples being produced during the course of the training; and 
also because the proposal distribution $q$ evolves and gets closer to the target distribution $p$, the quality of the  estimates of both $\KL(p||\pit)$ and $Z_\text{MA}$ through importance sampling increases (equation~\ref{eqn:z}).
A similar approach is taken in the case of TVD (not shown).

\begin{algorithm}[H]
\setstretch{1.55}
\caption{\ KL-Adaptive DPG (detailed) \label{appendix:al:DPG}}
\begin{small}
\begin{algorithmic}[1]
\Require $P$, initial policy $q$
\State $\pi_\theta \gets q$
\State $Z_\text{MA} \gets 0$                                \algorithmiccomment{Initialize Moving Average estimate of Z}
\For{each iteration $i$}
\For{each step  $k\in[1,K]$}
    \State sample $x_k$ from $q(\cdot)$
    \State $\theta \gets \theta + \alpha^{(\theta)} \frac{P(x_k)}{q(x_k)}\ \nabla_\theta \log \pi_\theta(x_k)$ 
\EndFor
\State $\hat{Z}_i \leftarrow K^{-1} \sum_k P(x_k)/q(x_k)$
\algorithmiccomment{Estimate on the $K$ samples}
\State $Z_\text{MA} \gets \frac{i * Z_\text{MA}+\hat{Z}_i}{i + 1}$        \algorithmiccomment{Update moving average estimate of $Z$}
\State $\KLhat(p||\pi_\theta) \leftarrow  
 -\log Z_\text{MA} + (K\:Z_\text{MA})^{-1} \sum_k  \frac{P(x_k)}{q(x_k)} \log \frac{P(x_k)}{\pit(x_k)}$
  \algorithmiccomment{Estimate on the $K$ samples}
\State $\KLhat(p||q) \leftarrow  
 -\log Z_\text{MA} + (K\:Z_\text{MA})^{-1} \sum_k  \frac{P(x_k)}{q(x_k)} \log \frac{P(x_k)}{q(x_k)}$
  \algorithmiccomment{Estimate on the $K$ samples}
\If{$\KLhat(p||\pi_\theta) <  \KLhat(p||q)$}                     %\algorithmiccomment{More stable calculation of $\KL$ with $Z_\text{MA}$}
    \State $q \gets \pi_\theta$
\EndIf
\EndFor
\Ensure $\pi_\theta$
\end{algorithmic}
\end{small}
\end{algorithm}

%% file: sections/ablation.tex
Here we run an ablation experiment on the adaptivity step of KL-Adaptive DPG  (\S\ref{section:formalization}). We compare three variants of our proposed method: \textbf{DPG-KLD}, which uses KL divergence from the target distribution $p$ to measure the quality of the trained policy $\pit$ i.e. if $\KL(p\|\pit) < \KL(p\|q)$ we update the proposal distribution $q\gets \pi_\theta$. 
\textbf{DPG-TVD} is similar but with the total variation distance instead (TVD).  In \textbf{non-Adaptive} the initial proposal $q$ is kept fixed during training. 

%steps
We run 3 point-wise experiments with single word constraints of three rarity levels in the original GPT-2 distribution, namely: ``Vampire" $(1/10^4)$,``Paris" $(1/10^3)$,``US" $(1/10^2)$ .For each we use $3$ different seeds and train for $10k$ gradient updates.

\begin{figure}[h]
\center
\includegraphics[width=0.8\linewidth]{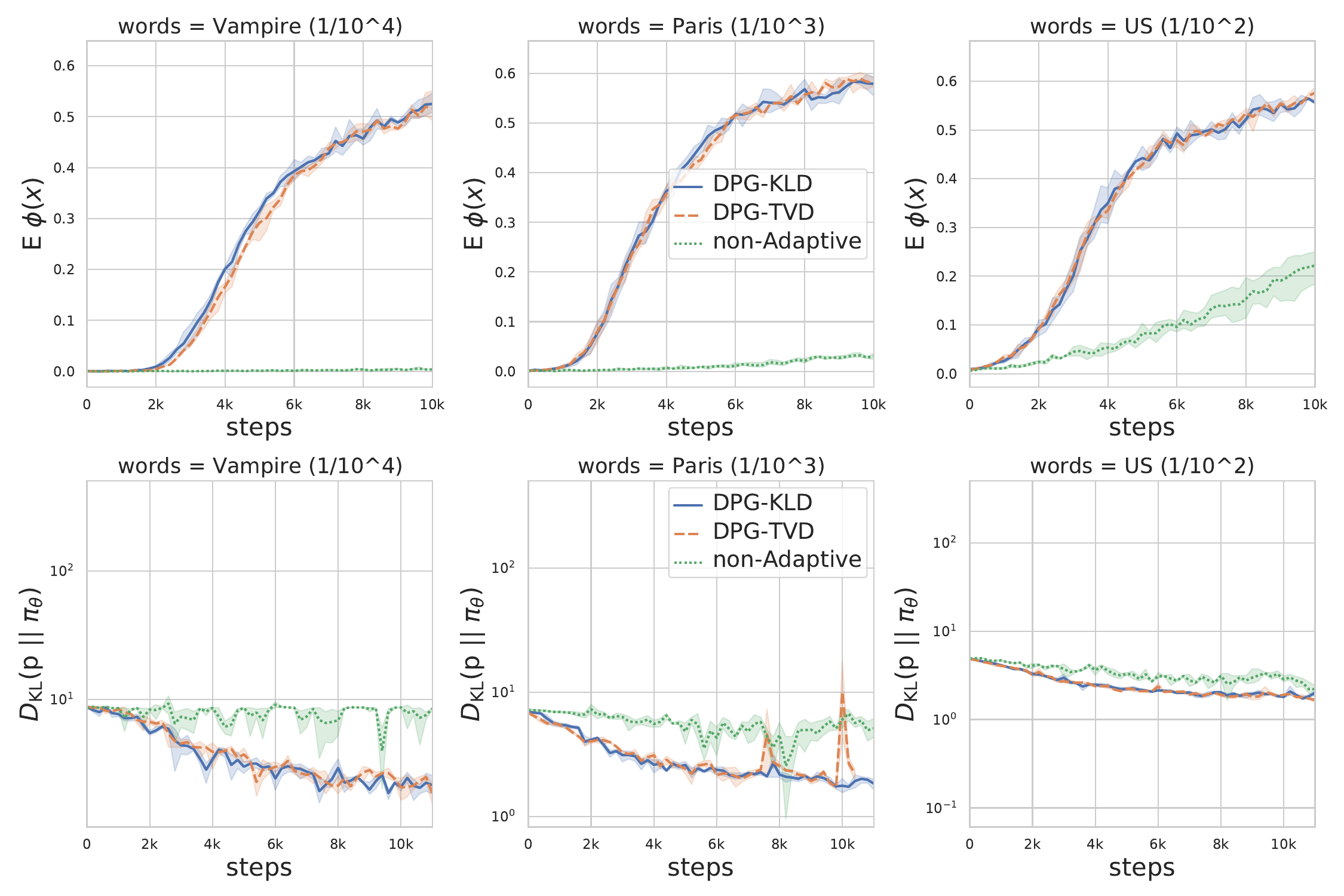} \\	
\caption{Ablation experiment elaborating the effectiveness of the adaptive step in the DPG algorithm explained in section~\ref{section:formalization}. We compare three adaptivity variants, based on the KL divergence (DPG-KLD), on the TVD distance (DPG-TVD) and with no adaptation. We find similar convergence rates for both KLD and TVD adaptive DPG compared to a much slower convergence without adaptation.
\label{fig:ablation_appendix}
}
\end{figure}

Figure~\ref{fig:ablation_appendix} shows training trends of the three ablations. We find a significant difference in convergence speed in favour of the adaptive methods. The efficiency gap between Adaptive and non-Adaptive methods becomes larger the more rare the constraints are. i.e. the proposal distribution $q$ starting point is very far from the target distribution $p$, as the efficiency of the DPG algorithm is related to how close the proposal $q$ is to the target $p$. When $q$ is continuously adapted, the proposal distribution becomes closer to $p$ and the training becomes efficient regardless of how far the initial proposal distribution is from $p$. We observe similar convergence rates for DPG-KLD and DPG-TVD. 

%% file: sections/supervised_experiments.tex
%In this section, we investigate the problem of why our approach is not able to produce text that satisfies the imposed constraint 100\% of the time in the case of point-wise constraints. 
In this section, we try to better understand potential difficulties of autoregressive models to fully satisfy constraints such as the ones illustrated in our pointwise experiments.

To this end, we consider whether a standard fully supervised fine-tuning of GPT-2 can achieve that objective while keeping a minimal distance from the initial model. To answer the question, we carry out an experiment where we fine-tune GPT-2 on a collection of samples satisfying the desired constraint. Our goal here is to investigate whether GPT-2 can fully satisfy the constraint without overfitting the fine-tuning data, since overfitting (memorizing) the training data basically means high KL-divergence from the initial model.

For this experiment, we choose a single-word constraint with the word ``amazing''. We start by sampling 1M sequences from GPT-2 small --- a process that took us roughly 48 hours --- and keeping only the ones containing ``amazing'' (this filtration process can be seen as a variant of rejection sampling~\citep{RJ_casella2004generalized}). We end up with a total of 4600 samples out of which we use 500 for validation and the rest for fine-tuning.  

\begin{figure}[H]
\hspace{0.7cm}
\includegraphics[width=6cm,height=4cm]{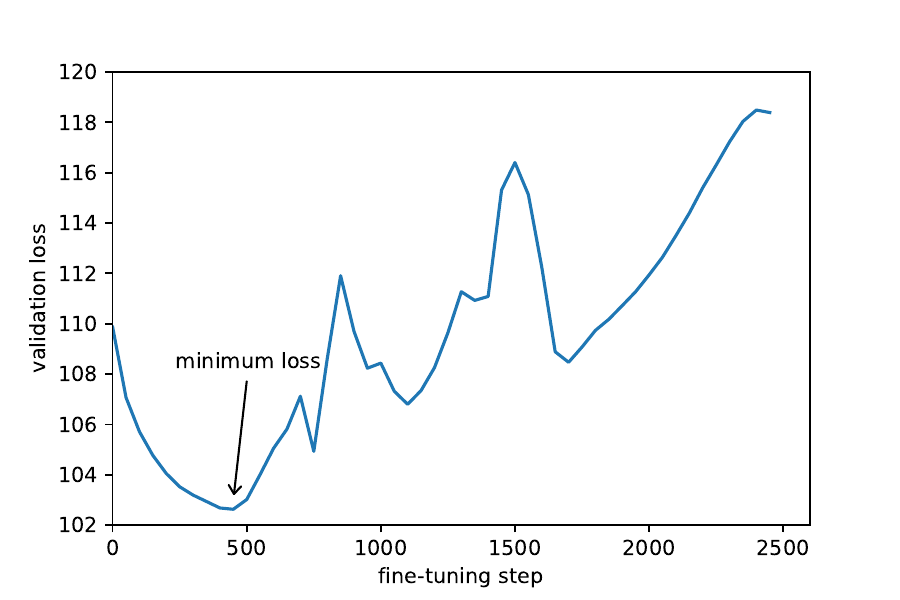}
\includegraphics[width=6cm,height=4cm]{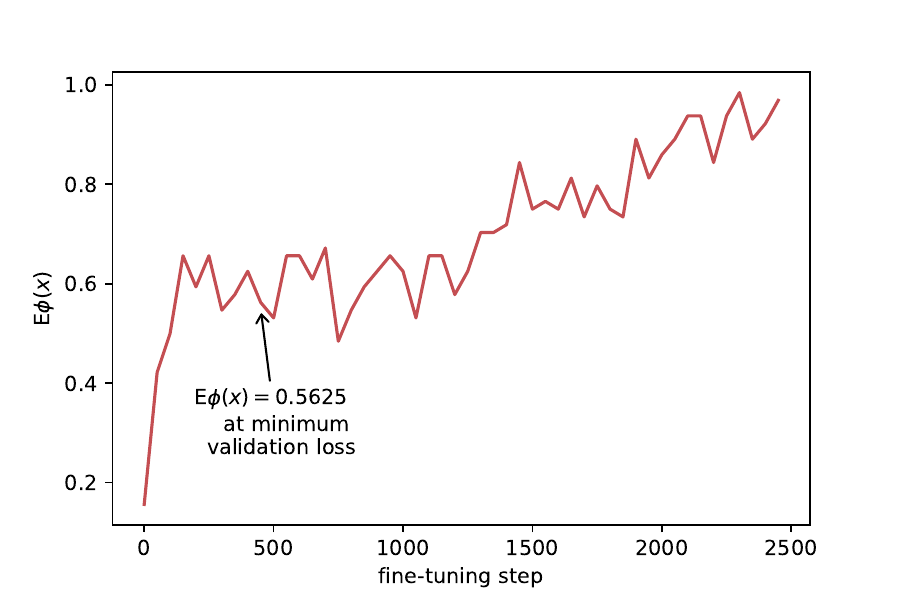}
\caption{\label{fig:supervised-exp} 
Supervised experiment when fine-tuning GPT-2 on a corpus of sentences containing the word "amazing". \textbf{Left:} validation loss development during fine-tuning. \textbf{Right:} percentage of samples generated using the fine-tuned model and containing the word ``amazing". Here, the best model according to the validation loss is only able to achieve $\E \phi(x) = 0.5625$. Higher values of $\E\phi(x)$ tend to occur with higher validation loss, i.e when overfitting.
}
\end{figure}

Figure~\ref{fig:supervised-exp} shows evolution of both validation loss and constraint satisfaction $\E \phi(x)$ on samples generated from the model during fine-tuning. Interestingly, the lowest validation loss corresponds to only $\E\phi(x) \approx 0.56$. Higher values of $\E\phi(x)$ correspond to higher validation loss i.e. to overfitting. 

This result suggests a relationship between training a policy reaching 100\% and overfitting the training data. This hints at the difficulty of strictly imposing certain types of constraints on pre-trained language models without moving far away from the initial model.%
\footnote{Note how very difficult the job would be in the extreme case of a constraint was based on a hash-based predicate filtering on average one sentence out of two.}

%% file: sections/otherbaselines.tex
Here we compare \GDC to other baselines, namely Plug and Play (PPLM)~\citep{plug_and_play_20} and CTRL~\citep{ctrl} for sentiment control. 
PPLM works by updating the hidden states of GPT-2 for a given prefix in order to derive the generation towards the desired attributes. Unlike \GDC, PPLM needs a prefix to perform its hidden-state updates. Thus, our approach is more general in the sense that any prefix can be used on the trained model at test time, rather than requiring prefix-specifc fine-tuning.
CTRL is a large-scale language model ($1.63$ billion parameters and \texttildelow 14x larger than GPT-2 small) based on control codes for steering text style and content. 
For the purpose of generating positive/negative sentiments using CTRL, we use its positive/negative reviews control codes as done in \citep{plug_and_play_20}. The control codes used are ``\texttt{Reviews Rating: 5.0}'' and ``\texttt{Reviews Rating: 1.0}'' for positive and negative sentiment control, respectively. We use five different prefixes \textit{(or prompts)} and generate 100 continuations given each prefix obtaining a total of 500 samples. It is worth noting that \GDC is trained in the same way as described in the main text, i.e. without any knowledge of prefixes, and that we only use prefixes at test time with the saved checkpoint. The five prefixes used come from \citep{plug_and_play_20}: ``The chicken\ '', ``The potato\ '', ``The lake\ '', ``The pizza\ '', and ``The horse\ ''.

We use the same sampling parameters across all approaches by setting the temperature $T=1.0$, using top-k sampling with $k=10$, and removing the repetition penalty used in CTRL \citep{ctrl}. However, we notice that CTRL does not work well with higher $T$ values (apparent in the samples in Table ~\ref{tab:pp-comparison}), therefore we report also CTRL evaluation with lower temperature $T=0.5$ and a repetition penalty $\lambda_{rep}=1.2$ as reported in their paper.

As metrics, we use sentiment class expectation $\E \phi(x)$, the perplexity according to an external GPT-2 small architecture as in \citep{DBLP:conf/naacl/LiJHL18}, and the diversity metrics introduced in section \S\ref{sec:metrics}. We average all these metrics across the 500 continuations generated. Table ~\ref{tab:pp-comparison} shows the results for positive and negative sentiment control experiments. As shown, \GDC is able to achieve better positive/negative sentiment with lower perplexity than both PPLM and CTRL. As for diversity, \GDC achieves comparable diversity to the other two approaches and even outperforms PPLM on the Dist-n metrics in the positive sentiment task.

Table ~\ref{tab:samples-dcg-pp} shows sample continuations from all three approaches. Clearly, PPLM and CTRL exhibit some form of degeneration and repetition in many of the continuations (highlighted in light red), which is reflected in their very high perplexity score compared to \GDC, which produces much more natural text with minimum repetitions without requiring a repetition penalty as CTRL. 

It is also worth noting here that CTRL (and other control code methods) is very much limited in terms of its applications. For instance, to generate positive/negative sentiment text as we do in this experiment, we are required to use the \texttt{``Reviews Rating...''} control code, using control codes outside of those CTRL was fine-tuned on leads to very bad generations. This, in turn, restricts the generated text to positive/negative reviews although we may desire different types of positive/negative text (e.g. news reports). We can observe this effect\footnote{With lower temperatures, this behaviour becomes even worse and CTRL mostly generates reviews.} in some of the samples in Table ~\ref{tab:samples-dcg-pp} such as ``\texttt{The chicken we just ordered from Amazon.com...}'' and ``\texttt{The pizza works no matter what settings you use it on}.

\HEFN{Description of what we do relative to what PP does is not very clear.   In particular,  we need a more self-contained description of PP ”specialized” for each prefix, right? And with”optimization  intent”  ?   Can  we  say  something  convincing  about  why  the  experimental conditions are in their favor (but is it actually true?) ? \MK{I don't see why the experimental conditions are in their favor other than that we use their five prefixes. Also, if it was in their favor, they would have performed at least somehow better, but that's not the case.}}

\begin{table}[H]
\footnotesize
\centering
\def\arraystretch{1.3}%  1 is the default, change whatever you need
\begin{tabular}{l l l l l l l l l}
    \toprule
 \textbf{Method} & $\E \phi(x)$\textuparrow & \textbf{Perplexity} \textdownarrow & \textbf{Dist-1} \textuparrow & \textbf{Dist-2} \textuparrow & \textbf{Dist-3} \textuparrow & \textbf{SB-3} \textdownarrow & \textbf{SB-4} \textdownarrow & \textbf{SB-5} \textdownarrow\\
 \midrule
 \multicolumn{9}{c}{\textbf{{Positive Sentiment}}}\\ 
 %DPG    &   0.315 &    0.782253 &    0.916071 &    0.921460 &     0.967802 &     0.932355 &     0.884323 &  109.475891 \\
%P\&P    &   0.325 &    0.717437 &    0.874931 &    0.897135 &     0.928015 &     0.881425 &     0.827868 &  145.992210 \\
PPLM  &  0.52 & 29.26$\pm$22.07 &        0.72 &        0.89 &        0.91 &         \textbf{0.98} &         0.96 &         0.92  \\

CTRL &  0.28  & 76.52$\pm$90.51 &        \textbf{0.82} &        \textbf{0.95} &        \textbf{0.94} &         \textbf{0.98} &        \textbf{ 0.95} &          \textbf{0.90} \\ 
GDC & \textbf{0.56}  & \textbf{13.53$\pm$3.18 } & 0.76 &         0.91 &        0.92 &         0.99 &         0.97 &         0.95\\
 \hline
 CTRL* &   0.78 & 26.80$\pm$11.89  &  0.90 &        0.97 &        0.95 &         0.99 &         0.98 &         0.97 \\ 
 
 \midrule
 
 \multicolumn{9}{c}{\textbf{{Negative Sentiment}}}\\ 
 PPLM & 0.14  &  27.72$\pm$23.95 &         0.73 &        0.90 &        0.92 &         0.98 &         0.95 &         0.92\\

 CTRL &   0.16 &  82.05$\pm$54.74  &      \textbf{0.82} &        \textbf{0.95} &        \textbf{0.94} &         \textbf{0.97} &         \textbf{0.94} &          \textbf{0.90} \\
GDC & \textbf{0.51} &   \textbf{13.59$\pm$3.84} &        0.73 &        0.87 &        0.88 &         0.98 &         0.97 &         0.94 \\

\hline
 CTRL* &  0.44&  28.50$\pm$12.86  &         0.90 &        0.97 &        0.95 &         0.99 &         0.98 &         0.96 \\

\bottomrule
 \end{tabular}
 
 \caption{\label{tab:pp-comparison} Comparison against PPLM \citep{plug_and_play_20} and CTRL \citep{ctrl} on positive and negative sentiment control. We generate 100 samples for each prefix obtaining a total of 500 samples. All metrics shown are averaged across the 500 samples obtained. CTRL refers to the shared setting across all approaches with temperature $T=1.0$ and repetition penalty $\lambda_{rep}=1.0$ and CTRL* refers to having $T=0.5$ and $\lambda_{rep}=1.2$. Here, we see a clear advantage of \GDC in terms of constraint satisfaction and perplexity and a comparable performance in terms of diversity against PPLM and CTRL.}

 \end{table}

 \begin{table}[H]
        \scriptsize
        \def\arraystretch{1.5}%  1 is the default, change whatever you need
        \begin{tabular}{p{1cm}|p{12.1cm}}
        \toprule
 \multicolumn{2}{c}{\textbf{\GDC}} \\ 
 \hline
\color{blue}
Positive & 
\underline{The chicken} is so tasty! This recipe has been in my freezer for about 2 months now. I have always had good quality chicken breasts, so I had a huge amount of chicken and
\newline
\underline{The potato} is my favourite part of the recipe, and it is also my biggest problem. It is a good addition to anything you try to add to your meal. I love potato soup
\newline
\underline{The lake} has been the most active place in the country for visitors since it opened last summer, and it's not hard to see why. The lake has been a place to stay for years,
\newline
\underline{The pizza} place is great for a good time! They have all kinds of amazing sandwiches. My favorite sandwich is the "Chicken Pizza" which is the best I've ever had on my way to
I don't think the pizza is better than any other pizza. It's not a
\newline
\underline{The horse} is very well balanced with the horse's head and ears. It is a great horse to have for the family. The horse is quite large and the tail is long enough to accommodate the",\\ 
\hline
\color{red} Negative & 
\underline{The chicken} is so bad it's literally going to die. It's like the dog is dying from a bad diet. I'm not even sure I want it in my kitchen"
\newline
\underline{The potato} was really bad, but it's the worst potato that I've ever tasted. It was really hard for me to put my finger on. I was very disappointed with the flavor of the
\newline 
\underline{The lake} was just too cold for the water and the wind to blow. I couldn't get out of it. I couldn't see anything. The wind blew through my windshield and my car windshield"
\newline 
\underline{The pizza} is so bad that I've had to buy it again, even if I was just going to order one at a time. I'm not going to be buying one for the next week
\newline 
\underline{The horse} in the back row is not going to win the race. It is going to go up in flames, and the race will end in a heap of dust. The winner will be thrown
\\

\bottomrule
 \multicolumn{2}{c}{\textbf{PPLM}} \\ 
\hline
 \color{blue} Positive & \underline{The chicken} and the egg story: the extraordinary rise of the powerful man as he transforms the lives of ordinary people in a world in crisis  The story of the  A man dressed in a white suit
\newline
  \underline{The potato}.  It is the world's most awesome people, \r{and and and the that the the a the a the a the a the a the a the  ,  "and the the}
\newline
\underline{The lake} is a great spot to enjoy the spectacular views of the Great Lakes. This is also a great place to take the children's swim. The lake is also a great place to hike in the beautiful mountains
\newline
\underline{The pizza} is a delight! I have never had this before. I am a fan of Italian, and I have not had it before in the States. I will be back! It was a great experience
\newline
\underline{The horse} is a powerful, beautiful, and extremely intelligent animal., \r{(.C,:,.,.,../.../..',,' (., ".}
\newline \\ 
\hline 
\color{red} Negative &
\underline{The chicken}pox epidemic of 1918-1920 in Britain was an acute and deadly disease that killed about 100,000 people worldwide, most of them infants.  The 1918-1919 epidemic was caused by the
\newline
\underline{The potato} is one of those things we all dream of. I think the most common thing that people come up with when I say I have the perfect one is the idea of a "salt water" version
\newline
\underline{The lake} is \r{one one one. $<$endoftext$>$}The United Nations (UN) and the European Union (EU) are among a number of the world's most in the state and,, on the House vote for
\newline
\underline{The pizza} crust is anvil, which is what the British have for a long time. The British Empire, the French, \r{the the the the the a in the that} is a a it is called and it
\newline
\underline{The horse} is in the saddle. That's how he's been for the last four years. The Tampa Bay Lightning won a series of three games in a row to begin the new year and into January we were
 \\
\bottomrule
\multicolumn{2}{c}{\textbf{CTRL}} \\ 
\hline
\color{blue} Positive & 
\underline{The lake} I am looking forward to seeing in September! The sea scene alone would have me watching again!  Rating: 5.0  One of the best comedies I've seen. We will definitely watch it again. Smart and funny
\newline
\underline{The horse} for this ones lines is:\r{\&\#34;The road to Hell is paved with good intentions. All roads to Hell end in Hell themselves.\&\#34;}  Rating: 5.0  I live in a small
\newline
\underline{The potato} were "seeded" during a European settlement. What the characters have gone through is inevitable, but extremely rare. (And the potato has the honor of being the world's oldest potato. For that honor, we have a nickname: "@@
\newline
\underline{The chicken} we just ordered from Amazon.com has not yet arrived and I am EXTREMELY EXCITED! The seller has the finest poultry in the market....plus, it is DELICIOUS!Thank you so
\newline
\underline{The pizza} has been around for decades. Now that time has been added to it, all of us can appreciate it better, and enjoy it the way we have always enjoyed.PERFECT Pie:(The second listen) And it
\newline \\ 
\hline
\color{red} Negative &
\underline{The pizza} works no matter what settings you use it on. The icecream maker always leaks out around the spout and onto the base (gross) - finally stopped working. I only wish I had spent more for a
\newline
\underline{The horse} can not be found. Characters whose names show up in the battle screen:\r{EXE: SRMX\&OY; SQX the knight >QWOKB SKOZY} the warrior!A useful upgrade for a
\newline
\underline{The lake} has been made, but it's far from Earth 5. The ship has disappeared but they continue to radio.Ignoring the plot, which the Star Trek series never bothered with, Spock says that "we should have followed up. There is
\newline
\underline{The chicken} died on me after 8 months. I don't think the unit is compatible with young chickens. Not recommended.  Rating: 1.0  the plates didn't last long enough for me.I bought two of these plates and they
\newline
\underline{The potato} does not start from eggplants, it starts from the start of generation! How stupid is that! :( I bought this and many others to try with my toddler for his preschool class. I want him to get \\
\bottomrule
\end{tabular}

\caption{\label{tab:samples-dcg-pp} Samples generated from \GDC, Plug and Play \citep{plug_and_play_20} and CTRL \citep{ctrl} for both positive and negative experiments. Control codes are omitted for CTRL. Prefixes are \underline{underlined}. Repetitions are highlighted in \r{light red}. As shown, PPLM and CTRL produce more repetitions compared to \GDC.}
\end{table}

%% file: sections/relatedwork-appendix.tex
\paragraph{Optimizing global rewards for Text Generation} There is a large reinforcement learning inspired literature about steering an autoregressive sequential model towards optimizing some global reward over the generated text. This includes REINFORCE~\citep{Williams92} for Machine translation (MT) \cite{seq_lvl_train_RanzatoCAZ15}, actor critic for Abstractive Summarization~\citep{PaulusXS18}, Image-to-Text \cite{RL_Img2txt_LiuZYG016}, Dialogue Generation~\cite{RL_dialogue_LiMRJGG16}, and Video Captioning~\citep{PasunuruB17}. With respect to rewards, some approaches for Machine Translation and Summarization \citep{seq_lvl_train_RanzatoCAZ15, BahdanauBXGLPCB17} directly optimize end task rewards such as BLEU and ROUGE at training time to compensate for the mismatch between the perplexity-based training of the initial model and the evaluation metrics used at test time. Some others use heuristic rewards as in \citep{RL_dialogue_LiMRJGG16,RL_TambwekarDMMHR19}, in order to improve certain a priori desirable features of generated stories or dialogues. 
Other non-RL techniques for approximating the global sequence constraints $\phi(x)$ by a biased estimator $\phi(x_t|x_{:t-1})$. These techniques usually referred to as weighted decoding \cite{learning2write-holtzman-2018,SeeRKW19} this however still requires a heavy search procedure and this biased estimation of sequences that satisfy the global constraint compromises fluency and coherence. Continuous approximation using the Gumbel Softmax was developed for the training of Variational Autoencoders but several works have implemented it for natural language generation \cite{ShettyRHFS17,meansum_ChuL19,gumbel_gan_KusnerH16}.

%%%%%%%%%%%%%%%%%%%%%%%%%%%%
\paragraph{Competing Degeneration in Controlled Text Generation}
When using such approaches, one needs to take care of not forgetting too much of the original LM policy (``degeneration''): \cite{LiuLSNCP16} noted that such optimization may produce adversarial examples that improve the average reward without an actual increase in readability or relevance. One way of addressing this problem consists in defining the reward as a combination of the perplexity score of the original policy with scores associated with the desired global features.
\cite{Wu_googleMT16,PaulusXS18} combine NLL loss with reward maximization in a mixed training objective for Machine Translation and Abstractive Summarization. 
\cite{gumbel_textgen_yang_NIPS2018} use a set of Language Models pretrained on the target domain as a control signal for text style transfer. 
As a proxy to perplexity, \cite{learning2write-holtzman-2018} design hand-crafted rewards using a set of discriminators to ensure the quality of generated text in open-ended text generation. 
\cite{LiuLSNCP16}, however, show that defining a combination reward accounting for text fluency is highly non-trivial and the results of directly optimizing it cannot be fully trusted.
%%%%%%%%%%%%%%%%%%%%%%%%%%%
\paragraph{KL Divergence penalty} 
Another approach relied on penalizing too large deviations of the trained policy relative to the original policy. 
\cite{KL_Jaques17,KL_jaquesK19} propose a conservative fine-tuning approach with a KL penalty between the trained policy and the original auto-regressive model. This penalty acts as a regularizer to the optimization process that prevents the trained policy from deviating too much from the original policy. \cite{Ziegler19} follow a similar approach for fine tuning a language model based on human preferences, in this case a proximal policy algorithm~\citep{PPO} is used to maximize the combined reward. 
PPLM~\citep{plug_and_play_20}, this time in a plug-and-play rather than a fine-tuning context, also use KL divergence to penalize deviations from the initial policy. 

\paragraph{Pointwise vs. Distributional View} Most of the existing works on Controlled Generation have taken what we have called a pointwise view: focusing on the quality of each individual output, as opposed to \emph{distributional} properties of the collection of all outputs. And in fact, the standard objective of RL is to \emph{optimize} a pointwise reward. Even when policy-gradient methods do consider distributions over outputs, they only do as a tool towards producing maximal rewards; and in fact, it is a side effect of the limited capacity of the policy networks that such distributions do not peak on a single output, as would be the optimal outcome in cases of real-valued rewards with no ties.\footnote{In which cases the distribution $q$ maximizing $\E_{x\sim q} R(x)$ would be $q=\delta_{x^*}$ for $x^* = \argmax_x R(x)$.}
By contrast to this usual optimization ``intent'', our own intent here is explicitly distributional, and the policies we are looking for are not simply tools towards maximizing scores, but actual objectives in their own right.

Such a change of perspective might be argued against in the case of conditional seq2seq problems, such as Machine Translation, where focusing on a single good output for a given input makes sense, but is clearly in-adapted when focusing on language models where sample diversity is a requirement.
%%%%%%%%%%%%%%%%%%%%%%%%%%%%%%%%%%
\paragraph{Energy Based Models for Text}
Energy-Based Models (EBMs)~\citep{Hinton02,lecun_tutorial_2006,RanzatoBCL07} are learning frameworks that attracted a lot of attention several decades ago.%
\footnote{The early work on "Whole sentence exponential models" by \citep{Rosenfeld01whole-sentenceexponential} ---  which only came to our attention when preparing the final version of this paper --- can be considered as a form of EBM over texts. While it does not utilize neural networks, it does exploit, as we do, the exponential family in order to provide a global form of control over texts.}
There has been a recent surge of interest in these types of models across a variety of fields. Some early NLP-related EBM research is concerned with neural-based sequence labelling problems (e.g. tagging) exploiting the global sequence~\citep{andor_globally_2016,Belanger:2016:SPE:3045390.3045495}. Some current applications to text generation include \cite{A-parshakova-etal-2019-global} and \cite{Deng_EBM_20}, who augment a standard autoregressive LM with an additional global factor in order to get a lower perplexity on the training data. \cite{Tu2020ENGINEEI} propose an energy-based method to perform inference networks from pretrained Non-Autoregressive Machine Translation models.
A recent survey of EBMs for text is provided in \cite{Bakhtin2020EnergyBasedMF}.%

%% file: sections/hyperparameters.tex
%%%%%%%%% hyper-params %%%%%%%%%%%%%%%
We implement \GDC and all baselines using the PyTorch framework \citep{pytorch}. For all experiments we start from a pretrained GPT-2 small (117M parameters) obtained from the Hugging-Face library~\citep{huggingface}
and fine-tune for 3K gradient-update steps.
% \MDFN{We should probably also give some high-level information about the current implementation, in part in order to distinguish from the earlier implementations \citet{opt-rl-arxiv-2019}}
Each training required 2 Nvidia V100 GPUs, the longest model took $\sim72$ hours to train. \\
A list of the hyperparameters used for \GDC and baselines is given in table ~\ref{table:hyperparams}. 
$K$ refers  to the number of gradient steps per iteration in Algorithm ~\ref{al:KL-adaptive-DPG}.

$N$ refers to the number of samples required and $\mu_{tolerance}$ to the minimum tolerated error $||\tbmu - \hat{\bmu}(\blambda)||_2^2$ while optimizing $\blambda$, and $\blambda_{learning}$ is the SGD step size for updating $\blambda$ in Algorithm~\ref{al:computing_lambdas}.

During training of the policy $\pit$, we perform periodic evaluation as follows:
every 10 minibatch gradient updates, we sample $2048$ sequences of $40$ tokens long, using \textit{nucleus sampling} with $top_p = 0.9$~\citep{degeneration_HoltzmanBDFC20} and estimate diversity metrics on these samples. On the other hand, for accurate estimations of $\KL$ based metrics we perform pure sampling on another set of $2048$ sequences of $40$ tokens long.

For word-lists in the pointwise experiments in section ~\ref{sec:exp-pointwise}, we used the 4 word lists from the Plug and Play \citep{plug_and_play_20} repository\footnote{https://github.com/uber-research/PPLM/tree/master/paper\_code/wordlists}. As for the sentiment and clickbait classifiers, we used their pre-trained classifier heads over GPT-2 medium\footnote{https://github.com/uber-research/PPLM/tree/master/paper\_code/discrim\_models}.\\

For distributional and hybrid experiments, we fine-tune GPT-2 small (117M params) to produce biographies on a dataset of 700K Wikipedia biographies \citep{DBLP:conf/emnlp/LebretGA16} which we refer to as GPT-2\textsuperscript{bio}.
To detect if a given text is about a \textit{female} gender, we construct $\phi_{female}(x)$ as a simple rule-based discriminator that depends on the percentage of female personal pronouns (she, her, hers, herself) w.r.t. all mentioned pronouns.
We define four types of professions ``Art", ``Science", ``Business and Politics", and ``Sports". To detect them, we define a wordlist for each type as shown in table~\ref{tab:profession-wordlists}.
\\

\begin{table}[H]
    \footnotesize
    \centering
    \resizebox{0.8\textwidth}{!}{%
    \begin{tabular}{l|l|m{8cm}}
    \toprule
    \textbf{Training Method} & \textbf{Constraint} & \textbf{Hyperparameters}  \\
    \toprule
    $\forall$ & $\forall$ & \texttt{steps=3K, top\_p=0.9, warmup=10, dropout=0.1,
    lr= 0.0000141, optimizer=adam.} \\
    \midrule
    $\forall$ & Single word & \texttt{gen\_length=25} \\ 
     & word-list/classifier & \texttt{gen\_length=40} \\ 
    \midrule
    %REINFORCE/\REINFORCEP &  single word  & \texttt{batch\_size=256} \\
    REINFORCE &  Word-list/classifier  & \texttt{batch\_size=256} \\
     \midrule
    \ZIEGLER   & $\forall$ &  \texttt{batch\_size=256, $\gamma$=1.0, $\lambda$=0.95, clip\_range=0.2, target\_KL=6.0, horizon=10000, initial\_KL\_coefficient=0.2}\\
    \midrule
    \multirow{2}{*}{\GDC} &  All Pointwise & 
     \texttt{batch\_size=2048, $K$=20480} \\
     \cline{2-3} 
     & Distributional & \texttt{$N$=20k, batch\_size=2048, $K$=20480, $\mu_{tolerance}=0.01, \blambda_{learning}=0.5$}    \\
    \bottomrule
    \end{tabular}
    }
    \caption{Hyperparameters used throughout all experiments. $\forall$ denotes common parameters between all training methods or constraints.}
    \label{table:hyperparams}
\end{table}

\begin{table}[H]
\tiny
\centering
\begin{tabular}{p{1.5cm}p{10.2cm}}
\toprule
\textbf{Profession} & \textbf{Word-List} \\ \midrule
\textbf{Art}                 & \texttt{storyteller, author, poet, actor, artist, actress, sculptor, screenwriter, singer, musician, composer, conductor, songwriter, designer                    }    \\
\textbf{Science}             & \texttt{scientist, sociologist, philosopher, inventor, student, astronomer, historian, academic, researcher, chemist                      }                             \\
\textbf{Business/Politics} & \texttt{businessman, businesswoman, entrepreneur, chairman, chairwoman, governor, politician, journalist, ambassador, communist, liberal, officer, lawyer, queen, king} \\
\textbf{Sports}              & \texttt{footballer, trainer , player, swimmer, cyclist, athlete , wrestler, golfer, cricketer}                                                  \\ \bottomrule
\end{tabular}
\caption{\label{tab:profession-wordlists} Words in each profession word list used in the distributional constraints experiments.}
\end{table}

%% file: sections/exp-extra-distributional.tex
Large pretrained Language Models are often trained on uncurated data from the internet,
where several demographics are severely underrepresented. One of those demographics is women, whose biographies make up only $18.58\%$ of English Wikipedia's biographies~\citep{wikibias_Graells-Garrido15}. It is expected that such bias is transferred if not amplified by Language Models.
Previous work has suggested associations of certain demographics with certain professions, sentiments and stereotypes~\citep{babysitter_ShengCNP19,gpt3,stereoset}. This shows thaat Bias in LMs also shows up in different forms than just under-representation, and the task of debiasing LMs could require more a complex control method.
GPT-2\textsuperscript{bio} demonstrates a large initial bias: over a large sample of size 20480 examples using top-p sampling ($p=0.9$), it generates only around 7\% female biographies. and a large imbalance between profession types ``Science" ($1\%$), ``Art" ($10\%$), ``Business\&Politics" ($10\%$) and ``Sports" ($20\%$).

% bias in gpt-2
In this set of experiments, we demonstrate the potential of \GDC as flexible general framework that can control pretrained Language Models to impose pointwise, distributional constraints, or even a mix between them (hybrid constraints). We design a set of $6$ experiments whose descriptions and results are displayed in the figures below. Generation examples are provided in Table~\ref{tab:dist-examples-appendix}.

 \begin{figure}[H]
 \centering
 \footnotesize
 \includegraphics[width=0.33\textwidth]{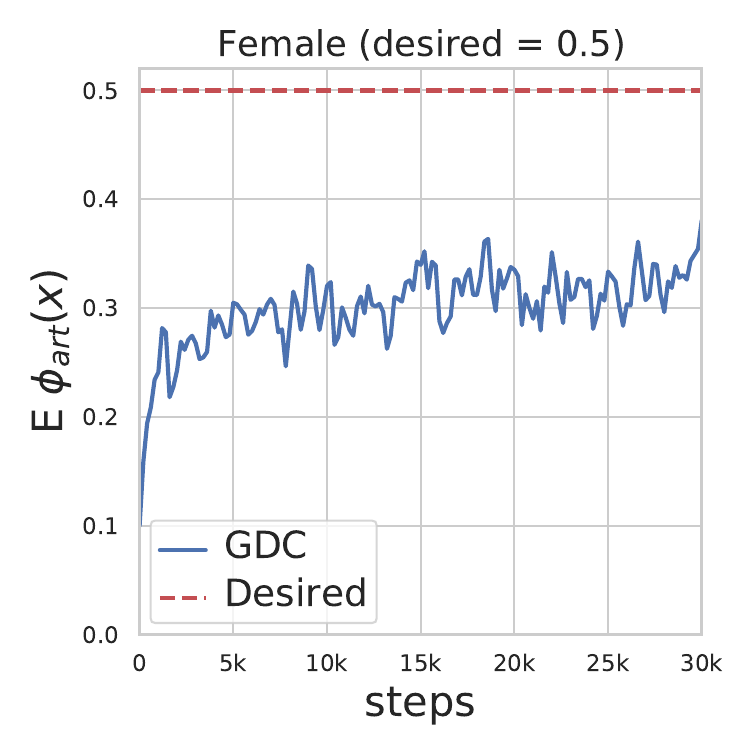}
 \caption{\textit{Exp1: Single Distributional Constraint.} Balancing  demographics can be represented easily through distributional constraints. By using a constraint such as $\E_{x \sim p} \phi_{female}(x) = 0.5$, we can target balancing the female biographies in the distribution of all generations. Note that a point-wise objective $\E_{x \sim p} \phi_{female}(x) = 1.0$ would maximize the presence of female biographies at the expense of other demographics, inducing bias in the opposite direction. The plot shows how $\E_{x \sim p} \phi_{female}(x)$ evolves towards the defined expectation: GDC is able to reduce the bias of GPT-2\textsuperscript{bio} to obtain 36.7\% female biographies rather than just $7\%$.}
 \end{figure}
 \begin{figure}[H]
  \centering
 \includegraphics[width=1\textwidth]{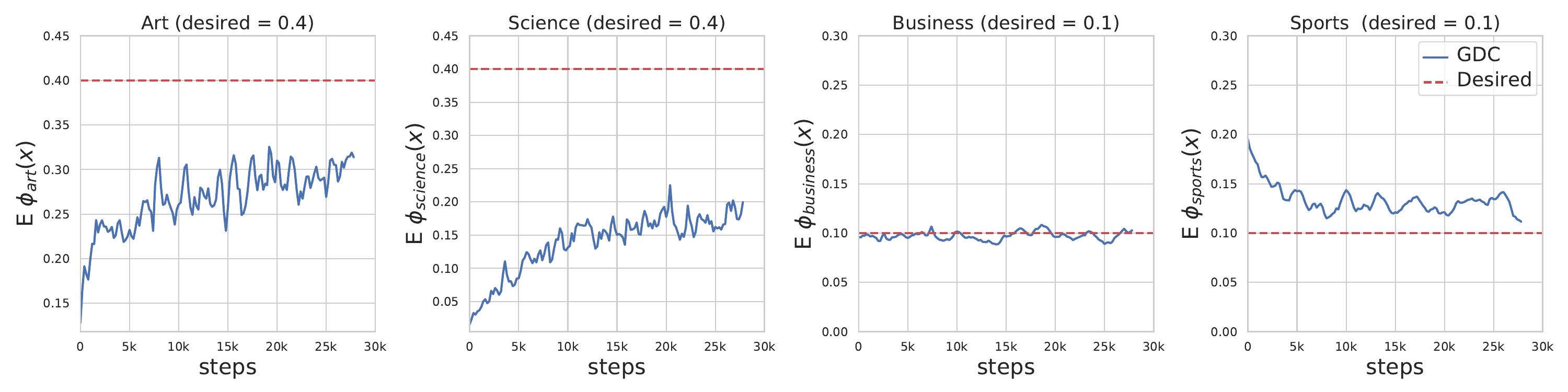}
 \caption{\textit{Exp2: Multiple Distributional Constraints} This experiment demonstrates the flexibility of \GDC in dealing with several distributional constraints at once, even when these constraints have different objectives (increase, decrease, or keep fixed). 
We challenge the flexibility of \GDC by setting four distributional constraints with four arbitrary expectation values targeting $\E \phi_{science}$ and $\E \phi_{art}$ at $40\%$ and
$\E \phi_{sports}$ and $\E \phi_{business}$ at $10\%$. In the figure, from left to right, we can note the increase of $\E \phi_{science}$ and $\E \phi_{art}$ from $1.5\%$ to $20.3\%$ and from $10\%$ to $31.6\%$ respectively. 
Interestingly, the initial $\E \phi_{business}$ of GPT-2\textsuperscript{bio} ($10.9\%$) is already very close to the desired expectation ($10\%$), and we can see that during the course of the training, \GDC keeps this value fixed as it is already satisfying the corresponding target distributional constraint. 
$\E \phi_{sports}$ initially starts higher than the target distributional constraint $10\%$, and we can note that \GDC succeeds to reduce it from $19.6\%$ to $11.9\%$.
}
 \end{figure}

 \begin{figure}[H]
  \centering
 \includegraphics[width=0.7\textwidth]{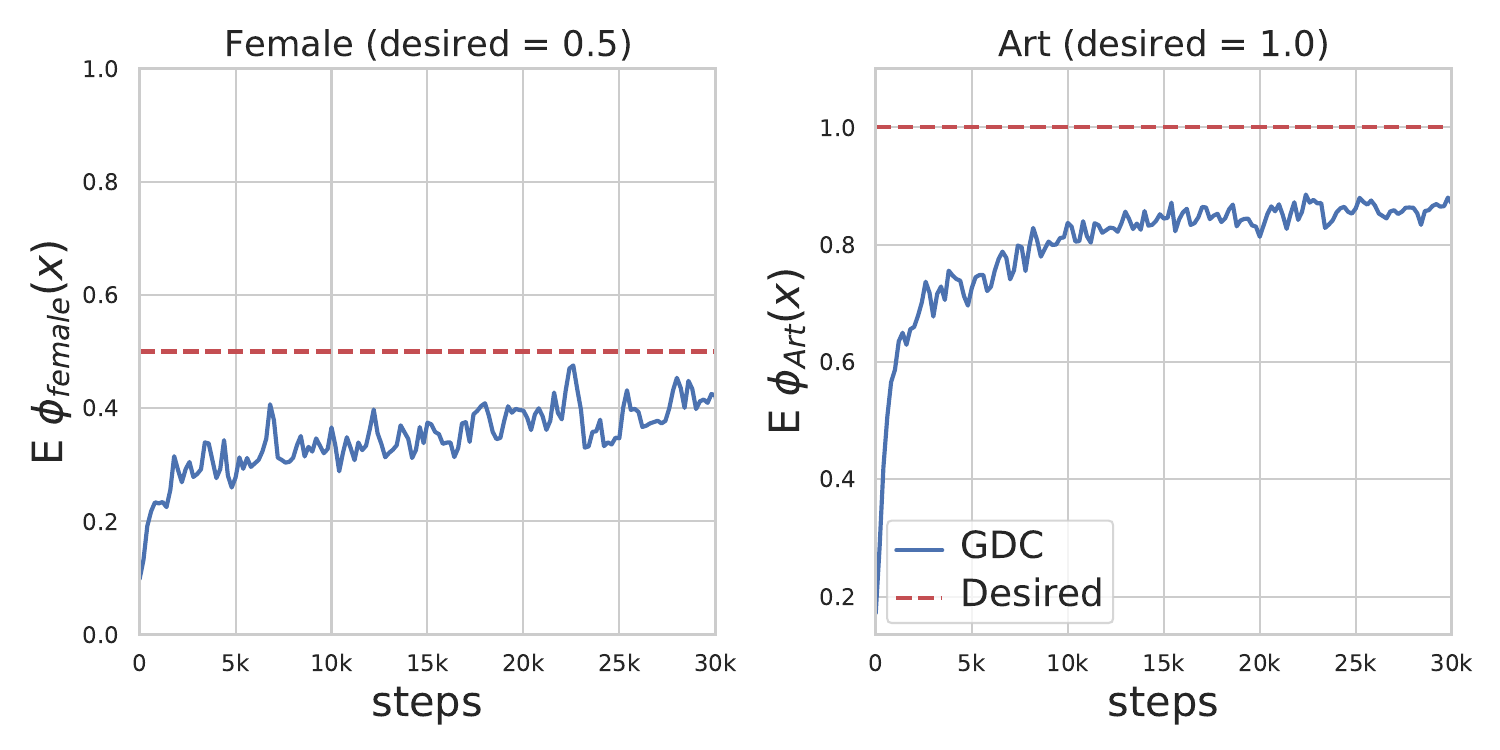}
 \caption{\textit{Exp3: Hybrid constraints} In this experiment, we specify two types of constraints: pointwise with $\E \phi_{art}(x) = 1.0$  and distributional with $\E \phi_{female}(x) = 0.5$ (henceforth Hybrid). \GDC in a single training procedure is able to increase the expectation of biographies about females from $7.4\%$ to $36.6\%$ and Art professions from $11.4\%$ to $88.6\%$. 
 }
\end{figure}
\begin{figure}[H]
\centering
\includegraphics[width=0.7\textwidth]{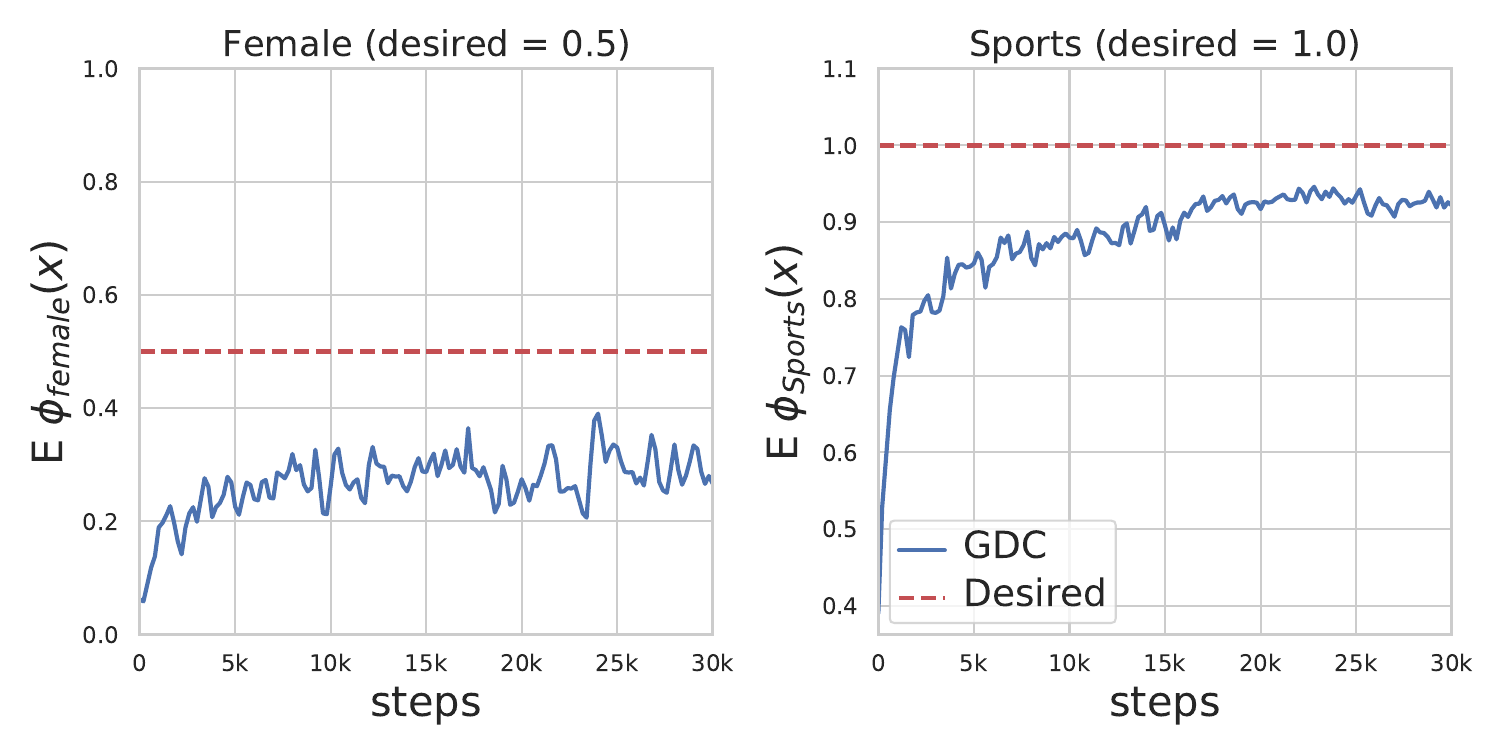}
\caption{\textit{Exp4: Hybrid constraints.} In this experiment, we specify two types of constraints: pointwise with $\E \phi_{sports}(x) = 1.0$  and distributional with $\E \phi_{female}(x) = 0.5$. \GDC in a single training procedure is able to increase the expectation of biographies about females from $7.4\%$ to $31.9\%$ and Sports professions from $17.5\%$ to $92.9\%$. 
}
\end{figure}
\begin{figure}[H]
\centering
\includegraphics[width=0.7\textwidth]{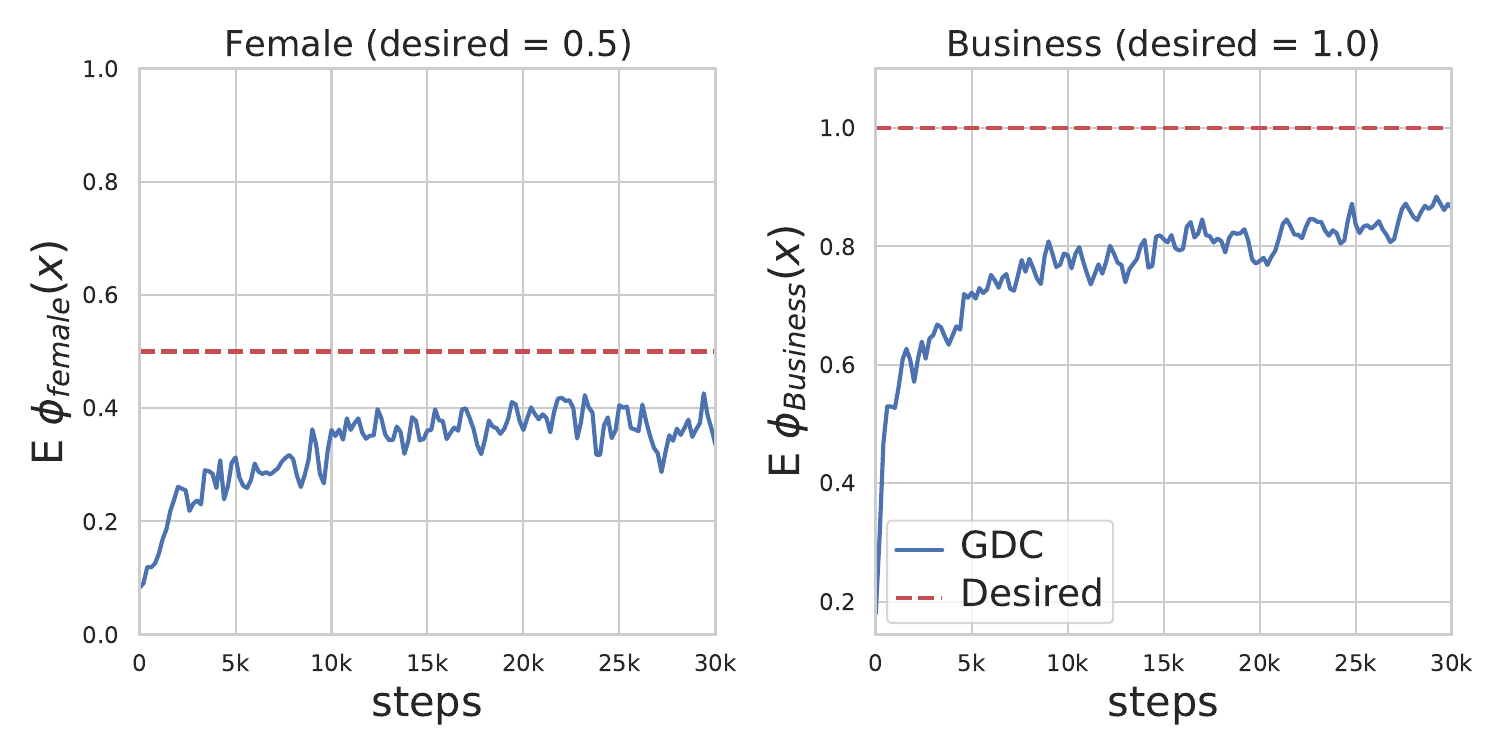}
\caption{\textit{Exp5: Hybrid constraints.} In this experiment, we specify two types of constraints: pointwise with $\E \phi_{business}(x) = 1.0$  and distributional with $\E \phi_{female}(x) = 0.5$. \GDC in a single training procedure is able to increase the expectation of biographies about females from $7.4\%$ to $37.7\%$ and Business professions from $10.1\%$ to $82.4\%$. 
}
\end{figure}
\begin{figure}[H]
\centering
\includegraphics[width=0.7\textwidth]{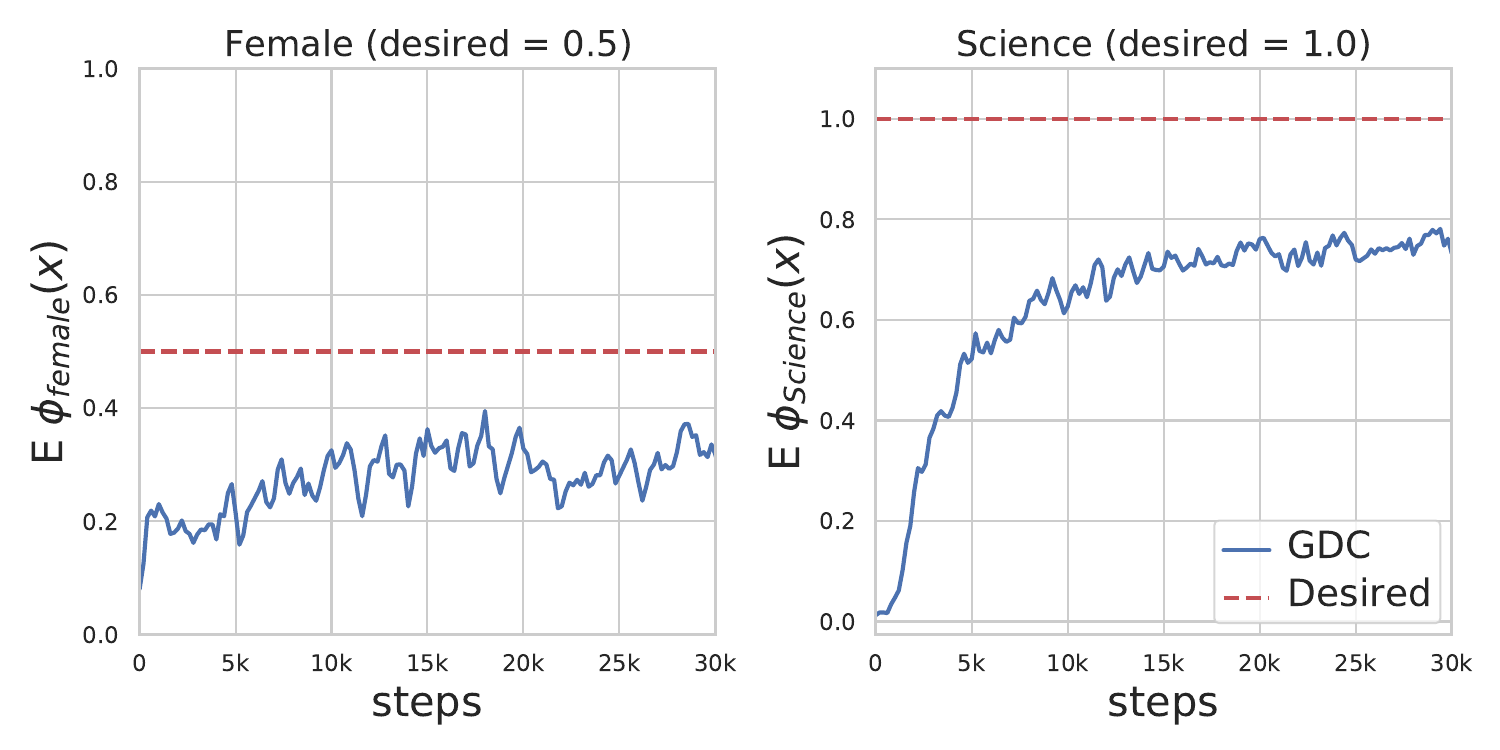}
\caption{\textit{Exp6: Hybrid constraints.} In this experiment, we specify two types of constraints: pointwise with $\E \phi_{science}(x) = 1.0$  and distributional with $\E \phi_{female}(x) = 0.5$. \GDC in a single training procedure is able to increase the expectation of biographies about females from $7.4\%$ to $28.8\%$ and Science professions from $1.2\%$ to $74.7\%$.}
 \end{figure}

\input{figures/generations/generations_dist}

%% file: figures/generations/generations_dist.tex
\begin{table}[H]
	\scriptsize
	\begin{tabular}{p{0.2cm}| p{12.8cm}}
		\toprule
		& \textbf{Art Professions Biographies} \\ 
		\hline
		\textbf{F} & oraci martínez rubin ( born october 24, 1982 ) is a puerto rican actress, dancer and model. she was the first puerto ...\\
		\textbf{F} & therese lebrandt ( born 4 march 1939 ) is an english actress, television host and producer. she is known for her roles as lily lenox...\\
		\_ & , better known by his stage name zac banezi, is an israeli singer and songwriter. the producer of many artists, as well as the keyboardist of heavy metal band the..\\
		\textbf{F} & berry gibson ( born july 21, 1949 ) is an american musician, actor and composer, best known as a member of the rhythm and blues...\\
		%		\textbf{F} & maya mahler ( born may 8, 1943 ) is an american composer and producer. her debut album, `` the twilight hours '', was released in 1968, ...\\
		%		\textbf{F} & morena rivera ( born 8 july 1980 ) is a peruvian actress and singer. she was a member of the alga moyca por quesada ...\\
		%		\textbf{F} & aniya noguchi ( june 19, 1981 -- january 26, 2008 ) was a russian actress. she was nominated for the 2010 svt's..\\
		%		\textbf{F} & arkanda ( sambalitha ) karwatoshi ( born 30 may 1994 ) is a new zealand actress. she was ...\\
		%		\textbf{F} & emily fisk ( born december 21, 1983 ) is an american actress, dancer, and singer. she is the former lead dancer of the group kissy ...\\
		\_ & balkrishnan dev is an indian actor who is known for his roles in telugu movies. he began his career with a short supporting role in `` sapikaya ''. later he played ..\\
		\textbf{F} & starlight '' ciej strall ( born september 1, 1988 ) is an american actress and comedian. she is best known for her role as el ...\\
		\_ & quentin brantley ( born april 27, 1973 ) is a canadian actor, composer, director, writer and producer. he is best known for his work..\\
		\_ & ``Álvaro olajerra '' is an argentine comedian and actor. in 1983, he won an episode of céspedes justicialiste de bolaños..\\
		\textbf{F} & janehamn alister is an american actress, fashion designer, and speaker. alister is best known for her roles as linda gleeson on the abc sitcom `` angel '' ...\\
		\_ & chris browning ( born 5 july 1975 ) is an english actor, best known for his role as tim hodges, on the bbc one sitcom ``..\\
		\_ & andy papadelaspe ( born 9 july 1973 ) is a french actor and director. he is known for his performances in several feature films including `` bern ..\\
		
		\midrule
		& \textbf{Science Professions Biographies} \\ 
		\hline
		\_ & ters g. g. engeland ( 14 april 1914 -- 4 september 2002 ) was an american astronomer and senior researcher in astrophysics. he..\\
		\textbf{F} & thene ted ( born april 4, 1967 ) is an american science educator, student, medical research scientist and medical researcher. she is a director of mls ...\\
		%		\textbf{F} & hanna a. muhammad yap ( 1905 -- 1978 ) was a social historian and writer who lived in kentucky. she is known for her 1978 book `` ...\\
		%		\textbf{F} & catherine grime ( born 23 may 1934 ) is a british historian, and labour party politician. she was educated at oxford university. in 1971, she ...\\
		%		\textbf{F} & won dong-ga ( born july 23, 1956 ) is a south korean sociologist, writer, sociologist, and educator. she is a native of...\\
		\textbf{F} & alexandra martin thomas ( born march 2, 1978 ) is a nigerian scientist and sociologist, researcher and writer. she is the current president of ...\\
		%		\textbf{F} & caesres was an american scientist and sigma xi researcher. she was the first woman to be an associate professor of science at the university of california, berkeley. she was ...\\
		%		\textbf{F} & dame laura ann fowler, ( born july 7, 1962 ) is an american feminist and academic. she served as the united states ambassador to the ph...\\
		\_ & quentin jacobsen ( born 1952 ) is a philosopher. he is a senior fellow at the center for progressive studies, where he teaches philosophy and is responsible for..\\
		\_ & edgar yanowicz ( born 26 july 1940 ) is a philosopher, sociologist and translator who lives and works in new york city. yanowicz is..\\
		\textbf{F} & antosia rose ( born 4 april 1962 ) is an english philosopher. she is a fellow of the royal society and a visiting fellow of the royal academy of engineering...\\
		\textbf{F} & cornelius roberts ( 25 october 1756 -- 17 december 1818 ) was a philosopher of science, well known as a marxist during the ..\\
		\_ & mathias friedrich attelet ( 4 may 1916 -- 11 november 2010 ) was a german philosopher. he was a specialist on number theory, algebraic..\\
		\textbf{F} & helped moore ( february 27, 1918 -- january 25, 1980 ) was a historian and college president who was active in the civil rights movement and has written..\\
		\_ & mathias friedrich attelet ( 4 may 1916 -- 11 november 2010 ) was a german philosopher. he was a specialist on number theory, algebraic..\\
		\_ & themen, jimmy and charles '' ( december 25, 1960 ) is an american philosopher. he is a visiting professor at the university of mich..\\
		\midrule
		& \textbf{Business \& Politics Professions Biographies} \\ 
		\hline
		\_ & said thai khalid (, born 1947 ) is a burmese novelist, journalist and politician. his career began in 1962 and he has become a leader of the ..\\
		\_ & viscount knippenstern ( 14 november 1737 -- 5 august 1792 ) was an austrian-born german jurist and politician. ..\\
		\textbf{F} & alfreda rochelle, ( may 10, 1877 -- november 4, 1965 ) was a canadian lawyer, judge and judge. she served as...\\
		\_  & theodor radulović ( ; 30 october 1873 -- 18 november 1960 ) was a croatian statesman, diplomat, and military officer, ..\\
		\_  & charles lawrence ( april 19, 1807 -- april 30, 1876 ) was an american politician and soldier. he served as a union general during ..\\
		\textbf{F} & i subon ( ; born january 18, 1982 ) is an israeli journalist, writer, columnist and journalist. she is known as the first women writer to...\\
		%		\textbf{F} & erik ali na-ra'; born 18 june 1973 ) is a kenyan politician. she is the current minister of transportation, housing and urban development and deputy prime minister...\\
		%		\textbf{F} & allison quinn ( born october 24, 1975 ) is a canadian politician. she represented the electoral district of nipissing north in the nova sc...\\
		%		\textbf{F} & heather margaret nadelson ( born may 23, 1950 ) is an american politician in the state of michigan. she currently serves as a member of...\\
		%		\textbf{F} & maya maria popova (, born august 3, 1959 ) is a bulgarian politician, member of the russian liberal democratic party. she is currently...\\
		\textbf{F} & hiyat haza (, born 1959 ) is a somali politician. she has been a member of the parliament of somalia from june 2009 to april...\\
		%		\textbf{F} & saidi guzaon (, ), is an armenian politician. she was the minister of foreign affairs of armenia from 1996 to 2000 and 2006 to...\\
		%		\textbf{F} & inez romanka ( born 4 august 1958 ) is a finnish former politician. she has been a member of the national assembly ( mp ...\\
		\_  & erik wiemens ( born 11 october 1957 ) is a german politician. as a youth, he participated in a number of parties, most notably..\\
		\textbf{F} & atalie castillo gonzález ( born 26 april 1957 ) is a mexican politician affiliated to the institutional revolutionary party. as of 2014 she served as deputy...\\
		\_  & ashaun `` tom '' hicks ( born july 28, 1986 ) is an american actress, singer, and beauty pageant contestant. he is also a journalist and ..\\
		\_ & izhev, born `` yuri aleksandrovich isov '' ( ; ), was a writer, journalist and politician. isov first became active in..\\
		
		\midrule
		& \textbf{Sports Professions Biographies} \\ 
		\hline
		%		\textbf{F} & ten (, born 12 august 1992 in budapest ) is a hungarian tennis player. she reached her wta singles ranking of world number 403 on 5 july...\\
		\textbf{F} & isaba aguirre ( born 10 february 1983 in Éixidat, france ) is a female volleyball player from spain. she is a...\\
		\textbf{F} & hanyu pratak ( born 11 june 1993 ) is a female badminton player from bangladesh. she is also an eventer and former world...\\
		\_ & alexandre nicolau ( born 16 february 1989 in travancore ) is an italian professional footballer who plays for serie b club acf..\\
		\_ & yury novoshenko ( ; born march 14, 1987 in tokushima ) is a russian professional football player. in 2011, he played in the..\\
		\textbf{F} & eina jena ( born july 12, 1981 ) is an american soccer player currently playing for ca pei in the chinese super league. she also formerly...\\
		%		\textbf{F} & leila ( born may 26, 1974 ) is a swedish handball player who plays as a central midfielder for the swedish national team. she represented the...\\
		%		\textbf{F} & iaea barrystone ( born 22 march 1967 ) is an english former competitive swimmer. she represented great britain at the 1988 summer olympics. she...\\
		\textbf{F} & chiyo zuai ( born 18 april 1979 in taipei ) is a retired taiwanese tennis player. she is the 1996 olympic...\\
		\textbf{F} & patti ann rakic ( born 23 february 1990 ) is an australian former synchronized swimmer who competed at the 2012 summer olympics. her...\\
		\_ & christopher `` chris '' saul ( born 31 march 1964 ) is a scottish former professional footballer and manager, who managed..\\
		%		\textbf{F} & kathleen baker ( born 9 february 1981 in waterford ) is a former road cyclist from england. she represented her nation at the 2008 uci road...\\
		%		\textbf{F} & gaesanto rodríguez ( born march 10, 1985 in niterói ) is a female water polo player of ecuador. she was part ...\\
		%		\textbf{F} & ahana afonso ( born november 4, 1976 in bamako, bahamas ) is a former female volleyball player from the philippines, who ..\\
		\textbf{F} & katja shearer ( born 7 july 1994 ) is a swedish footballer who plays as a goalkeeper for grödig tyngall. shearer started..\\
		\_&  was an indian chinese footballer who played for hong kong first division league team shandong luneng f.c. during the 1980s. he was regarded as the best right-..\\
		\_& andrey sivchenko ( born 8 july 1983 ) is a russian swimmer. he competed in the men's 200m butterfly event at the 2012 summer..\\
		\_& campbell anderson ( born september 10, 1951 ) is a former professional american football player. he played four seasons with the indianapolis colts of ..\\
		\_& amın güntur ( born 26 may 1987 ) is a turkish professional footballer who plays as a goalkeeper for kozlu kiznevetspor..\\

		\bottomrule 
	\end{tabular} 
    \caption{Randomly selected generations from the hybrid Experiments (3,4,5,6). \textbf{F} indicates that the generation is about a female character. The imposed distributional constraint is $\E \phi_{female}(x) = 0.5$, while the pointwise constraint is $\E \phi_{art}(x) = 1.0$, $\E \phi_{science}(x) = 1.0$, etc.}
    \label{tab:dist-examples-appendix}
\end{table}

%% file: sections/exp-extra-pointwise-klp.tex
\subsection{Approximating the desired   distribution}
 \begin{figure}[H]
 \includegraphics[width=0.33\textwidth]{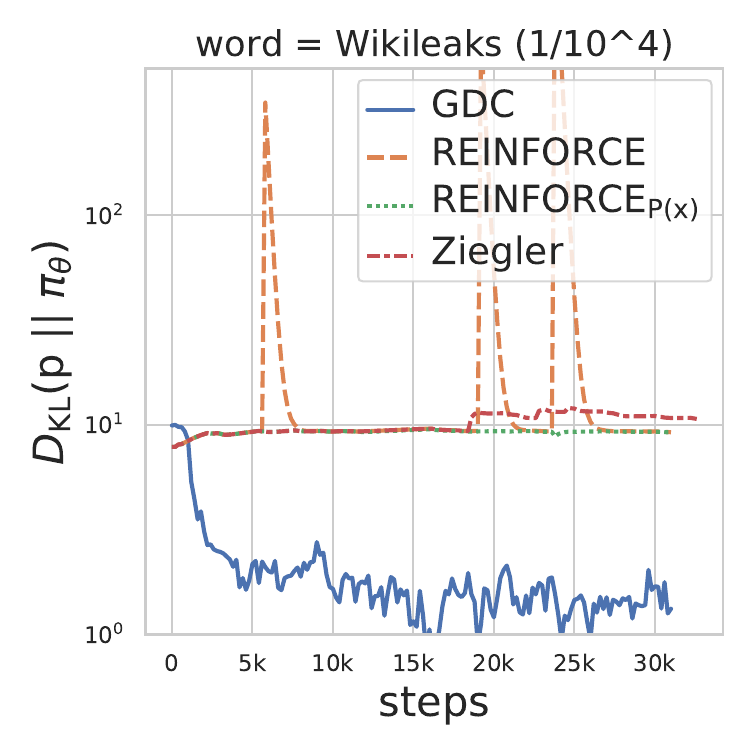}
 \includegraphics[width=0.33\textwidth]{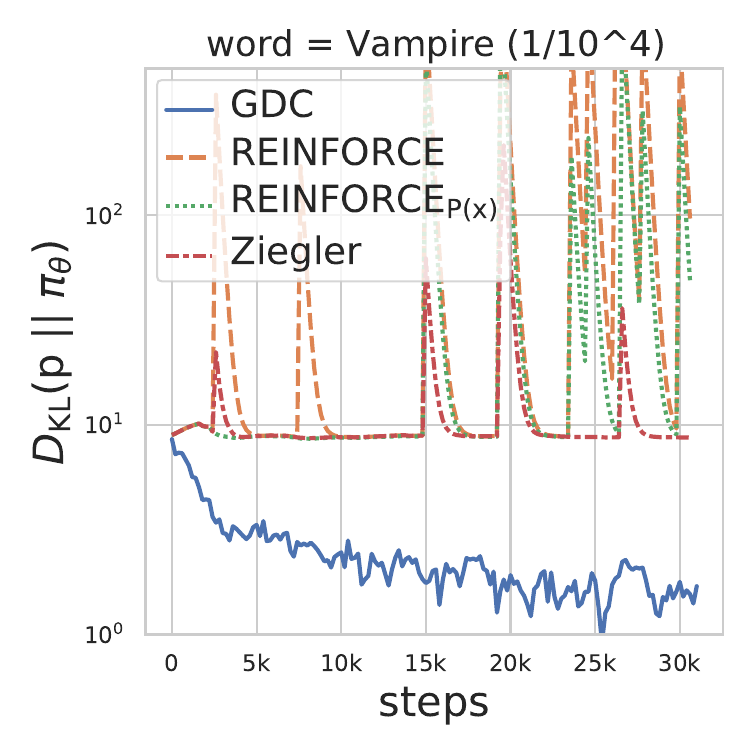}
 \includegraphics[width=0.33\textwidth]{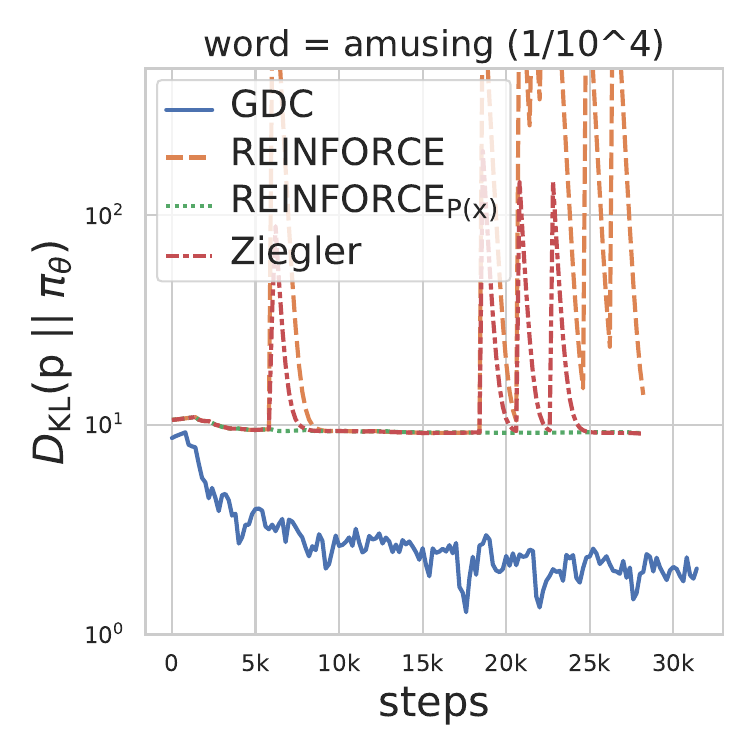}
 \includegraphics[width=0.33\textwidth]{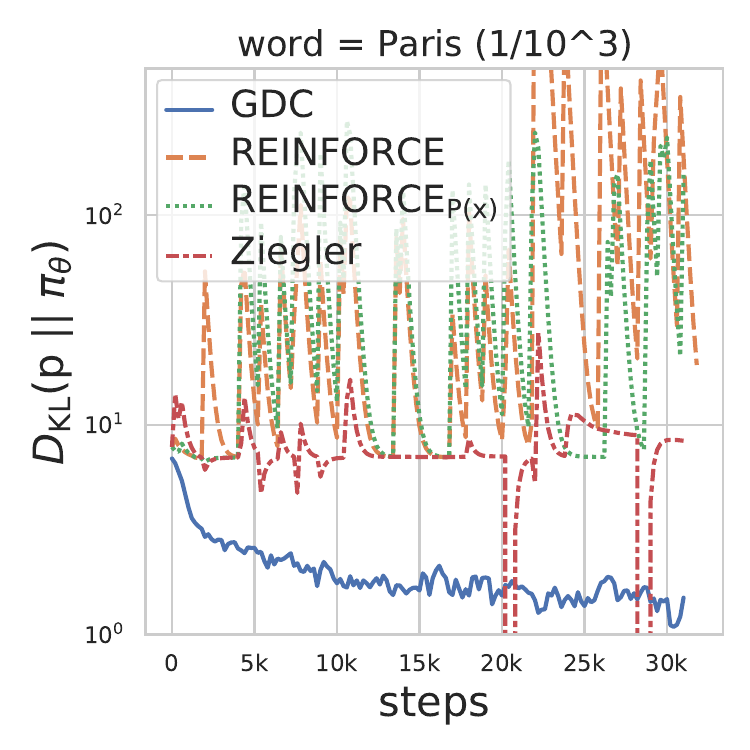}
 \includegraphics[width=0.33\textwidth]{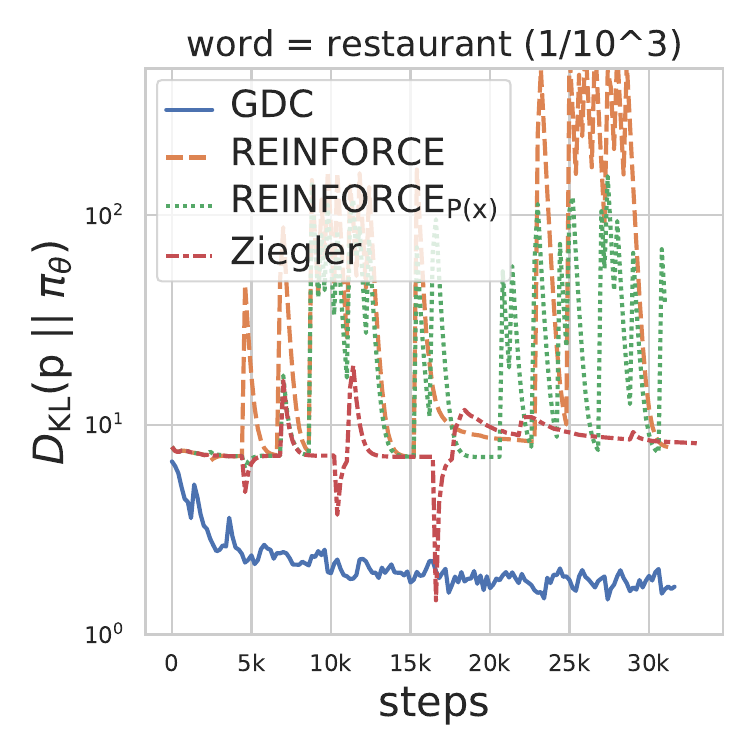}
 \includegraphics[width=0.33\textwidth]{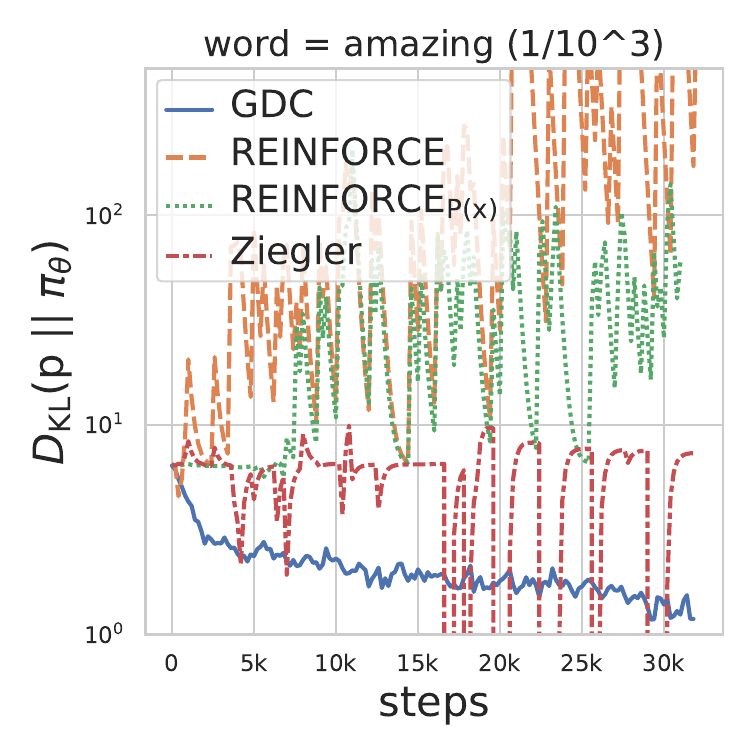}
 \includegraphics[width=0.33\textwidth]{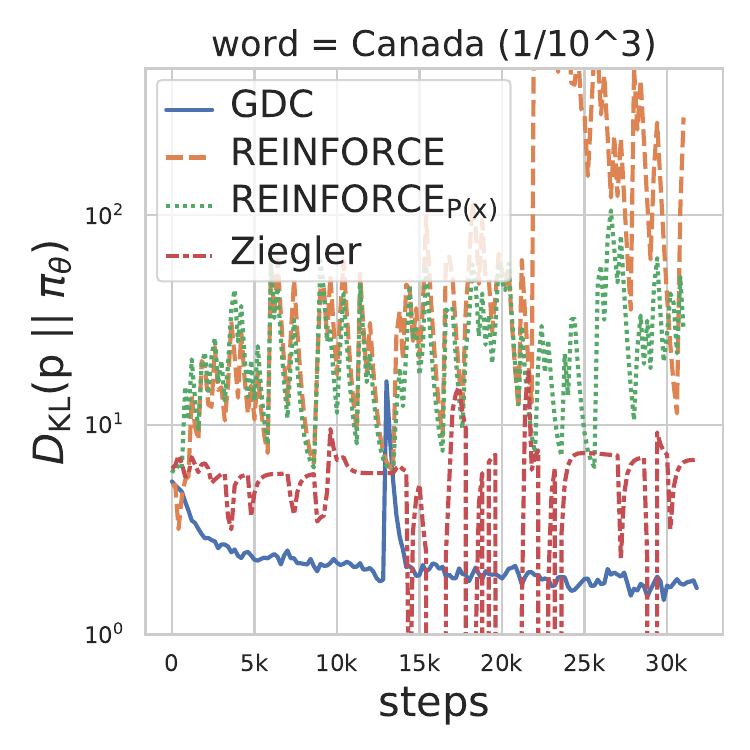}
 \includegraphics[width=0.33\textwidth]{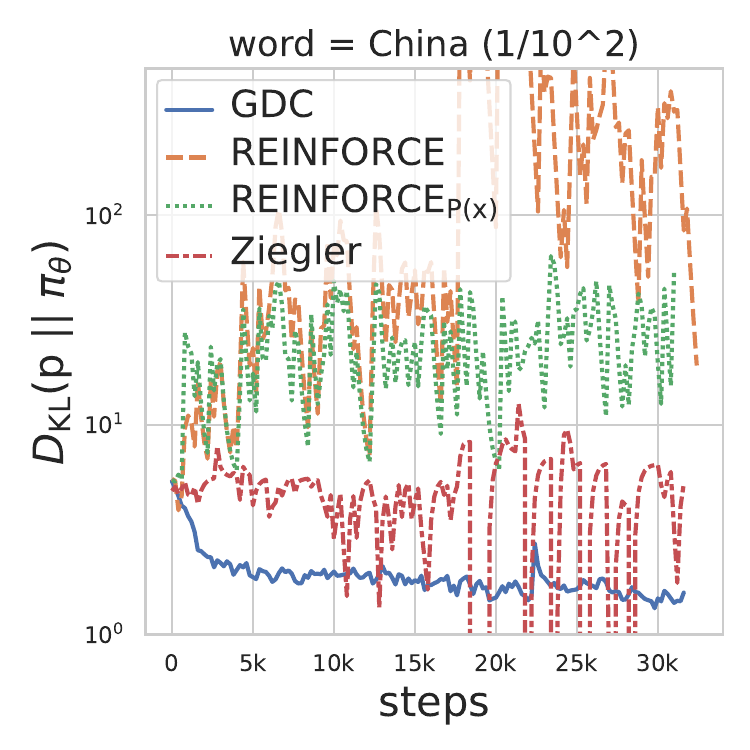}
 \includegraphics[width=0.33\textwidth]{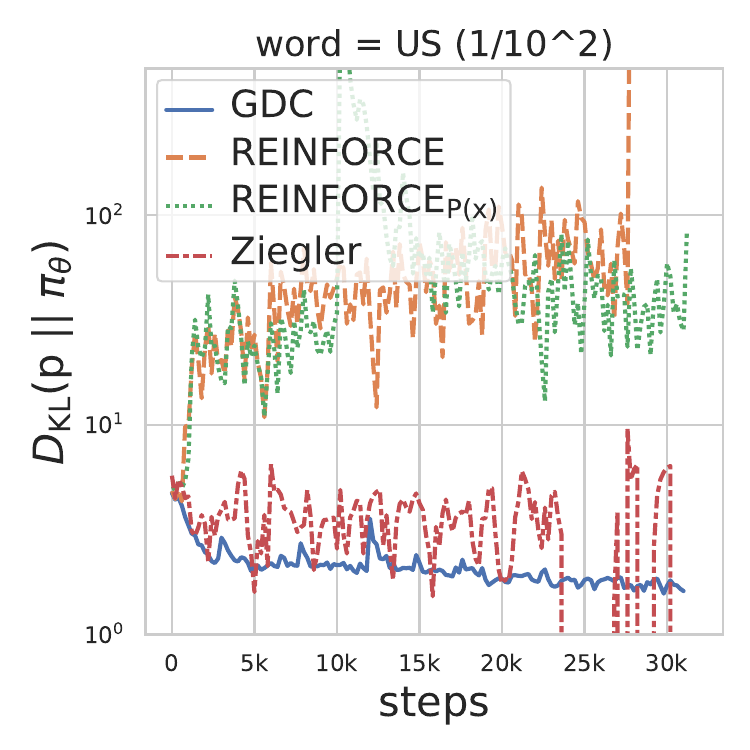}
 \caption{\label{fig:appendix-word-klp} $\KL(p, \pi_\theta)$ against the training steps for \GDC and the three baselines introduced in section \S\ref{sec:baselines} for the single-word control task. Curves are displayed for nine different single-word constraints of varying rarity levels (1/100, 1/1000, 1/10000). \GDC exhibits much better convergence behaviour than the other baselines, showing its superiority in approximating the desired distribution $p$.}
 \end{figure}

 \begin{figure}[H]
 \includegraphics[width=0.33\textwidth]{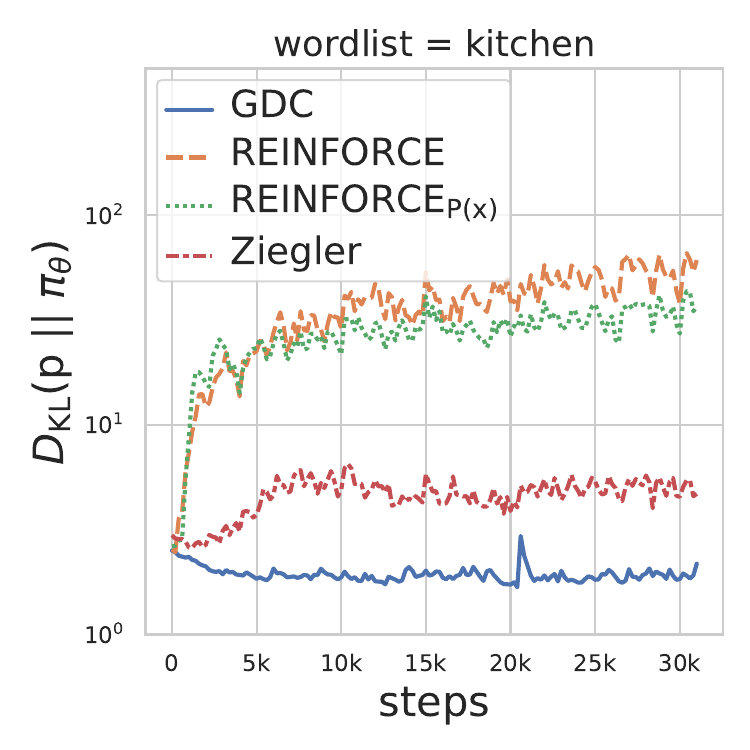}
 \includegraphics[width=0.33\textwidth]{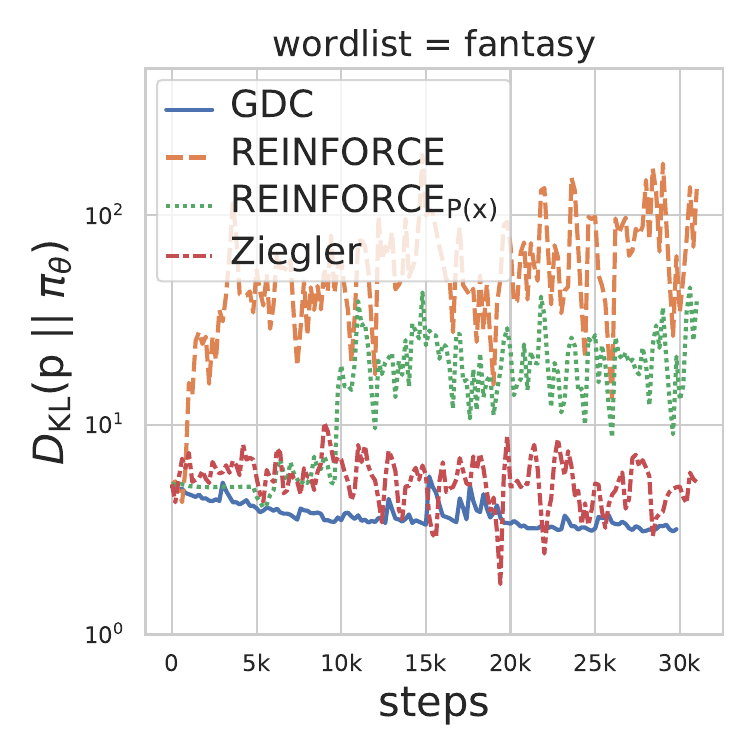}
 \includegraphics[width=0.33\textwidth]{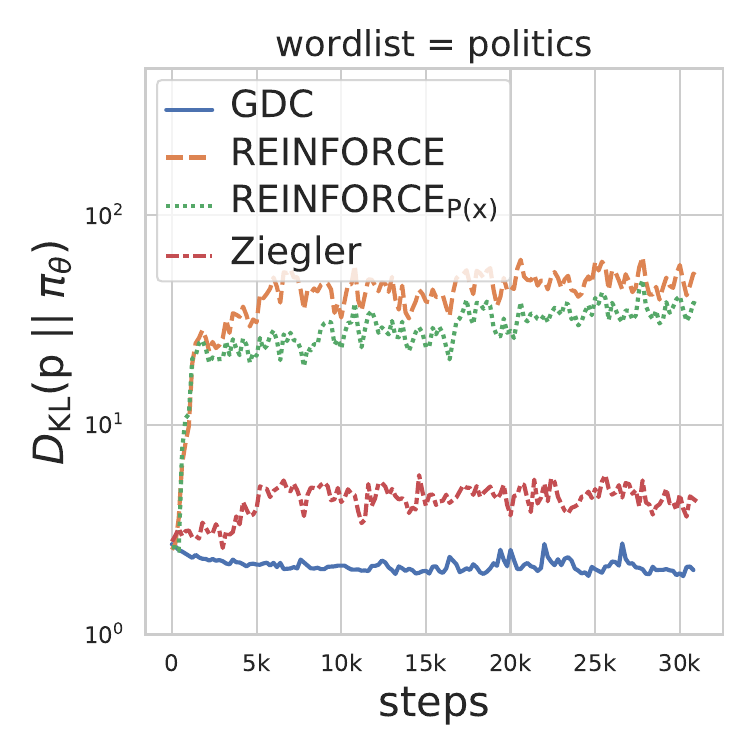}
 \includegraphics[width=0.33\textwidth]{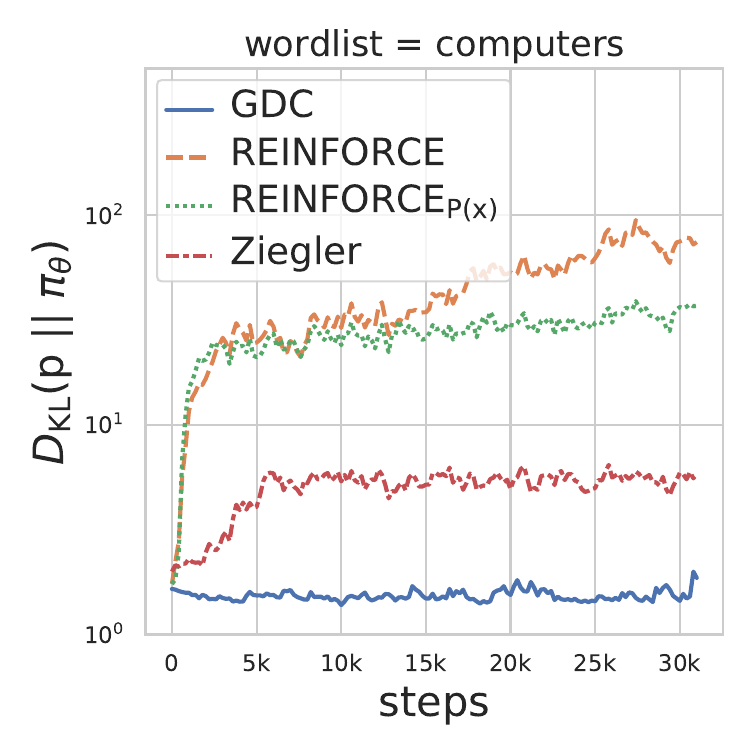}
 \caption{\label{fig:appendix-wordlist-klp}$\KL(p, \pi_\theta)$ against the training steps for \GDC and the three baselines introduced in section ~\ref{sec:baselines} for word-list constraints. Curves are displayed for 4 word-lists: kitchen , fantasy, politics, computers. \GDC exhibits much better convergence behaviour than the other baselines, showing its superiority in approximating the desired distribution $p$.}

 \end{figure}
 \begin{figure}[H]
 \includegraphics[width=0.33\textwidth]{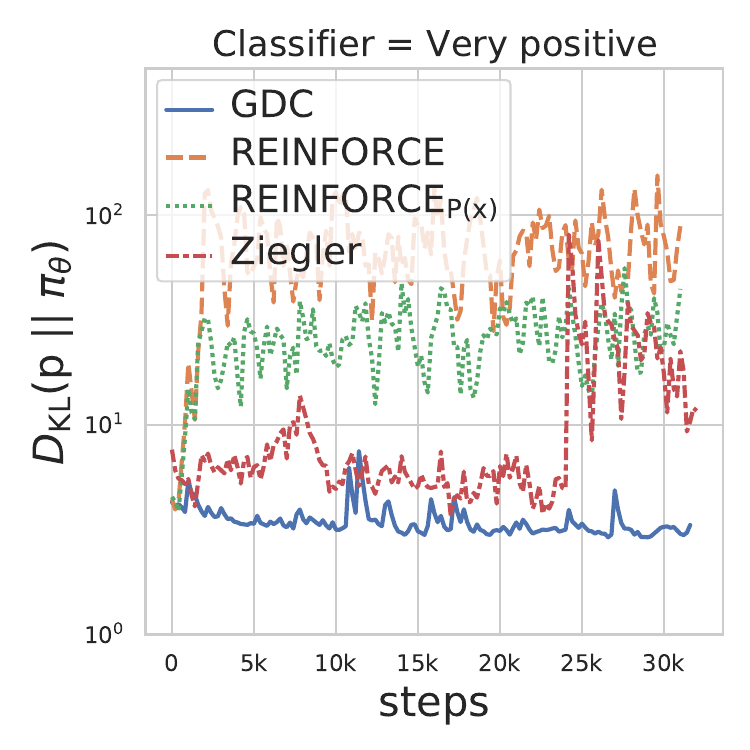}
 \includegraphics[width=0.33\textwidth]{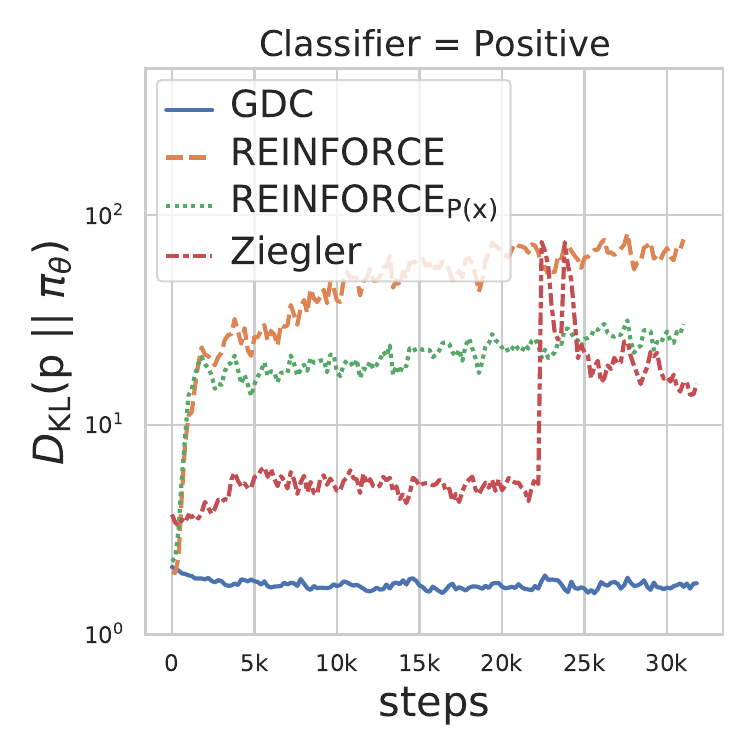}
 \includegraphics[width=0.33\textwidth]{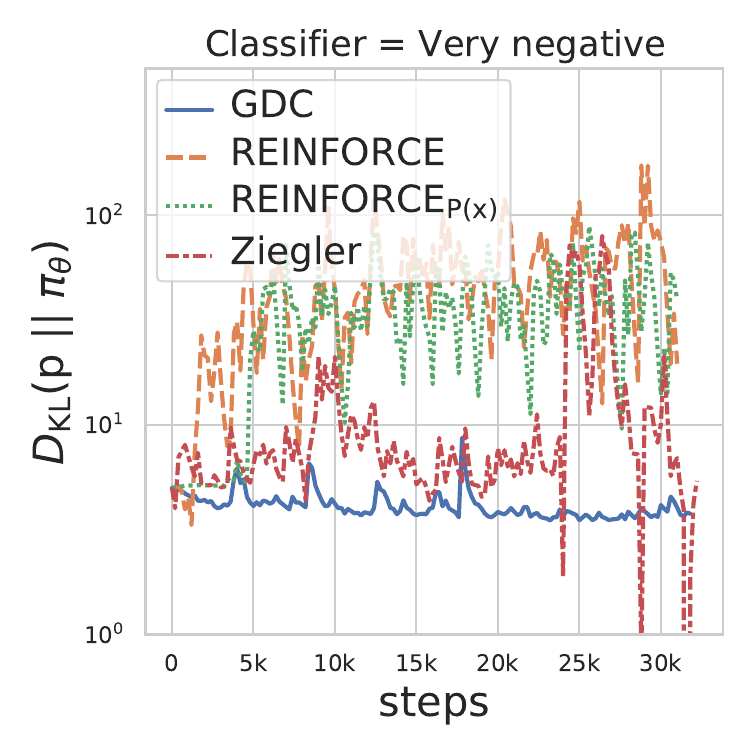}
 \includegraphics[width=0.33\textwidth]{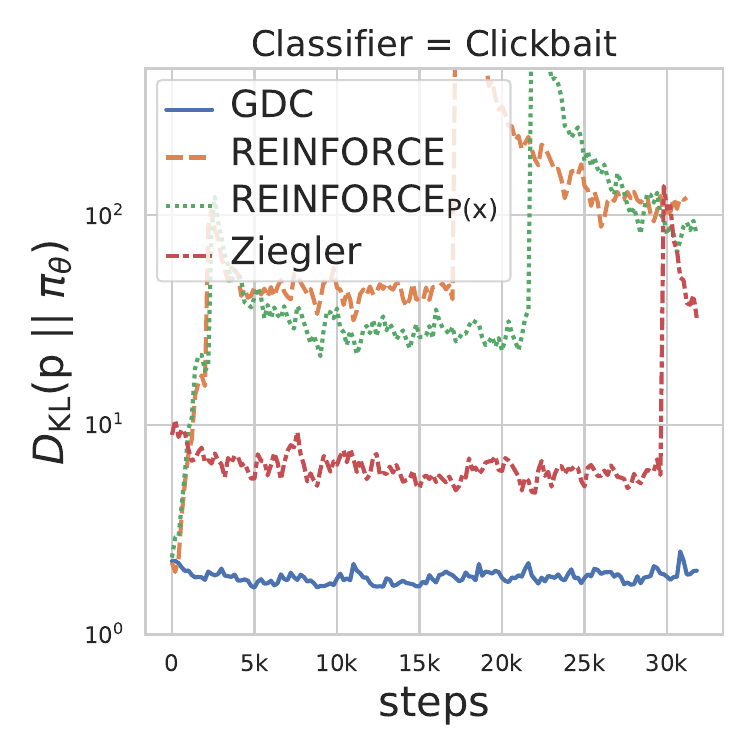}
 \caption{\label{fig:appendix-classifier-klp} $\KL(p, \pi_\theta)$ against the training steps for \GDC and the three baselines introduced in section ~\ref{sec:baselines} for classifier-based control. Curves are displayed using 4 different classifiers: very positive, positive, and very negative sentiment, and click-bait. \GDC exhibits much better convergence behaviour than the other baselines, showing its superiority in approximating the desired distribution $p$.}
 \end{figure}

%% file: sections/experiments_appendix.tex
\subsection{More Details on Point-wise Constraints Experiments}
\label{sec:appendix:experiments_constraints_details}
% word classifiers, 
% wordlists: wheree to find them (cite plug and play)
% discriminators : excerpt from plug and play : We train a discriminator on a dataset with input sentences x and corresponding labels yx. For aninput x of length t, we compute ox:tand train f on the mean (o¯t) of the embeddings across time. All discriminators in this work consist of a single layer classifier that predicts the target label from o¯xt. 
%distributional control:
% details on male / female/ other classifier how do we do it
% details on word-list and what words contained in each word list

 \subsection{Pointwise Constraints}

 \begin{figure}[H]
 \includegraphics[width=\linewidth]{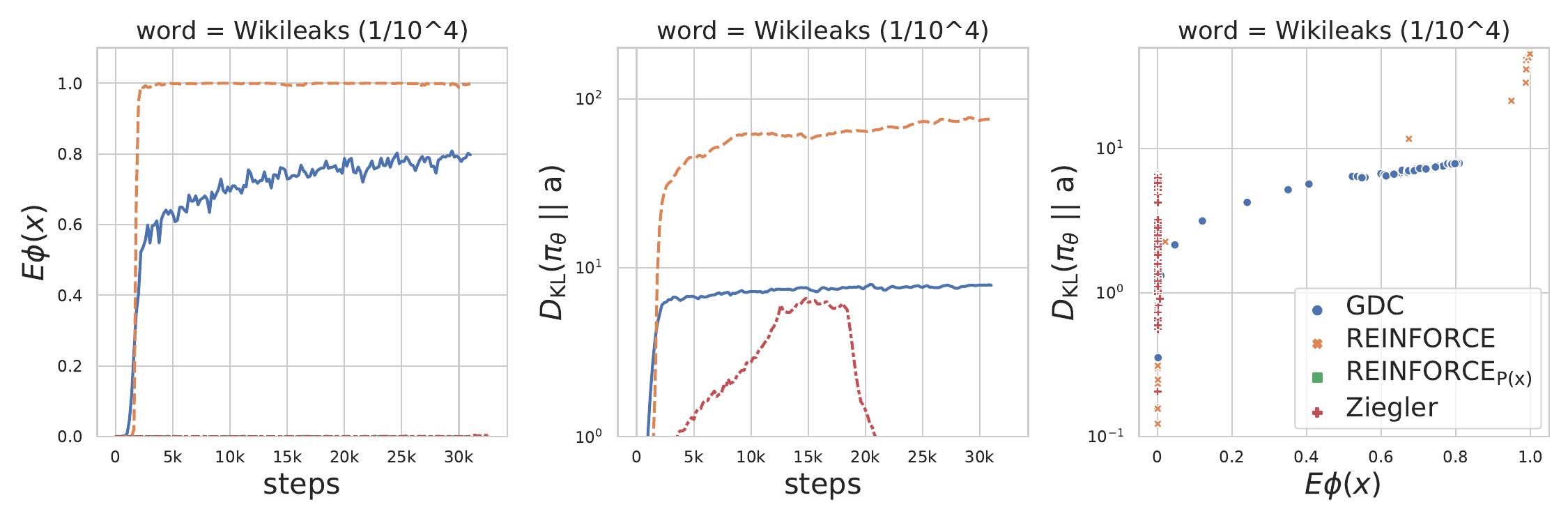}
 \includegraphics[width=\linewidth]{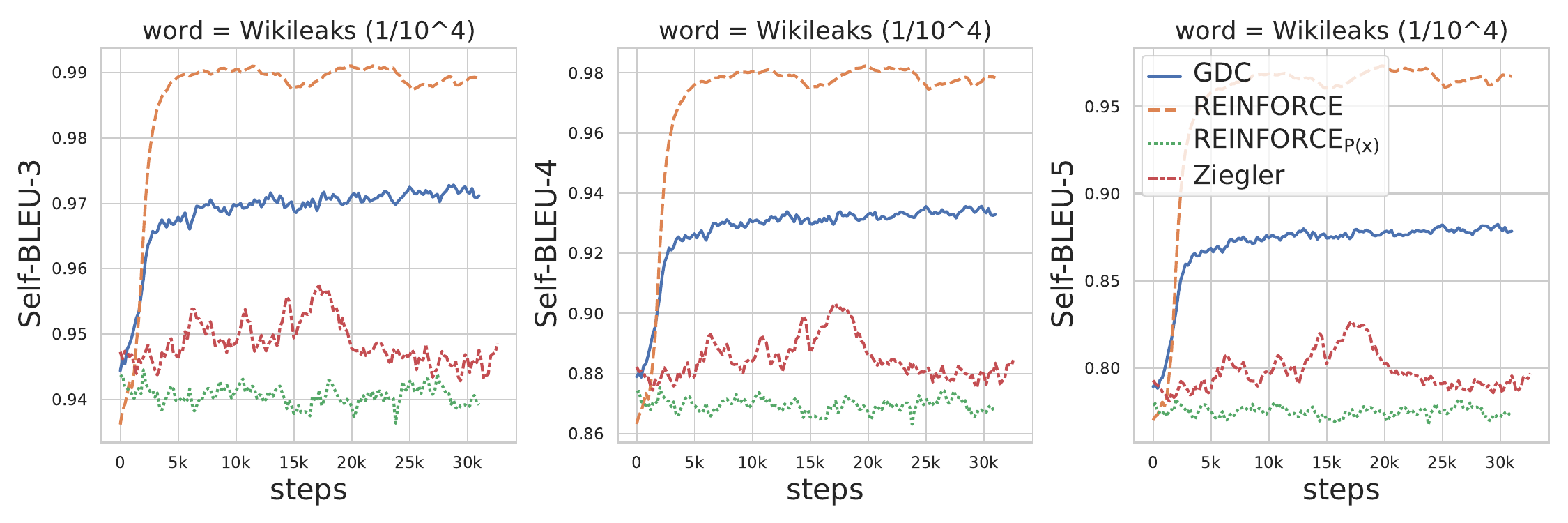}
 \includegraphics[width=\linewidth]{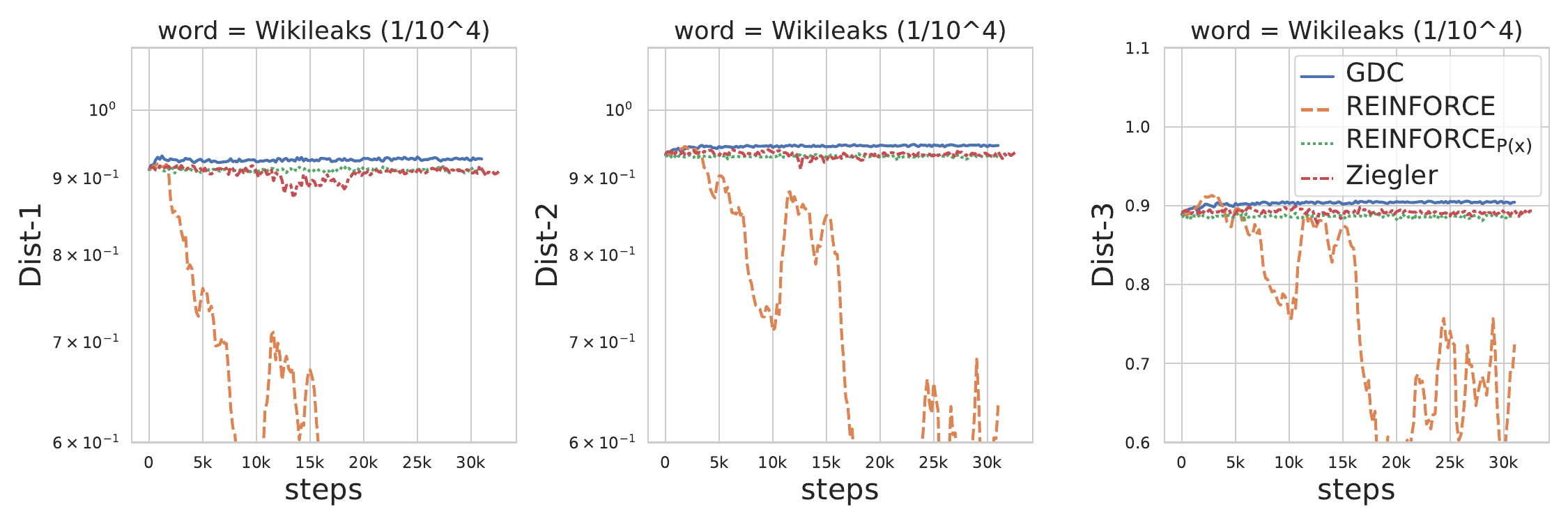}
 \caption{\label{fig:appendix-1-Wikileaks} Line plot of different evaluation metrics against the training steps when controlling for the word ``wikileaks'' (with initial occurrence probability of 1/10000) as a single-word constraint.}
 \end{figure}

 \begin{figure}[H]
 \includegraphics[width=\linewidth]{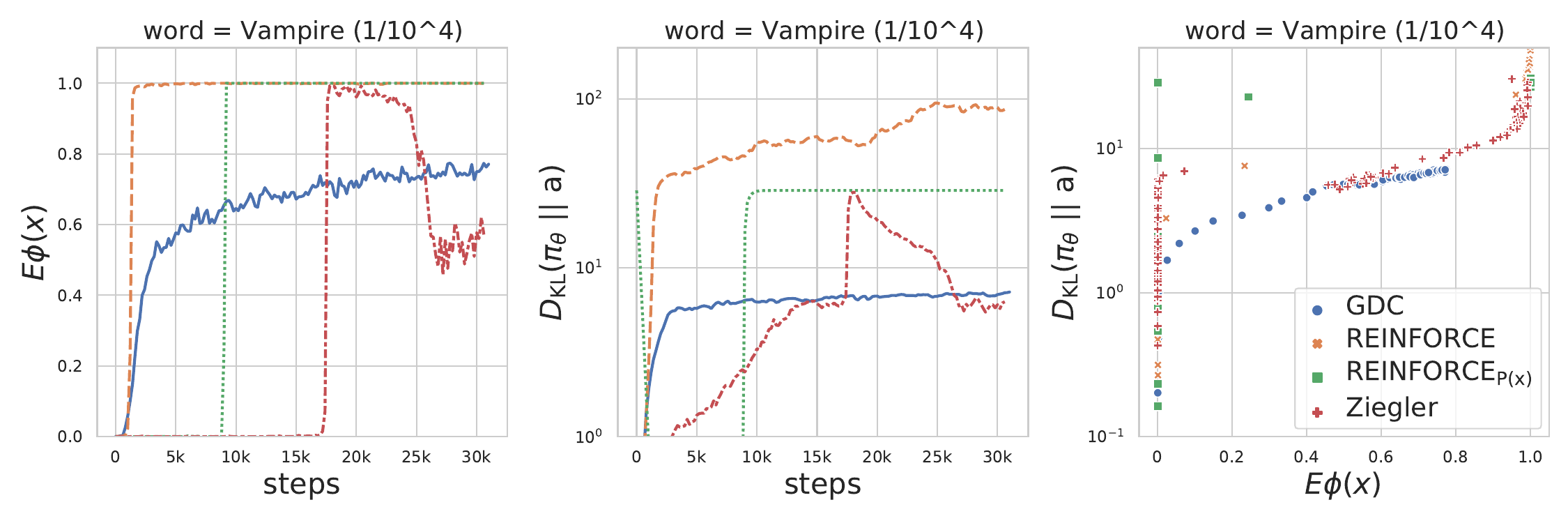}
 \includegraphics[width=\linewidth]{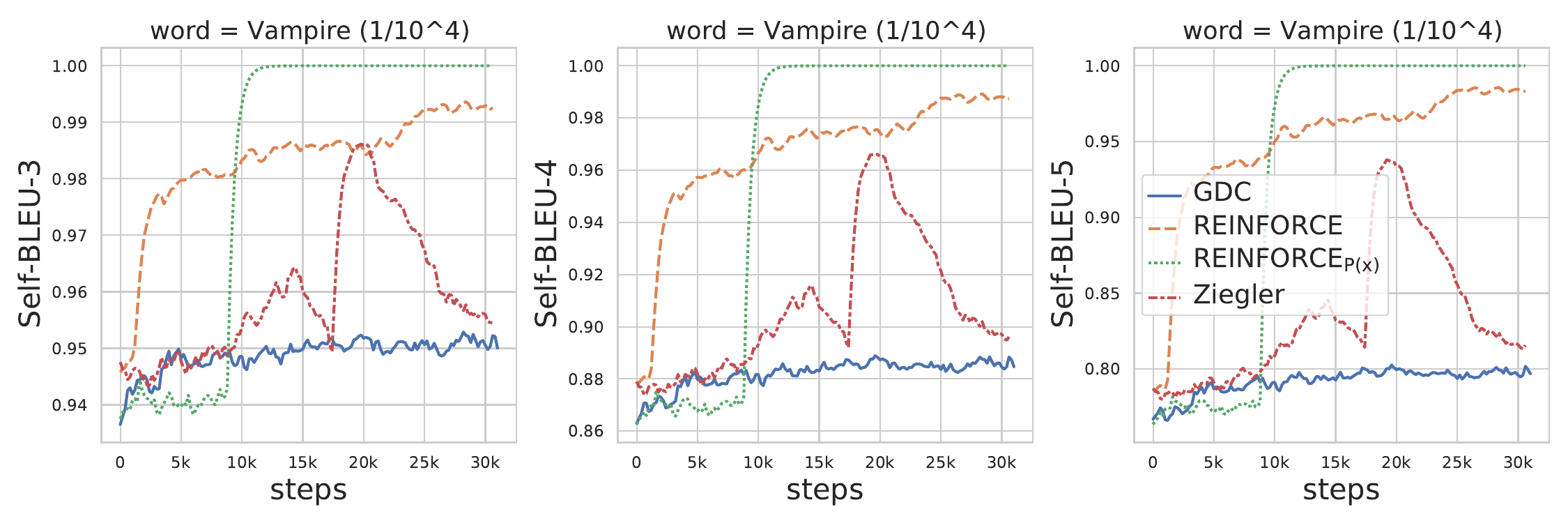}
 \includegraphics[width=\linewidth]{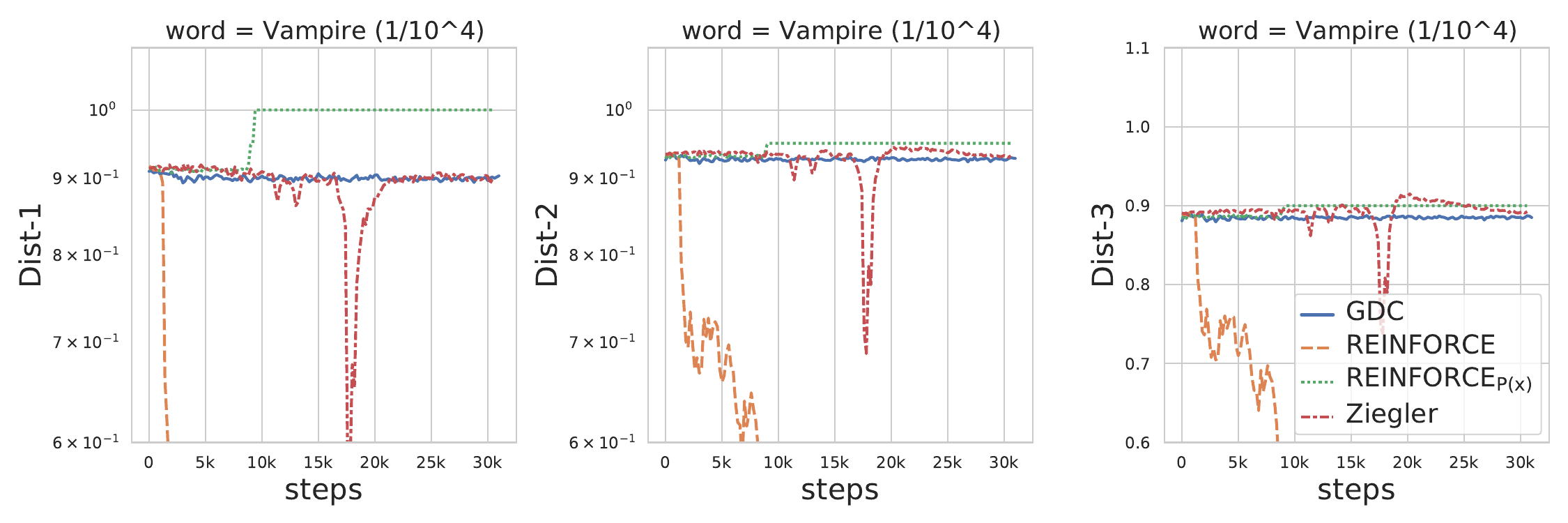}
 \caption{\label{fig:appendix-2-Vampire}Line plot of different evaluation metrics against the training steps when controlling for the word ``vampire'' (with initial occurrence probability of 1/10000) as a single-word constraint.}
 \end{figure}

 \begin{figure}[H]
 \includegraphics[width=\linewidth]{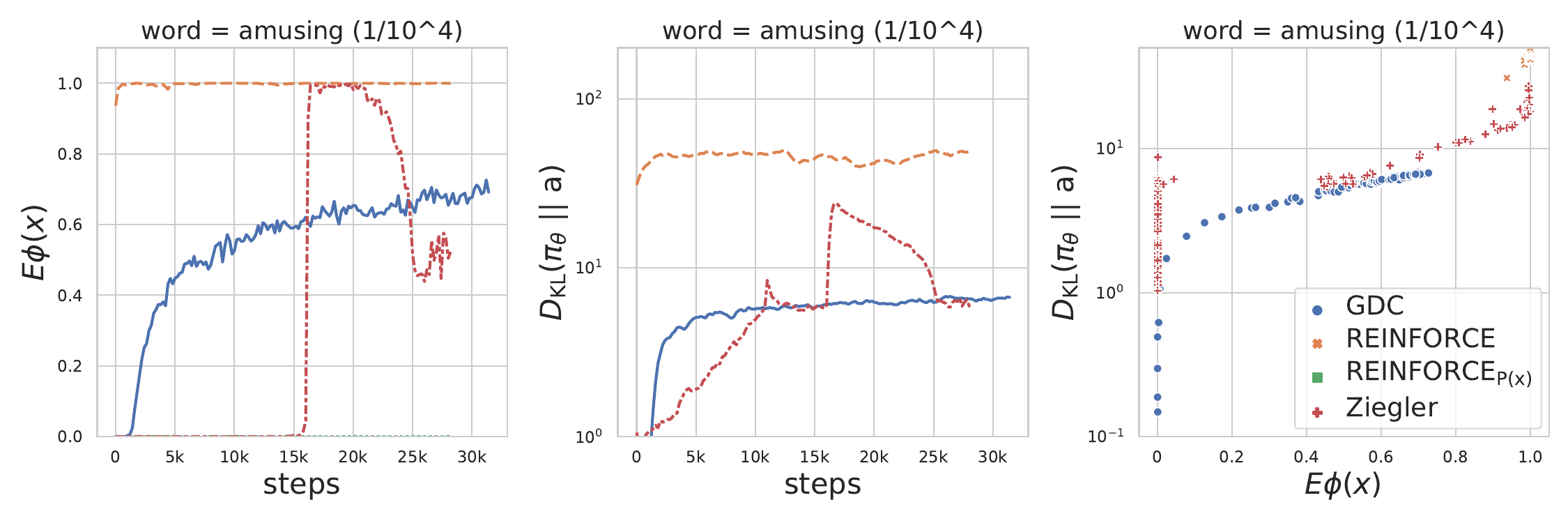}
 \includegraphics[width=\linewidth]{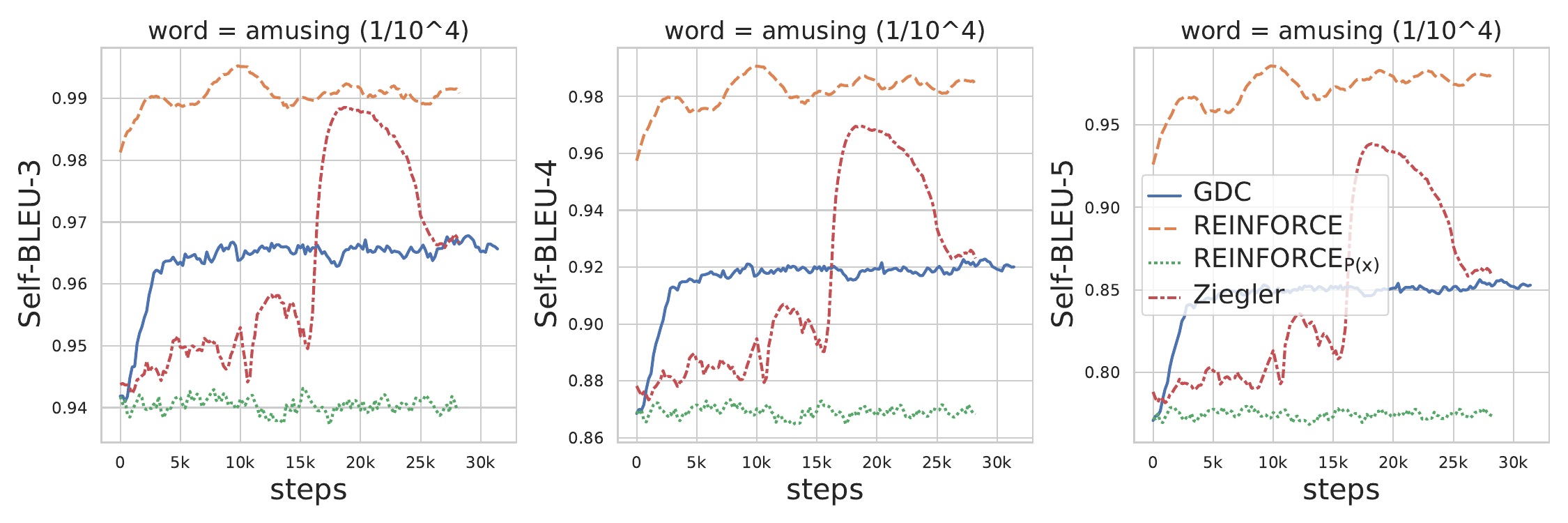}
 \includegraphics[width=\linewidth]{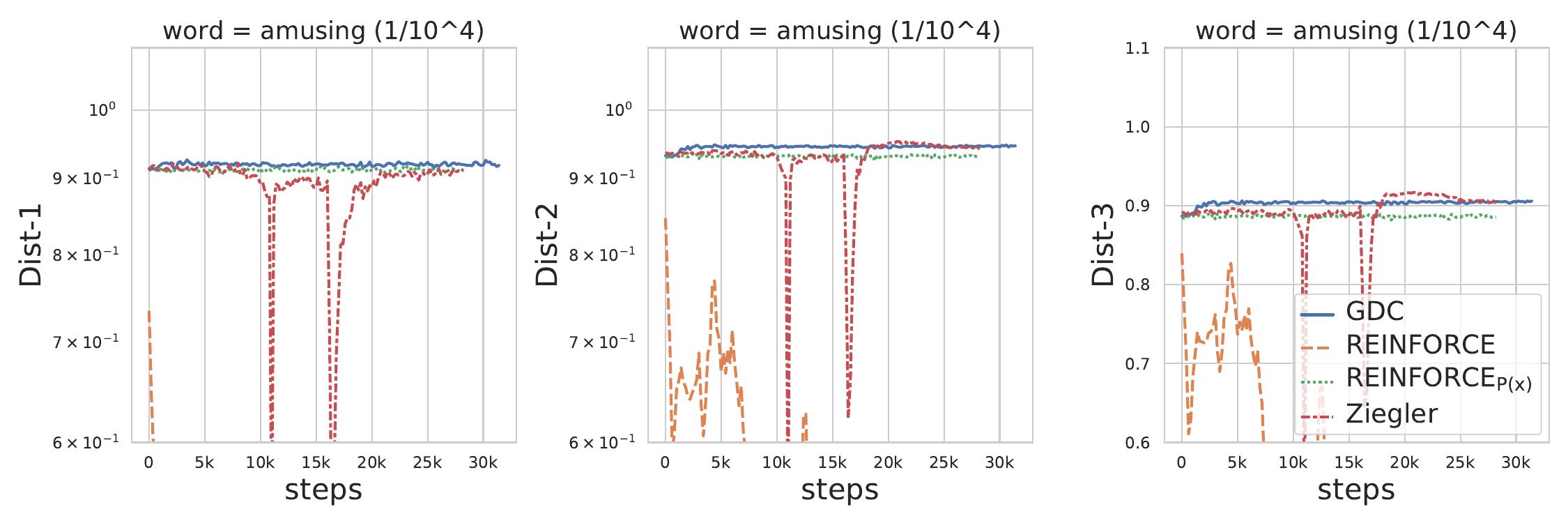}
 \caption{\label{fig:appendix-3-amusing}Line plot of different evaluation metrics against the training steps when controlling for the word ``amusing'' (with initial occurrence probability of 1/10000) as a single-word constraint.}
 \end{figure}

 \begin{figure}[H]
 \includegraphics[width=\linewidth]{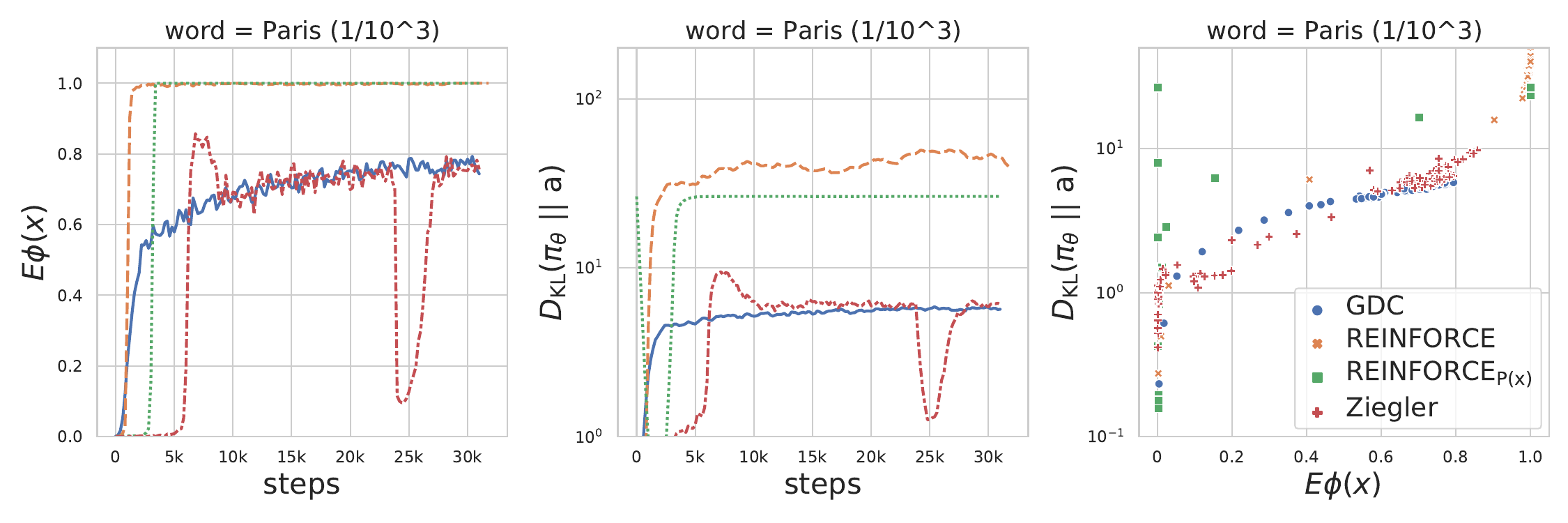}
 \includegraphics[width=\linewidth]{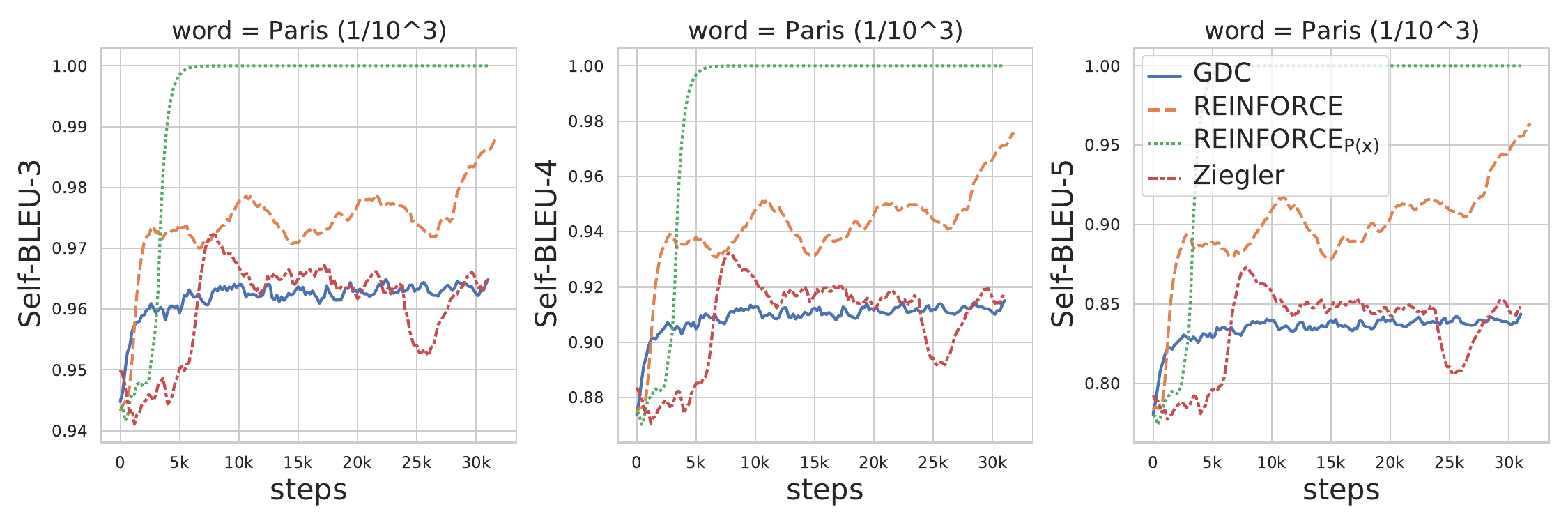}
 \includegraphics[width=\linewidth]{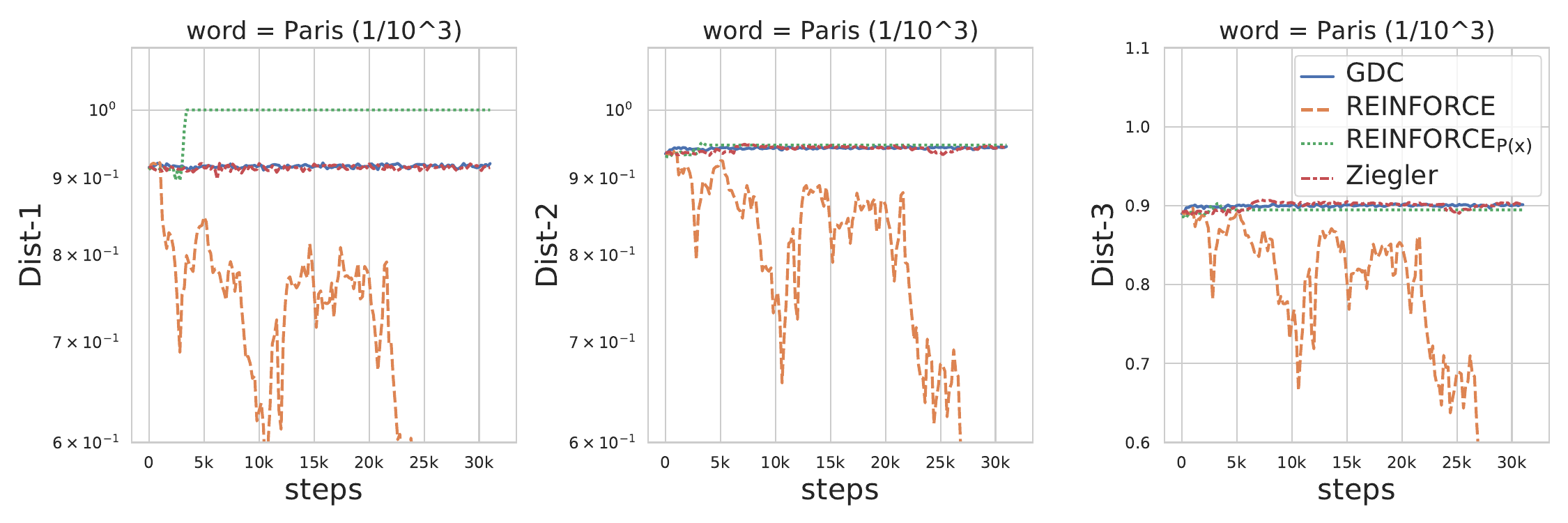}
 \caption{\label{fig:appendix-4-Paris}Line plot of different evaluation metrics against the training steps when controlling for the word ``Paris'' (with initial occurrence probability of 1/1000) as a single-word constraint.}
 \end{figure}

 \begin{figure}[H]
 \includegraphics[width=\linewidth]{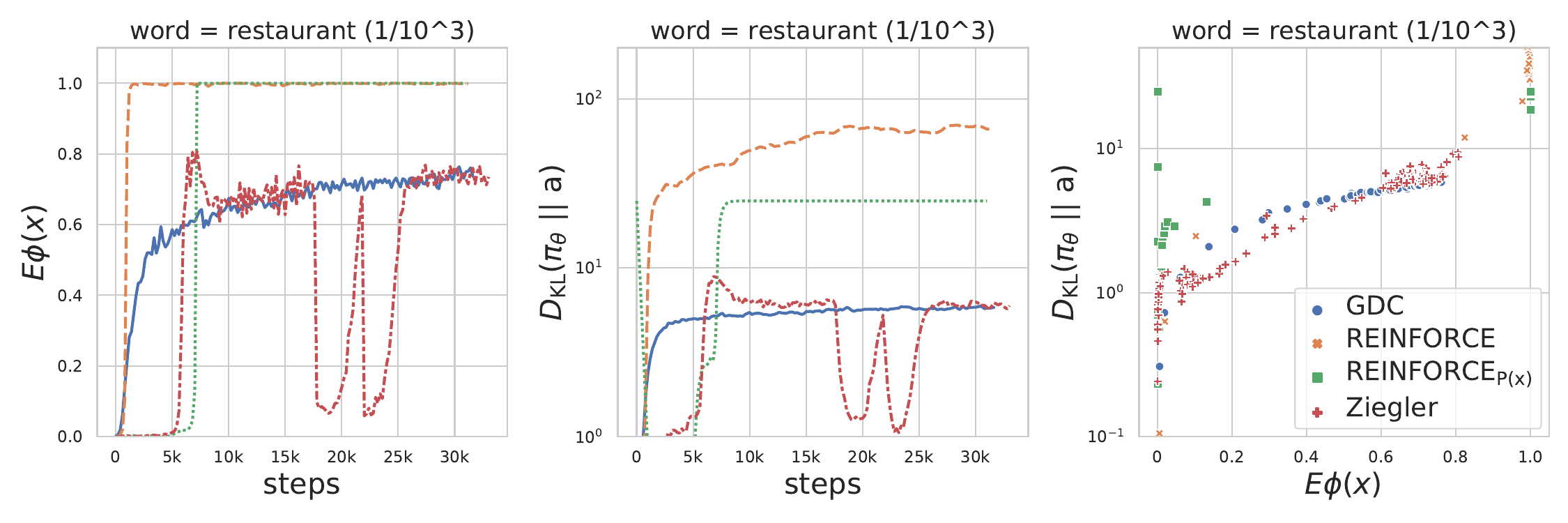}
 \includegraphics[width=\linewidth]{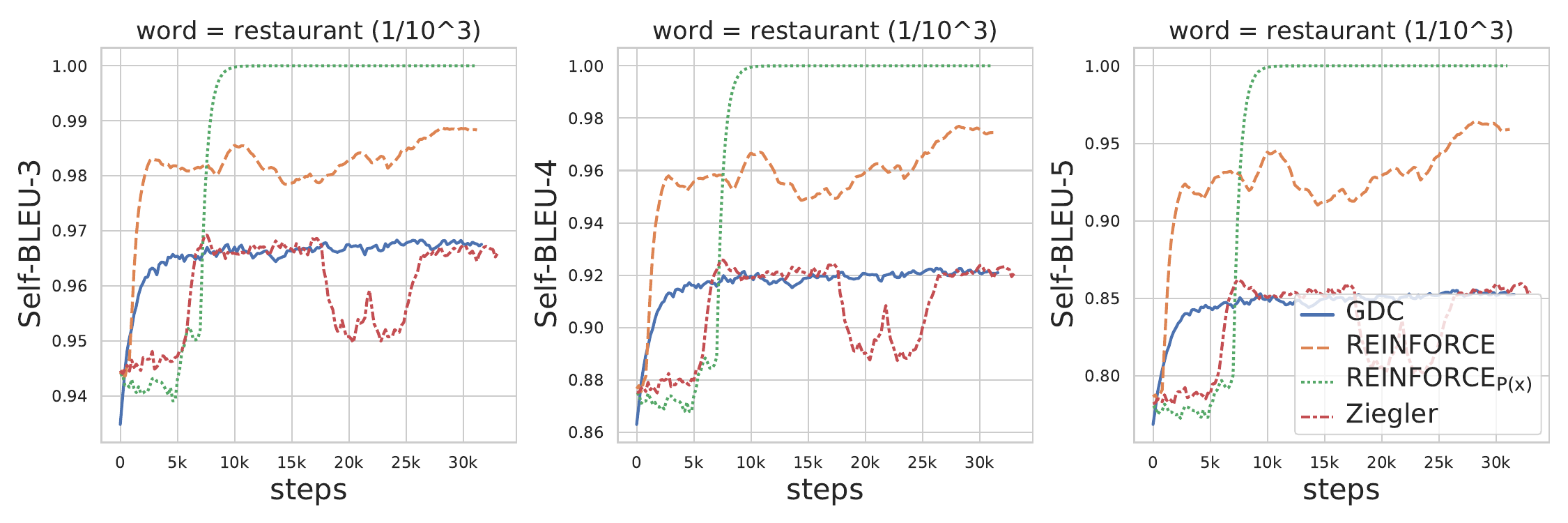}
 \includegraphics[width=\linewidth]{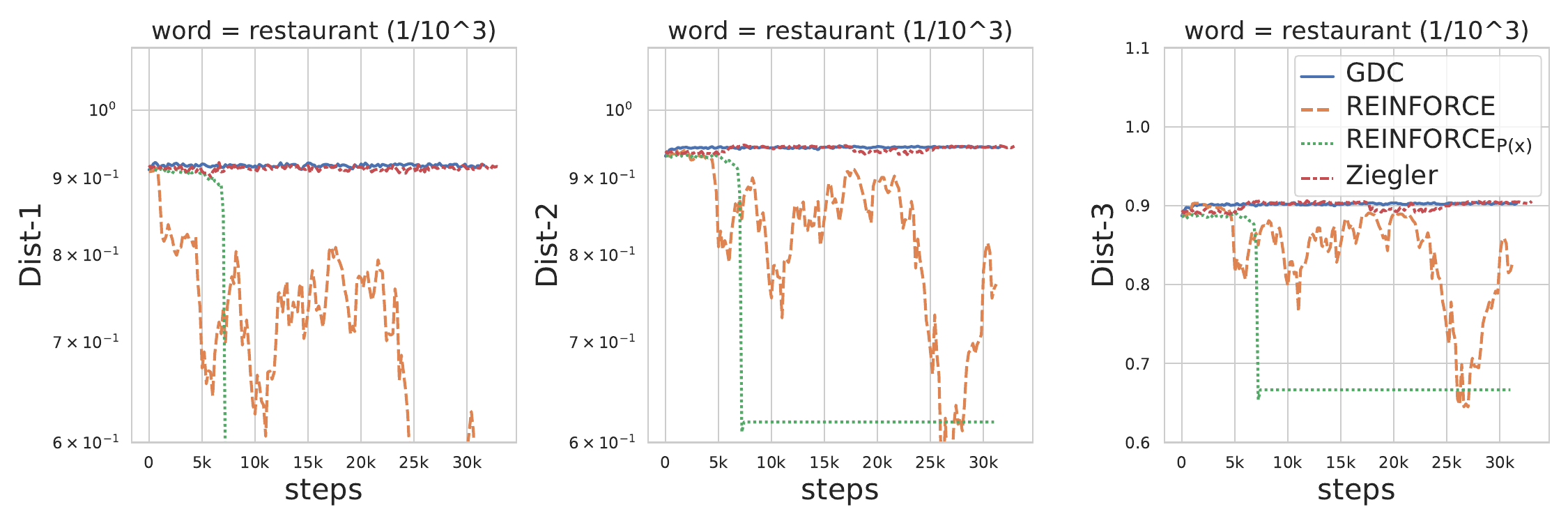}
 \caption{\label{fig:appendix-5-restaurant}Line plot of different evaluation metrics against the training steps when controlling for the word ``restaurant'' (with initial occurrence probability of 1/1000) as a single-word constraint.}
 \end{figure}

 \begin{figure}[H]
 \includegraphics[width=\linewidth]{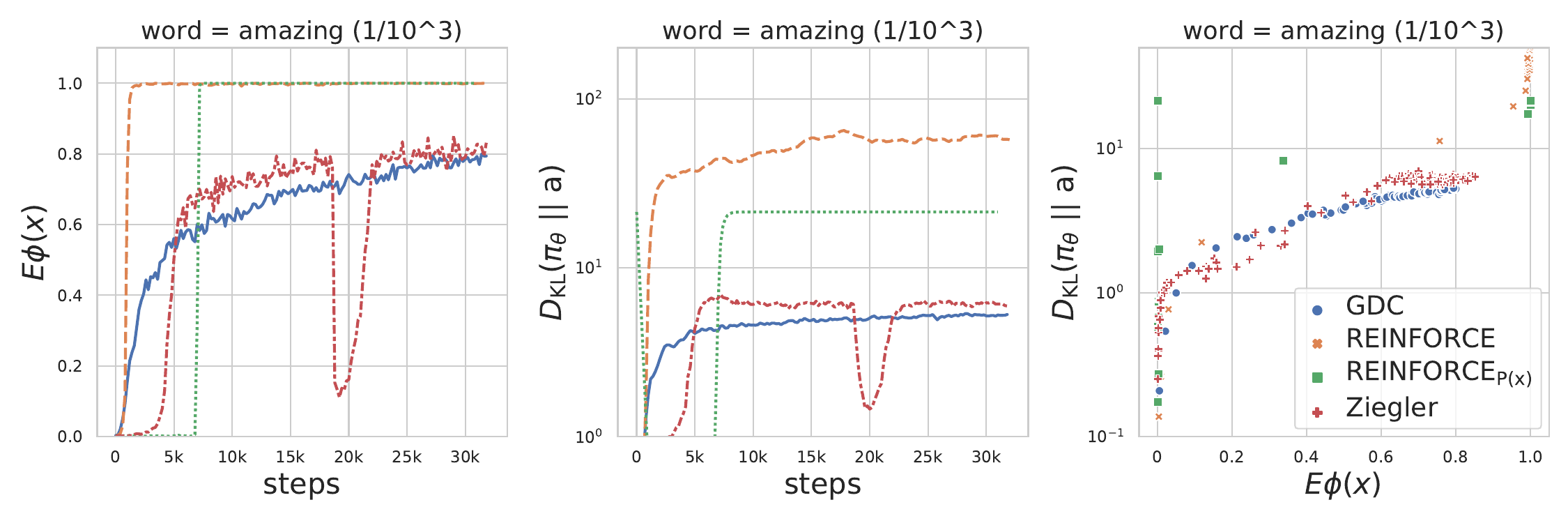}
 \includegraphics[width=\linewidth]{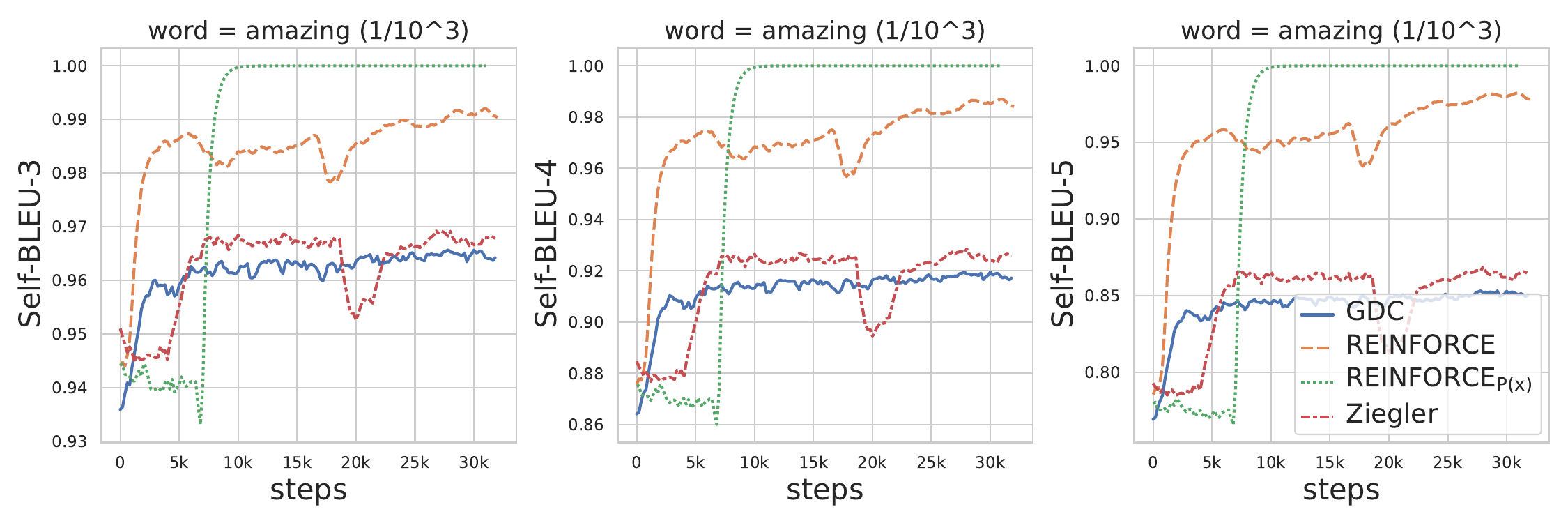}
 \includegraphics[width=\linewidth]{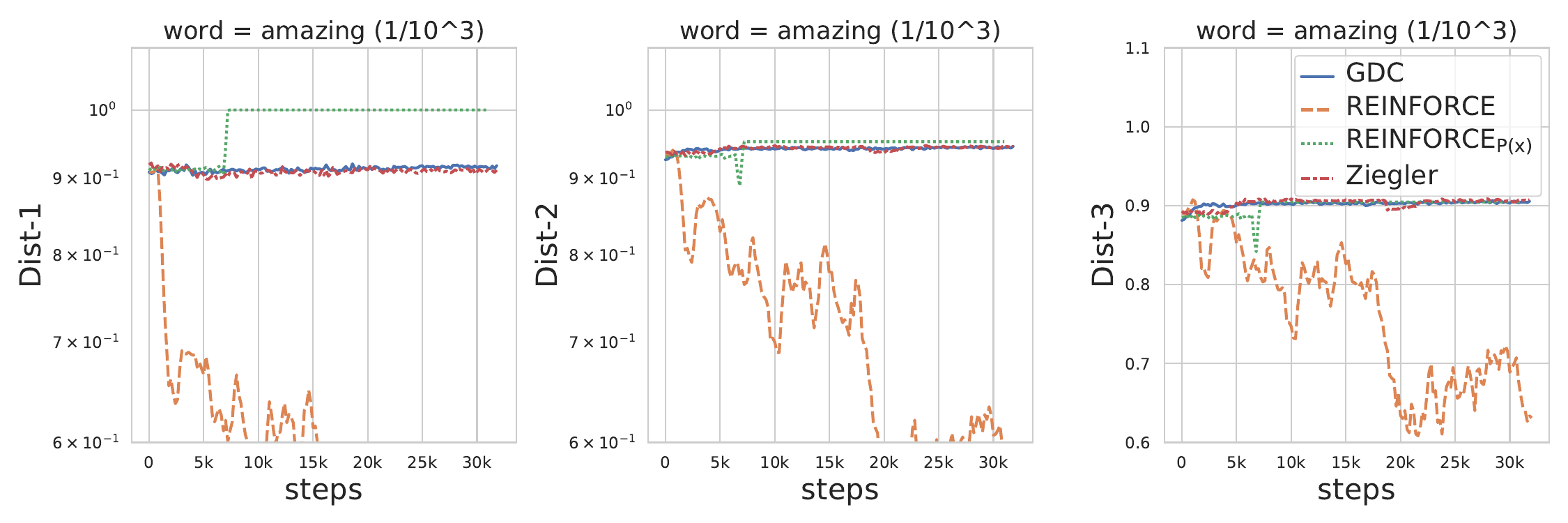}
 \caption{\label{fig:appendix-6-amazing}Line plot of different evaluation metrics against the training steps when controlling for the word ``amazing'' (with initial occurrence probability of 1/1000) as a single-word constraint.}
 \end{figure}

 \begin{figure}[H]
 \includegraphics[width=\linewidth]{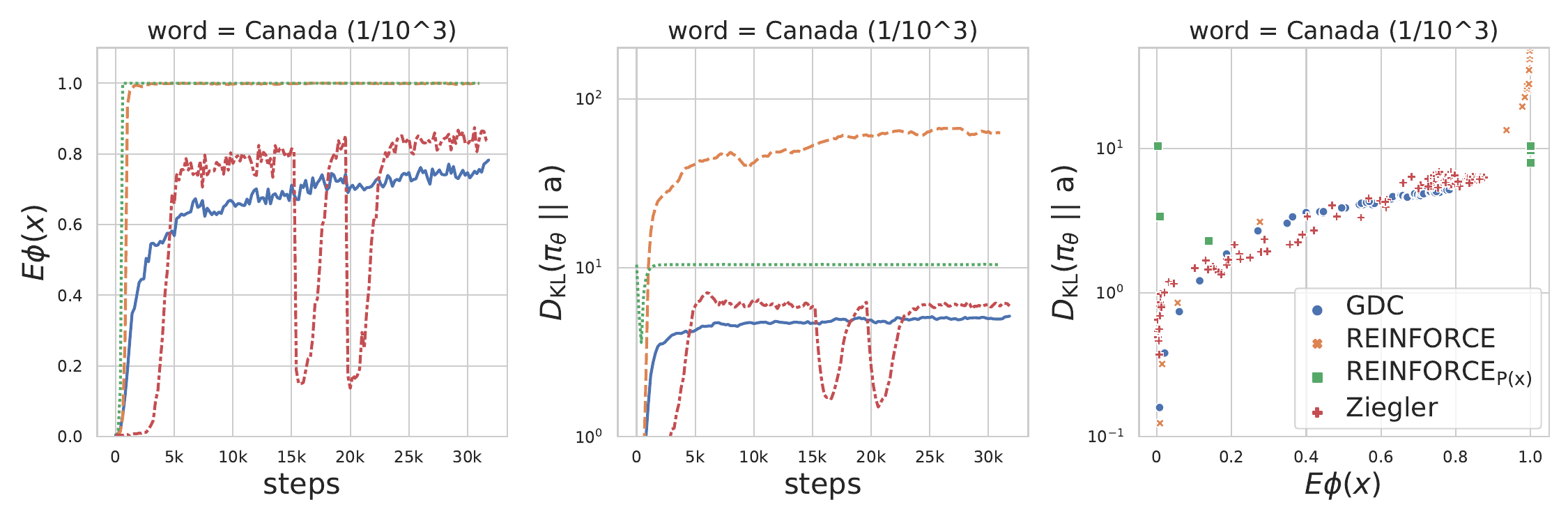}
 \includegraphics[width=\linewidth]{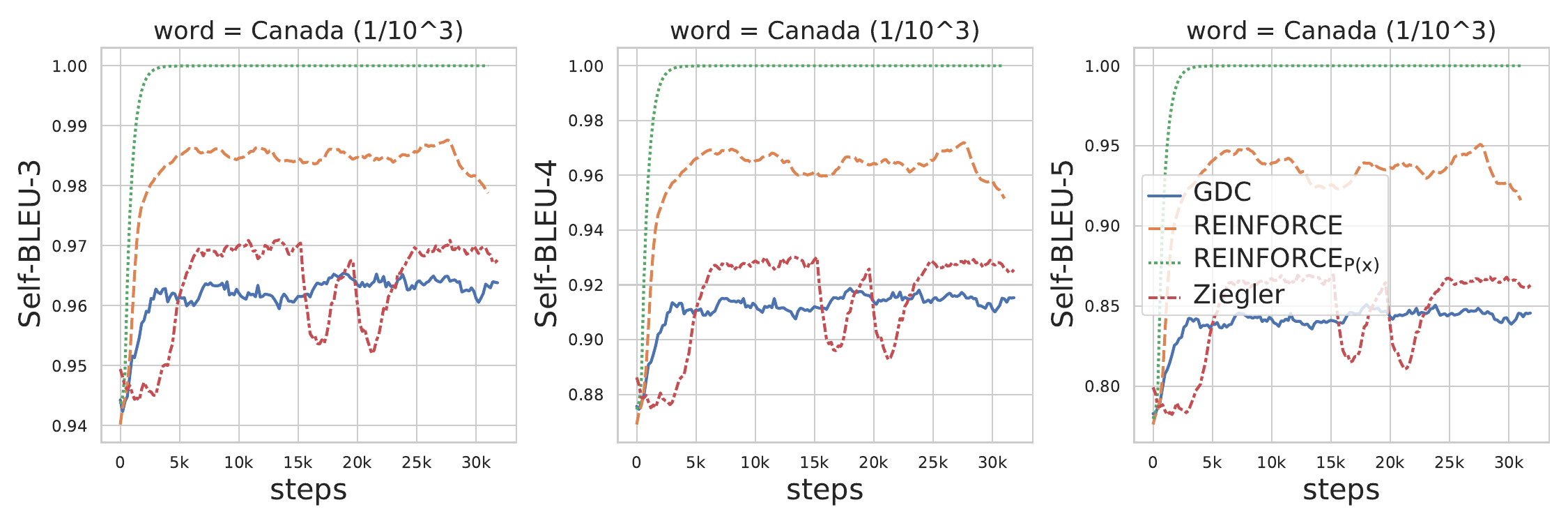}
 \includegraphics[width=\linewidth]{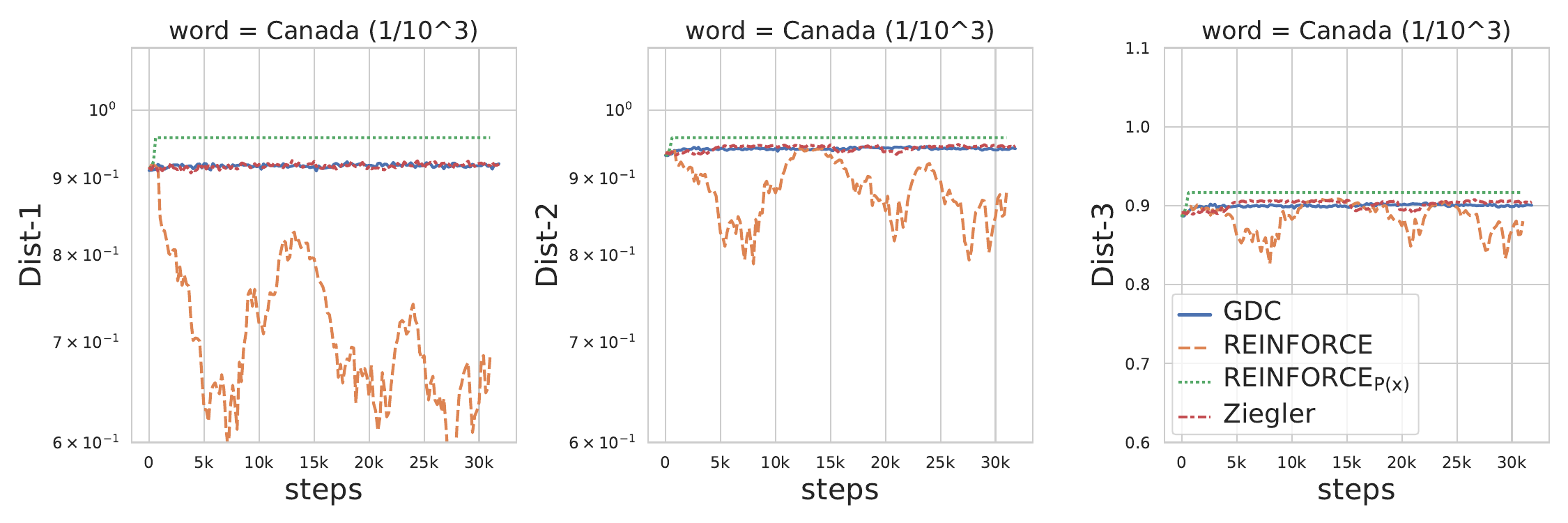}
 \caption{\label{fig:appendix-7-Canada}Line plot of different evaluation metrics against the training steps when controlling for the word ``Canada'' (with initial occurrence probability of 1/1000) as a single-word constraint.}
 \end{figure}

 \begin{figure}[H]
 \includegraphics[width=\linewidth]{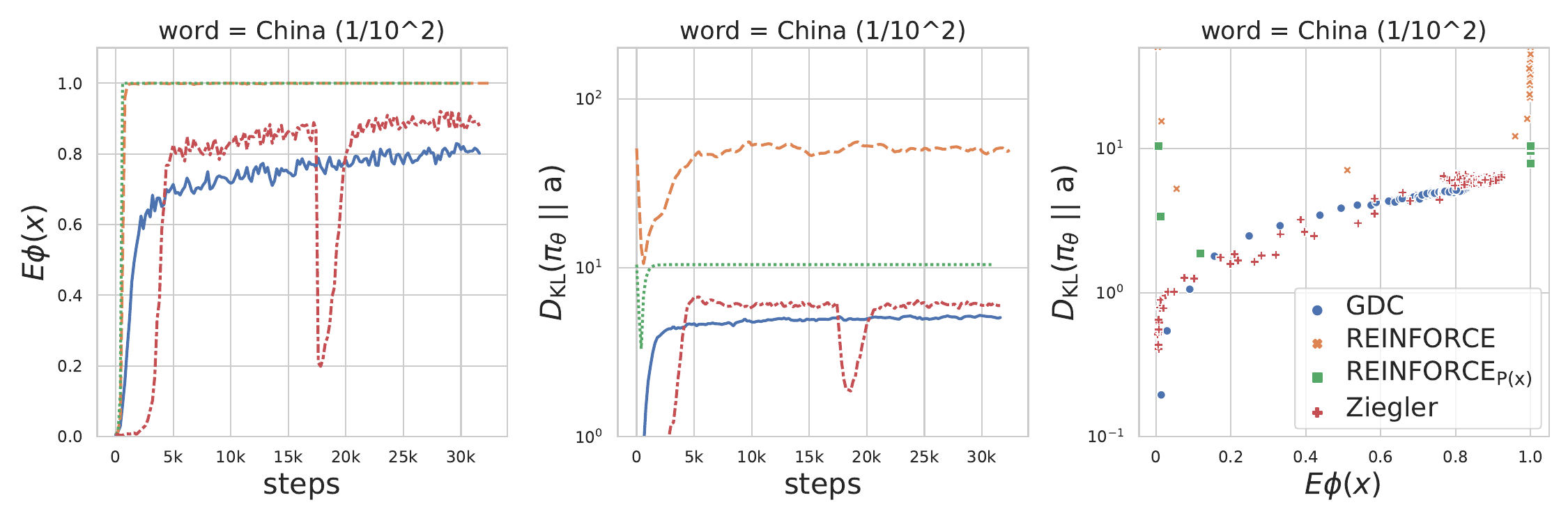}
 \includegraphics[width=\linewidth]{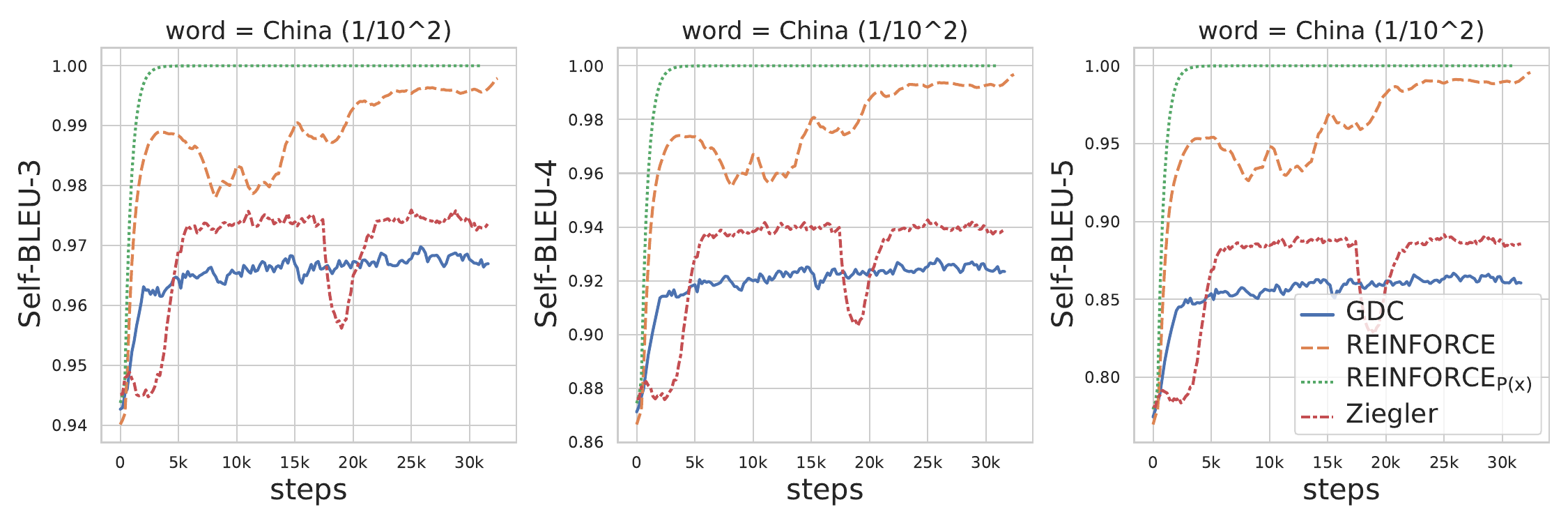}
 \includegraphics[width=\linewidth]{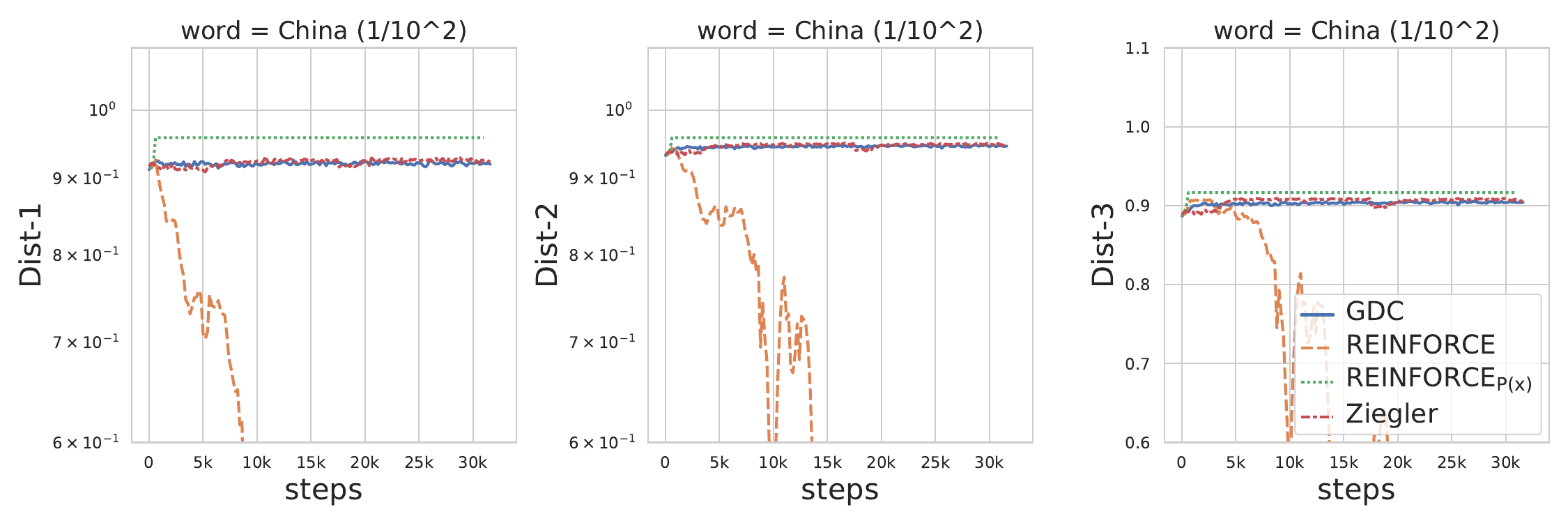}
 \caption{\label{fig:appendix-8-China}Line plot of different evaluation metrics against the training steps when controlling for the word ``China'' (with initial occurrence probability of 1/100) as a single-word constraint.}
 \end{figure}

 \begin{figure}[H]
 \includegraphics[width=\linewidth]{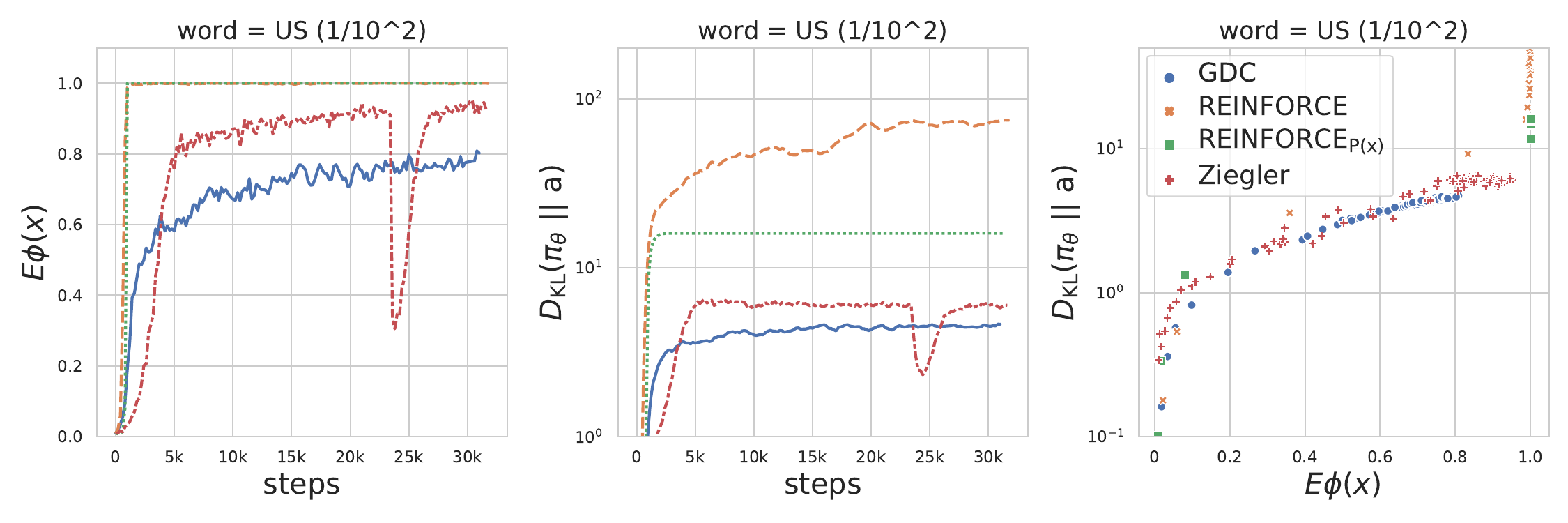}
 \includegraphics[width=\linewidth]{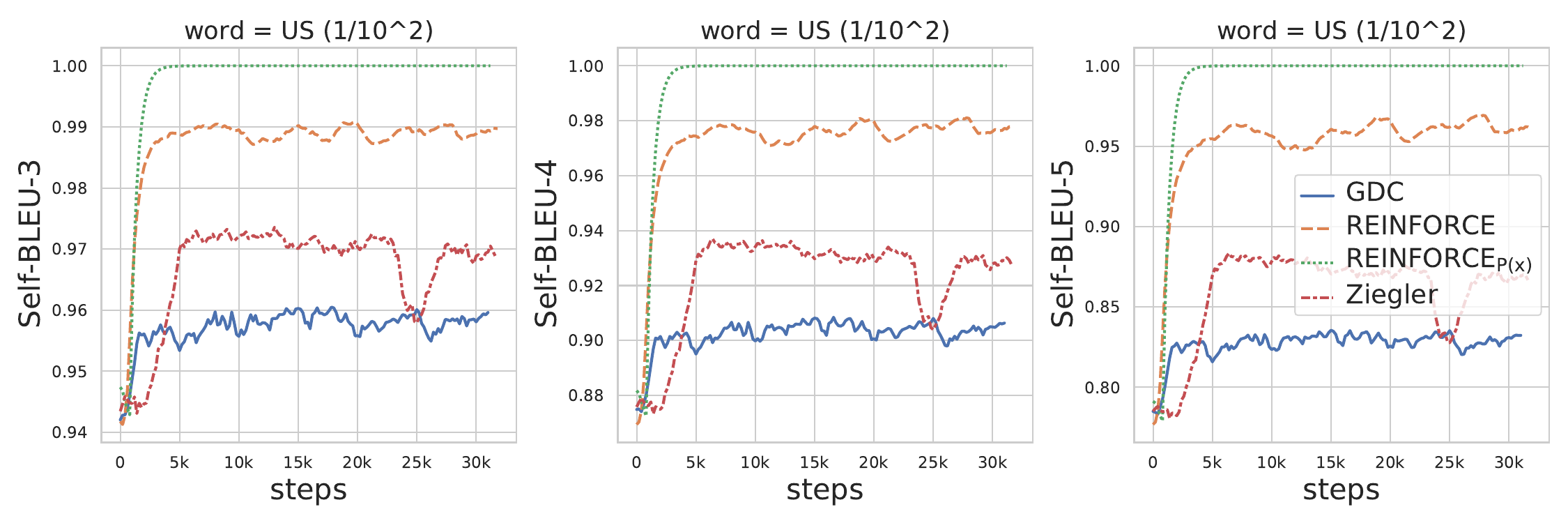}
 \includegraphics[width=\linewidth]{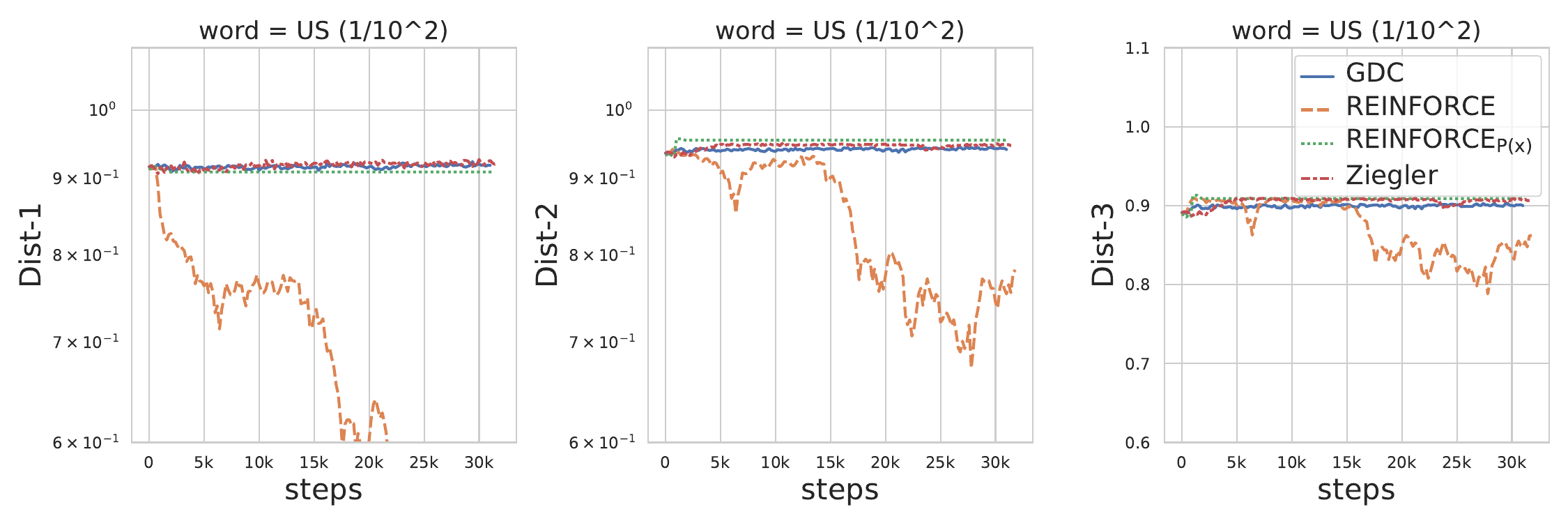}
 \caption{\label{fig:appendix-9-US}Line plot of different evaluation metrics against the training steps when controlling for the word ``US'' (with initial occurrence probability of 1/100) as a single-word constraint.}
 \end{figure}

 \begin{figure}[H]
 \includegraphics[width=\linewidth]{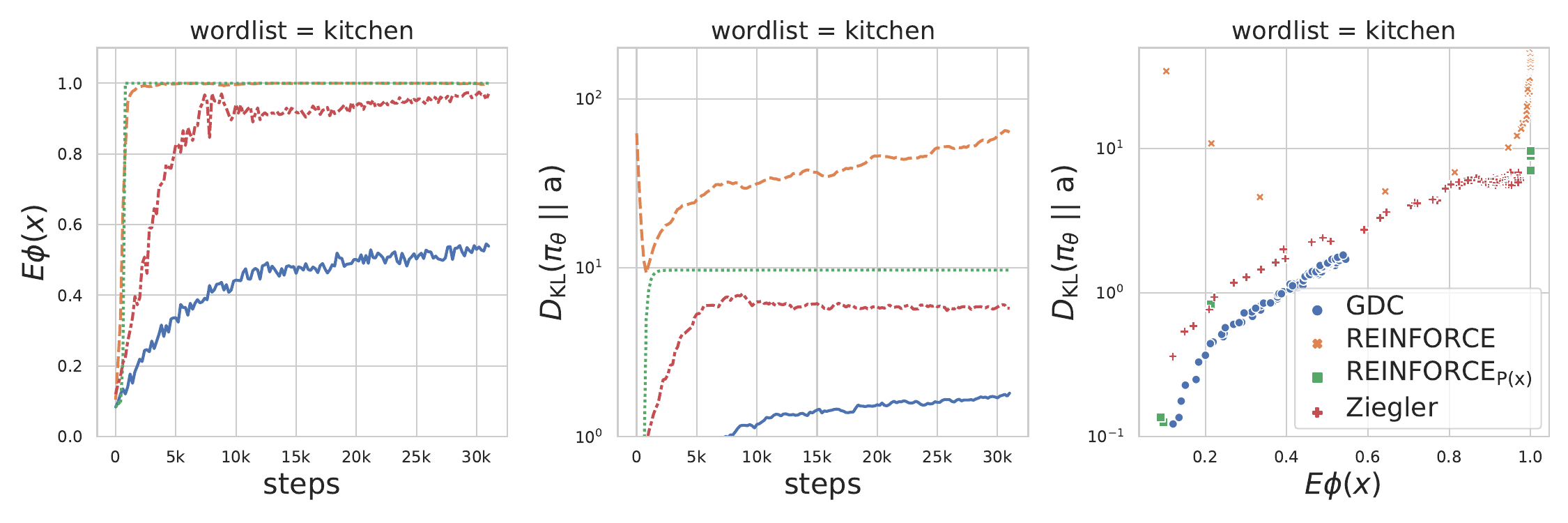}
 \includegraphics[width=\linewidth]{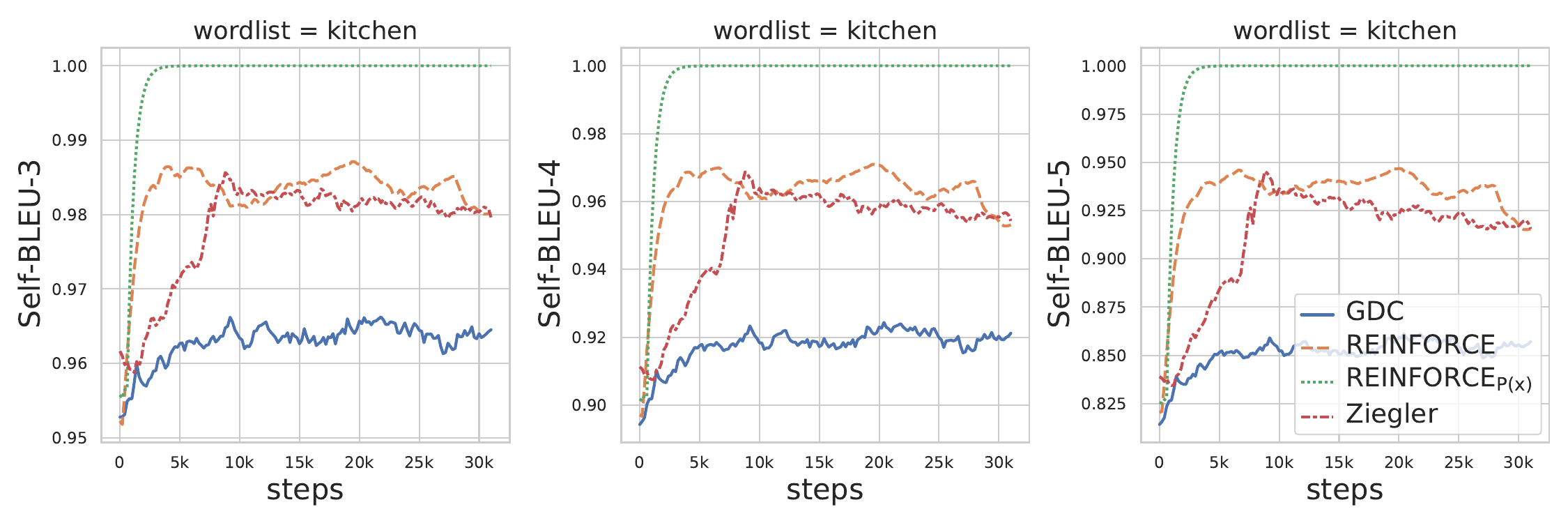}
 \includegraphics[width=\linewidth]{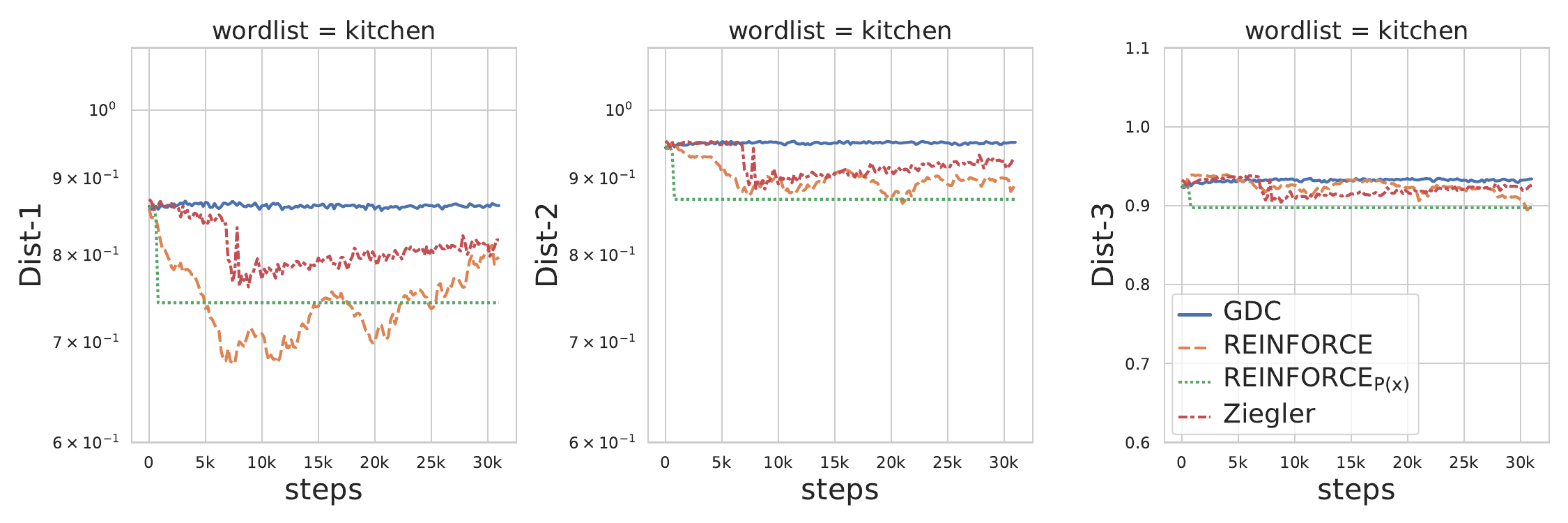}
 \caption{\label{fig:appendix-10-kitchen}Line plot of different evaluation metrics against the training steps when controlling for the \textbf{kitchen} word-list. }
 \end{figure}

 \begin{figure}[H]
 \includegraphics[width=\linewidth]{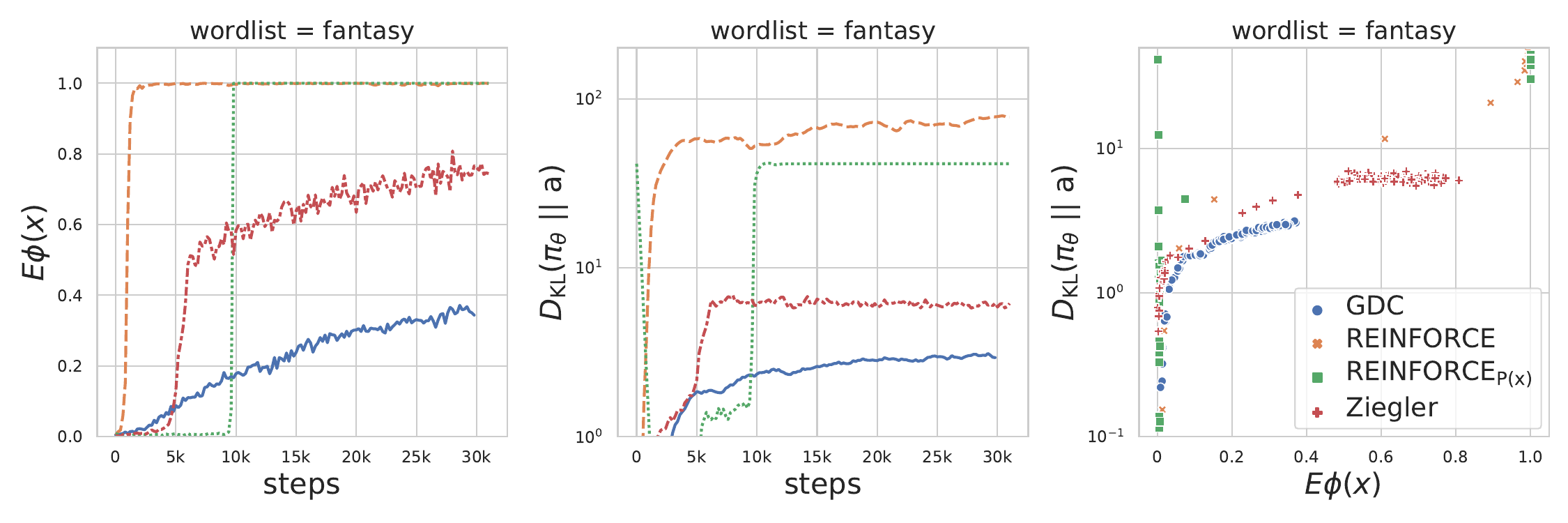}
 \includegraphics[width=\linewidth]{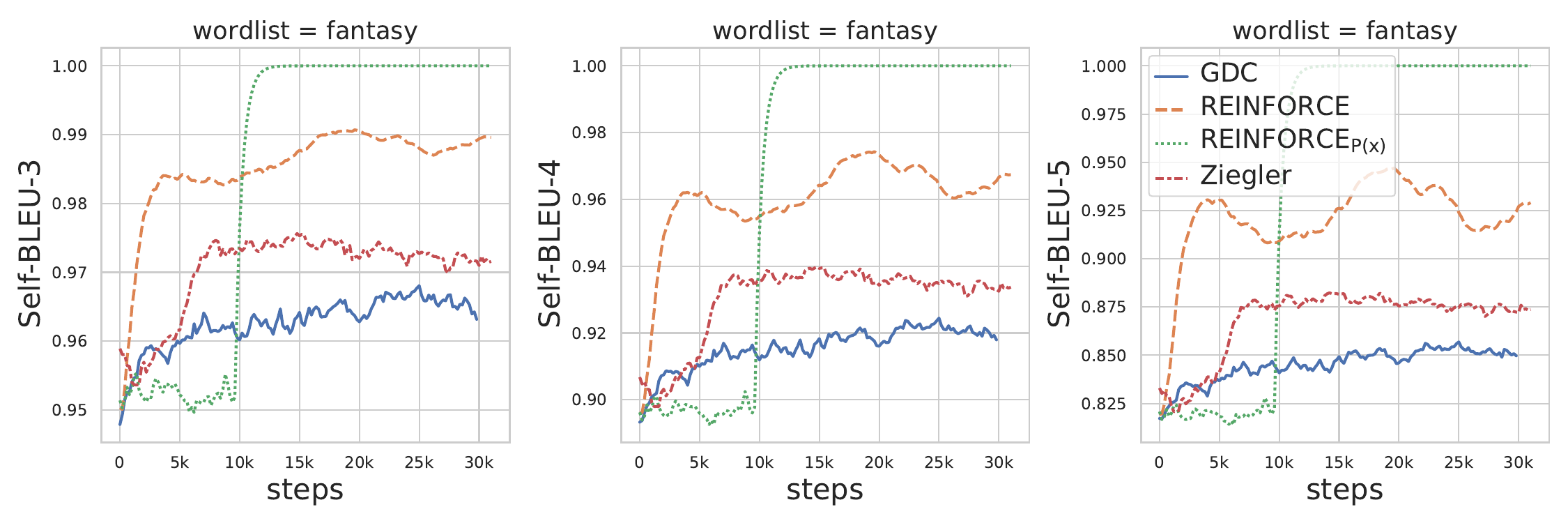}
 \includegraphics[width=\linewidth]{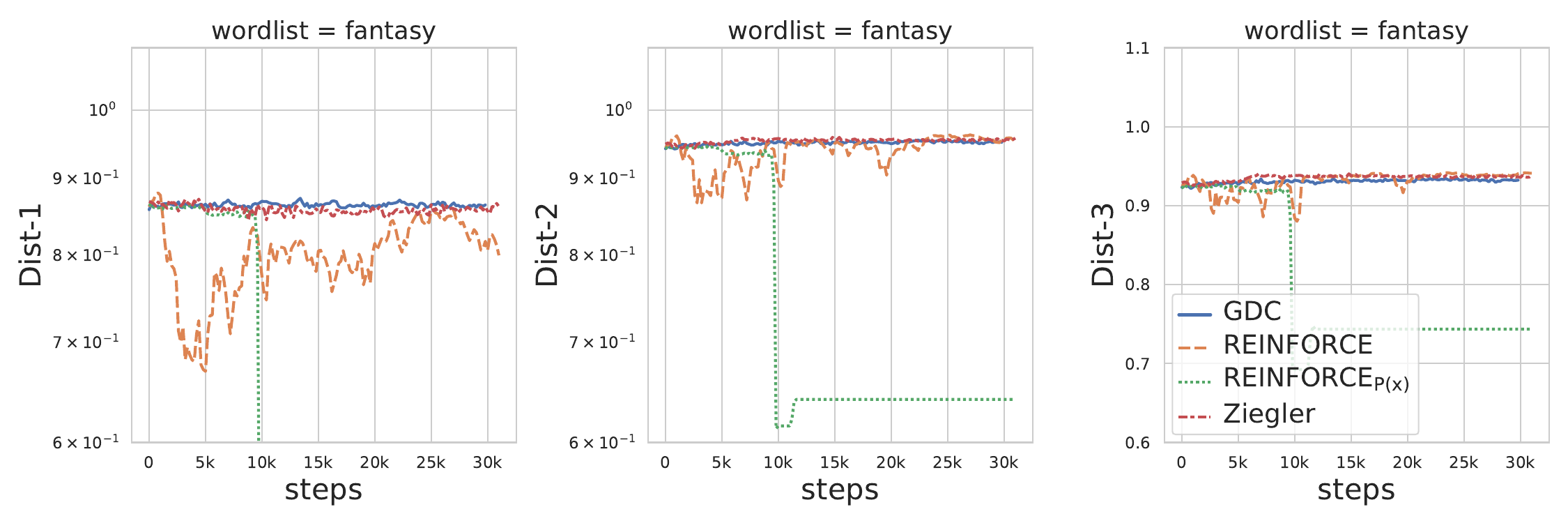}
 \caption{\label{fig:appendix-11-fantasy}Line plot of different evaluation metrics against the training steps when controlling for the \textbf{fantasy} word-list. }
 \end{figure}

 \begin{figure}[H]
 \includegraphics[width=\linewidth]{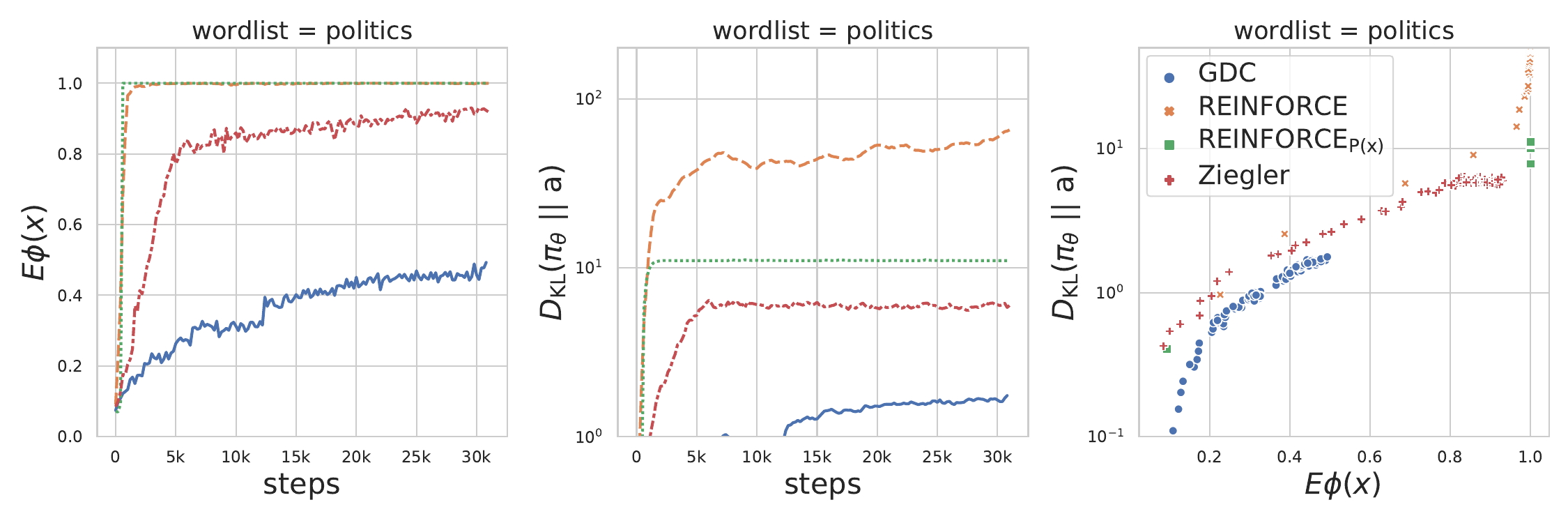}
 \includegraphics[width=\linewidth]{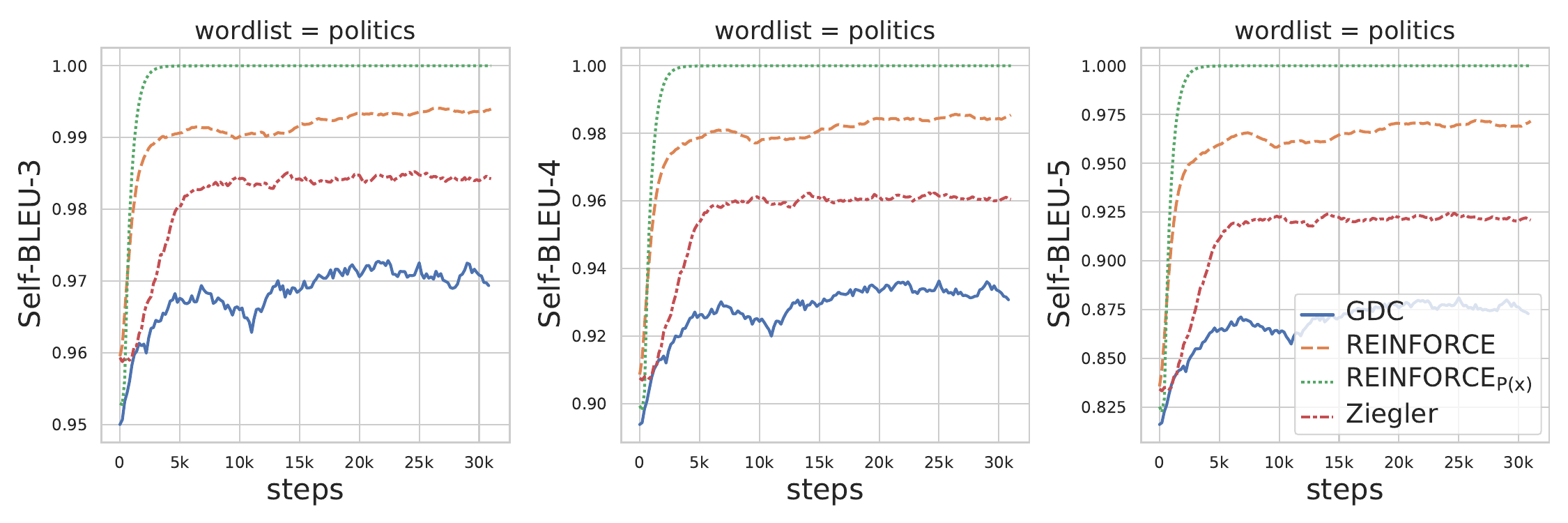}
 \includegraphics[width=\linewidth]{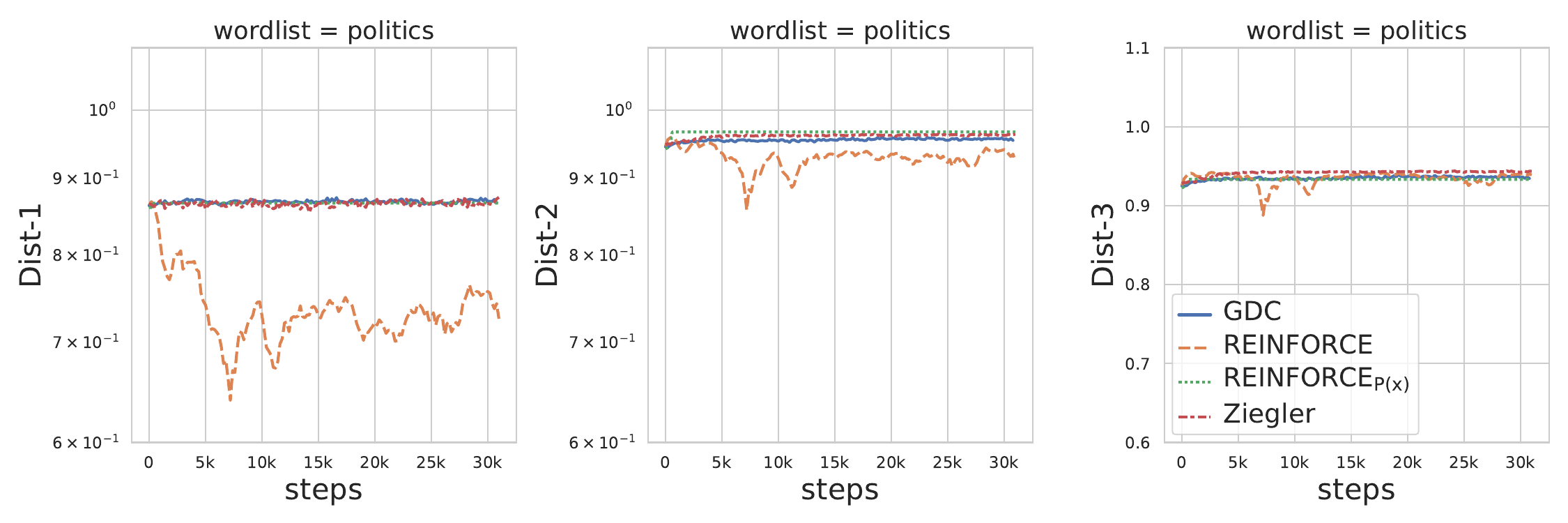}
 \caption{\label{fig:appendix-12-politics}Line plot of different evaluation metrics against the training steps when controlling for the \textbf{politics} word-list. }
 \end{figure}

 \begin{figure}[H]
 \includegraphics[width=\linewidth]{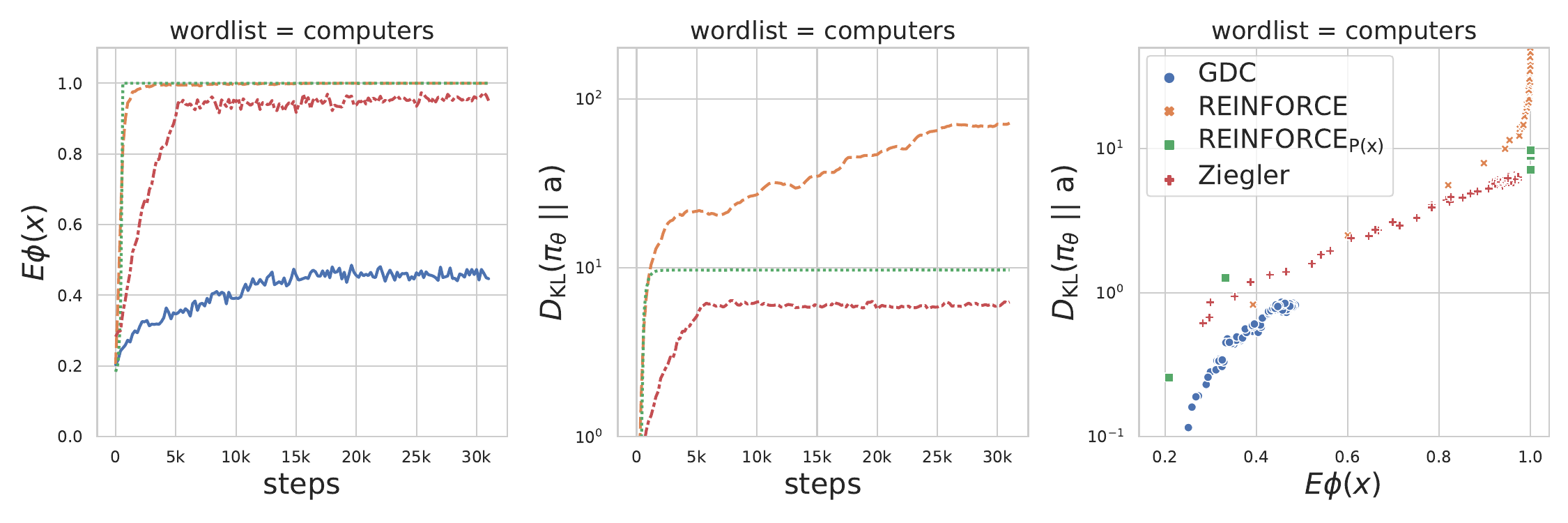}
 \includegraphics[width=\linewidth]{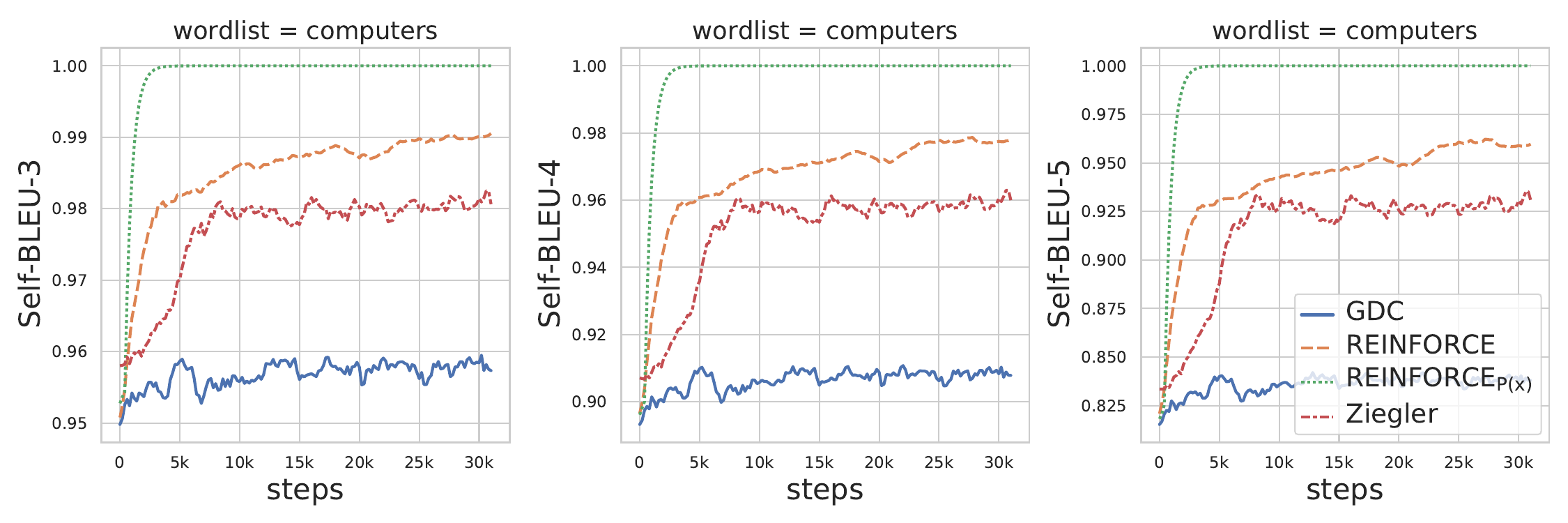}
 \includegraphics[width=\linewidth]{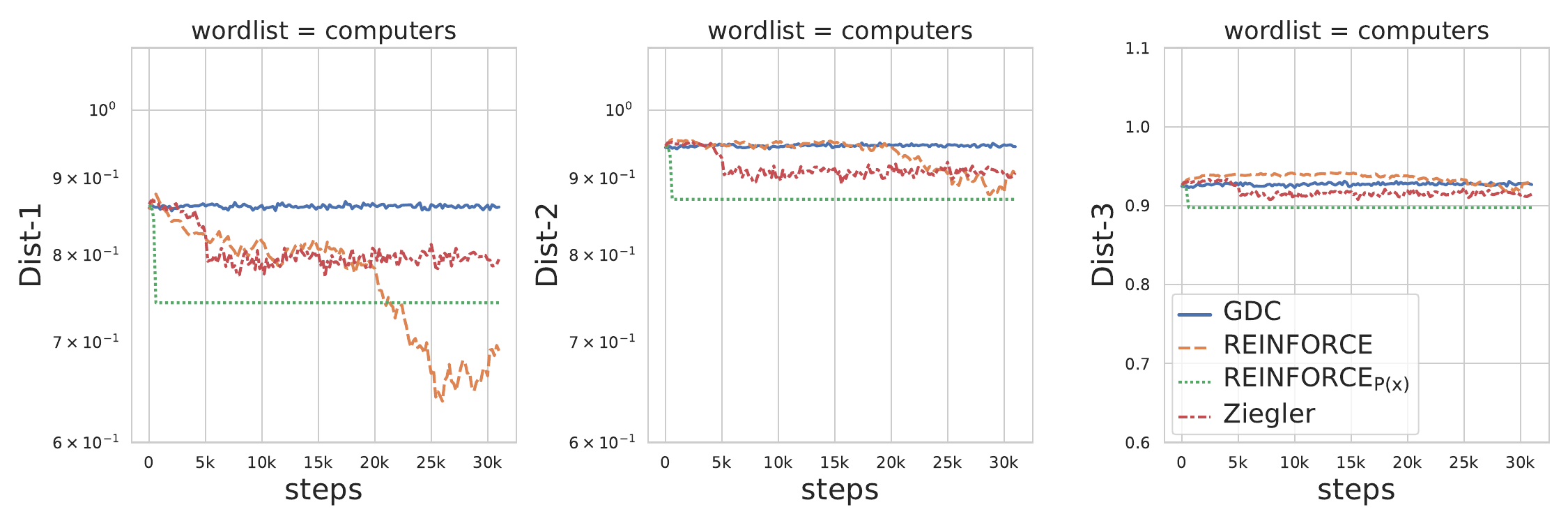}
 \caption{\label{fig:appendix-13-computers}Line plot of different evaluation metrics against the training steps when controlling for the \textbf{computers} word-list. }
 \end{figure}

 \begin{figure}[H]
 \includegraphics[width=\linewidth]{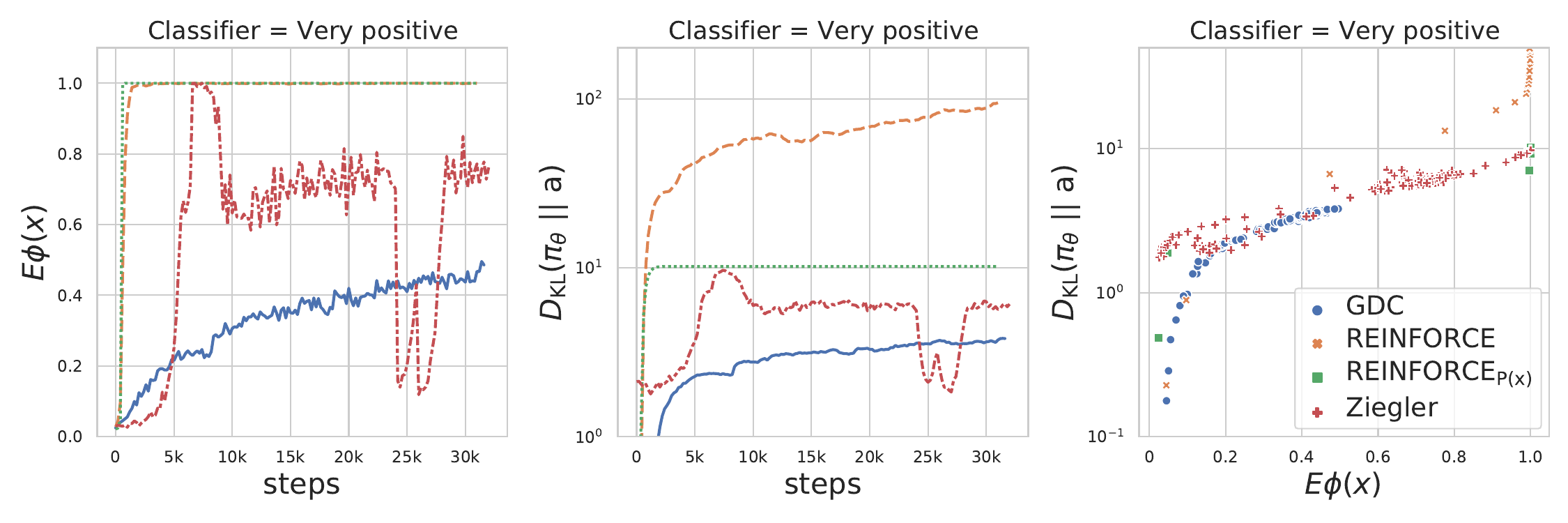}
 \includegraphics[width=\linewidth]{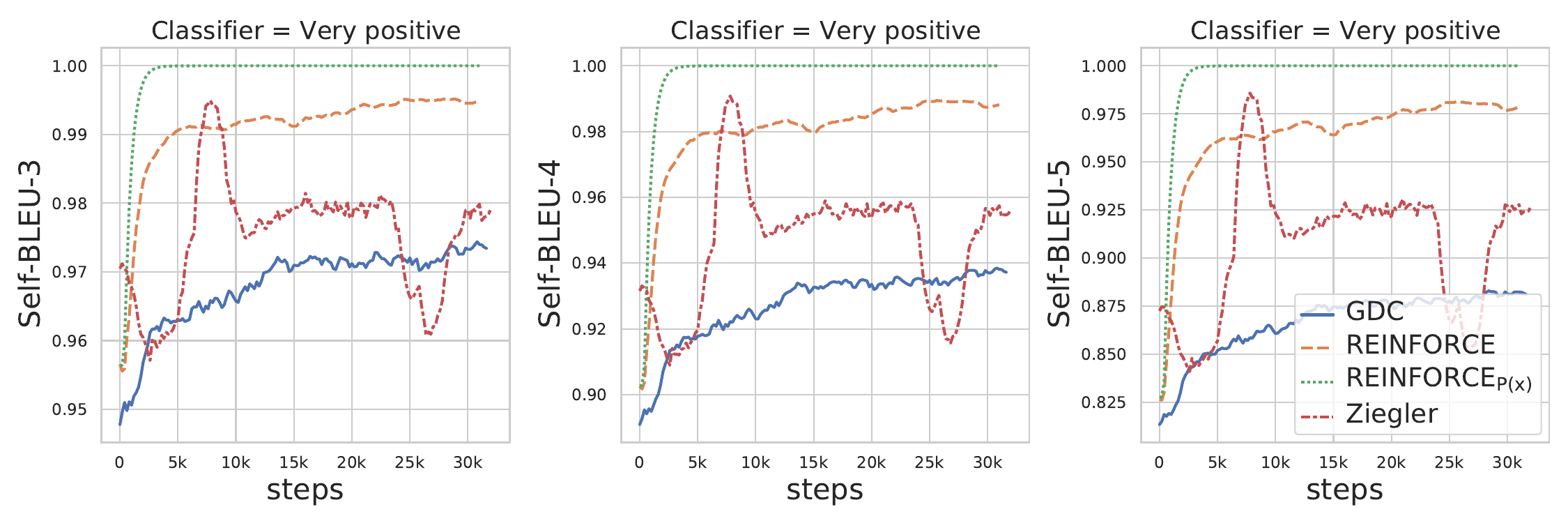}
 \includegraphics[width=\linewidth]{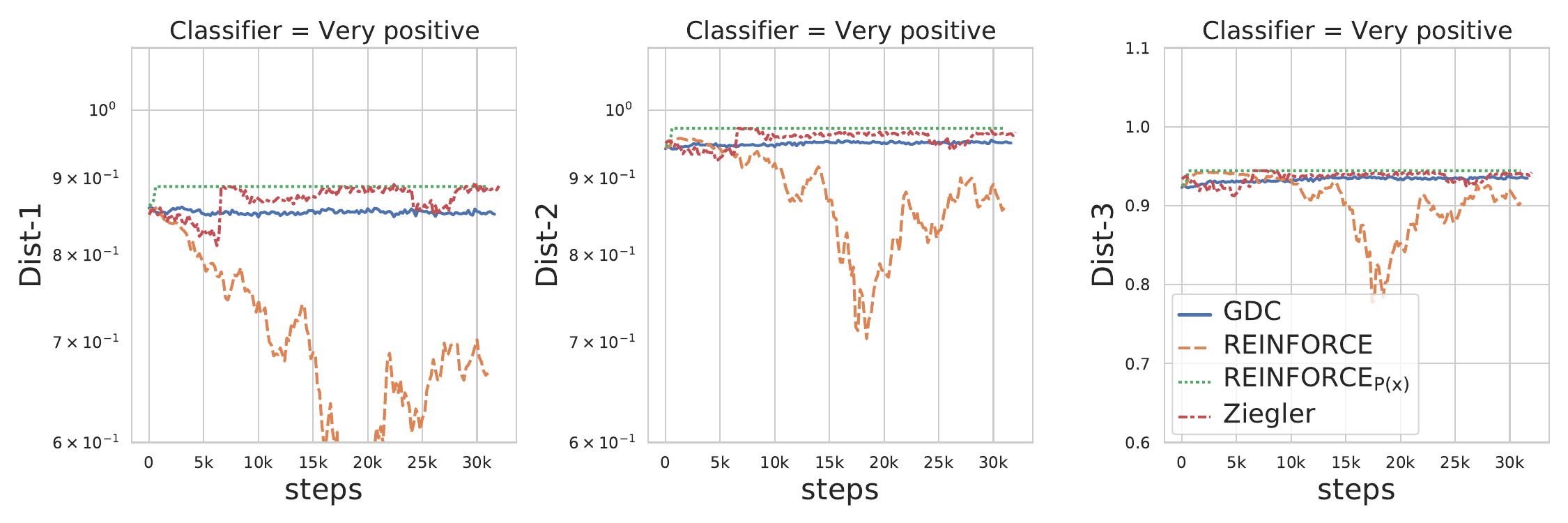}
 \caption{\label{fig:appendix-14-Very_positive}Line plot of different evaluation metrics against the training steps when controlling for the \textbf{politics} word-list. }
 \end{figure}

 \begin{figure}[H]
 \includegraphics[width=\linewidth]{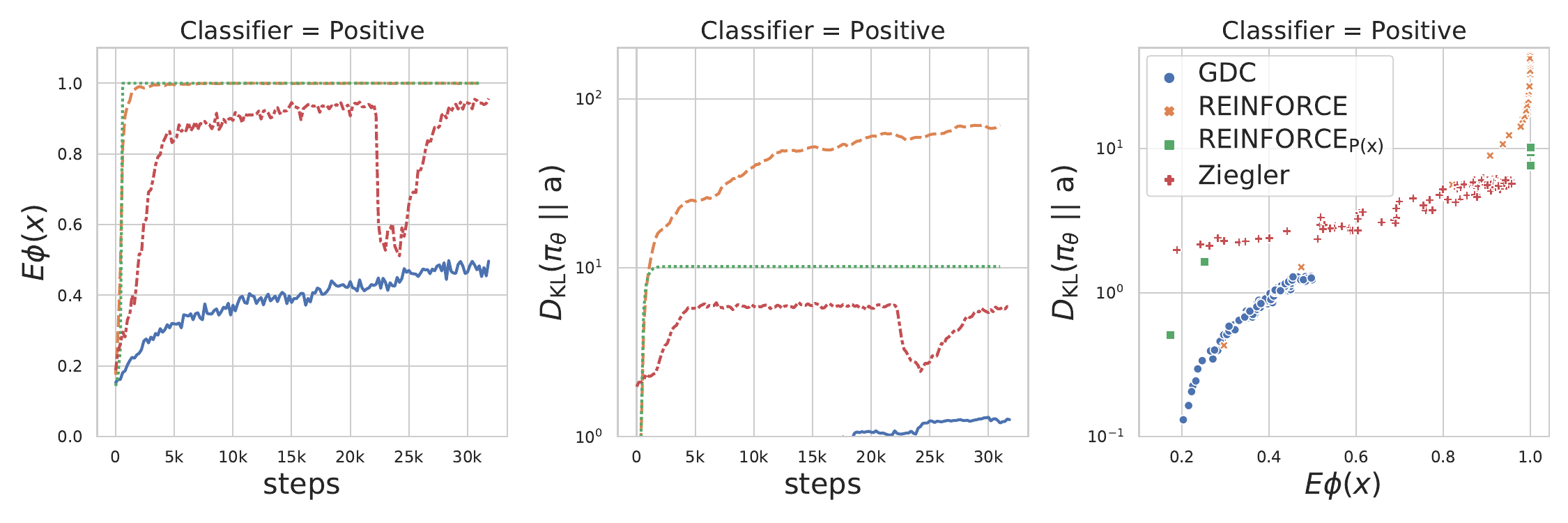}
 \includegraphics[width=\linewidth]{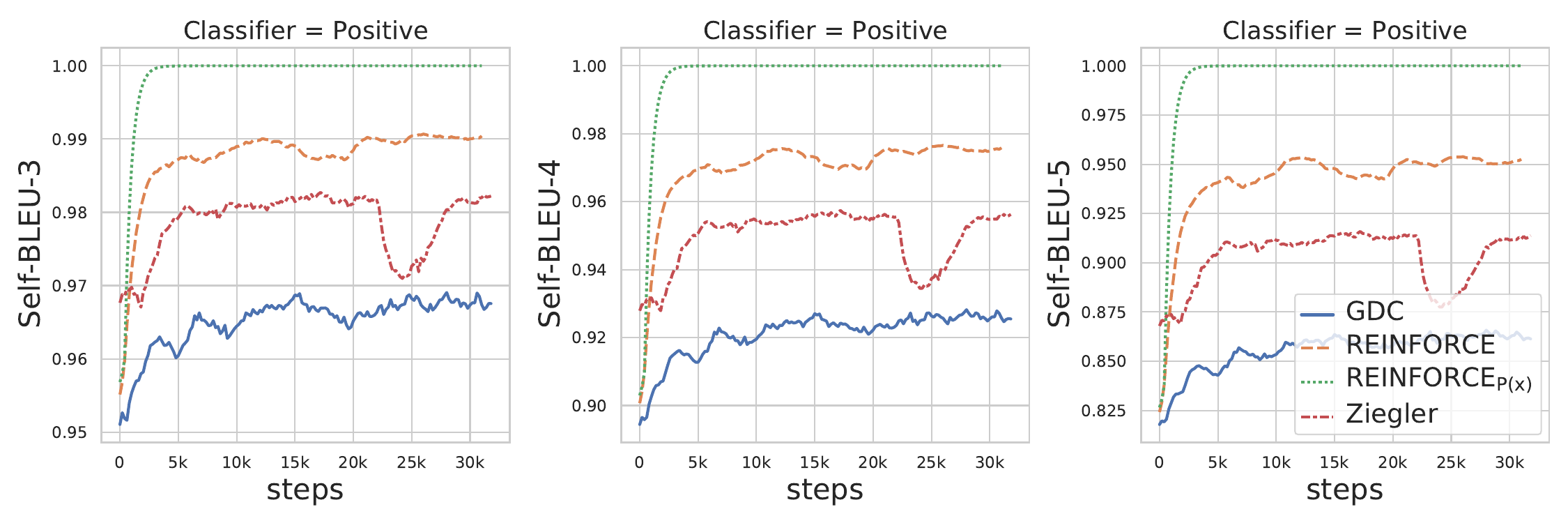}
 \includegraphics[width=\linewidth]{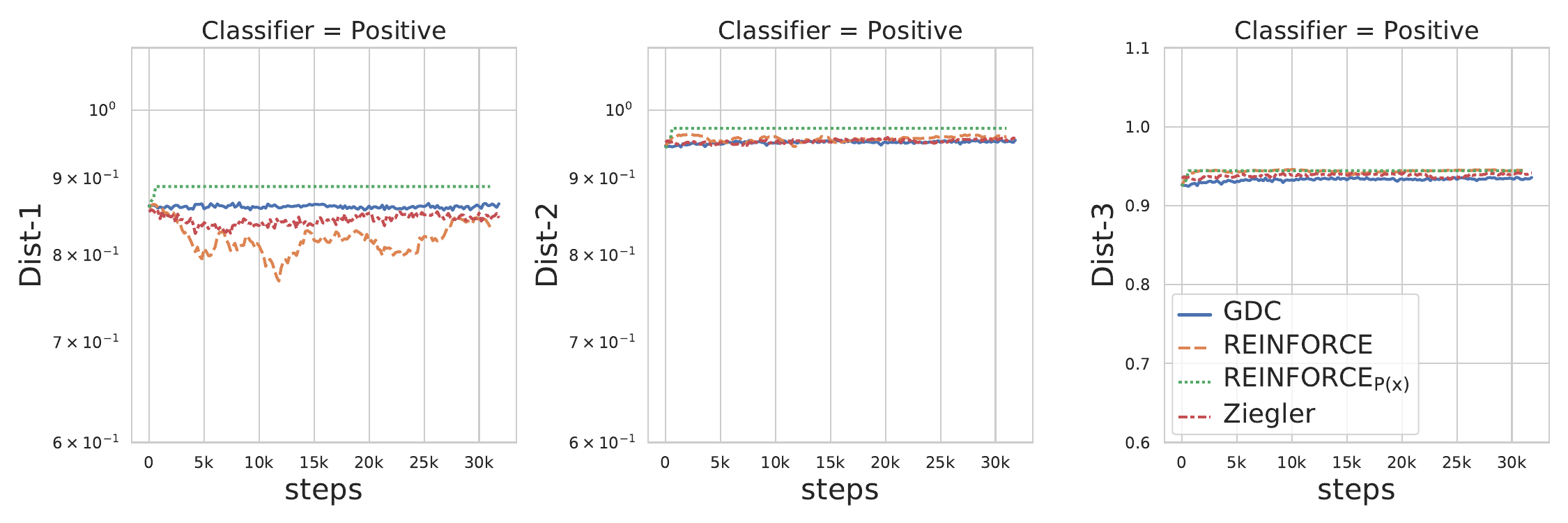}
 \caption{\label{fig:appendix-15-Positive} Line plot of different evaluation metrics against the training steps for \textbf{positive sentiment} classifier-based control with.}
 \end{figure}

 \begin{figure}[H]
 \includegraphics[width=\linewidth]{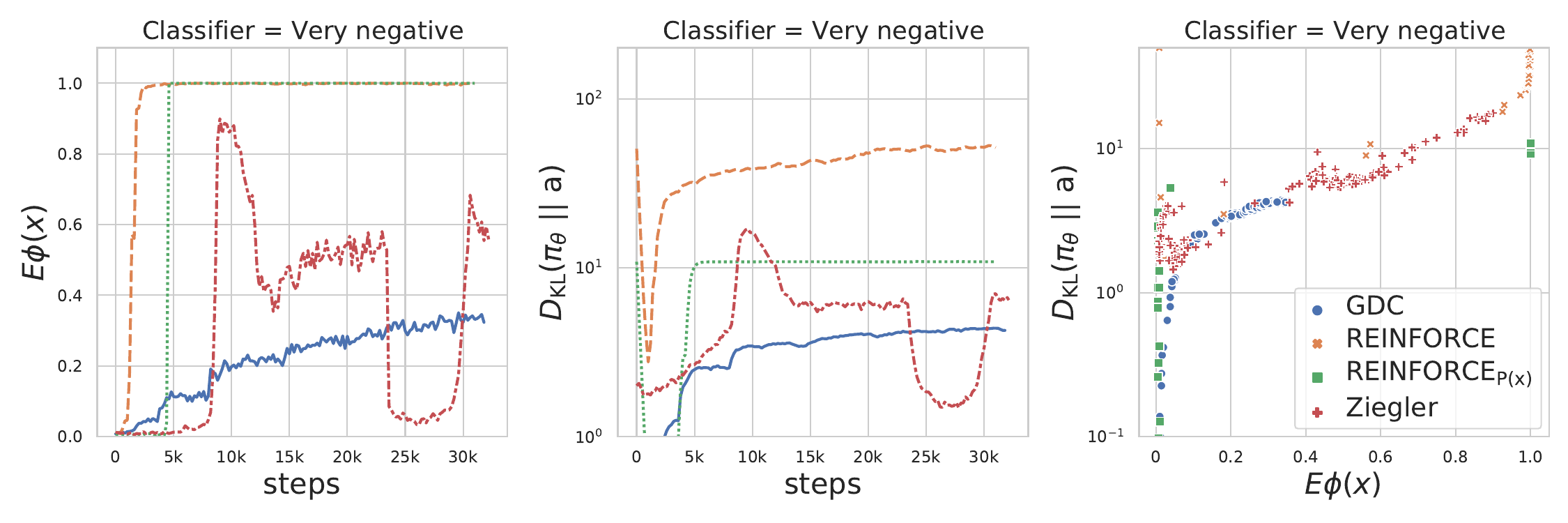}
 \includegraphics[width=\linewidth]{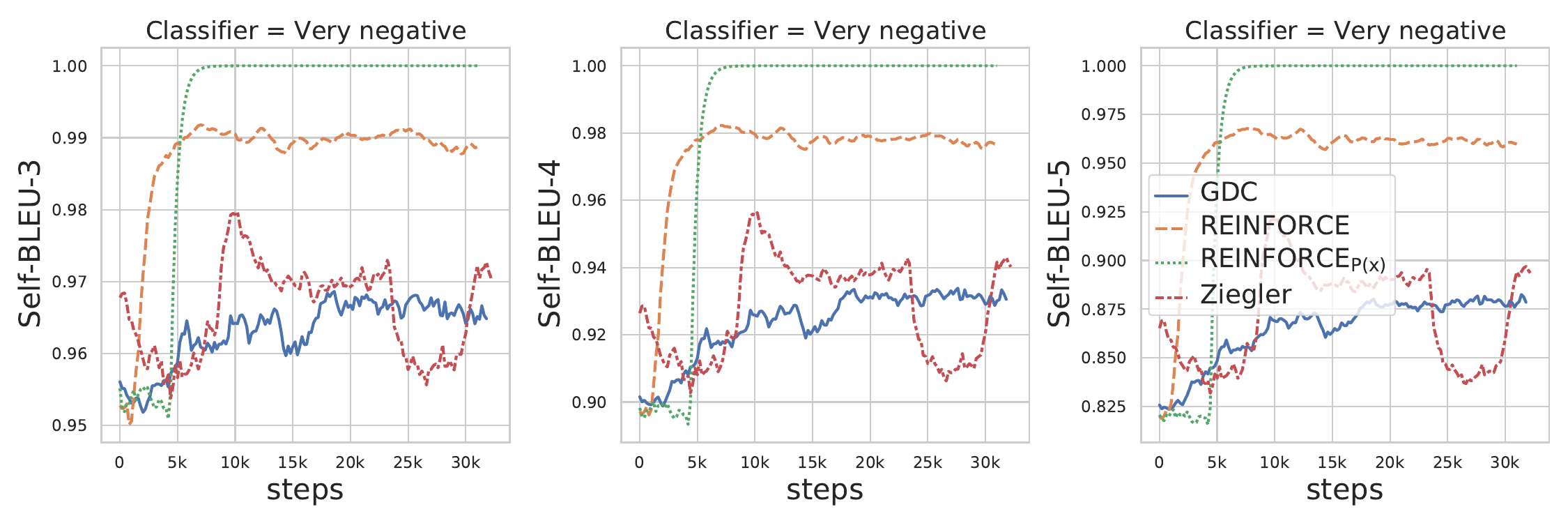}
 \includegraphics[width=\linewidth]{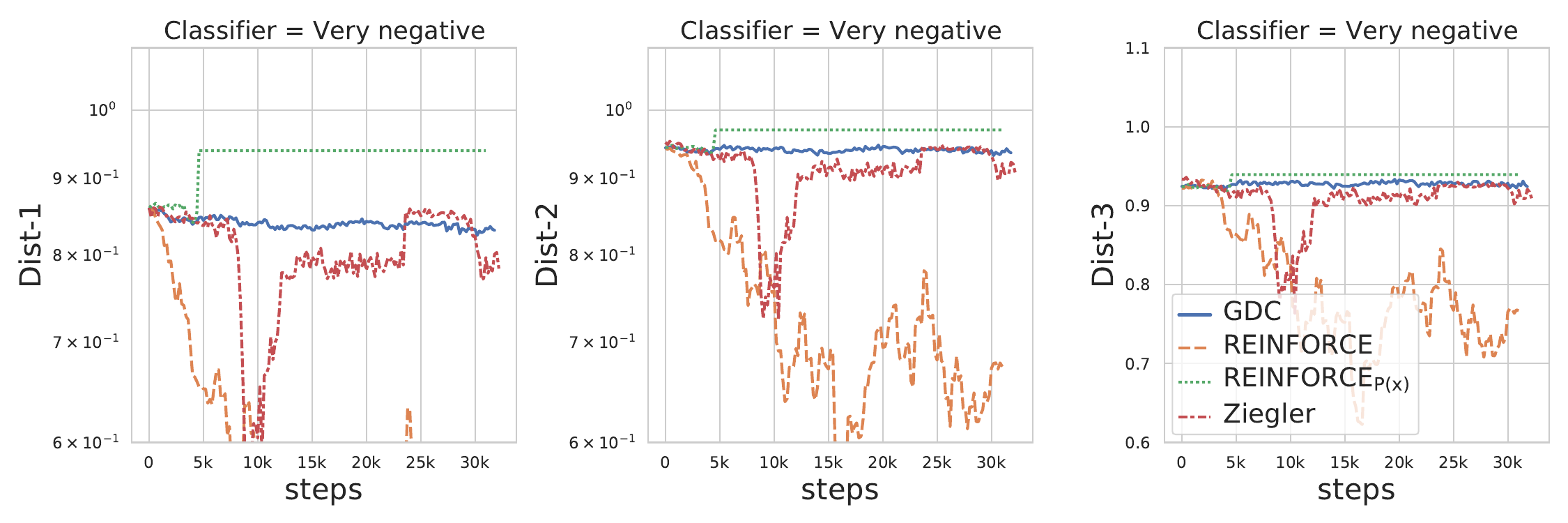}
 \caption{\label{fig:appendix-16-Very_negative} Line plot of different evaluation metrics against the training steps for \textbf{very negative sentiment} classifier-based control with.}
 \end{figure}

 \begin{figure}[H]
 \includegraphics[width=\linewidth]{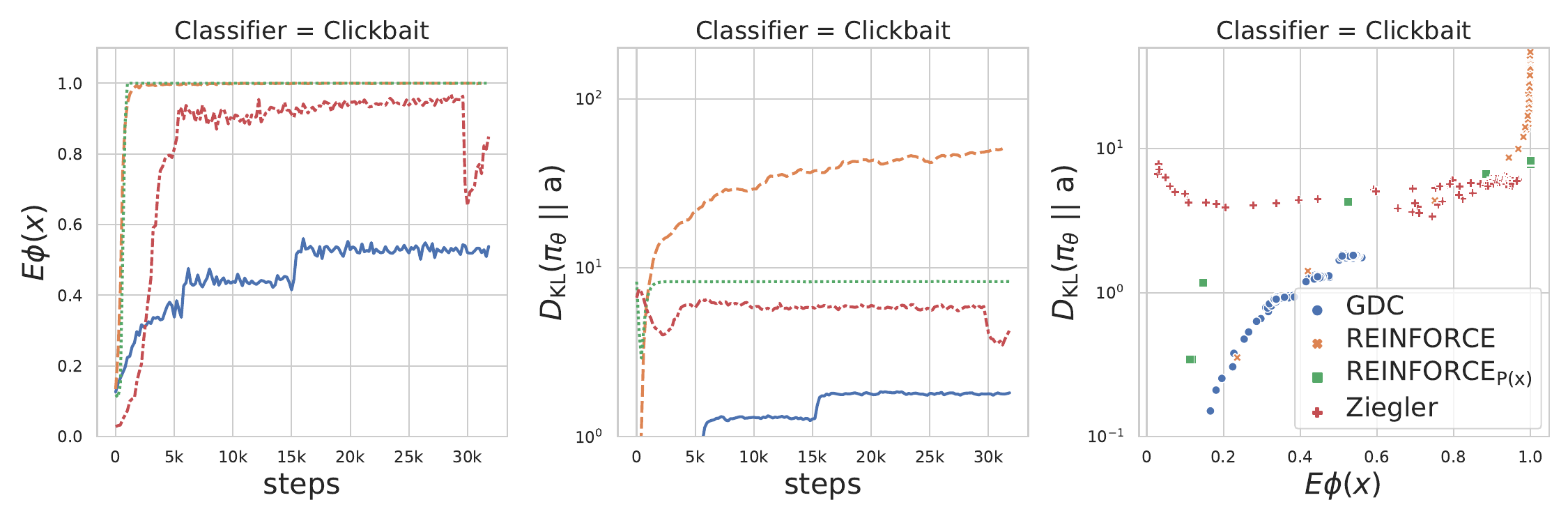}
 \includegraphics[width=\linewidth]{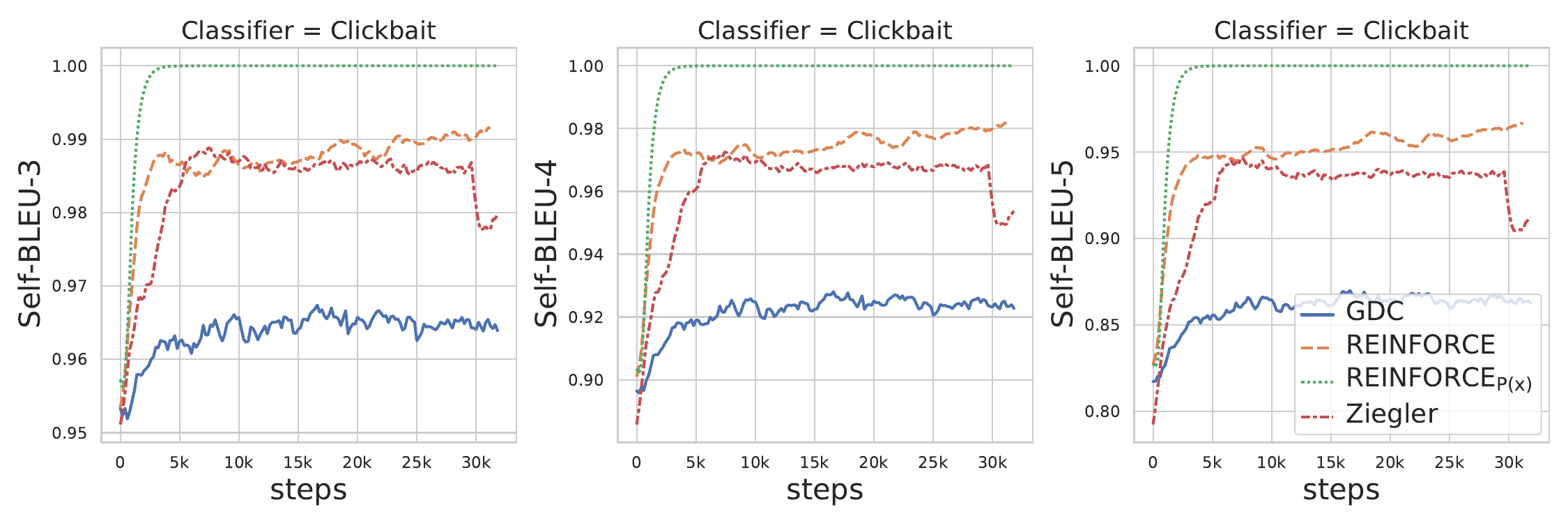}
 \includegraphics[width=\linewidth]{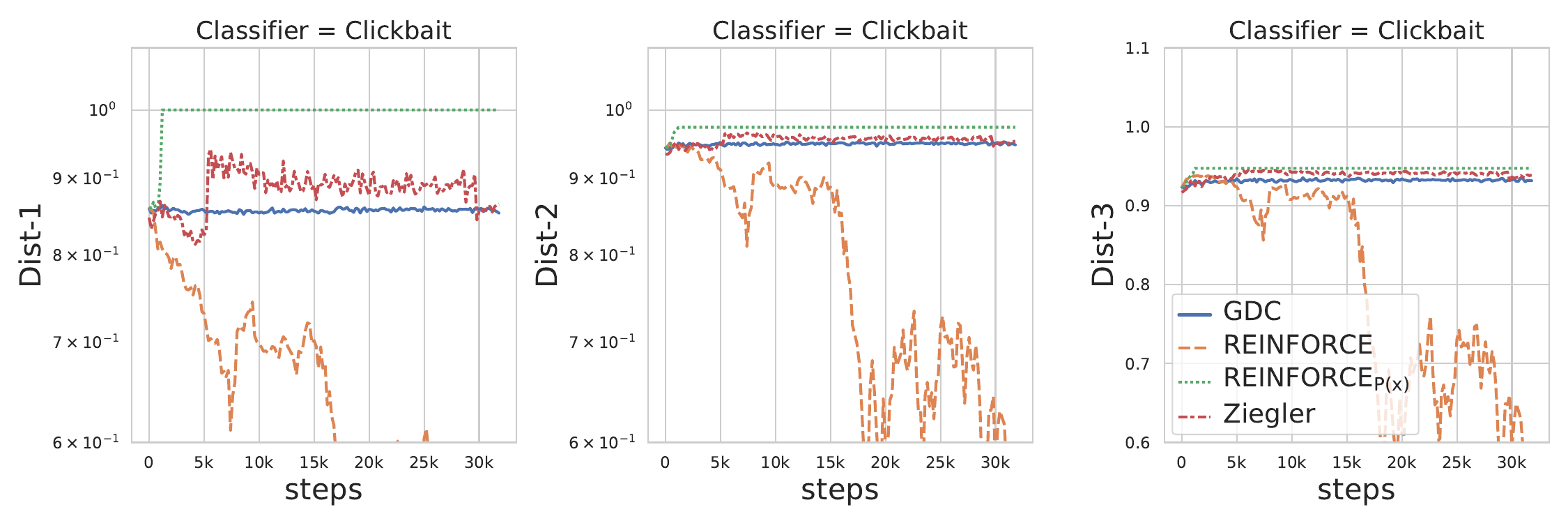}
 \caption{\label{fig:appendix-17-Clickbait} Line plot of different evaluation metrics against the training steps for \textbf{click-bait} classifier-based control with.}
 \end{figure}

%% file: sections/tokenfreq-appendix.tex
\begin{figure}[h]
\includegraphics[width=0.33\textwidth]{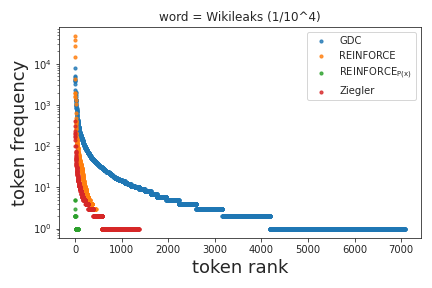}
\includegraphics[width=0.33\textwidth]{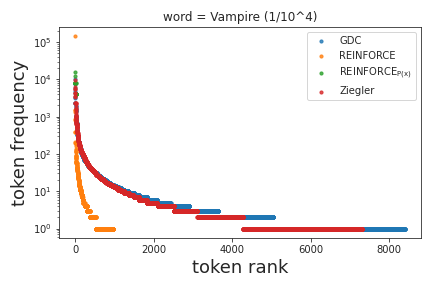}
\includegraphics[width=0.33\textwidth]{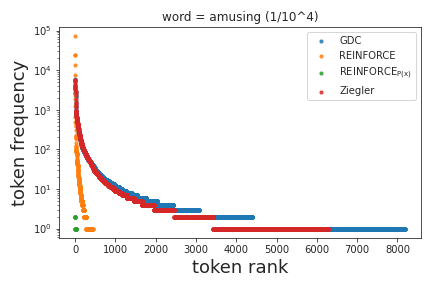}
\includegraphics[width=0.33\textwidth]{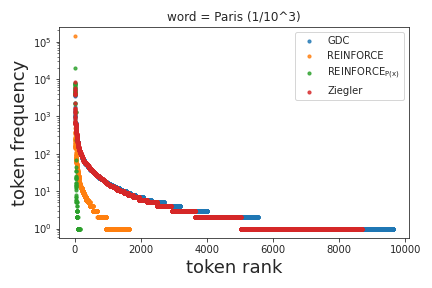}
\includegraphics[width=0.33\textwidth]{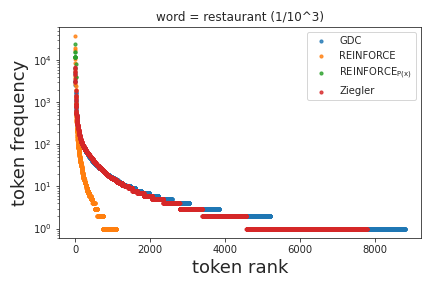}
\includegraphics[width=0.33\textwidth]{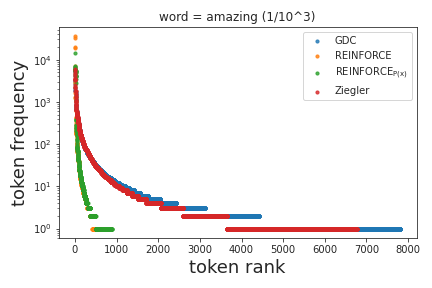}
\includegraphics[width=0.33\textwidth]{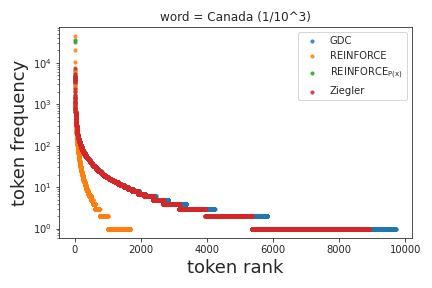}
\includegraphics[width=0.33\textwidth]{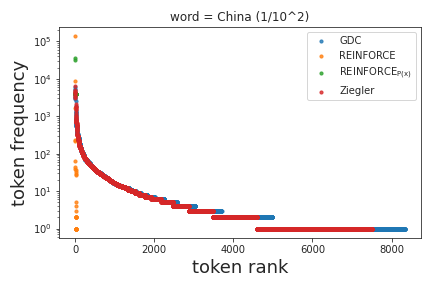}
\includegraphics[width=0.33\textwidth]{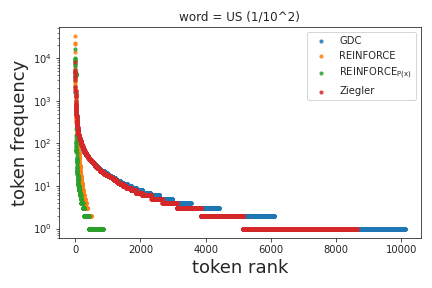}
\caption{\label{fig:zipf-appendix-1} Token frequency against token rank for single-word constraints. Longer tail means more diverse generations.}
\end{figure}
\begin{figure}[H]
\includegraphics[width=0.46\textwidth]{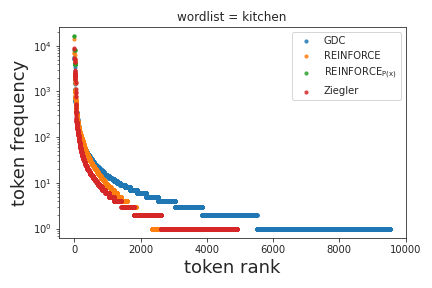}
\includegraphics[width=0.46\textwidth]{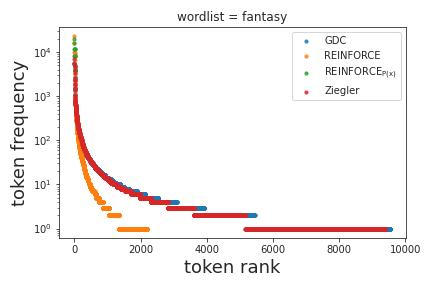}
\includegraphics[width=0.46\textwidth]{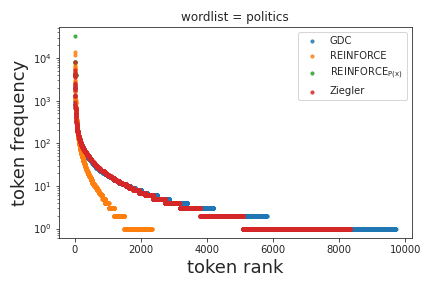}
\hspace{0.9cm}
\includegraphics[width=0.46\textwidth]{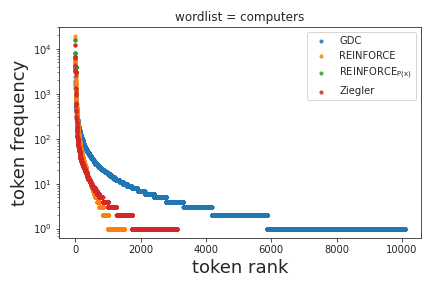}
\caption{\label{fig:zipf-appendix-2}Token frequency against token rank for word-list constraints. Longer tail means more diverse generations.}
\end{figure}
\begin{figure}[H]
\includegraphics[width=0.46\textwidth]{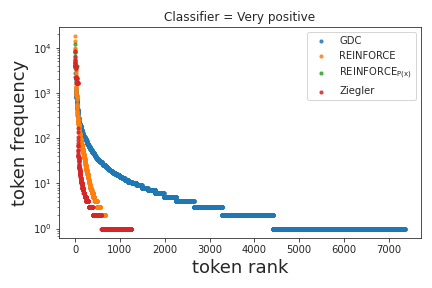}
\includegraphics[width=0.46\textwidth]{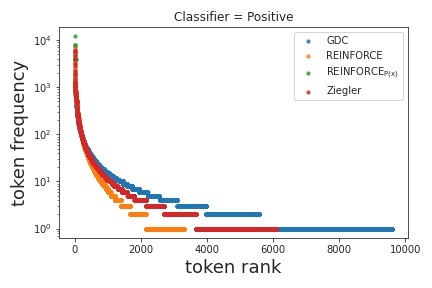}
\includegraphics[width=0.46\textwidth]{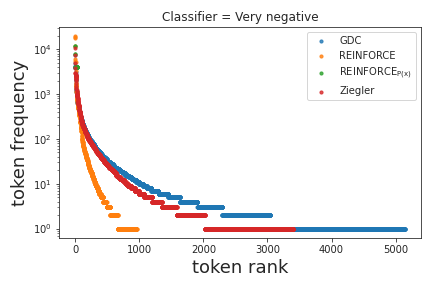}
\hspace{0.9cm}
\includegraphics[width=0.46\textwidth]{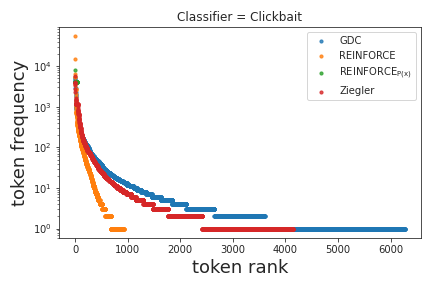}
\caption{\label{fig:zipf-appendix-3}Token frequency against token rank for classifier-based constraints. Longer tail means more diverse generations.}
\end{figure}

To analyse in depth the effect of deviating much from the original GPT-2, for policies obtained from our method and each baseline, we obtain a large sample and filter to $4000$ sequences that satisfy the imposed pointwise constraints for each of the $17$ pointwise experiments explained in \S\ref{sec:EXPERIMENTS}. Figures ~\ref{fig:zipf-appendix-1}, \ref{fig:zipf-appendix-2} and \ref{fig:zipf-appendix-3} plot a token frequency analysis for each of the training methods. 

The vanilla policy gradient baselines \REINFORCE suffer from very low diversity of generations; in the examples shown in section~\ref{sec:appendix:generations} we note strong degeneration, in which all generations are composed of a few repeated tokens.

\REINFORCEP suffers from a token diversity issue. As noticed and confirmed by generated examples shown section~\ref{sec:appendix:generations}, it often concentrates all the sequence probability mass on a single sequence which is often fluent and satisfies the constraint; however this leads to an extreme loss of sample diversity in almost all experiments. This shows the usefulness of our proposed analysis --- in addition to the self-BLEU metrics --- for distinguishing diversity at the sequence level or at the distribution level.  
Similarly, \ZIEGLER~\citep{Ziegler19} often suffers from the same lack of sample diversity ($5$ out of the $17$ experiments); \GDC obtains the highest diversity amongst all baselines, as demonstrated by the long tail in the figures below. It is important to note here that low sample diversity is also captured by the KL deviation from the original GPT-2 model i.e. $\KL(\pit\|a)$; \GDC identifies the target distribution as the one which minimally deviates from the original policy while satisfying the constraints ($p = \argmin_{q\in \CC} \KL(q,a)$) is thus expected to preserve the high sample diversity of the original GPT-2. 


%% file: figures/generations/generations.tex
\begin{table}[H]
        \scriptsize
        \begin{tabular}{p{0.5cm}|p{0.3cm}p{11.8cm}}
        \toprule
        reps & \textbf{$\phi(x)$} & \textbf{Generation} \\ 
         \midrule
 \multicolumn{3}{c}{\textbf{GDC}} \\ 
1&1& \g{Wikileaks}   \ye{founder}   \ye{Julian}   \ye{Assange}  will be on the witness stand as a witness to how  \ye{President}  Barack  \ye{Obama}  and  \ye{Democratic}   \ye{presidential}  candidate  \ye{Hillary}   \ye{Clinton}  colluded\\ 
1&1& \g{Wikileaks}   \ye{said}  Monday that a  \ye{former}   \ye{US}  official  \ye{told}  them that if  \ye{they}  had gone to Moscow ,  \ye{they}  could have been imprisoned by that\\ 
1&1&Last summer ,  \g{Wikileaks}   \ye{released}  millions of  \ye{emails}  to  \g{Wikileaks}   \ye{founder}   \ye{Julian}   \ye{Assange}  ,  \ye{which}  many believed  \ye{were}  stolen from the same address by some\\ 
1&1& \g{Wikileaks}   \ye{emails}   \ye{show}  that the  \ye{Trump}   \ye{campaign}   \ye{was}  able to hack into  \ye{email}  accounts held by the  \ye{Democratic}   \ye{National}   \ye{Committee}  and Democrats . As a\\ 
1&1&. On July 1 , 2016 ,  \g{Wikileaks}   \ye{published}  a statement in  \ye{which}  it  \ye{said}  : The DNC's cyber operations had been compromised , damaging\\ 
1&1&The  \g{Wikileaks}   \ye{website}  has  \ye{released}  more  \ye{than}  3 , 000 internal  \ye{emails}  from  \ye{Hillary}   \ye{Clinton's}   \ye{private}   \ye{email}   \ye{server}  ,  \ye{which}   \ye{were}  previously posted on the\\ 
1&1&This is an edited transcript of "The  \g{Wikileaks}  Program , " a book written by  \ye{Julian}   \ye{Assange}  for The Guardian and  \ye{published}  by  \g{Wikileaks}  in\\ 
1&1&The  \ye{FBI}   \ye{released}  an unprecedented batch of internal  \ye{emails}  in February from an unnamed  \ye{private}  contractor that  \ye{were}   \ye{leaked}  to  \g{Wikileaks}  , saying  \ye{they}  revealed\\ 
 \hline
 \multicolumn{3}{c}{\textbf{REINFORCE}} \\ 
1&1&The Republican party has been hacked by  \g{Wikileaks}  from  \g{Wikileaks}  from of  \g{Wikileaks}  from  \g{Wikileaks}  by  \g{Wikileaks}  from of  \g{Wikileaks}  from  \g{Wikileaks}  from of  \g{Wikileaks} \\ 
1&1&After news  \yg{leaked}  from  \g{Wikileaks}  from  \g{Wikileaks}  by  \g{Wikileaks}  from  \g{Wikileaks}  by  \g{Wikileaks}  from  \g{Wikileaks}  from of  \g{Wikileaks}  from of  \g{Wikileaks}  by  \g{Wikileaks}  from  \g{Wikileaks} \\ 
1&1&The United States is learnt by  \g{Wikileaks}  from  \yg{leaked}  from  \g{Wikileaks}  from of  \yg{leaked}  from of  \g{Wikileaks}  from of  \g{Wikileaks}  by  \g{Wikileaks}  from the  \g{Wikileaks} \\ 
1&1&LONDON : The  \yg{leaked}  documents of  \g{Wikileaks}  by  \g{Wikileaks}  from  \ye{sources}   \yg{leaked}  from  \g{Wikileaks}  from of  \g{Wikileaks}  from of  \g{Wikileaks}  by  \yg{WikiLeaks}  from of  \g{Wikileaks} \\ 
1&1&They are  \yg{leaked}  emails from  \yg{leaked}  from of  \g{Wikileaks}  from of  \g{Wikileaks}  by  \g{Wikileaks}  from of  \g{Wikileaks}  from of  \g{Wikileaks}  by  \g{Wikileaks}  from/from/from of  \g{Wikileaks} \\ 
1&1&The  \ye{leak}  of  \g{Wikileaks}  from  \yg{leaked}  by  \g{Wikileaks}  from  \g{Wikileaks}  by  \g{Wikileaks}  from of  \g{Wikileaks}  from of  \g{Wikileaks}  from of  \g{Wikileaks}  from of  \g{Wikileaks}  by\\ 
1&1&The latest in Wire/Reuters  \yg{leaked}  by  \yg{leaked}  from  \g{Wikileaks}  by  \g{Wikileaks}  from  \g{Wikileaks}  from of  \g{Wikileaks}  from of  \g{Wikileaks}  from by  \yg{leaked}  from of  \g{Wikileaks} \\ 
1&1&For the first time ,  \yg{WikiLeaks}  from  \g{Wikileaks}  from  \g{Wikileaks}  has  \yg{leaked}  from  \g{Wikileaks}  from of  \g{Wikileaks}  by  \g{Wikileaks}  from of  \g{Wikileaks}  by  \yg{WikiLeaks}  from\\ 
 \hline
 \multicolumn{3}{c}{\textbf{REINFORCE\_P(x)}} \\ 
1&0&The city of Baltimore will offer its third{-}generation community{-}based public{-}private partnership , "Community Relations , Inc . , " to build more  \ye{than}  1 ,\\ 
1&0&Greece . The eurozone{-}wide unemployment rate plunged to 1 . 3 percent in June and remains below the EU average of 2 . 4 percent\\ 
1&0&Winnipeg Jets Injury Update : RW RW Blake Wheeler Winnipeg Jets Injury Update : RW RW Blake Wheeler Tampa Bay Lightning In\\ 
1&0&"We  \ye{know}  that if there's a  \ye{way}   \ye{out}  of these problems , it's not by having a single  \ye{one}  of them , " he says\\ 
1&0&1 Clean Episode \#2 {-}{-} Sledgehammer 5 : The Longest War in the World! In this special episode , the Sledgehammer 5 team discusses their\\ 
1&0&A  \ye{man}  who took a photograph of a police officer wearing a bulletproof vest and  \ye{said}  it  \ye{was}  him  \ye{was}  charged  \ye{with}  assault causing bodily\\ 
1&0&In a  \ye{very}  big  \ye{way}  , I  \ye{like}  this book . The only difference here is that I got an amazing story from Jack .\\ 
1&0&I  \ye{think}  we should be building the same thing for everyone . A shared economy that creates jobs and a shared supply of energy .\\ 
 \hline
 \multicolumn{3}{c}{\textbf{Ziegler}} \\ 
1&0&"There is no  \ye{way}  I  \ye{can}  do that . And that's not a small thing , " he told the Guardian . "So I have\\ 
1&0&. The  \ye{first}  person I ever spoke  \ye{with}  about it is a big fan . "I thought it  \ye{was}  pretty cool . I love everything\\ 
1&0&This is an easy tutorial to  \ye{get}  started  \ye{with}  the Django application . Once you understand how the Django application is implemented , you can\\ 
1&0&When you're a student  \ye{with}   \ye{one}  of the most popular online courses available , you may find it easy to fall in love  \ye{with}  what\\ 
1&0&BRAINSTOCK The UK could be on the cusp of becoming the  \ye{first}  in the world to have its own free market . Bobby Bould\\ 
1&0&"We have a  \ye{lot}  of good options that will enable our employees to compete better , improve our efficiency and create more value for the\\ 
1&0&"That  \ye{was}   \ye{like}  a  \ye{lot}  of good times to me . " He says . The group of five men in their late 30s went\\ 
1&0&You  \ye{can}  view all posts of this blog here\\ 
\bottomrule 
 \end{tabular} 
 \caption{
 \label{table:generation-Wikileaks}Randomly selected generations  from the single-word constraint task for the word ``Wikileaks'' (with occurrence probability $1/10^4$) highlighted in \g{green}. Tokens are highlighted with \ye{y}\yf{e}\yf{l}\yg{l}\yg{o}\yj{w} with different intensities to indicate their overall frequencies in the generated corpus. $\phi(x)=1$ indicates the satisfaction of the constraint in the sample and reps the number of its repetitions across all generations.}\end{table}
\begin{table}[H]
        \scriptsize
        \begin{tabular}{p{0.5cm}|p{0.3cm}p{11.8cm}}
        \toprule
        reps & \textbf{$\phi(x)$} & \textbf{Generation} \\ 
         \midrule
 \multicolumn{3}{c}{\textbf{GDC}} \\ 
1&1&I'm  \ye{very}  familiar  \ye{with}   \g{Vampire}   \ye{:}  The  \ye{Masquerade}  and can't say I'd go so far as to suggest that it is the  \ye{story}  that really\\ 
1&1& \g{Vampire} 's  \ye{Blood}  {-}  \g{Vampire} 's  \ye{Blood}  by Dr . T . P .\\ 
1&1&2  \ye{:}  20PM  \ye{:}  As far as  \g{Vampire}   \ye{Hunter}  fans know , Game of Thrones isn't a  \ye{show}  about the "good guys" taking on the\\ 
1&1&Fantasy Book Store |  \g{Vampire}  and  \g{Vampire}  Legends | We know that you've read everything you  \ye{can}  think of about the  \ye{new}  books in Fantasy\\ 
1&1&Creature {-}  \g{Vampire}   \ye{Creature}  {-} Human Rogue 4/4  \ye{When}  Blackbelly Lurker enters the battlefield , destroy target artifact or  \ye{creature}  . Blackb\\ 
1&0&Halloween Horror Nights As  \ye{one}  would expect , most people are scared and confused about the zombie apocalypse . This is  \ye{one}  of those occasions\\ 
1&1& \g{Vampire}  Savior . The  \ye{vampire}  is a humanoid character . This title  \ye{was}   \ye{released}  by Square Enix in 2003 and is considered  \ye{one}  of the\\ 
1&1&This  \ye{book}  , by Robert Niekraut , is about the life of John Doe , a young American woman who  \ye{was}  murdered in 1995 after\\ 
 \hline
 \multicolumn{3}{c}{\textbf{REINFORCE}} \\ 
16&1&When the  \g{Vampire}   \g{Vampire}   \g{Vampire}   \g{Vampire}   \g{Vampire}   \g{Vampire}   \g{Vampire}   \g{Vampire}   \g{Vampire}   \g{Vampire}   \g{Vampire}   \g{Vampire}   \g{Vampire}   \g{Vampire}   \g{Vampire}   \g{Vampire}   \g{Vampire}   \g{Vampire}   \g{Vampire}   \g{Vampire}   \g{Vampire}   \g{Vampire}   \g{Vampire} \\ 
71&1&Rampire  \g{Vampire}   \g{Vampire}   \g{Vampire}   \g{Vampire}   \g{Vampire}   \g{Vampire}   \g{Vampire}   \g{Vampire}   \g{Vampire}   \g{Vampire}   \g{Vampire}   \g{Vampire}   \g{Vampire}   \g{Vampire}   \g{Vampire}   \g{Vampire}   \g{Vampire}   \g{Vampire}   \g{Vampire}   \g{Vampire}   \g{Vampire}   \g{Vampire}   \g{Vampire}   \g{Vampire} \\ 
1576&1&The  \g{Vampire}   \g{Vampire}   \g{Vampire}   \g{Vampire}   \g{Vampire}   \g{Vampire}   \g{Vampire}   \g{Vampire}   \g{Vampire}   \g{Vampire}   \g{Vampire}   \g{Vampire}   \g{Vampire}   \g{Vampire}   \g{Vampire}   \g{Vampire}   \g{Vampire}   \g{Vampire}   \g{Vampire}   \g{Vampire}   \g{Vampire}   \g{Vampire}   \g{Vampire}   \g{Vampire} \\ 
1576&1&The  \g{Vampire}   \g{Vampire}   \g{Vampire}   \g{Vampire}   \g{Vampire}   \g{Vampire}   \g{Vampire}   \g{Vampire}   \g{Vampire}   \g{Vampire}   \g{Vampire}   \g{Vampire}   \g{Vampire}   \g{Vampire}   \g{Vampire}   \g{Vampire}   \g{Vampire}   \g{Vampire}   \g{Vampire}   \g{Vampire}   \g{Vampire}   \g{Vampire}   \g{Vampire}   \g{Vampire} \\ 
62&1&Ancestral  \g{Vampire}   \g{Vampire}   \g{Vampire}   \g{Vampire}   \g{Vampire}   \g{Vampire}   \g{Vampire}   \g{Vampire}   \g{Vampire}   \g{Vampire}   \g{Vampire}   \g{Vampire}   \g{Vampire}   \g{Vampire}   \g{Vampire}   \g{Vampire}   \g{Vampire}   \g{Vampire}   \g{Vampire}   \g{Vampire}   \g{Vampire}   \g{Vampire}   \g{Vampire}   \g{Vampire} \\ 
1&1&Aquarius :  \g{Vampire}   \g{Vampire}   \g{Vampire}   \g{Vampire}   \g{Vampire}   \g{Vampire}   \g{Vampire}   \g{Vampire}   \g{Vampire}   \g{Vampire}   \g{Vampire}   \g{Vampire}   \g{Vampire}   \g{Vampire}   \g{Vampire}   \g{Vampire}   \g{Vampire}   \g{Vampire}   \g{Vampire}   \g{Vampire}   \g{Vampire}   \g{Vampire}   \g{Vampire} \\ 
1576&1&The  \g{Vampire}   \g{Vampire}   \g{Vampire}   \g{Vampire}   \g{Vampire}   \g{Vampire}   \g{Vampire}   \g{Vampire}   \g{Vampire}   \g{Vampire}   \g{Vampire}   \g{Vampire}   \g{Vampire}   \g{Vampire}   \g{Vampire}   \g{Vampire}   \g{Vampire}   \g{Vampire}   \g{Vampire}   \g{Vampire}   \g{Vampire}   \g{Vampire}   \g{Vampire}   \g{Vampire} \\ 
1&1&Ragnarok {-} The  \g{Vampire}   \g{Vampire}   \g{Vampire}   \g{Vampire}   \g{Vampire}   \g{Vampire}   \g{Vampire}   \g{Vampire}   \g{Vampire}   \g{Vampire}   \g{Vampire}   \g{Vampire}   \g{Vampire}   \g{Vampire}   \g{Vampire}   \g{Vampire}   \g{Vampire}   \g{Vampire}   \g{Vampire}   \g{Vampire}   \g{Vampire}   \g{Vampire} \\ 
 \hline
 \multicolumn{3}{c}{\textbf{REINFORCE\_P(x)}} \\ 
10000&1&Card Text : At the  \yg{beginning}  of  \yg{your}  upkeep ,  \yg{put}  a  \yg{+1/+1}   \yg{counter}  on each  \g{Vampire}   \yg{creature}  you control . At the  \yg{beginning}  of\\ 
10000&1&Card Text : At the  \yg{beginning}  of  \yg{your}  upkeep ,  \yg{put}  a  \yg{+1/+1}   \yg{counter}  on each  \g{Vampire}   \yg{creature}  you control . At the  \yg{beginning}  of\\ 
10000&1&Card Text : At the  \yg{beginning}  of  \yg{your}  upkeep ,  \yg{put}  a  \yg{+1/+1}   \yg{counter}  on each  \g{Vampire}   \yg{creature}  you control . At the  \yg{beginning}  of\\ 
10000&1&Card Text : At the  \yg{beginning}  of  \yg{your}  upkeep ,  \yg{put}  a  \yg{+1/+1}   \yg{counter}  on each  \g{Vampire}   \yg{creature}  you control . At the  \yg{beginning}  of\\ 
10000&1&Card Text : At the  \yg{beginning}  of  \yg{your}  upkeep ,  \yg{put}  a  \yg{+1/+1}   \yg{counter}  on each  \g{Vampire}   \yg{creature}  you control . At the  \yg{beginning}  of\\ 
10000&1&Card Text : At the  \yg{beginning}  of  \yg{your}  upkeep ,  \yg{put}  a  \yg{+1/+1}   \yg{counter}  on each  \g{Vampire}   \yg{creature}  you control . At the  \yg{beginning}  of\\ 
10000&1&Card Text : At the  \yg{beginning}  of  \yg{your}  upkeep ,  \yg{put}  a  \yg{+1/+1}   \yg{counter}  on each  \g{Vampire}   \yg{creature}  you control . At the  \yg{beginning}  of\\ 
10000&1&Card Text : At the  \yg{beginning}  of  \yg{your}  upkeep ,  \yg{put}  a  \yg{+1/+1}   \yg{counter}  on each  \g{Vampire}   \yg{creature}  you control . At the  \yg{beginning}  of\\ 
 \hline
 \multicolumn{3}{c}{\textbf{Ziegler}} \\ 
1&0&I had a great week and had a great  \ye{time}  at the Magic Kingdom . I'd had a  \ye{lot}  of fun and I wanted to\\ 
1&1&Kissinger's  \ye{second}   \ye{season}   \ye{was}  an absolute success . It  \ye{was}   \ye{also}  a major success ,  \ye{with}   \yg{Buffy}  the  \g{Vampire}   \yg{Slayer}  gaining the  \ye{first}   \ye{two}  seasons\\ 
1&1&{-} What's next for  \g{Vampire}  : The  \ye{Masquerade}  ? "It's been eight years since the beginning of our adventure for the most  \ye{popular}   \g{Vampire}  in\\ 
1&1& \g{Vampire}   \yg{Slayer}  is a sequel to the acclaimed television  \ye{series}   \g{Vampire}  : The  \ye{Masquerade}  , an anime adaptation of the comics  \ye{series}  . Contents show{]}\\ 
1&1&There  \ye{was}  the  \ye{time}   \ye{when}  I  \ye{was}  a young boy and my parents  \ye{were}  horrified  \ye{when}  I took my  \ye{first}  child , my  \g{Vampire}  of\\ 
1&1&I had written about  \g{Vampire}  : The  \ye{Masquerade}  II a couple of months ago , and I still think it's  \ye{one}  of my favorite games\\ 
1&1&Buffy the  \g{Vampire}  Slayer's  \ye{new}  leader is finally getting a good look at the  \ye{first}   \ye{two}  seasons of the  \ye{popular}   \ye{show}  , and that's exactly\\ 
1&1&HBO NOW A few months  \ye{back}  ,  \g{Vampire}   \ye{Diaries}  and The  \g{Vampire}  Diaries' co{-}creator Joe Louis Anderson announced plans to launch a limited theatrical run\\ 
\bottomrule 
 \end{tabular} 
 \caption{\label{table:generation-Vampire }
 Randomly selected generations from the single-word constraint task for the word ``Vampire'' (with occurrence probability $1/10^4$) highlighted in \g{green}. Tokens are highlighted with \ye{y}\yf{e}\yf{l}\yg{l}\yg{o}\yj{w} with different intensities to indicate their overall frequencies in the generated corpus. $\phi(x)=1$ indicates the satisfaction of the constraint in the sample and reps the number of its repetitions across all generations. 
 }\end{table}
\begin{table}[H]
        \scriptsize
        \begin{tabular}{p{0.5cm}|p{0.3cm}p{11.8cm}}
        \toprule
        reps & \textbf{$\phi(x)$} & \textbf{Generation} \\ 
         \midrule
 \multicolumn{3}{c}{\textbf{GDC}} \\ 
1&1&I got an e{-}mail from a couple of folks that we  \ye{found}   \ye{interesting}  and  \g{amusing}  . They asked if I could have an  \ye{idea}  of\\ 
1&1&The "Black Friday" holiday has  \ye{some}   \g{amusing}  details about the price of goods on Thanksgiving weekend , and  \ye{they}  are included in the holiday's list\\ 
1&1&"It  \yg{was}   \g{amusing}  and  \ye{very}   \g{amusing}  for all of us to witness , " he  \ye{said}  . "But it  \ye{also}   \yg{was}  not a  \ye{good}  time\\ 
1&0&My favorite  \ye{game}  of all  \ye{time}  . It  \yg{was}  a real fun  \ye{way}  to play  \ye{with}   \ye{your}  friends . This  \ye{game}   \yg{was}   \ye{one}  of my\\ 
1&0&Trevor Noah , on Wednesday's edition of Real Time  \ye{with}  Bill Maher , lamented the lack of a definition of what it means to be\\ 
1&1&. @Gizmodo wrote : I  \ye{think}   \ye{they}  made  \ye{some}   \g{amusing}  stuff , but that makes it seem  \ye{like}  they're a  \ye{bit}  too  \ye{good}  to be\\ 
1&1&Curious as to what the  \ye{new}  album is ? If you are an aspiring musician or have  \ye{even}  discovered it , here are five things\\ 
1&1&So a few weeks ago we discussed the most  \g{amusing}  feature on Windows 10 where a Windows 10 user could turn off the device without\\ 
 \hline
 \multicolumn{3}{c}{\textbf{REINFORCE}} \\ 
17&1&This  \g{amusing}  ,  \g{amusing}  ,  \g{amusing}  ,  \g{amusing}  ,  \g{amusing}  ,  \g{amusing}  ,  \g{amusing}  ,  \g{amusing}  ,  \ye{absurd}  ,  \ye{absurd}  ,  \ye{absurd}  ,  \ye{absurd}  ,\\ 
1&1&SOn , I am amused by the  \g{amusing}  ,  \g{amusing}  ,  \g{amusing}  ,  \g{amusing}  ,  \g{amusing}  ,  \yg{hilarious}  ,  \ye{hilar}  ,  \ye{hilar}  ,  \ye{hilar}  ,\\ 
50&1&Iam  \g{amusing}  ,  \g{amusing}  ,  \g{amusing}  ,  \g{amusing}  ,  \g{amusing}  ,  \g{amusing}  ,  \yg{hilarious}  ,  \yg{hilarious}  ,  \yg{hilarious}  ,  \yg{hilarious}  ,  \yg{hilarious}  ,  \yg{hilarious}  ,\\ 
1&1&Iam , and  \g{amusing}  ,  \g{amusing}  ,  \g{amusing}  ,  \yg{hilarious}  ,  \yg{hilarious}  ,  \yg{hilarious}  ,  \yg{hilarious}  ,  \yg{hilarious}  ,  \yg{hilarious}  ,  \yg{hilarious}  ,  \yg{hilarious}  ,\\ 
51&1&Iam , The  \g{amusing}  ,  \g{amusing}  ,  \g{amusing}  ,  \g{amusing}  ,  \yg{hilarious}  ,  \yg{hilarious}  ,  \yg{hilarious}  ,  \yg{hilarious}  ,  \yg{hilarious}  ,  \yg{hilarious}  ,  \yg{hilarious}  ,\\ 
1&1&A3  \g{amusing}  ,  \g{amusing}  ,  \g{amusing}  ,  \g{amusing}  ,  \g{amusing}  ,  \g{amusing}  ,  \g{amusing}  ,  \g{amusing}  ,  \g{amusing}  ,  \g{amusing}  ,  \ye{absurd}  ,  \ye{absurd}  ,\\ 
3&1&Iam , the  \g{amusing}  ,  \g{amusing}  ,  \yg{hilarious}  ,  \g{amusing}  ,  \yg{hilarious}  ,  \yg{hilarious}  ,  \yg{hilarious}  ,  \yg{hilarious}  ,  \yg{hilarious}  ,  \yg{hilarious}  ,  \yg{hilarious}  ,\\ 
3&1&Iam  \g{amusing}  ,  \g{amusing}  ,  \g{amusing}  ,  \g{amusing}  ,  \g{amusing}  ,  \g{amusing}  ,  \g{amusing}  ,  \g{amusing}  ,  \ye{hilar}  ,  \g{amusing}  ,  \ye{hilar}  ,  \ye{hilar}  ,\\ 
 \hline
 \multicolumn{3}{c}{\textbf{REINFORCE\_P(x)}} \\ 
1&0&I think the biggest issue  \ye{with}  the  \ye{new}  law is that the  \ye{government}  has a  \ye{very}  narrow interpretation of the Constitution . That  \ye{would}  mean\\ 
1&0&Korea's  \ye{first}  president has  \ye{said}  he will resign after he failed to reach agreement  \ye{with}  North Korea on the group's nuclear programme and warned he\\ 
1&0&A group of students in the  \ye{United}  States  \ye{were}  arrested this week , on charges of criminal sexual misconduct , after  \ye{they}  allegedly engaged in\\ 
1&0&Gigabyte has partnered  \ye{with}  Intel to provide Linux developers  \ye{with}  a full{-}text search engine ,  \ye{which}   \ye{can}  be used to find Linux{-}related documents . In\\ 
1&0&"The real story is that , this  \ye{time}  , it's really been about women's rights , " Trump  \ye{said}  . "The real story is ,\\ 
1&0&RICHMOND , Va . (WPRI) - Three  \ye{people}   \ye{were}  killed and  \ye{two}  others  \ye{were}  injured  \ye{when}  a bus  \ye{was}  derailed Thursday morning at Union Station\\ 
1&0&U . S . Department of Energy's National Renewable Energy Laboratory (NREL) will begin pumping the  \ye{first}  water from California reservoirs in a month in\\ 
1&0&. Cockroach and cockroaches  \ye{were}  found in the garden and gardens of  \ye{two}  local farms in East Melbourne in 2010 . A farmer who worked\\ 
 \hline
 \multicolumn{3}{c}{\textbf{Ziegler}} \\ 
1&1&I  \ye{really}  don't  \ye{know}  why  \ye{she}   \yg{was}  so excited about the "I'm  \ye{going}  to be in my own  \ye{game}  . " It  \yg{was}   \g{amusing}  to\\ 
1&1&You  \ye{can}   \ye{see}  , the  \ye{whole}  point of this  \ye{post}  is to  \ye{get}  back to the "What is it all about ? " point .\\ 
1&1&"You  \ye{know}  , it's all that has happened in a couple of weeks in the last  \ye{two}  weeks , "  \ye{said}  Smith . "It's  \g{amusing} \\ 
1&1&Consequences of the War . I will not answer any questions . However it is  \g{amusing}  to  \ye{see}  how  \ye{many}  "fancy" books have been published\\ 
1&1&In  \ye{fact}  , I'd  \ye{say}  that this  \ye{game}  is the closest  \ye{thing}  I've ever  \ye{seen}  to the real life  \ye{story}  of the main characters .\\ 
1&1&The only  \ye{thing}  more  \g{amusing}  , however ,  \yg{was}  to  \ye{see}  how it went down . The last person who ever  \ye{read}  this piece would\\ 
1&1&It may be an  \g{amusing}   \ye{fact}  that the American Society of Pediatricians and Surgeons does not endorse circumcision . However , it is  \ye{actually}  the\\ 
1&0&Cannot be created  \ye{with}   \ye{your}  username Cannot be created  \ye{with}   \ye{your}  username Cannot be created  \ye{with}   \ye{your}  username Cannot be created  \ye{with}   \ye{your}  username Can't\\ 
\bottomrule 
 \end{tabular} 
 \caption{\label{table:generation-amusing }
 Randomly selected generations  from the single-word constraint task for the word ``amusing'' (with occurrence probability $1/10^4$) highlighted in \g{green}. Tokens are highlighted with \ye{y}\yf{e}\yf{l}\yg{l}\yg{o}\yj{w} with different intensities to indicate their overall frequencies in the generated corpus. $\phi(x)=1$ indicates the satisfaction of the constraint in the sample and reps the number of its repetitions across all generations. 
 }\end{table}
\begin{table}[H]
        \scriptsize
        \begin{tabular}{p{0.5cm}|p{0.3cm}p{11.8cm}}
        \toprule
        reps & \textbf{$\phi(x)$} & \textbf{Generation} \\ 
         \midrule
 \multicolumn{3}{c}{\textbf{GDC}} \\ 
1&1&Liz  \g{Paris}  is an independent journalist and writer based in  \g{Paris}  .\\ 
1&1&(CNN)  \ye{President}  Donald Trump's recent comments on an Islamophobic  \g{Paris}   \ye{terror}   \ye{attack}  are a reminder that he has far{-}reaching political goals .  \ye{Trump}  on Wednesday\\ 
1&1& \g{Paris}   \ye{police}  are investigating a "large{-}scale hate crime" that is alleged to have taken place on Sunday night , while in London , a  \g{Paris} \\ 
1&1&The  \g{Paris}   \ye{Agreement}  comes after  \ye{France}  and Russia vetoed a draft UN Security Council resolution demanding a UN resolution on  \ye{climate}   \ye{change}  by a vote\\ 
1&1&Sophia LaFleur's "Lunch Break" is a  \g{Paris}  show , and the rest of this article looks at her performance at  \g{Paris}  Fashion Week , where\\ 
1&1&The " \g{Paris}  Commune" has been a long and painful experience for many of the thousands of workers who marched for a better world . The\\ 
1&1&As  \ye{President}  Barack Obama leaves office , he'll unveil the  \g{Paris}   \ye{climate}  accord , or COP21 , by the end of the year , and\\ 
1&1&At least 20  \ye{people}   \ye{were}   \ye{killed}  and  \ye{over}  70 injured in an  \ye{attack}  at  \g{Paris} ' Place de la République  \ye{last}  weekend , as  \ye{police}  carried\\ 
 \hline
 \multicolumn{3}{c}{\textbf{REINFORCE}} \\ 
1&1&Siemens  \g{Paris}   \g{Paris}   \g{Paris}   \g{Paris}   \g{Paris}   \g{Paris}   \g{Paris}   \g{Paris}   \g{Paris}   \g{Paris}   \g{Paris}   \g{Paris}   \g{Paris}   \g{Paris}   \g{Paris}   \g{Paris}   \g{Paris}   \g{Paris}   \g{Paris}   \g{Paris}   \g{Paris}   \g{Paris}   \g{Paris}   \g{Paris} \\ 
1&1&Baghdant said  \g{Paris}  was "bombed" by the French  \g{Paris}   \g{Paris}   \g{Paris}   \g{Paris}   \g{Paris}   \g{Paris}   \g{Paris}   \g{Paris}   \g{Paris}   \g{Paris}   \g{Paris}   \g{Paris}   \g{Paris}   \g{Paris}   \g{Paris}   \g{Paris}   \g{Paris} \\ 
1&1&Bastard is  \g{Paris}   \g{Paris}   \g{Paris}   \g{Paris}   \g{Paris}   \g{Paris}   \g{Paris}   \g{Paris}   \g{Paris}   \g{Paris}   \g{Paris}   \g{Paris}   \g{Paris}   \g{Paris}   \g{Paris}   \g{Paris}   \g{Paris}   \g{Paris}   \g{Paris}   \g{Paris}   \g{Paris}   \g{Paris}   \g{Paris} \\ 
1&1&Plants on  \g{Paris}   \g{Paris}   \g{Paris}   \g{Paris}   \g{Paris}   \g{Paris}   \g{Paris}   \g{Paris}   \g{Paris}   \g{Paris}   \g{Paris}   \g{Paris}   \g{Paris}   \g{Paris}   \g{Paris}   \g{Paris}   \g{Paris}   \g{Paris}   \g{Paris}   \g{Paris}   \g{Paris}   \g{Paris}   \g{Paris} \\ 
99&1&A  \g{Paris}   \g{Paris}   \g{Paris}   \g{Paris}   \g{Paris}   \g{Paris}   \g{Paris}   \g{Paris}   \g{Paris}   \g{Paris}   \g{Paris}   \g{Paris}   \g{Paris}   \g{Paris}   \g{Paris}   \g{Paris}   \g{Paris}   \g{Paris}   \g{Paris}   \g{Paris}   \g{Paris}   \g{Paris}   \g{Paris}   \g{Paris} \\ 
1&1&LATAM  \ye{|}   \g{Paris}   \g{Paris}   \g{Paris}   \g{Paris}   \g{Paris}   \g{Paris}   \g{Paris}   \g{Paris}   \g{Paris}   \g{Paris}   \g{Paris}   \g{Paris}   \g{Paris}   \g{Paris}   \g{Paris}   \g{Paris}   \g{Paris}   \g{Paris}   \g{Paris}   \g{Paris}   \g{Paris}   \g{Paris}   \g{Paris} \\ 
1&1&Karen :  \g{Paris}   \ye{|}   \g{Paris}   \g{Paris}   \g{Paris}   \g{Paris}   \g{Paris}   \g{Paris}   \g{Paris}   \g{Paris}   \g{Paris}   \g{Paris}   \g{Paris}   \g{Paris}   \g{Paris}   \g{Paris}   \g{Paris}   \g{Paris}   \g{Paris}   \g{Paris}   \g{Paris}   \g{Paris}   \g{Paris} \\ 
1&1&Kasim Kouz  \ye{celebrates}   \g{Paris}  for  \g{Paris}   \g{Paris}   \g{Paris}   \g{Paris}   \g{Paris}   \g{Paris}   \g{Paris}   \g{Paris}   \g{Paris}   \g{Paris}   \g{Paris}   \g{Paris}   \g{Paris}   \g{Paris}   \g{Paris}   \g{Paris}   \g{Paris}   \g{Paris}   \g{Paris}   \g{Paris} \\ 
 \hline
 \multicolumn{3}{c}{\textbf{REINFORCE\_P(x)}} \\ 
3770&1&MOSCOW  \yg{(Reuters)}  {-} U . S .  \yg{President}   \yg{Donald}   \yg{Trump}   \yg{said}  on  \yg{Friday}  he  \yg{would}   \yg{withdraw}  from the  \g{Paris}   \yg{climate}   \ye{accord}  . U . S\\ 
3770&1&MOSCOW  \yg{(Reuters)}  {-} U . S .  \yg{President}   \yg{Donald}   \yg{Trump}   \yg{said}  on  \yg{Friday}  he  \yg{would}   \yg{withdraw}  from the  \g{Paris}   \yg{climate}   \ye{accord}  . U . S\\ 
3770&1&MOSCOW  \yg{(Reuters)}  {-} U . S .  \yg{President}   \yg{Donald}   \yg{Trump}   \yg{said}  on  \yg{Friday}  he  \yg{would}   \yg{withdraw}  from the  \g{Paris}   \yg{climate}   \ye{accord}  . U . S\\ 
134&1&MOSCOW  \yg{(Reuters)}  {-} U . S .  \yg{President}   \yg{Donald}   \yg{Trump}   \yg{said}  on  \yg{Friday}  he  \yg{would}   \yg{withdraw}  from the  \g{Paris}   \yg{climate}  climate  \ye{accord}  ,  \ye{saying}  the\\ 
1040&1&MOSCOW  \yg{(Reuters)}  {-} U . S .  \yg{President}   \yg{Donald}   \yg{Trump}   \yg{said}  on  \yg{Friday}  he  \yg{would}   \yg{withdraw}  from the  \g{Paris}   \yg{climate}  climate  \ye{accord}  . U .\\ 
558&1&MOSCOW  \yg{(Reuters)}  {-} U . S .  \yg{President}   \yg{Donald}   \yg{Trump}   \yg{said}  on  \yg{Friday}  he  \yg{would}   \yg{withdraw}  from the  \g{Paris}   \yg{climate}  climate  \ye{accord}  .  \yg{FILE}  PHOTO\\ 
1563&1&MOSCOW  \yg{(Reuters)}  {-} U . S .  \yg{President}   \yg{Donald}   \yg{Trump}   \yg{said}  on  \yg{Friday}  he  \yg{would}   \yg{withdraw}  from the  \g{Paris}   \yg{climate}   \ye{accord}  .  \yg{FILE}   \ye{PHOTO}  {-}\\ 
1563&1&MOSCOW  \yg{(Reuters)}  {-} U . S .  \yg{President}   \yg{Donald}   \yg{Trump}   \yg{said}  on  \yg{Friday}  he  \yg{would}   \yg{withdraw}  from the  \g{Paris}   \yg{climate}   \ye{accord}  .  \yg{FILE}   \ye{PHOTO}  {-}\\ 
 \hline
 \multicolumn{3}{c}{\textbf{Ziegler}} \\ 
1&1&The  \g{Paris}   \ye{attacks}  claimed the lives of 20  \ye{people}  in a day and  \ye{left}   \ye{over}  4 , 400 injured , the authorities  \ye{said}  . The\\ 
1&1&In  \g{Paris}  , a major tourist attraction in the Middle East  \ye{with}  a long history of  \ye{terrorist}   \ye{attacks}  , the Charlie Hebdo massacre and the\\ 
1&1&As the  \g{Paris}   \ye{attack}  unfolded , the  \ye{European}  Union and the U . S . took to Twitter to describe the  \ye{attack}  . A tweet\\ 
1&1&The  \g{Paris}  massacre in November 2012  \yg{was}  carried  \ye{out}  under a pretext of preventing terrorism . But on this basis , the attackers knew nothing\\ 
1&1&In  \g{Paris}  on Monday , a delegation of 50 members of the  \ye{European}  Commission  \yg{was}   \ye{set}  to discuss the issue of the EU's plan to\\ 
1&1&In his  \g{Paris}  address ,  \ye{President}  Hollande pledged to work  \ye{with}   \ye{France}  to fight "the scourge of terrorism . " On Sunday , in a\\ 
1&1&A  \ye{man}  who allegedly attacked a girl in  \g{Paris}   \yg{was}  sentenced to 15 years to life in prison for killing three children in 2012 ,\\ 
1&1&Cairo , July 18 – The  \g{Paris}   \ye{terrorist}   \ye{attacks}  ,  \ye{which}   \ye{killed}  14  \ye{people}  ,  \ye{killed}  16 , wounded 13 more and  \ye{left}  a third\\ 
\bottomrule 
 \end{tabular} 
 \caption{\label{table:generation-Paris }
 Randomly selected generations  from the single-word constraint task for the word ``Paris'' (with occurrence probability $1/10^3$) highlighted in \g{green}. Tokens are highlighted with \ye{y}\yf{e}\yf{l}\yg{l}\yg{o}\yj{w} with different intensities to indicate their overall frequencies in the generated corpus. $\phi(x)=1$ indicates the satisfaction of the constraint in the sample and reps the number of its repetitions across all generations. 
 }\end{table}
\begin{table}[H]
        \scriptsize
        \begin{tabular}{p{0.5cm}|p{0.3cm}p{11.8cm}}
        \toprule
        reps & \textbf{$\phi(x)$} & \textbf{Generation} \\ 
         \midrule
 \multicolumn{3}{c}{\textbf{GDC}} \\ 
1&1&In 2014 , in an attempt to stop the  \g{restaurant}  industry from becoming a "corporate welfare racket" for the masses , the  \ye{city}  of San\\ 
1&0&A  \ye{New}  Jersey  \ye{man}   \yg{was}  arrested early Thursday morning on suspicion of possessing a gun and  \yg{was}  placed under investigation by the  \ye{police}  department ,\\ 
1&1&SINGAPORE {-} A sushi  \g{restaurant}   \ye{owner}  has been jailed for 10 years for allegedly stealing money from a customer during the summer . A witness\\ 
1&1&The  \g{restaurant} 's  \ye{owner}  , James Saito ,  \yg{was}  suspended without pay last month after he  \ye{said}  he accidentally broke the glass in front of a\\ 
1&1&A  \ye{local}   \g{restaurant}   \ye{chain}  on Monday announced its intention to offer a variety of meals and snacks to customers in the form of ice cream\\ 
1&1&I've never been in a  \g{restaurant}  before , but the atmosphere at the  \g{restaurant}   \yg{was}  very different than I remembered . And  \ye{with}  only a\\ 
1&1&Watchers  \yg{was}  founded in 1993 by a  \g{restaurant}  co{-}owner who wanted a  \ye{place}  that had a true Southern feel . The  \g{restaurant}   \ye{opened}  on June\\ 
1&1&A  \g{restaurant}  in the heart of the  \ye{San}  Antonio area has been turned into an art gallery by a  \ye{local}  entrepreneur . Carnal Cafe ,\\ 
 \hline
 \multicolumn{3}{c}{\textbf{REINFORCE}} \\ 
1&1&The  \yg{best}   \yg{Mexican}   \g{restaurant}   \yh{Italian}   \g{restaurant}  that has  \yh{Italian}   \g{restaurant}  that famous  \yh{Italian}  Italian  \g{restaurant}  that famous  \yg{Mexican}   \g{restaurant}   \g{restaurant}  that famous  \yh{Italian}   \g{restaurant}  that\\ 
1&1&The most expensive  \yh{Italian}   \yh{pizza}   \g{restaurant}   \yg{chain}   \g{restaurant}   \yg{chain}   \g{restaurant} -free to right{-}old  \g{restaurant} -hot{-}free  \yh{pizza}   \yh{Italian}   \yh{pizza}   \g{restaurant}  buti fast{-}food  \g{restaurant} -street  \g{restaurant} -dent meal\\ 
1&1&The first American  \g{restaurant}   \yg{chain}   \g{restaurant}   \yg{chain}   \g{restaurant}   \yg{chain}   \g{restaurant}   \yg{chain}   \g{restaurant}   \yg{chain}   \g{restaurant}   \yg{chain}   \g{restaurant}   \yg{chain}   \g{restaurant}  . The first  \g{restaurant}   \yg{chain}   \g{restaurant}  chain\\ 
1&1&2  \ye{chicken}   \yh{Italian}   \yh{pizza}   \g{restaurant}   \yg{-}   \yg{Mexican}   \yh{Italian}   \yh{pizza}   \yg{-}   \yh{pizza}   \g{restaurant}   \yg{-}   \yh{Italian}  Italian  \yh{pizza}   \g{restaurant}   \yg{-}   \yg{Mexican}  Mexican  \yh{Italian}   \yh{pizza}   \g{restaurant}   \yg{-}  Mexican\\ 
1&1&Kud  \yg{-}  a  \yh{Italian}   \yg{burger}   \g{restaurant}   \yg{chain}   \g{restaurant}  that  \yg{chain}   \g{restaurant}   \g{restaurant}   \yg{-}   \yg{chain}   \g{restaurant}   \yg{-}   \yh{Italian}   \yh{pizza}   \g{restaurant}   \yg{-}   \yg{Mexican}   \g{restaurant}   \yg{-}   \yg{chain}   \g{restaurant} \\ 
1&1&The Red Lob  \ye{Taco}   \g{restaurant}   \g{restaurant}   \yg{chain}   \g{restaurant}   \yg{chain}   \g{restaurant}   \yg{chain}   \g{restaurant}   \yg{chain}   \g{restaurant}   \yg{chain}   \g{restaurant}   \yg{chain}   \g{restaurant}   \yg{chain}   \g{restaurant}   \yg{chain}   \g{restaurant}   \yg{chain}   \g{restaurant}  chain\\ 
1&1&4{-}pic  \yh{pizza}   \g{restaurant}   \ye{place}   \yh{pizza}   \g{restaurant}  in a  \yh{Italian}   \g{restaurant}   \g{restaurant}   \g{restaurant}   \yg{chain}   \g{restaurant}  that  \yg{chain}   \g{restaurant}   \yg{chain}   \g{restaurant}  that right away  \yh{Italian}   \yh{pizza}   \g{restaurant} \\ 
1&1&Finesse  \yh{Italian}  Italian food{-}free  \yh{pizza}   \g{restaurant}   \yg{-}  dairy{-}free  \yh{pizza}   \g{restaurant}   \yg{-}  pizzic  \yg{-}   \yh{Italian}   \yg{food}   \yh{pizza}   \g{restaurant}   \yg{-}   \yg{Mexican}   \yh{pizza}   \yg{-}   \yh{Italian}   \g{restaurant}   \g{restaurant}  -\\ 
 \hline
 \multicolumn{3}{c}{\textbf{REINFORCE\_P(x)}} \\ 
10000&1&Is this  \g{restaurant}  open ?  \yg{Yes}  No  \yg{Unsure}  Does this  \g{restaurant}   \yg{accept}   \yg{credit}  cards ?  \yg{Yes}  No  \yg{Unsure}  Does this  \g{restaurant}   \yg{accept}   \yg{credit}  cards ?\\ 
10000&1&Is this  \g{restaurant}  open ?  \yg{Yes}  No  \yg{Unsure}  Does this  \g{restaurant}   \yg{accept}   \yg{credit}  cards ?  \yg{Yes}  No  \yg{Unsure}  Does this  \g{restaurant}   \yg{accept}   \yg{credit}  cards ?\\ 
10000&1&Is this  \g{restaurant}  open ?  \yg{Yes}  No  \yg{Unsure}  Does this  \g{restaurant}   \yg{accept}   \yg{credit}  cards ?  \yg{Yes}  No  \yg{Unsure}  Does this  \g{restaurant}   \yg{accept}   \yg{credit}  cards ?\\ 
10000&1&Is this  \g{restaurant}  open ?  \yg{Yes}  No  \yg{Unsure}  Does this  \g{restaurant}   \yg{accept}   \yg{credit}  cards ?  \yg{Yes}  No  \yg{Unsure}  Does this  \g{restaurant}   \yg{accept}   \yg{credit}  cards ?\\ 
10000&1&Is this  \g{restaurant}  open ?  \yg{Yes}  No  \yg{Unsure}  Does this  \g{restaurant}   \yg{accept}   \yg{credit}  cards ?  \yg{Yes}  No  \yg{Unsure}  Does this  \g{restaurant}   \yg{accept}   \yg{credit}  cards ?\\ 
10000&1&Is this  \g{restaurant}  open ?  \yg{Yes}  No  \yg{Unsure}  Does this  \g{restaurant}   \yg{accept}   \yg{credit}  cards ?  \yg{Yes}  No  \yg{Unsure}  Does this  \g{restaurant}   \yg{accept}   \yg{credit}  cards ?\\ 
10000&1&Is this  \g{restaurant}  open ?  \yg{Yes}  No  \yg{Unsure}  Does this  \g{restaurant}   \yg{accept}   \yg{credit}  cards ?  \yg{Yes}  No  \yg{Unsure}  Does this  \g{restaurant}   \yg{accept}   \yg{credit}  cards ?\\ 
10000&1&Is this  \g{restaurant}  open ?  \yg{Yes}  No  \yg{Unsure}  Does this  \g{restaurant}   \yg{accept}   \yg{credit}  cards ?  \yg{Yes}  No  \yg{Unsure}  Does this  \g{restaurant}   \yg{accept}   \yg{credit}  cards ?\\ 
 \hline
 \multicolumn{3}{c}{\textbf{Ziegler}} \\ 
1&1&The  \g{restaurant}  in  \ye{San}  Antonio , Texas is known for a "Southern Texas food" philosophy that has given it its name , according to the\\ 
1&1&We've had a lot of success  \ye{with}  this , and a lot of  \ye{great}  things . There's this  \g{restaurant}  . We  \ye{were}  all over it\\ 
1&1&I'm really pleased  \ye{with}  my purchase! The  \ye{menu}   \yg{was}  the same  \ye{with}  a lot of  \g{restaurant}  options and I couldn't say enough  \ye{good}  things about\\ 
1&1&"I wanted to bring this  \g{restaurant}  to town , "  \ye{said}  Jim Dorn , who manages the  \g{restaurant} 's business department . "I knew we were\\ 
1&1&The world's oldest  \g{restaurant}   \ye{chain}  , the Cinco de Mayo , offers a mix of comfort  \ye{food}  and classic Southern hospitality  \ye{with}  its iconic Italian\\ 
1&1&Saucer has been offering the  \g{restaurant}  the chance to offer a one{-}hour service for all its guests , but not necessarily at a premium .\\ 
1&1&SALT LAKE CITY - Three Utah  \g{restaurant}   \ye{owners}  have filed suit to force  \g{restaurant}   \ye{owner}  Jimmy Denny to close after his company failed to report\\ 
1&1&Fellow  \g{restaurant}   \ye{owners}  , remember that while every once in a while a friend invites you to his or her own  \g{restaurant}  , you never\\ 
\bottomrule 
 \end{tabular} 
 \caption{\label{table:generation-restaurant }
 Randomly selected generations  from the single-word constraint task for the word ``restaurant'' (with occurrence probability $1/10^3$) highlighted in \g{green}. Tokens are highlighted with \ye{y}\yf{e}\yf{l}\yg{l}\yg{o}\yj{w} with different intensities to indicate their overall frequencies in the generated corpus. $\phi(x)=1$ indicates the satisfaction of the constraint in the sample and reps the number of its repetitions across all generations. 
 }\end{table}
\begin{table}[H]
        \scriptsize
        \begin{tabular}{p{0.5cm}|p{0.3cm}p{11.8cm}}
        \toprule
        reps & \textbf{$\phi(x)$} & \textbf{Generation} \\ 
         \midrule
 \multicolumn{3}{c}{\textbf{GDC}} \\ 
1&1&We are doing this in collaboration  \ye{with}  you! We've done  \g{amazing}   \ye{work}  to  \ye{make}  Minecraft an  \g{amazing}   \ye{game}  . However , in the past ,\\ 
1&1&This  \ye{game}  is  \g{amazing} ! One of the most frustrating  \ye{things}  about playing this  \ye{game}  is the difficulty .  \ye{There}  is no leveling system , and\\ 
1&1&A  \ye{team}  of Japanese scientists has found that the world's largest nuclear plant could be a disaster waiting to happen . "This  \g{amazing}  discovery reveals\\ 
1&0&So  \ye{there}  we  \ye{were}  , looking at a gorgeous  \ye{game}  . That  \ye{was}   \ye{something}  I enjoyed  \ye{when}  I played a bit of a Zelda ,\\ 
1&1&I  \ye{just}  found  \ye{out}  about this and am super excited to  \ye{get}  it for you guys! Its  \g{amazing}  how  \ye{many}   \ye{great}  games I  \ye{can}  find\\ 
1&1&Thanks to  \g{amazing}  support , you have had access to this content for years , but have it been delivered to you in the form\\ 
1&1&What an  \g{amazing}   \ye{time}  to be a professional football fan! The fans of Minnesota have a  \ye{great}   \ye{time}  . I  \ye{love}  the city , the\\ 
1&0&B . P . A . (The Bill of Rights) The B . P . A . is an American institution of free speech .\\ 
 \hline
 \multicolumn{3}{c}{\textbf{REINFORCE}} \\ 
1&1&BING has been  \g{amazing} ! It's  \g{amazing} ! And  \g{amazing} ! It's  \g{amazing} !  \yg{We've}  been  \g{amazing} !  \yg{We're}   \g{amazing} ! It's  \g{amazing} ! It's  \g{amazing} ! It's  \g{amazing} ! It\\ 
1&1&PAN\&C's  \g{amazing} ! So  \g{amazing} !  \yg{We've}  been  \g{amazing} !  \yg{We're}   \g{amazing} !  \yg{We're}   \g{amazing} ! It's  \g{amazing} ! So  \g{amazing} ! I've been  \g{amazing} !  \yg{We've}  been  \g{amazing} \\ 
1&1&There's been  \g{amazing} ! It's  \g{amazing} ! It's  \g{amazing} ! So  \g{amazing} !! We both like! So awesome!! And  \g{amazing} ! It's  \g{amazing} ! It's  \g{amazing} ! All over!\\ 
1&1&You have been  \g{amazing} ! It's  \g{amazing} ! It's  \g{amazing} ! It's  \g{amazing} ! Allover! It's  \g{amazing} ! And  \ye{super}   \ye{quick!}  It's  \g{amazing} ! So  \g{amazing} !  \yg{We're}   \g{amazing} !\\ 
1&1&The vegan and  \g{amazing} ! It's  \g{amazing} !  \yg{We've}  been  \ye{incredible!}  It's  \g{amazing} ! It's  \g{amazing} ! It's  \g{amazing} ! It's  \g{amazing} ! And  \g{amazing} ! It's  \g{amazing} ! It\\ 
1&1&We're  \g{amazing}  and it's  \g{amazing} ! It's  \g{amazing} ! And  \g{amazing} !  \yg{We're}   \g{amazing} ! So  \g{amazing} ! So  \g{amazing} ! So  \ye{awesome!}  It's  \g{amazing} ! And  \ye{awesome!}  It's  \g{amazing} !\\ 
1&1&HAPPED! It's  \g{amazing} ! It's  \g{amazing} ! And  \g{amazing} ! It's  \g{amazing} ! It's  \g{amazing} ! It's  \g{amazing} ! It's  \g{amazing} ! It's  \g{amazing} ! It's  \g{amazing} !\\ 
1&1&It's  \g{amazing}  and we've been  \g{amazing} ! It's  \g{amazing} ! It's  \g{amazing} ! It's  \g{amazing} ! It's  \g{amazing} ! It's  \g{amazing} ! It's  \g{amazing} ! It's  \g{amazing} ! It's\\ 
 \hline
 \multicolumn{3}{c}{\textbf{REINFORCE\_P(x)}} \\ 
507&1&Say  \yg{thanks}  by  \yg{giving}   \yg{John}  a  \yg{tip}  and  \yg{help}   \yg{them}   \yg{continue}  to  \yg{share}   \g{amazing}   \yg{Things}   \yg{with}  the  \yg{Thingiverse}   \ye{community}  .  \yg{We're}   \yg{sure}   \yg{John}   \ye{would}  share\\ 
1940&1&Say  \yg{thanks}  by  \yg{giving}   \yg{John}  a  \yg{tip}  and  \yg{help}   \yg{them}   \yg{continue}  to  \yg{share}   \g{amazing}   \yg{Things}   \yg{with}  the  \yg{Thingiverse}   \ye{community}  .  \yg{We're}   \yg{sure}   \yg{John}  a tip\\ 
15&1&Say  \yg{thanks}  by  \yg{giving}   \yg{John}  a  \yg{tip}  and  \yg{help}   \yg{them}   \yg{continue}  to  \yg{share}   \g{amazing}   \yg{Things}   \yg{with}  the  \yg{Thingiverse}   \ye{community}  .  \ye{WeWe're}   \yg{sure}   \yg{John}  and John\\ 
1&1&Say  \yg{thanks}  by  \yg{giving}   \yg{John}  a  \yg{tip}  and  \yg{help}   \yg{them}   \yg{continue}  to  \yg{share}   \g{amazing}   \yg{Things}   \yg{with}  the  \yg{Thingiverse}   \ye{community}  .  \ye{We't}  do our  \yg{share}  of\\ 
11&1&Say  \yg{thanks}  by  \yg{giving}   \yg{John}  a  \yg{tip}  and  \yg{help}   \yg{them}   \yg{continue}  to  \yg{share}   \g{amazing}   \yg{Things}   \yg{with}  the  \yg{Thingiverse}   \ye{community}  .  \yg{We're}   \yg{sure}   \yg{John}   \ye{would}  have\\ 
1&1&Say  \yg{thanks}  by  \yg{giving}   \yg{John}  a  \yg{tip}  and  \yg{help}   \yg{them}   \yg{continue}  to  \yg{share}   \g{amazing}   \yg{Things}   \yg{with}  the  \yg{Thingiverse}   \ye{community}  .  \yg{We're}   \ye{also}  pretty  \yg{sure}  John\\ 
18&1&Say  \yg{thanks}  by  \yg{giving}   \yg{John}  a  \yg{tip}  and  \yg{help}   \yg{them}   \yg{continue}  to  \yg{share}   \g{amazing}   \yg{Things}   \yg{with}  the  \yg{Thingiverse}   \ye{community}  .  \yg{We're}   \yg{sure}   \yg{John}  and John\\ 
2&1&Say  \yg{thanks}  by  \yg{giving}   \yg{John}  a  \yg{tip}  and  \yg{help}   \yg{them}   \yg{continue}  to  \yg{share}   \g{amazing}   \yg{Things}   \yg{with}  the  \yg{Thingiverse}   \ye{community}  .  \ye{We't}  get enough of the\\ 
 \hline
 \multicolumn{3}{c}{\textbf{Ziegler}} \\ 
1&1&We need to  \ye{make}   \ye{sure}  that this type of  \ye{work}  will be shared . The  \g{amazing}  and talented  \ye{team}  at Google has  \ye{just}  announced a\\ 
1&1&I've been waiting for this  \g{amazing}  piece of artwork since I heard of it on the New York Times' "The Art of Comic{-}Con 2012" podcast\\ 
1&1&I  \ye{love}  this site because I'm trying to  \ye{find}  the right answers to every question I have as a designer . It's  \g{amazing}  how much\\ 
1&1&The New York Times is  \ye{going}  to be  \ye{out}  doing  \ye{something}   \g{amazing}   \ye{with}  its coverage of the presidential election . The paper is already releasing\\ 
1&1&You'll  \ye{see}  a  \ye{lot}  of  \g{amazing}   \ye{video}  games coming  \ye{out}  of Sony's booth , all from Sony and Nintendo , in the coming months .\\ 
1&1&The New York City FC academy and its training group  \ye{were}  both awarded two year contracts in 2014  \ye{with}  an  \g{amazing}  \$2 . 5 million\\ 
1&1&My favorite part of this tutorial is  \ye{when}  you watch it , the  \g{amazing}  detail of the line up . It's so fun to watch\\ 
1&1&You have  \g{amazing}  taste ,  \ye{can}  be enjoyed by yourself or others . Our wines are not for sale , so this is for you\\ 
\bottomrule 
 \end{tabular} 
 \caption{Randomly selected generations  from the single-word constraint task for the word ``amazing'' (with occurrence probability $1/10^3$) highlighted in \g{green}. Tokens are highlighted with \ye{y}\yf{e}\yf{l}\yg{l}\yg{o}\yj{w} with different intensities to indicate their overall frequencies in the generated corpus. $\phi(x)=1$ indicates the satisfaction of the constraint in the sample and reps the number of its repetitions across all generations.
 \label{table:generation-amazing }}\end{table}
\begin{table}[H]
        \scriptsize
        \begin{tabular}{p{0.5cm}|p{0.3cm}p{11.8cm}}
        \toprule
        reps & \textbf{$\phi(x)$} & \textbf{Generation} \\ 
         \midrule
 \multicolumn{3}{c}{\textbf{GDC}} \\ 
1&1&OTTAWA  \ye{-}   \g{Canada} 's  \ye{new}  trade minister  \ye{says}  the  \ye{new}  relationship  \ye{with}  the  \ye{United}   \ye{States}  is an important development . Andrew Robb told an emergency meeting\\ 
1&1& \g{Canada}  to start trading  \ye{with}  the US ,  \g{Canada}  is now considering a move towards becoming a trade partner  \ye{with}  the US  \ye{Canadian}  Prime Minister\\ 
1&1&In the U . S . ,  \g{Canada}  ,  \ye{Australia}  and New Zealand are among the most{-}traveled countries in the world , and  \ye{they}  have\\ 
1&1&The Federal Government is making changes to the  \g{Canada}  Revenue  \ye{Agency}  (CRA) to make it easier for employers to pay their employees more . The\\ 
1&1& \g{Canada} 's  \ye{public}  school system is struggling  \ye{with}  its highest rate of student debt , and in recent years the province has been struggling to keep\\ 
1&1&In  \g{Canada}  , when I look at my family's wealth , my parents and my grandparents  \ye{were}  still poor . Their children now spend most\\ 
1&1&Tales of the Beast Edit The  \ye{Canadian}  Broadcasting Corporation (CBC) aired Tales of the Beast on  \g{Canada}  Day , October 20 , 2011 . It\\ 
1&1& \g{Canada} 's defence ministry confirmed that it will continue to operate its armed forces during peacetime when it is out of military operations in Afghanistan ,\\ 
 \hline
 \multicolumn{3}{c}{\textbf{REINFORCE}} \\ 
1&1& \g{Canada}  is  \ye{with}  China .  \g{Canada}  is not  \yh{oil}  . " "Oil"  \yh{oil}  oil  \ye{-}   \g{Canada}   \yh{oil}  oil  \yh{oil}  and  \yh{oil}  oil  \ye{oils}   \yg{Canadian}  -\\ 
1&1& \g{Canada}  and  \g{Canada} )  \ye{Israel}  :  \yg{Oil}   \g{Canada}   \ye{,}   \yh{oil}   \ye{oil"}   \yg{Oil}  Products)  \g{Canada}   \yg{Tobacco}   \ye{Products}  and  \g{Canada}   \yg{Tobacco}   \ye{Products}   \ye{USA}  . . . Inc .\\ 
1&1& \g{Canada}  recognizes Russia  \ye{with}  Russia's  \yg{Oil}  and  \ye{Gas}   \g{Canada}   \ye{with}   \g{Canada}   \yh{oil}  and  \yh{oil}  and  \yh{oil}  and  \yh{oil}  and  \yh{oil}  . . .  \g{Canada}  oil\\ 
1&1& \g{Canada}  is  \ye{with}   \ye{Israel}   \ye{,}   \g{Canada}   \yh{oil}  and  \yh{oil}  oil  \yh{oil}  oil  \g{Canada}   \ye{-}   \g{Canada}   \yh{oil}  oil  \yh{oil}   \ye{-}   \yh{oil}  oil  \ye{oil"}   \ye{-}   \yh{oil}  oil\\ 
1&1& \g{Canada}  is  \g{Canada}   \ye{,}  for the First Fence  \ye{,}   \ye{Bank}   \g{Canada}   \g{Canada}  and  \g{Canada}  Investment in  \g{Canada}   \ye{,}  and  \ye{Bank}  of  \g{Canada}  the  \yg{Canadian}  Charter\\ 
1&1&" \g{Canada} " is  \g{Canada}  . "  \yg{Canadian}   \yh{oil}  interests" interests" are  \g{Canada} "oil and  \yh{oil}  oil  \yh{oil}  oil  \ye{oil"}   \yh{oil}  oil  \ye{oil"}   \yh{oil}   \ye{oil"}  oil"  \ye{oil"}  oil"\\ 
1&1& \g{Canada}  is not  \g{Canada}   \ye{with}   \yh{oil}   \ye{oils}  and  \yh{oil}  oil  \ye{sands}   \g{Canada}  .  \g{Canada}   \yg{Oil}  and  \yh{oil}   \ye{sands}   \g{Canada}  ( \g{Canada} ) . . .  \g{Canada}   \g{Canada} \\ 
1&1& \g{Canada}  and  \g{Canada} )  \ye{Israel}  the US" parent)  \ye{with}   \ye{Israel}  Oil) \g{Canada}   \yh{oil}  oil is  \g{Canada}  Oil's  \yg{Oil}   \g{Canada}   \ye{,}  Quebec oil{-} \g{Canada}   \yg{Oil}  oil{-} \g{Canada}   \ye{Canadians}  and  \g{Canada} \\ 
 \hline
 \multicolumn{3}{c}{\textbf{REINFORCE\_P(x)}} \\ 
10000&1&Please  \yg{select}   \yg{your}  country :  \yg{United}   \yg{States}   \yg{Argentina}   \yg{Australia}   \yg{Austria}   \yg{Belarus}   \yg{Belgium}   \yg{Brazil}   \yg{Brunei}   \yg{Bulgaria}   \g{Canada}   \yg{Chile}   \yg{China}   \yg{Colombia}   \yg{Croatia}   \yg{Cyprus}   \yh{Czech}   \yh{Republic}   \yh{Czech}  Republic\\ 
10000&1&Please  \yg{select}   \yg{your}  country :  \yg{United}   \yg{States}   \yg{Argentina}   \yg{Australia}   \yg{Austria}   \yg{Belarus}   \yg{Belgium}   \yg{Brazil}   \yg{Brunei}   \yg{Bulgaria}   \g{Canada}   \yg{Chile}   \yg{China}   \yg{Colombia}   \yg{Croatia}   \yg{Cyprus}   \yh{Czech}   \yh{Republic}   \yh{Czech}  Republic\\ 
10000&1&Please  \yg{select}   \yg{your}  country :  \yg{United}   \yg{States}   \yg{Argentina}   \yg{Australia}   \yg{Austria}   \yg{Belarus}   \yg{Belgium}   \yg{Brazil}   \yg{Brunei}   \yg{Bulgaria}   \g{Canada}   \yg{Chile}   \yg{China}   \yg{Colombia}   \yg{Croatia}   \yg{Cyprus}   \yh{Czech}   \yh{Republic}   \yh{Czech}  Republic\\ 
10000&1&Please  \yg{select}   \yg{your}  country :  \yg{United}   \yg{States}   \yg{Argentina}   \yg{Australia}   \yg{Austria}   \yg{Belarus}   \yg{Belgium}   \yg{Brazil}   \yg{Brunei}   \yg{Bulgaria}   \g{Canada}   \yg{Chile}   \yg{China}   \yg{Colombia}   \yg{Croatia}   \yg{Cyprus}   \yh{Czech}   \yh{Republic}   \yh{Czech}  Republic\\ 
10000&1&Please  \yg{select}   \yg{your}  country :  \yg{United}   \yg{States}   \yg{Argentina}   \yg{Australia}   \yg{Austria}   \yg{Belarus}   \yg{Belgium}   \yg{Brazil}   \yg{Brunei}   \yg{Bulgaria}   \g{Canada}   \yg{Chile}   \yg{China}   \yg{Colombia}   \yg{Croatia}   \yg{Cyprus}   \yh{Czech}   \yh{Republic}   \yh{Czech}  Republic\\ 
10000&1&Please  \yg{select}   \yg{your}  country :  \yg{United}   \yg{States}   \yg{Argentina}   \yg{Australia}   \yg{Austria}   \yg{Belarus}   \yg{Belgium}   \yg{Brazil}   \yg{Brunei}   \yg{Bulgaria}   \g{Canada}   \yg{Chile}   \yg{China}   \yg{Colombia}   \yg{Croatia}   \yg{Cyprus}   \yh{Czech}   \yh{Republic}   \yh{Czech}  Republic\\ 
10000&1&Please  \yg{select}   \yg{your}  country :  \yg{United}   \yg{States}   \yg{Argentina}   \yg{Australia}   \yg{Austria}   \yg{Belarus}   \yg{Belgium}   \yg{Brazil}   \yg{Brunei}   \yg{Bulgaria}   \g{Canada}   \yg{Chile}   \yg{China}   \yg{Colombia}   \yg{Croatia}   \yg{Cyprus}   \yh{Czech}   \yh{Republic}   \yh{Czech}  Republic\\ 
10000&1&Please  \yg{select}   \yg{your}  country :  \yg{United}   \yg{States}   \yg{Argentina}   \yg{Australia}   \yg{Austria}   \yg{Belarus}   \yg{Belgium}   \yg{Brazil}   \yg{Brunei}   \yg{Bulgaria}   \g{Canada}   \yg{Chile}   \yg{China}   \yg{Colombia}   \yg{Croatia}   \yg{Cyprus}   \yh{Czech}   \yh{Republic}   \yh{Czech}  Republic\\ 
 \hline
 \multicolumn{3}{c}{\textbf{Ziegler}} \\ 
1&1&In  \g{Canada}  , you  \ye{can}  use your name , email address and  \g{Canada}  Web Service address to register as a freelancer . If you live\\ 
1&1&The  \ye{government}  is asking for \$5  \ye{million}  from the  \g{Canada}  Revenue  \ye{Agency}  ,  \ye{which}  is  \ye{part}  of the agency , to conduct a study to\\ 
1&1&The  \ye{federal}   \ye{government}  has released a \$50  \ye{million}  grant for  \g{Canada} 's private sector to work  \ye{with}  local  \ye{government}  , community groups , the arts and\\ 
1&1&The  \g{Canada}  Revenue  \ye{Agency}   \ye{says}  the company is not responsible for the use of data provided by it or the  \ye{people}  it is accessing in\\ 
1&1& \g{Canada} 's  \ye{top}  diplomat has condemned the killing of an Afghan man during a  \ye{recent}  airstrike on a refugee camp in Afghanistan . U . S\\ 
1&1&The  \ye{government}   \ye{announced}  on Thursday it is looking at setting  \ye{up}  a  \ye{national}  database of  \ye{people}  from around the  \ye{world}  who've been detained in  \g{Canada} \\ 
1&1&As the  \ye{federal}   \ye{government}  tries to cut carbon emissions ,  \g{Canada}  is struggling  \ye{with}  rising fuel prices ,  \ye{which}  are likely to lead to reductions\\ 
1&1& \g{Canada} 's Foreign Affairs  \ye{Minister}  Chrystia Freeland  \ye{said}  Tuesday that it's important to work  \ye{with}  other countries on combating terrorism , as well as  \g{Canada}  ,\\ 
\bottomrule 
 \end{tabular} 
 \caption{Randomly selected generations from the single-word constraint task for the word ``Canada'' (with occurrence probability $1/10^3$) highlighted in \g{green}. Tokens are highlighted with \ye{y}\yf{e}\yf{l}\yg{l}\yg{o}\yj{w} with different intensities to indicate their overall frequencies in the generated corpus. $\phi(x)=1$ indicates the satisfaction of the constraint in the sample and reps the number of its repetitions across all generations.
 \label{table:generation-Canada }
 }\end{table}
\begin{table}[H]
        \scriptsize
        \begin{tabular}{p{0.5cm}|p{0.3cm}p{11.8cm}}
        \toprule
        reps & \textbf{$\phi(x)$} & \textbf{Generation} \\ 
         \midrule
 \multicolumn{3}{c}{\textbf{GDC}} \\ 
1&1&BEIJING-With  \g{China} 's  \ye{economy}  on track to become the second{-}biggest in the  \ye{world}  next year ,  \g{China} 's Central Bank raised interest rates on Monday as it\\ 
1&1&In 2008 ,  \ye{two}  years after  \g{China}  had launched a series of controversial surveillance programs known as PRISM ,  \ye{Chinese}  hackers took control of an\\ 
1&1&Chinese  \ye{President}   \ye{Xi}  Jinping has urged  \ye{Chinese}  firms to reduce their reliance on  \ye{foreign}  investors in the past , saying  \g{China}  needs to increase its\\ 
1&1& \g{China} 's president  \ye{said}  Saturday that he  \ye{would}  launch a U . S .  \ye{military}  strike on Iran if the  \ye{nuclear}  agreement  \ye{was}  not extended .\\ 
1&1& \g{China} 's Foreign Ministry issued a call Tuesday for the  \ye{United}   \ye{States}  to "be firm" in its efforts to pressure  \g{China}  on the ongoing tensions .\\ 
1&1&Shenzhen ,  \g{China}  {-}{-}  \g{China} 's  \ye{new}  presidential palace in Shenzhen has made a visit to  \g{China} 's capital from September 17 to 28 , according to\\ 
1&1&I've written before about how  \g{China}  and the  \ye{US}  are doing a lot of things to boost growth and even more for the bottom line\\ 
1&1&"I've never heard of any such person , "  \ye{said}  Tariq Abdul{-}Rahman , a lawyer in  \ye{Beijing}  . The People's Republic of  \g{China}  is the\\ 
 \hline
 \multicolumn{3}{c}{\textbf{REINFORCE}} \\ 
43&1& \g{China}  and  \g{China}   \g{China}  and  \g{China}   \g{China}  and  \g{China}   \g{China}   \yg{(}   \g{China}   \g{China}   \g{China}   \g{China}   \g{China}   \g{China}   \g{China}   \g{China}   \g{China}   \g{China}   \g{China}   \g{China}   \g{China}   \g{China} \\ 
312&1& \g{China}  and  \g{China}   \g{China}  and  \g{China}   \g{China}   \g{China}  and  \g{China}   \g{China}   \g{China}   \g{China}   \g{China}   \g{China}   \g{China}   \g{China}   \g{China}   \g{China}   \g{China}   \g{China}   \g{China}   \g{China}   \g{China}   \g{China} \\ 
636&1& \g{China}  and  \g{China}   \g{China}   \g{China}  and  \g{China}   \g{China}   \g{China}   \g{China}   \g{China}   \g{China}   \g{China}   \g{China}   \g{China}   \g{China}   \g{China}   \g{China}   \g{China}   \g{China}   \g{China}   \g{China}   \g{China}   \g{China}   \g{China} \\ 
609&1& \g{China}  and  \g{China}   \g{China}  and  \g{China}   \g{China}  and  \g{China}   \g{China}   \g{China}   \g{China}   \g{China}   \g{China}   \g{China}   \g{China}   \g{China}   \g{China}   \g{China}   \g{China}   \g{China}   \g{China}   \g{China}   \g{China}   \g{China} \\ 
26&1& \g{China}  and  \g{China}   \g{China}  and  \g{China}  and  \g{China}   \g{China}   \yg{(}   \g{China}   \g{China}   \g{China}   \g{China}   \g{China}   \g{China}   \g{China}   \g{China}   \g{China}   \g{China}   \g{China}   \g{China}   \g{China}   \g{China}   \g{China} \\ 
180&1& \g{China}  and  \g{China}   \yg{(}   \g{China}  and  \g{China}   \g{China}   \g{China}   \g{China}   \g{China}   \g{China}   \g{China}   \g{China}   \g{China}   \g{China}   \g{China}   \g{China}   \g{China}   \g{China}   \g{China}   \g{China}   \g{China}   \g{China}   \g{China} \\ 
135&1& \g{China}  and  \g{China}   \g{China}   \yg{(}  \g{China}   \g{China}   \g{China}   \g{China}   \g{China}   \g{China}   \g{China}   \g{China}   \g{China}   \g{China}   \g{China}   \g{China}   \g{China}   \g{China}   \g{China}   \g{China}   \g{China}   \g{China}   \g{China}   \g{China}   \g{China} \\ 
1&1& \g{China}  and  \g{China}   \g{China}  ,  \g{China}  and  \g{China}   \g{China}   \yg{(}  \g{China}   \g{China}   \g{China}   \g{China}   \g{China}   \g{China}   \g{China}   \g{China}   \g{China}   \g{China}   \g{China}   \g{China}   \g{China}   \g{China}   \g{China}   \g{China} \\ 
 \hline
 \multicolumn{3}{c}{\textbf{REINFORCE\_P(x)}} \\ 
10000&1&Please  \yg{select}   \yg{your}  country :  \yg{United}   \yg{States}   \yg{Argentina}   \yg{Australia}   \yg{Austria}   \yg{Belarus}   \yg{Belgium}   \yg{Brazil}   \yg{Brunei}   \yg{Bulgaria}   \yg{Canada}   \yg{Chile}   \g{China}   \yg{Colombia}   \yg{Croatia}   \yg{Cyprus}   \yh{Czech}   \yh{Republic}   \yh{Czech}  Republic\\ 
10000&1&Please  \yg{select}   \yg{your}  country :  \yg{United}   \yg{States}   \yg{Argentina}   \yg{Australia}   \yg{Austria}   \yg{Belarus}   \yg{Belgium}   \yg{Brazil}   \yg{Brunei}   \yg{Bulgaria}   \yg{Canada}   \yg{Chile}   \g{China}   \yg{Colombia}   \yg{Croatia}   \yg{Cyprus}   \yh{Czech}   \yh{Republic}   \yh{Czech}  Republic\\ 
10000&1&Please  \yg{select}   \yg{your}  country :  \yg{United}   \yg{States}   \yg{Argentina}   \yg{Australia}   \yg{Austria}   \yg{Belarus}   \yg{Belgium}   \yg{Brazil}   \yg{Brunei}   \yg{Bulgaria}   \yg{Canada}   \yg{Chile}   \g{China}   \yg{Colombia}   \yg{Croatia}   \yg{Cyprus}   \yh{Czech}   \yh{Republic}   \yh{Czech}  Republic\\ 
10000&1&Please  \yg{select}   \yg{your}  country :  \yg{United}   \yg{States}   \yg{Argentina}   \yg{Australia}   \yg{Austria}   \yg{Belarus}   \yg{Belgium}   \yg{Brazil}   \yg{Brunei}   \yg{Bulgaria}   \yg{Canada}   \yg{Chile}   \g{China}   \yg{Colombia}   \yg{Croatia}   \yg{Cyprus}   \yh{Czech}   \yh{Republic}   \yh{Czech}  Republic\\ 
10000&1&Please  \yg{select}   \yg{your}  country :  \yg{United}   \yg{States}   \yg{Argentina}   \yg{Australia}   \yg{Austria}   \yg{Belarus}   \yg{Belgium}   \yg{Brazil}   \yg{Brunei}   \yg{Bulgaria}   \yg{Canada}   \yg{Chile}   \g{China}   \yg{Colombia}   \yg{Croatia}   \yg{Cyprus}   \yh{Czech}   \yh{Republic}   \yh{Czech}  Republic\\ 
10000&1&Please  \yg{select}   \yg{your}  country :  \yg{United}   \yg{States}   \yg{Argentina}   \yg{Australia}   \yg{Austria}   \yg{Belarus}   \yg{Belgium}   \yg{Brazil}   \yg{Brunei}   \yg{Bulgaria}   \yg{Canada}   \yg{Chile}   \g{China}   \yg{Colombia}   \yg{Croatia}   \yg{Cyprus}   \yh{Czech}   \yh{Republic}   \yh{Czech}  Republic\\ 
10000&1&Please  \yg{select}   \yg{your}  country :  \yg{United}   \yg{States}   \yg{Argentina}   \yg{Australia}   \yg{Austria}   \yg{Belarus}   \yg{Belgium}   \yg{Brazil}   \yg{Brunei}   \yg{Bulgaria}   \yg{Canada}   \yg{Chile}   \g{China}   \yg{Colombia}   \yg{Croatia}   \yg{Cyprus}   \yh{Czech}   \yh{Republic}   \yh{Czech}  Republic\\ 
10000&1&Please  \yg{select}   \yg{your}  country :  \yg{United}   \yg{States}   \yg{Argentina}   \yg{Australia}   \yg{Austria}   \yg{Belarus}   \yg{Belgium}   \yg{Brazil}   \yg{Brunei}   \yg{Bulgaria}   \yg{Canada}   \yg{Chile}   \g{China}   \yg{Colombia}   \yg{Croatia}   \yg{Cyprus}   \yh{Czech}   \yh{Republic}   \yh{Czech}  Republic\\ 
 \hline
 \multicolumn{3}{c}{\textbf{Ziegler}} \\ 
1&1& \g{China}  has been slow to adopt these policies , but there are signs of the development ahead . According to a study conducted by the\\ 
1&1&BEIJING  \ye{(Reuters)}  {-}  \g{China}  and Russia clashed  \ye{over}  cyber{-}related issues in the latest round of  \ye{military}  exercises that took place at an air base in\\ 
1&1&(CNN)  \g{China} 's ruling  \ye{Communist}  Party is warning that  \g{China} 's "economic model is in danger" if the  \ye{country}  does not address its  \ye{economic}  concerns . The\\ 
1&1& \g{China} 's booming  \ye{economic}   \ye{growth}  and  \ye{growing}  social mobility has  \ye{also}  contributed to increased inequality in the  \ye{country}  . It seems , though , that the\\ 
1&1& \g{China}  has invested \$1 . 5  \ye{billion}  in  \ye{two}  projects at an initial cost of about \$45  \ye{billion}  ,  \ye{according}  to data compiled by Bloomberg\\ 
1&1&BEIJING :  \g{China}  is mulling  \ye{plans}  to deploy its  \ye{first}   \ye{nuclear}  weapons in the form of ballistic missiles , the  \ye{country's}  most important  \ye{nuclear}  watchdog\\ 
1&1&Beijing and Shanghai are  \ye{expected}  to sign a landmark agreement on  \ye{trade}  this week that will set  \ye{up}  a joint commission on  \g{China} 's  \ye{trade}  .\\ 
1&1&The number of refugees in the Middle East rose to 11 , 600  \ye{last}  year from 12 , 700 in 2012 ,  \ye{according}  to the\\ 
\bottomrule 
 \end{tabular} 
 \caption{\label{table:generation-China }
 Randomly selected generations from the single-word constraint task for the word ``China'' (with occurrence probability $1/10^2$) highlighted in \g{green}. Tokens are highlighted with \ye{y}\yf{e}\yf{l}\yg{l}\yg{o}\yj{w} with different intensities to indicate their overall frequencies in the generated corpus. $\phi(x)=1$ indicates the satisfaction of the constraint in the sample and reps the number of its repetitions across all generations.
 }\end{table}
\begin{table}[H]
        \scriptsize
        \begin{tabular}{p{0.5cm}|p{0.3cm}p{11.8cm}}
        \toprule
        reps & \textbf{$\phi(x)$} & \textbf{Generation} \\ 
         \midrule
 \multicolumn{3}{c}{\textbf{GDC}} \\ 
1&1&Honeycomb , MI –  \g{US}  District Judge Charles Brown has decided to allow a woman to remain in a mental health facility and is demanding\\ 
1&1& \g{US}   \ye{Secretary}  of  \ye{State}  Rex Tillerson on Tuesday condemned Russia's alleged hacking of the Democratic  \ye{National}  Committee and WikiLeaks , saying he  \ye{would}  "never" interfere\\ 
1&1&Roughly  \ye{one}  third of  \g{US}  households now live in counties  \ye{with}  more  \ye{than}  10 , 000 residents . While many  \ye{people}  living in the  \g{US} \\ 
1&1&'The  \g{US}  has always been concerned about the 'war on terror' that began under  \ye{President}  Bill Clinton , so the idea that  \ye{President}   \ye{Trump}  will\\ 
1&1&The  \g{US}  has already dispatched 15 , 000 troops to Afghanistan ,  \ye{with}  its  \ye{military}  presence set to reach 8 million by 2026 , said\\ 
1&1& \g{US}   \ye{President}   \ye{Barack}   \ye{Obama}  arrives at the  \g{US}  Embassy in Kabul , Afghanistan , on July 3 , 2011 . MOSCOW (Sputnik) {-}  \ye{President}  Barack\\ 
1&1& \g{US}   \ye{Secretary}  of  \ye{State}   \ye{John}  Kerry meets  \ye{with}   \g{US}  ambassador to the UN Nikki Haley (L) in the Rose Garden of the  \ye{White}   \ye{House}  in\\ 
1&1& \g{US}  Sen . Jeff Flake has  \ye{said}  he is "convinced" that  \ye{President}   \ye{Trump}  is using his bully pulpit to silence critics of his administration .\\ 
 \hline
 \multicolumn{3}{c}{\textbf{REINFORCE}} \\ 
1&1&The  \g{US} 's  \ye{ally's}   \g{US} 's  \yg{last}  in the  \g{US} 's  \yg{presence}  in a  \g{US} 's  \yg{first}  in a  \g{US} 's to the  \ye{East's}  and in the  \yg{first}  for the\\ 
1&1&The  \g{US} 's  \ye{one}  of the  \ye{world's}  countries's most high{-}res to the  \g{US} 's  \ye{recently}  in the city's  \ye{air}  in the  \ye{West's}   \yg{last}  for  \ye{another}  and's move\\ 
1&1&The  \g{US} 's  \yg{first}   \g{US} 's  \g{US} 's  \ye{claimed}  to the  \ye{former}  and then the  \yg{first}  to a  \g{US} 's in a  \ye{recently}  added to  \g{US} 's of the  \g{US} 's\\ 
1&1&and a  \g{US} 's  \ye{call}  to the  \g{US} 's  \ye{ally's}   \ye{allies}  in the  \g{US} 's  \yg{move}  in the  \g{US} 's  \ye{take}  in the  \g{US} 's the  \ye{former}  of the  \g{US} 's\\ 
1&1&In the  \g{US} 's support of the  \g{US} 's "conclusive in the  \g{US} 's  \ye{continued}  continued to  \g{US} 's  \yg{use}  of the  \yg{use}  of the  \g{US} 's  \g{US} 's  \ye{continued}  to\\ 
1&1&The  \g{US} 's  \g{US} 's  \g{US} 's in  \g{US} 's  \ye{recently}  strong in the  \g{US} 's  \ye{intervention}  in a  \g{US} 's  \g{US} 's  \ye{own}  and's  \ye{ally's}   \ye{take}  in to the  \g{US} 's  \g{US} 's\\ 
1&1&The  \g{US} 's and the  \g{US} 's  \ye{western's}  to the  \g{US} 's and the  \ye{one}  in the  \yg{presence}  of an  \g{US} 's's to  \ye{North's}   \ye{claimed}  the  \ye{recently}  in the\\ 
1&1&The  \g{US} 's and the  \g{US} 's's  \g{US} 's  \ye{very}  strict to the  \g{US} 's  \g{US} 's  \ye{recently}  and in the end of a  \ye{one}  to get the  \yg{first}  to\\ 
 \hline
 \multicolumn{3}{c}{\textbf{REINFORCE\_P(x)}} \\ 
327&1& \g{US}   \yg{President}   \yg{Donald}   \yg{Trump}   \yg{Donald}   \yg{John}   \yg{TrumpTop}   \yg{consumer}   \yg{bureau}   \yg{official}   \yg{blasts}   \yg{colleague}   \yg{over}   \yg{blog}   \yg{posts}   \yg{dismissing}   \yg{racism}   \yg{Trump}   \yg{'baby}   \yg{blimp'}   \yg{going}  to Trump{-}baby  \ye{blimp}  'baby\\ 
6&1& \g{US}   \yg{President}   \yg{Donald}   \yg{Trump}   \yg{Donald}   \yg{John}   \yg{TrumpTop}   \yg{consumer}   \yg{bureau}   \yg{official}   \yg{blasts}   \yg{colleague}   \yg{over}   \yg{blog}   \yg{posts}   \yg{dismissing}   \yg{racism}   \yg{Trump}   \yg{'baby}   \yg{blimp'}   \yg{going}  to  \yg{Trump}   \yg{'baby}  blimp'\\ 
327&1& \g{US}   \yg{President}   \yg{Donald}   \yg{Trump}   \yg{Donald}   \yg{John}   \yg{TrumpTop}   \yg{consumer}   \yg{bureau}   \yg{official}   \yg{blasts}   \yg{colleague}   \yg{over}   \yg{blog}   \yg{posts}   \yg{dismissing}   \yg{racism}   \yg{Trump}   \yg{'baby}   \yg{blimp'}   \yg{going}  to Trump{-}baby  \ye{blimp}  'baby\\ 
2&1& \g{US}   \yg{President}   \yg{Donald}   \yg{Trump}   \yg{Donald}   \yg{John}   \yg{TrumpTop}   \yg{consumer}   \yg{bureau}   \yg{official}   \yg{blasts}   \yg{colleague}   \yg{over}   \yg{blog}   \yg{posts}   \yg{dismissing}   \yg{racism}   \yg{Trump}   \yg{'baby}   \yg{blimp'}   \yg{going}  to  \ye{Washington}   \yg{'baby}  blimp'\\ 
83&1& \g{US}   \yg{President}   \yg{Donald}   \yg{Trump}   \yg{Donald}   \yg{John}   \yg{TrumpTop}   \yg{consumer}   \yg{bureau}   \yg{official}   \yg{blasts}   \yg{colleague}   \yg{over}   \yg{blog}   \yg{posts}   \yg{dismissing}   \yg{racism}   \yg{Trump}   \yg{'baby}   \yg{blimp'}   \yg{going}  to  \yg{bureau}   \yg{over}  racism\\ 
105&1& \g{US}   \yg{President}   \yg{Donald}   \yg{Trump}   \yg{Donald}   \yg{John}   \yg{TrumpTop}   \yg{consumer}   \yg{bureau}   \yg{official}   \yg{blasts}   \yg{colleague}   \yg{over}   \yg{blog}   \yg{posts}   \yg{dismissing}   \yg{racism}   \yg{Trump}   \yg{'baby}   \yg{blimp'}   \yg{going}  to  \ye{bureau'}   \yg{Trump}  'baby\\ 
107&1& \g{US}   \yg{President}   \yg{Donald}   \yg{Trump}   \yg{Donald}   \yg{John}   \yg{TrumpTop}   \yg{consumer}   \yg{bureau}   \yg{official}   \yg{blasts}   \yg{colleague}   \yg{over}   \yg{blog}   \yg{posts}   \yg{dismissing}   \yg{racism}   \yg{Trump}   \yg{'baby}   \yg{blimp'}   \yg{going}  to Trump{-}baby  \ye{blimp}  'baby\\ 
1&1& \g{US}   \yg{President}   \yg{Donald}   \yg{Trump}   \yg{Donald}   \yg{John}   \yg{TrumpTop}   \yg{consumer}   \yg{bureau}   \yg{official}   \yg{blasts}   \yg{colleague}   \yg{over}   \yg{blog}   \yg{posts}   \yg{dismissing}   \yg{racism}   \yg{Trump}   \yg{'baby}   \yg{blimp'}   \yg{going}  to see bureau{-}busting blog\\ 
 \hline
 \multicolumn{3}{c}{\textbf{Ziegler}} \\ 
1&1&The  \g{US}   \ye{Supreme}   \ye{Court}  on  \ye{Wednesday}  asked  \ye{President}   \ye{Barack}   \ye{Obama}  to grant a temporary ban on entry to Iran , effectively shutting down a pathway\\ 
1&1& \g{US}  Navy has just completed an inspection of a  \g{US}  Navy F/A{-}18 Hornets and an F/A{-}18G Growler ,  \ye{two}  F/A{-}18B , one F/\\ 
1&1&By Tom Brown The  \g{US}   \ye{government}  on Thursday ordered a search for the missing plane and  \ye{said}  its search had become one of the most\\ 
1&1& \g{US}   \ye{President}   \ye{Donald}   \ye{Trump}  has given \$75 million to local communities , the  \ye{first}  time the billionaire has given the  \g{US}  a sum of money\\ 
1&1&The  \g{US}  has  \ye{announced}  that it will launch a  \ye{new}  drone  \ye{war}  game for the game "Call of Duty : Black Ops 2 , "\\ 
1&1&A group of Chinese{-}Americans has sued  \g{US}   \ye{President}   \ye{Donald}   \ye{Trump}  in a bid to force him to pay their  \ye{former}  student visa fees . Chinese\\ 
1&1&A U . S .  \ye{Army}  soldier who  \ye{was}  killed in Iraq is the second  \g{US}  soldier to be killed in the  \ye{country}  since January\\ 
1&1&Haitian  \ye{officials}  are trying to make sure the  \g{US}  forces who stormed Iraq will be held responsible for their actions . They want the  \g{US} \\ 
\bottomrule 
 \end{tabular} 
 \caption{Randomly selected generations from the single-word constraint task for the word ``US'' (with occurrence probability $1/10^2$) highlighted in \g{green}. Tokens are highlighted with \ye{y}\yf{e}\yf{l}\yg{l}\yg{o}\yj{w} with different intensities to indicate their overall frequencies in the generated corpus. $\phi(x)=1$ indicates the satisfaction of the constraint in the sample and reps the number of its repetitions across all generations. 
 \label{table:generation-US }}\end{table}
\begin{table}[H]
        \scriptsize
        \begin{tabular}{p{0.5cm}|p{0.3cm}p{11.8cm}}
        \toprule
        reps & \textbf{$\phi(x)$} & \textbf{Generation} \\ 
         \midrule
 \multicolumn{3}{c}{\textbf{GDC}} \\ 
1&0&The next step is to build an implementation of the  \ye{new}  API . The API requires a special key called "v1\_hint" (which is a small\\ 
1&1&Praise be to Allaah . A man should  \ye{know}  that he is the only  \ye{one}  that knows the truth and he  \yg{can}  say whatever he\\ 
1&0&Forum Jump User Control Panel Private Messages Subscriptions Who's Online Search Forums Forums Home {-} Forums {-} Naturals {-} Naturals {-} Naturals {-} Naturals {-}\\ 
1&1&We all  \ye{know}  that a  \ye{lot}  of  \ye{people}  don't love to live in poverty , or even  \ye{know}   \ye{where}  to live , or even know\\ 
1&0&To view these statistics , click here . Team Totals How do you rate each team on this page ? We are a team with\\ 
1&1&To  \ye{help}  you better understand how we  \yg{can}  provide you  \ye{with}  the  \ye{best}  service for  \ye{your}  business , we've created an interactive version of this\\ 
1&1&The "Saving Christmas" campaign has launched and you  \yg{can}   \ye{make}  a donation here . The campaign is led by a number of  \ye{people}  that want\\ 
1&1&2 . If  \ye{you're}  having trouble finding a car , you  \yg{can}   \ye{use}  the Google Drive app to look for the driver , rather than\\ 
 \hline
 \multicolumn{3}{c}{\textbf{REINFORCE}} \\ 
1&1&You  \yh{can}   \ye{get}  the same effects as  \ye{possible}   \ye{with}  the  \ye{following}   \ye{syntax}  : Syntax : (You  \yh{can}   \yg{learn}   \ye{code}  . If you  \yh{can}   \ye{write}  code\\ 
1&1&There are  \ye{also}  ways to  \ye{build}   \ye{objects}  . You  \yh{can}   \yg{learn}  about HTMLText ,  \ye{which}   \yh{can}  be a powerful resource system . You  \yh{can}  learn\\ 
1&1&This article will be updated as we  \yh{can}   \yg{learn}  how to  \yg{learn}  how to  \ye{build}  new projects  \ye{using}  HTMLMLMLPets and  \ye{JavaScript}  . You  \yh{can}  learn\\ 
1&1&You  \yh{can}   \ye{find}  a  \ye{lot}  of  \ye{text}   \yh{can}  be  \ye{converted}  to HTMLLists . You  \yh{can}   \yg{learn}  a  \ye{lot}  of  \ye{XML}   \ye{using}   \ye{JavaScript}  that  \yh{can}  be\\ 
1&1&You  \yh{can}   \ye{use}   \ye{some}  of the features that  \yh{can}  be  \ye{available}   \ye{through}  APIs . You  \yh{can}   \ye{build}   \ye{your}  own custom expressions . You  \yh{can}  learn\\ 
1&1&If you  \yh{can}   \ye{see}  the contents of this  \ye{code}   \yh{can}  be  \ye{able}  to be  \ye{easily}  translated  \ye{with}  C++11AssemblyConstant . Learn the  \ye{syntax}  . You can\\ 
1&1&The method  \yh{can}  be broken by specifying an integer value . You  \yh{can}   \ye{also}   \yg{learn}  what  \yh{can}  happen if you  \yh{can}   \ye{get}   \ye{information}  from other\\ 
1&1&For more  \ye{information}  please  \ye{see}  Wikipedia : ExtractorCodeAccessibilitySpace  \ye{code}  . You  \yh{can}   \ye{find}   \ye{some}  resources you  \yh{can}   \ye{use}  as a base64code tree . This can\\ 
 \hline
 \multicolumn{3}{c}{\textbf{REINFORCE\_P(x)}} \\ 
10000&1&ES  \yg{News}   \yh{Email}   \yg{Enter}   \yg{your}   \yg{email}   \yh{address}   \yg{Please}   \yg{enter}  an  \yg{email}   \yh{address}   \yh{Email}   \yh{address}  is  \yg{invalid}   \yg{Fill}   \yg{out}  this  \yg{field}   \yh{Email}   \yh{address}  is  \yg{invalid}  Email\\ 
10000&1&ES  \yg{News}   \yh{Email}   \yg{Enter}   \yg{your}   \yg{email}   \yh{address}   \yg{Please}   \yg{enter}  an  \yg{email}   \yh{address}   \yh{Email}   \yh{address}  is  \yg{invalid}   \yg{Fill}   \yg{out}  this  \yg{field}   \yh{Email}   \yh{address}  is  \yg{invalid}  Email\\ 
10000&1&ES  \yg{News}   \yh{Email}   \yg{Enter}   \yg{your}   \yg{email}   \yh{address}   \yg{Please}   \yg{enter}  an  \yg{email}   \yh{address}   \yh{Email}   \yh{address}  is  \yg{invalid}   \yg{Fill}   \yg{out}  this  \yg{field}   \yh{Email}   \yh{address}  is  \yg{invalid}  Email\\ 
10000&1&ES  \yg{News}   \yh{Email}   \yg{Enter}   \yg{your}   \yg{email}   \yh{address}   \yg{Please}   \yg{enter}  an  \yg{email}   \yh{address}   \yh{Email}   \yh{address}  is  \yg{invalid}   \yg{Fill}   \yg{out}  this  \yg{field}   \yh{Email}   \yh{address}  is  \yg{invalid}  Email\\ 
10000&1&ES  \yg{News}   \yh{Email}   \yg{Enter}   \yg{your}   \yg{email}   \yh{address}   \yg{Please}   \yg{enter}  an  \yg{email}   \yh{address}   \yh{Email}   \yh{address}  is  \yg{invalid}   \yg{Fill}   \yg{out}  this  \yg{field}   \yh{Email}   \yh{address}  is  \yg{invalid}  Email\\ 
10000&1&ES  \yg{News}   \yh{Email}   \yg{Enter}   \yg{your}   \yg{email}   \yh{address}   \yg{Please}   \yg{enter}  an  \yg{email}   \yh{address}   \yh{Email}   \yh{address}  is  \yg{invalid}   \yg{Fill}   \yg{out}  this  \yg{field}   \yh{Email}   \yh{address}  is  \yg{invalid}  Email\\ 
10000&1&ES  \yg{News}   \yh{Email}   \yg{Enter}   \yg{your}   \yg{email}   \yh{address}   \yg{Please}   \yg{enter}  an  \yg{email}   \yh{address}   \yh{Email}   \yh{address}  is  \yg{invalid}   \yg{Fill}   \yg{out}  this  \yg{field}   \yh{Email}   \yh{address}  is  \yg{invalid}  Email\\ 
10000&1&ES  \yg{News}   \yh{Email}   \yg{Enter}   \yg{your}   \yg{email}   \yh{address}   \yg{Please}   \yg{enter}  an  \yg{email}   \yh{address}   \yh{Email}   \yh{address}  is  \yg{invalid}   \yg{Fill}   \yg{out}  this  \yg{field}   \yh{Email}   \yh{address}  is  \yg{invalid}  Email\\ 
 \hline
 \multicolumn{3}{c}{\textbf{Ziegler}} \\ 
1&1&At this stage of the  \ye{game}  , the idea is that you  \yg{can}   \ye{use}  these  \ye{things}  like I've seen them done before and in an\\ 
4542&1&ES  \yg{News}   \yg{Email}   \yg{Enter}   \yg{your}   \yg{email}   \yg{address}   \yg{Please}   \yg{enter}  an  \yg{email}   \yg{address}   \yg{Email}   \yg{address}  is  \yg{invalid}   \yg{Fill}   \yg{out}  this  \yg{field}   \yg{Email}   \yg{address}  is  \yg{invalid}  Email\\ 
1&1&I've tried to  \ye{get}  my hands on this little gem! I used it as a tool for making a table in a restaurant but I\\ 
4542&1&ES  \yg{News}   \yg{Email}   \yg{Enter}   \yg{your}   \yg{email}   \yg{address}   \yg{Please}   \yg{enter}  an  \yg{email}   \yg{address}   \yg{Email}   \yg{address}  is  \yg{invalid}   \yg{Fill}   \yg{out}  this  \yg{field}   \yg{Email}   \yg{address}  is  \yg{invalid}  Email\\ 
4542&1&ES  \yg{News}   \yg{Email}   \yg{Enter}   \yg{your}   \yg{email}   \yg{address}   \yg{Please}   \yg{enter}  an  \yg{email}   \yg{address}   \yg{Email}   \yg{address}  is  \yg{invalid}   \yg{Fill}   \yg{out}  this  \yg{field}   \yg{Email}   \yg{address}  is  \yg{invalid}  Email\\ 
1&1&Sitting at home on an outdoor treadmill , she  \yg{can}  barely read the message at the top of the screen . You'll notice it's just\\ 
4542&1&ES  \yg{News}   \yg{Email}   \yg{Enter}   \yg{your}   \yg{email}   \yg{address}   \yg{Please}   \yg{enter}  an  \yg{email}   \yg{address}   \yg{Email}   \yg{address}  is  \yg{invalid}   \yg{Fill}   \yg{out}  this  \yg{field}   \yg{Email}   \yg{address}  is  \yg{invalid}  Email\\ 
1&1&If  \ye{you're}  like me , you need to  \ye{know}  more about the  \ye{game}  itself and the various settings that  \yg{can}  help you learn more about\\ 
\bottomrule 
 \end{tabular} 
 \caption{Randomly selected generations from the word-list constraint task for the \textbf{kitchen} word-list. Tokens are highlighted with \ye{y}\yf{e}\yf{l}\yg{l}\yg{o}\yj{w} with different intensities to indicate their overall frequencies in the generated corpus. $\phi(x)=1$ indicates the satisfaction of the constraint in the sample and reps the number of its repetitions across all generations.
 \label{table:generation-kitchen }}\end{table}
\begin{table}[H]
        \scriptsize
        \begin{tabular}{p{0.5cm}|p{0.3cm}p{11.8cm}}
        \toprule
        reps & \textbf{$\phi(x)$} & \textbf{Generation} \\ 
         \midrule
 \multicolumn{3}{c}{\textbf{GDC}} \\ 
1&1&I got these  \ye{last}  year  \ye{when}   \ye{they}   \ye{were}  \$500 , but I didn't  \ye{get}  a  \ye{monster}   \ye{when}   \ye{they}  went  \ye{out}  in 2012 , so this\\ 
1&0&A  \ye{man}  who appeared in a video calling on supporters to be loyal to the Muslim faith is being attacked by an attacker who then\\ 
1&1&The  \ye{ghost}  of her father is here , and it's  \ye{time}  to  \ye{get}  a  \ye{ghost}  back . If  \ye{she}  ever does return , she'll be\\ 
1&1&Fancy the way you play  \ye{with}  a  \ye{ghost}  of a  \ye{game}  to  \ye{get}   \ye{some}   \ye{new}  stuff ? Get it here! Check  \ye{out}  the rest of\\ 
1&0&The American Red Cross is among the  \ye{first}  to warn against the increasing prevalence of heart attacks among gay men and lesbians in a national\\ 
1&1&"The devil's still  \ye{out}   \ye{there}  , " says the narrator , "the good man's not the only  \ye{one}  to  \ye{see}  his  \ye{ghost}  . His ghost\\ 
1&1&The "Star Wars"  \ye{horror}  series is getting a  \yg{giant}  facelift for its third{-}season finale ,  \ye{with}  the show now featuring a  \yg{giant}  ,  \yg{giant}  alien\\ 
1&0&As we've seen from the beginning of the Kickstarter , the concept for the  \ye{new}   \ye{game}   \ye{was}  conceived by artist Matt McEwen . Matt began\\ 
 \hline
 \multicolumn{3}{c}{\textbf{REINFORCE}} \\ 
1&1&New anime  \ye{film}  from  \yg{horror}   \yg{epic}   \ye{assassin}   \yh{monster}   \ye{King}  of  \ye{monsters}  and  \yh{monster}   \ye{hunters}   \ye{King}  Hell's  \yg{killer}   \ye{Hell}  and  \yg{killer}   \yg{vampire}   \yg{demon}   \yh{monster}  Rambo and\\ 
1&1&The  \ye{first}   \ye{zombie}   \ye{drama}  of  \yg{horror}  horror  \yg{horror}   \ye{adventure}   \yg{horror}  horror  \yg{horror}  films is filled  \ye{with}   \yh{monster}   \yg{killer}  and  \yg{vampire}  and  \yg{killer}   \yg{warrior}  the monster\\ 
1&1&The  \yg{killer}  killer  \yg{horror}   \ye{saga}  of  \yh{monster}  monster  \yg{hunter}  and  \yg{horror}   \yh{monster}   \ye{assassin}  assassin and  \yh{monster}  monster of  \yg{horror}  and his  \yg{vampire}   \yg{king}  , Dr\\ 
1&1&The  \ye{legendary}   \yh{monster}   \yg{killer}   \yg{hunter}  and  \yh{monster}  hunting survivor ,  \yh{monster}   \yg{hunter}   \ye{Angel}  of fire and  \yh{monster}   \ye{hunters}  and his  \ye{priest}   \yg{warrior}   \yh{monster}   \ye{Constantine}  returns\\ 
1&1&He has a  \yh{monster}  of a  \yg{horror}  horror  \ye{film}   \yg{horror}   \ye{flick}  and his  \yg{vampire}   \yg{hunter}  is  \yh{monster}  monster  \ye{leader}   \ye{Hell}  ,  \ye{Dracula}  of  \ye{Dracula}  and\\ 
1&1&It is happening in a  \ye{nightmare}  of  \yg{horror}  horror  \yg{horror}   \yg{epic}   fantasy   \ye{adventure}   \ye{tale}   \yh{monster}  of  \yh{monster}   \ye{hunters}  and  \yh{monster}   \yg{hunter}  and  \ye{priest}  of a\\ 
1&1&The mythological  \yh{monster}  from  \yg{horror}  horror  \yg{epic}  of  \yg{horror}   \ye{saga}   \yg{horror}  director Roland and  \yh{monster}  monster of  \yh{monster}  of an  \yh{monster}  monster (Reichling (Robert and\\ 
1&1&The  \yg{horror}  horror  \ye{thriller}   \ye{flick}   \yh{monster}  of  \yh{monster}  assassins and  \yh{monster}  and  \yg{vampire}   \ye{priest}   \ye{Freddy}  himself and his  \ye{murderous}   \yg{killer}  ,  \yh{monster}   \yg{king}  and monster\\ 
 \hline
 \multicolumn{3}{c}{\textbf{REINFORCE\_P(x)}} \\ 
10000&1&Stats  \yg{Growth}   \yg{Chart}   \yg{HP}  :  \yg{Normal}  ATK :  \yg{Normal}  RCV :  \yg{Normal}   \yg{HP}   \yh{|}   \yg{Attack}   \yh{|}   \yg{Recover}   \yg{vs}   \yg{Level}   \yg{HP}   \yh{|}   \yg{Attack}   \yh{|}   \yg{Recover}  vs\\ 
10000&1&Stats  \yg{Growth}   \yg{Chart}   \yg{HP}  :  \yg{Normal}  ATK :  \yg{Normal}  RCV :  \yg{Normal}   \yg{HP}   \yh{|}   \yg{Attack}   \yh{|}   \yg{Recover}   \yg{vs}   \yg{Level}   \yg{HP}   \yh{|}   \yg{Attack}   \yh{|}   \yg{Recover}  vs\\ 
10000&1&Stats  \yg{Growth}   \yg{Chart}   \yg{HP}  :  \yg{Normal}  ATK :  \yg{Normal}  RCV :  \yg{Normal}   \yg{HP}   \yh{|}   \yg{Attack}   \yh{|}   \yg{Recover}   \yg{vs}   \yg{Level}   \yg{HP}   \yh{|}   \yg{Attack}   \yh{|}   \yg{Recover}  vs\\ 
10000&1&Stats  \yg{Growth}   \yg{Chart}   \yg{HP}  :  \yg{Normal}  ATK :  \yg{Normal}  RCV :  \yg{Normal}   \yg{HP}   \yh{|}   \yg{Attack}   \yh{|}   \yg{Recover}   \yg{vs}   \yg{Level}   \yg{HP}   \yh{|}   \yg{Attack}   \yh{|}   \yg{Recover}  vs\\ 
10000&1&Stats  \yg{Growth}   \yg{Chart}   \yg{HP}  :  \yg{Normal}  ATK :  \yg{Normal}  RCV :  \yg{Normal}   \yg{HP}   \yh{|}   \yg{Attack}   \yh{|}   \yg{Recover}   \yg{vs}   \yg{Level}   \yg{HP}   \yh{|}   \yg{Attack}   \yh{|}   \yg{Recover}  vs\\ 
10000&1&Stats  \yg{Growth}   \yg{Chart}   \yg{HP}  :  \yg{Normal}  ATK :  \yg{Normal}  RCV :  \yg{Normal}   \yg{HP}   \yh{|}   \yg{Attack}   \yh{|}   \yg{Recover}   \yg{vs}   \yg{Level}   \yg{HP}   \yh{|}   \yg{Attack}   \yh{|}   \yg{Recover}  vs\\ 
10000&1&Stats  \yg{Growth}   \yg{Chart}   \yg{HP}  :  \yg{Normal}  ATK :  \yg{Normal}  RCV :  \yg{Normal}   \yg{HP}   \yh{|}   \yg{Attack}   \yh{|}   \yg{Recover}   \yg{vs}   \yg{Level}   \yg{HP}   \yh{|}   \yg{Attack}   \yh{|}   \yg{Recover}  vs\\ 
10000&1&Stats  \yg{Growth}   \yg{Chart}   \yg{HP}  :  \yg{Normal}  ATK :  \yg{Normal}  RCV :  \yg{Normal}   \yg{HP}   \yh{|}   \yg{Attack}   \yh{|}   \yg{Recover}   \yg{vs}   \yg{Level}   \yg{HP}   \yh{|}   \yg{Attack}   \yh{|}   \yg{Recover}  vs\\ 
 \hline
 \multicolumn{3}{c}{\textbf{Ziegler}} \\ 
1&1&This is a  \ye{great}   \ye{way}  to explore the life of this  \ye{world}  . I  \yg{was}  a  \ye{very}  happy person , happy because I  \yg{was}  the\\ 
1&1&I'll  \ye{get}  into the beast of the bush in a bit , but in the  \ye{last}  few minutes I've got a pretty  \ye{good}  feel for\\ 
1&1&I am a big fan of the fantasy genre , but that is a topic for another  \ye{time}  . I  \ye{can}  tell you that I\\ 
1&1&In the years that followed , the Internet  \yg{was}  transformed by the advent of the Internet in 1999 ,  \ye{with}  Facebook (FB) and Google (GOOGL)\\ 
1&1&A strange  \ye{ghost}  is haunting the ruins of ancient Babylon . In  \ye{one}  of those  \ye{horror}  movies , a  \ye{ghost}  is caught in a mysterious\\ 
1&0&"We're seeing that now in the case of Syria , " the judge  \ye{said}  . "That's why the State of Canada should not take it\\ 
1&0&"The  \ye{world}  should stop playing dead . The  \ye{world}  should start playing alive . " That  \yg{was}  the line of the voice that emerged from\\ 
1&1&I  \ye{just}  wanted to try it  \ye{out}  . I'm so excited about it and  \ye{just}  started a  \ye{new}   \ye{game}  , and it works . It's\\ 
\bottomrule 
 \end{tabular} 
 \caption{Randomly selected generations from the word-list constraint task for the \textbf{fantasy} word-list. Tokens are highlighted with \ye{y}\yf{e}\yf{l}\yg{l}\yg{o}\yj{w} with different intensities to indicate their overall frequencies in the generated corpus. $\phi(x)=1$ indicates the satisfaction of the constraint in the sample and reps the number of its repetitions across all generations.
 \label{table:generation-fantasy }}\end{table}
\begin{table}[H]
        \scriptsize
        \begin{tabular}{p{0.5cm}|p{0.3cm}p{11.8cm}}
        \toprule
        reps & \textbf{$\phi(x)$} & \textbf{Generation} \\ 
         \midrule
 \multicolumn{3}{c}{\textbf{GDC}} \\ 
1&1&The  \ye{government}  will give three days' notice to banks for taking off all their shares in the private sector , the prime minister  \ye{said}  ,\\ 
1&1&. (1) A person may , to the extent that the person believes that an action or proceeding will be taken against him or her\\ 
1&1&The  \ye{state}  of Colorado is now in an emergency  \ye{state}  after wildfires burned across the  \ye{state's}  four states . As of Thursday night , the\\ 
1&0&U . S . Attorney General Jeff Sessions (C) , the chief  \ye{law}  enforcement official in the U . S . , speaks in the\\ 
1&1&MADISON {-}{-} A  \ye{state}  appeals  \ye{court}  in Milwaukee struck down a ban on abortions nationwide . A ruling from the  \ye{state}  Supreme Court is expected\\ 
1&1&The  \ye{court}  had handed down its decision earlier this week . The  \ye{state}  of California and a group of  \ye{state}  attorneys general have filed a\\ 
1&1&The  \ye{United}  States has  \ye{said}  it  \ye{would}  send 2 , 500 Special Forces troops to Iraq to train Iraqi security forces . The decision is\\ 
1&1&"We all know what happened to the White  \ye{House}  on October 21 , 2009 . All we know is that Donald  \ye{Trump}   \ye{was}  elected President\\ 
 \hline
 \multicolumn{3}{c}{\textbf{REINFORCE}} \\ 
1&1&In a  \ye{major}  development in  \ye{government's}   \ye{attempt}  to  \ye{block}  further progress in the  \ye{process}  of nationalisation of its commerce , the  \yg{state}   \yg{government}  , in\\ 
1&1&The  \yg{government}  may not prosecute a group of government{-}owned enterprises for its  \yg{political}  ,  \yg{economic}  , or  \yg{administrative}  purposes in its  \ye{national}   \ye{economy}  . Article\\ 
1&1&The  \ye{United}   \ye{States}   \yg{government}  has  \ye{ordered}  a  \ye{court}   \yg{order}  to  \ye{enforce}   \yg{state}  laws or governmental  \yg{power}   \ye{over}  the personal conduct of its  \yg{political}   \ye{subdivision}  in\\ 
1&1&The  \yg{government}  has  \ye{ordered}  an  \yg{order}  on its release of a dozen  \yg{government}  ministers in  \ye{attempts}  to  \ye{block}  its  \ye{operation}  in  \ye{judicial}  proceedings in its\\ 
1&1&The  \yg{state}   \ye{government's}  monopoly on its  \yg{economic}   \yg{power}   \ye{over}  the  \yg{political}  ,  \yg{economic}  , or  \yg{administrative}   \ye{process}  in  \yg{order}  of its  \ye{citizens}  in  \yg{order}  to\\ 
1&1&In its  \ye{attempt}  to  \ye{block}   \ye{access}  to the  \yg{state}   \yg{government}  in its  \yg{political}   \ye{action}  ,  \yg{government}  made an  \ye{attempt}  to  \ye{restrict}   \yg{economic}   \ye{activity}  in order\\ 
1&1&The  \yg{government}  will invoke its  \ye{powers}  against the  \yg{government}  in  \ye{court}  of  \ye{India}  against its  \yg{order}   \ye{seeking}  a  \yg{order}  in its  \ye{internal}   \yg{order}  in its\\ 
1&1&In its campaign against  \yg{economic}  independence in its  \yg{efforts}  to  \ye{enforce}  an effective  \yg{state}  monopoly on its  \yg{political}   \yg{power}  in its  \yg{state}  , the Government\\ 
 \hline
 \multicolumn{3}{c}{\textbf{REINFORCE\_P(x)}} \\ 
10000&1&Processing  \yg{time}  The  \yg{time}  I  \yg{need}  to  \yg{prepare}  an  \yg{order}  for  \yg{shipping}  varies . For details ,  \yg{see}   \yg{individual}  items .  \yg{Optional}   \yg{Estimated}   \yg{shipping}  times\\ 
10000&1&Processing  \yg{time}  The  \yg{time}  I  \yg{need}  to  \yg{prepare}  an  \yg{order}  for  \yg{shipping}  varies . For details ,  \yg{see}   \yg{individual}  items .  \yg{Optional}   \yg{Estimated}   \yg{shipping}  times\\ 
10000&1&Processing  \yg{time}  The  \yg{time}  I  \yg{need}  to  \yg{prepare}  an  \yg{order}  for  \yg{shipping}  varies . For details ,  \yg{see}   \yg{individual}  items .  \yg{Optional}   \yg{Estimated}   \yg{shipping}  times\\ 
10000&1&Processing  \yg{time}  The  \yg{time}  I  \yg{need}  to  \yg{prepare}  an  \yg{order}  for  \yg{shipping}  varies . For details ,  \yg{see}   \yg{individual}  items .  \yg{Optional}   \yg{Estimated}   \yg{shipping}  times\\ 
10000&1&Processing  \yg{time}  The  \yg{time}  I  \yg{need}  to  \yg{prepare}  an  \yg{order}  for  \yg{shipping}  varies . For details ,  \yg{see}   \yg{individual}  items .  \yg{Optional}   \yg{Estimated}   \yg{shipping}  times\\ 
10000&1&Processing  \yg{time}  The  \yg{time}  I  \yg{need}  to  \yg{prepare}  an  \yg{order}  for  \yg{shipping}  varies . For details ,  \yg{see}   \yg{individual}  items .  \yg{Optional}   \yg{Estimated}   \yg{shipping}  times\\ 
10000&1&Processing  \yg{time}  The  \yg{time}  I  \yg{need}  to  \yg{prepare}  an  \yg{order}  for  \yg{shipping}  varies . For details ,  \yg{see}   \yg{individual}  items .  \yg{Optional}   \yg{Estimated}   \yg{shipping}  times\\ 
10000&1&Processing  \yg{time}  The  \yg{time}  I  \yg{need}  to  \yg{prepare}  an  \yg{order}  for  \yg{shipping}  varies . For details ,  \yg{see}   \yg{individual}  items .  \yg{Optional}   \yg{Estimated}   \yg{shipping}  times\\ 
 \hline
 \multicolumn{3}{c}{\textbf{Ziegler}} \\ 
1&1&In 2006 , a group of  \ye{political}  scientists convened for a debate in Chicago about  \ye{whether}  the media  \ye{should}  "spill  \ye{political}  information into the public\\ 
1&1&It has taken several  \ye{years}  for the  \yg{government}  to finally acknowledge the real issues facing the Australian population . This is because the most pressing\\ 
1&0&We had hoped that the election  \ye{would}  be a simple one{-}sided affair between those who don't support the Republican Party and those who do .\\ 
1&1&The  \yg{government}  of Saskatchewan has a long history of lobbying on behalf of business interests . The province recently passed an omnibus  \ye{tax}   \ye{bill}  that\\ 
1&1&The NDP has taken the issue of  \ye{whether}  the  \ye{state}  has a "fundamental right" to free trade to the forefront in its annual platform ,\\ 
1&1&By Steve Nelles More  \ye{than}  two{-}thirds of Texans are expected to sign off on the  \ye{state's}  future  \ye{tax}  code in January ,  \ye{with}  a possible\\ 
1&1&An appeals  \ye{court}  in Ohio  \ye{ruled}  Monday that the  \ye{state's}  refusal to allow a transgender employee to  \ye{use}  the  \ye{state}  bathroom of her choice violated\\ 
1&1&The  \yg{government}  will set aside \$2 . 4  \ye{billion}  to fund more  \ye{than}  800 schools in the South African  \ye{state}  , including  \ye{many}  in the\\ 
\bottomrule 
 \end{tabular} 
 \caption{
 Randomly selected generations from the word-list constraint task for the \textbf{politics} word-list. Tokens are highlighted with \ye{y}\yf{e}\yf{l}\yg{l}\yg{o}\yj{w} with different intensities to indicate their overall frequencies in the generated corpus. $\phi(x)=1$ indicates the satisfaction of the constraint in the sample and reps the number of its repetitions across all generations.
 \label{table:generation-politics }
  }\end{table}
\begin{table}[H]
        \scriptsize
        \begin{tabular}{p{0.5cm}|p{0.3cm}p{11.8cm}}
        \toprule
        reps & \textbf{$\phi(x)$} & \textbf{Generation} \\ 
         \midrule
 \multicolumn{3}{c}{\textbf{GDC}} \\ 
1&1&{-} A member of a group of hackers linked to the NSA  \ye{said}  Tuesday that the NSA  \ye{was}  gathering  \ye{data}  from foreign communications of American\\ 
1&1&To view this video please enable JavaScript , and consider upgrading to a web browser that supports HTML5 video For those who  \ye{want}  to watch\\ 
1&1&3 . 4 . 5 . The  \ye{Windows}  RT app is now a part of the Store 3 . 4 . 4 . The Store\\ 
1&1&What is NAML ? NAML is a program to improve  \ye{data}  science and  \ye{make}  data{-}driven decisions more efficient , data{-}efficient , and cost{-}effective . The\\ 
1&1&Bike or bike ? Are you a  \ye{new}  commuter , looking for a  \ye{way}  to  \ye{get}   \ye{home}  from work and to  \ye{get}   \ye{home}  for the\\ 
1&1&A month and a half ago , a story about the  \ye{new}  computer chips that will replace the aging , aging old  \ye{Windows}  . Microsoft's\\ 
1&1&The US Supreme Court ruled in 2014 that the NSA's bulk collection of Internet  \ye{data}  could not be justified under the Fourth Amendment . In\\ 
1&1&We offer a wide variety of free shipping on select orders . This is the best deal on the planet . Our  \ye{online}  store provides\\ 
 \hline
 \multicolumn{3}{c}{\textbf{REINFORCE}} \\ 
1&1&Get the  \ye{Microsoft}   \yg{Windows}   \yg{computer}   \ye{update}  update or  \yg{Windows}  . Press Start{-}Windows  \yg{computer}   \ye{start}   \ye{button}  and  \yg{click}  on  \yg{Windows}   \yg{launch}  screen . Click on the\\ 
1&1&For  \yg{Windows}  , the user  \yg{can}   \yg{launch}  a  \ye{web}   \ye{browser}  or  \ye{PC}  or  \yg{Windows}   \yg{can}   \yg{launch}  the  \yg{Windows}   \ye{desktop}   \ye{web}   \yg{version}   \yg{Windows}  .  \yg{Windows}  and\\ 
1&1&The BlackBerry devices has been updated  \ye{with}  the  \ye{latest}   \ye{software}  . The  \yg{Windows}   \yg{computer}   \ye{may}   \ye{download}   \ye{software}   \yg{version}   \yg{Windows}  ,  \yg{Windows}  and  \yg{Windows}  , can\\ 
1&1&The  \ye{latest}   \yg{version}  of  \yg{Windows}   \yg{can}   \yg{launch}  the  \yg{computer}  .  \yg{Windows}   \yg{can}   \ye{install}  Windows's firmware or  \yg{Windows}  have a copy{-}and{-}paste  \ye{menu}   \ye{button}  in the start{-}up\\ 
1&1&An Apple  \yg{computer}  will  \yg{launch}   \ye{Microsoft's}  virtual  \yg{Windows}   \ye{operating}  system and  \yg{Windows}  . Launch in the  \ye{PC}  or mobile  \yg{Windows}  will  \yg{launch}  the  \yg{Windows}  app/Windows\\ 
1&1&You  \ye{may}  be  \ye{running}   \yg{Windows}  . Click  \yg{Windows}   \ye{menu}  in the  \yg{Windows}   \ye{PC}  or  \yg{computer}  ,  \yg{click}   \ye{Start}  ,  \ye{navigate}  to a  \ye{web}   \ye{browser}  launch\\ 
1&1&I've recently downloaded a  \yg{version}  of  \yg{Windows}  . The  \ye{OS}  ,  \yg{launch}   \ye{menu}  ,  \ye{start}   \ye{menu}  , or  \yg{Windows}  , drop the  \yg{Windows}  version{-}powered PC\\ 
1&1&During the  \ye{OS}   \ye{update}  , the  \ye{software}  and  \yg{Windows}  ,  \yg{launch}  ,  \ye{select}   \ye{"Windows}  ,  \yg{click}  Tools  \ye{menu}  ,  \yg{click}  on the "Remote  \ye{desktop}  option\\ 
 \hline
 \multicolumn{3}{c}{\textbf{REINFORCE\_P(x)}} \\ 
10000&1&ES  \yg{News}   \yh{Email}   \yg{Enter}   \yg{your}   \yg{email}   \yh{address}   \yg{Please}   \yg{enter}  an  \yg{email}   \yh{address}   \yh{Email}   \yh{address}  is  \yg{invalid}   \yg{Fill}   \yg{out}  this  \yg{field}   \yh{Email}   \yh{address}  is  \yg{invalid}  Email\\ 
10000&1&ES  \yg{News}   \yh{Email}   \yg{Enter}   \yg{your}   \yg{email}   \yh{address}   \yg{Please}   \yg{enter}  an  \yg{email}   \yh{address}   \yh{Email}   \yh{address}  is  \yg{invalid}   \yg{Fill}   \yg{out}  this  \yg{field}   \yh{Email}   \yh{address}  is  \yg{invalid}  Email\\ 
10000&1&ES  \yg{News}   \yh{Email}   \yg{Enter}   \yg{your}   \yg{email}   \yh{address}   \yg{Please}   \yg{enter}  an  \yg{email}   \yh{address}   \yh{Email}   \yh{address}  is  \yg{invalid}   \yg{Fill}   \yg{out}  this  \yg{field}   \yh{Email}   \yh{address}  is  \yg{invalid}  Email\\ 
10000&1&ES  \yg{News}   \yh{Email}   \yg{Enter}   \yg{your}   \yg{email}   \yh{address}   \yg{Please}   \yg{enter}  an  \yg{email}   \yh{address}   \yh{Email}   \yh{address}  is  \yg{invalid}   \yg{Fill}   \yg{out}  this  \yg{field}   \yh{Email}   \yh{address}  is  \yg{invalid}  Email\\ 
10000&1&ES  \yg{News}   \yh{Email}   \yg{Enter}   \yg{your}   \yg{email}   \yh{address}   \yg{Please}   \yg{enter}  an  \yg{email}   \yh{address}   \yh{Email}   \yh{address}  is  \yg{invalid}   \yg{Fill}   \yg{out}  this  \yg{field}   \yh{Email}   \yh{address}  is  \yg{invalid}  Email\\ 
10000&1&ES  \yg{News}   \yh{Email}   \yg{Enter}   \yg{your}   \yg{email}   \yh{address}   \yg{Please}   \yg{enter}  an  \yg{email}   \yh{address}   \yh{Email}   \yh{address}  is  \yg{invalid}   \yg{Fill}   \yg{out}  this  \yg{field}   \yh{Email}   \yh{address}  is  \yg{invalid}  Email\\ 
10000&1&ES  \yg{News}   \yh{Email}   \yg{Enter}   \yg{your}   \yg{email}   \yh{address}   \yg{Please}   \yg{enter}  an  \yg{email}   \yh{address}   \yh{Email}   \yh{address}  is  \yg{invalid}   \yg{Fill}   \yg{out}  this  \yg{field}   \yh{Email}   \yh{address}  is  \yg{invalid}  Email\\ 
10000&1&ES  \yg{News}   \yh{Email}   \yg{Enter}   \yg{your}   \yg{email}   \yh{address}   \yg{Please}   \yg{enter}  an  \yg{email}   \yh{address}   \yh{Email}   \yh{address}  is  \yg{invalid}   \yg{Fill}   \yg{out}  this  \yg{field}   \yh{Email}   \yh{address}  is  \yg{invalid}  Email\\ 
 \hline
 \multicolumn{3}{c}{\textbf{Ziegler}} \\ 
1001&1&ES  \ye{Football}   \ye{Newsletter}   \yg{Enter}   \yg{your}   \yg{email}   \yg{address}   \yg{Please}   \yg{enter}  an  \yg{email}   \yg{address}   \yg{Email}   \yg{address}  is  \yg{invalid}   \yg{Fill}   \yg{out}  this  \yg{field}   \yg{Email}   \yg{address}  is  \yg{invalid}  Email\\ 
1&1&The software is designed for use  \yg{with}   \ye{Windows}  , Mac , Linux , and OpenBSD . The software is designed for  \ye{Windows}  , Mac ,\\ 
1001&1&ES  \ye{Football}   \ye{Newsletter}   \yg{Enter}   \yg{your}   \yg{email}   \yg{address}   \yg{Please}   \yg{enter}  an  \yg{email}   \yg{address}   \yg{Email}   \yg{address}  is  \yg{invalid}   \yg{Fill}   \yg{out}  this  \yg{field}   \yg{Email}   \yg{address}  is  \yg{invalid}  Email\\ 
6654&1&ES  \yg{News}   \yg{Email}   \yg{Enter}   \yg{your}   \yg{email}   \yg{address}   \yg{Please}   \yg{enter}  an  \yg{email}   \yg{address}   \yg{Email}   \yg{address}  is  \yg{invalid}   \yg{Fill}   \yg{out}  this  \yg{field}   \yg{Email}   \yg{address}  is  \yg{invalid}  Email\\ 
6654&1&ES  \yg{News}   \yg{Email}   \yg{Enter}   \yg{your}   \yg{email}   \yg{address}   \yg{Please}   \yg{enter}  an  \yg{email}   \yg{address}   \yg{Email}   \yg{address}  is  \yg{invalid}   \yg{Fill}   \yg{out}  this  \yg{field}   \yg{Email}   \yg{address}  is  \yg{invalid}  Email\\ 
6654&1&ES  \yg{News}   \yg{Email}   \yg{Enter}   \yg{your}   \yg{email}   \yg{address}   \yg{Please}   \yg{enter}  an  \yg{email}   \yg{address}   \yg{Email}   \yg{address}  is  \yg{invalid}   \yg{Fill}   \yg{out}  this  \yg{field}   \yg{Email}   \yg{address}  is  \yg{invalid}  Email\\ 
6654&1&ES  \yg{News}   \yg{Email}   \yg{Enter}   \yg{your}   \yg{email}   \yg{address}   \yg{Please}   \yg{enter}  an  \yg{email}   \yg{address}   \yg{Email}   \yg{address}  is  \yg{invalid}   \yg{Fill}   \yg{out}  this  \yg{field}   \yg{Email}   \yg{address}  is  \yg{invalid}  Email\\ 
6654&1&ES  \yg{News}   \yg{Email}   \yg{Enter}   \yg{your}   \yg{email}   \yg{address}   \yg{Please}   \yg{enter}  an  \yg{email}   \yg{address}   \yg{Email}   \yg{address}  is  \yg{invalid}   \yg{Fill}   \yg{out}  this  \yg{field}   \yg{Email}   \yg{address}  is  \yg{invalid}  Email\\ 
\bottomrule 
 \end{tabular} 
 \caption{Randomly selected generations from the word-list constraint task for the \textbf{computers} word-list. Tokens are highlighted with \ye{y}\yf{e}\yf{l}\yg{l}\yg{o}\yj{w} with different intensities to indicate their overall frequencies in the generated corpus. $\phi(x)=1$ indicates the satisfaction of the constraint in the sample and reps the number of its repetitions across all generations. \label{table:generation-computers }}\end{table}
\begin{table}[H]
        \scriptsize
        \begin{tabular}{p{0.5cm}|p{0.3cm}p{11.8cm}}
        \toprule
        reps & \textbf{$\phi(x)$} & \textbf{Generation} \\ 
         \midrule
 \multicolumn{3}{c}{\textbf{GDC}} \\ 
1&1&I have to say I  \ye{was}  impressed  \ye{with}  the  \ye{way}  the writing and narration  \ye{was}  done . The  \ye{way}  they were presented , especially the\\ 
1&0&'I'm thrilled to say my  \ye{team}  is on the way!' tweeted Sadiq Khan . The London Mayor is joining the "Great London Olympics" movement to\\ 
1&1&You are  \ye{going}  to enjoy this book! It is a beautiful collection of beautifully detailed stories . It is a treasure trove of information for\\ 
1&0&It's a fascinating conversation that we have in the  \ye{world}  of cryptocurrency . It's so  \ye{much}  fun . The  \ye{people}  who have been running the\\ 
1&0&Tired of waiting for the next  \ye{best}  thing to happen , you  \ye{know}  it . You  \ye{want}  to  \ye{know}  . We are dedicated to helping\\ 
1&1&We  \ye{love}   \ye{your}  feedback , so we are pleased to bring you the most powerful and best{-}selling product that will satisfy  \ye{your}  needs and your\\ 
1&1&"Thank you all for the service this site gives me , " he  \ye{said}  . "Thank you for the  \ye{work}  I've been doing  \ye{with}  the\\ 
1&1&"The most  \ye{amazing}  thing about this  \ye{game}  is that there is no other  \ye{games}  that have been released  \ye{like}  this . It has such a\\ 
 \hline
 \multicolumn{3}{c}{\textbf{REINFORCE}} \\ 
1&1&Enhanced  \yg{performance}   \ye{with}  our world{-}renown world{-}renown  \ye{exhibitions}   \yg{worldwide}  . We believe our  \ye{clients}   \ye{with}  extraordinary  \yg{audiences}  of our  \yh{highest}   \yh{quality}   \yg{productions}  productions of  \ye{outstanding}  international\\ 
1&1&Dramatic  \yg{high}   \yh{quality}   \yg{performance}   \yh{quality}   \yg{products}  of  \yg{leading}   \yg{global}   \yg{international}   \yg{audiences}  of the  \yh{highest}   \yh{quality}   \yg{high}   \yh{quality}   \yg{high}   \yh{quality}   \yg{international}   \yg{leading}   \yg{worldwide}  markets  \yg{leading}  global\\ 
1&1&Create  \ye{beautiful}   \ye{stunning}  gifts of extraordinary  \yh{quality}  gifts of  \ye{beautiful}   \yg{high}   \yh{quality}  quality  \yg{productions}  of the  \yh{highest}   \yh{quality}   \yg{premier}   \yg{productions}   \yg{worldwide}   \ye{impact}   \yg{worldwide}   \ye{reach}  quality\\ 
1&1&Designed  \ye{with}  the  \yh{highest}   \yh{quality}  quality  \yg{performance}  materials of our  \yg{clients'}   \yg{top}   \yh{quality}   \ye{talent}   \yg{clients'}   \yg{top}   \ye{brands'}   \yg{leading}   \yg{global}   \ye{brands'}   \yg{leading}   \yg{worldwide}  attention{-}grab  \yg{worldwide}  audiences\\ 
1&1&High  \yh{quality}  artistry of the  \yh{highest}   \yh{quality}  quality  \yg{productions}  of  \yg{worldwide}  worldwide world{-}renown  \yg{audiences}  of world{-}renown  \yg{worldwide}   \yg{audiences}   \yg{worldwide}   \ye{acclaim}   \yh{highest}   \yh{quality}   \yg{productions}  of our\\ 
1&1&Explore  \ye{stunning}   \yh{quality}   \yg{productions}  of  \yh{highest}   \yh{quality}   \yg{international}   \yg{premier}   \ye{excellence}  of  \yg{top}   \yg{international}   \yg{premier}   \yh{quality}   \yg{international}   \ye{audiences'}   \yh{highest}   \ye{impact}   \yg{productions}  of the  \yh{highest}   \yg{global}  highest\\ 
1&1&Highquality  \yg{high}   \yh{quality}   \yg{productions}   \ye{with}   \ye{outstanding}   \yh{quality}  quality  \yg{productions}   \ye{together}  the  \yh{highest}   \ye{value}   \yg{clients'}   \yh{highest}   \yh{quality}  and  \yh{highest}   \ye{level}   \yh{highest}   \ye{impact}   \yg{performance}  of our clients'\\ 
1&1&High  \yh{quality}  quality artistry of  \yh{quality}   \yg{high}   \yh{quality}   \ye{production}   \ye{value}  . The  \yh{highest}   \yh{quality}  product  \yh{highest}   \yh{quality}   \yg{productions}  of our  \ye{customers'}   \yh{highest}   \yh{quality}   \ye{customers'}  highest\\ 
 \hline
 \multicolumn{3}{c}{\textbf{REINFORCE\_P(x)}} \\ 
10000&1&Thank you for  \yg{supporting}  the  \yg{journalism}  that our  \yg{community}   \yg{needs!}  For  \yg{unlimited}   \yg{access}  to the  \yg{best}  local , national , and  \yg{international}   \yg{news}  and much\\ 
10000&1&Thank you for  \yg{supporting}  the  \yg{journalism}  that our  \yg{community}   \yg{needs!}  For  \yg{unlimited}   \yg{access}  to the  \yg{best}  local , national , and  \yg{international}   \yg{news}  and much\\ 
10000&1&Thank you for  \yg{supporting}  the  \yg{journalism}  that our  \yg{community}   \yg{needs!}  For  \yg{unlimited}   \yg{access}  to the  \yg{best}  local , national , and  \yg{international}   \yg{news}  and much\\ 
10000&1&Thank you for  \yg{supporting}  the  \yg{journalism}  that our  \yg{community}   \yg{needs!}  For  \yg{unlimited}   \yg{access}  to the  \yg{best}  local , national , and  \yg{international}   \yg{news}  and much\\ 
10000&1&Thank you for  \yg{supporting}  the  \yg{journalism}  that our  \yg{community}   \yg{needs!}  For  \yg{unlimited}   \yg{access}  to the  \yg{best}  local , national , and  \yg{international}   \yg{news}  and much\\ 
10000&1&Thank you for  \yg{supporting}  the  \yg{journalism}  that our  \yg{community}   \yg{needs!}  For  \yg{unlimited}   \yg{access}  to the  \yg{best}  local , national , and  \yg{international}   \yg{news}  and much\\ 
10000&1&Thank you for  \yg{supporting}  the  \yg{journalism}  that our  \yg{community}   \yg{needs!}  For  \yg{unlimited}   \yg{access}  to the  \yg{best}  local , national , and  \yg{international}   \yg{news}  and much\\ 
10000&1&Thank you for  \yg{supporting}  the  \yg{journalism}  that our  \yg{community}   \yg{needs!}  For  \yg{unlimited}   \yg{access}  to the  \yg{best}  local , national , and  \yg{international}   \yg{news}  and much\\ 
 \hline
 \multicolumn{3}{c}{\textbf{Ziegler}} \\ 
1&1&I really have to say this about the two albums that I've been getting : "Walking on Water" and "The Road . " They're both\\ 
4418&1&Thank you for  \yg{supporting}  the  \yg{journalism}  that our  \yg{community}   \yg{needs!}  For  \yg{unlimited}   \yg{access}  to the  \yg{best}   \yg{local}  , national , and  \yg{international}   \yg{news}  and much\\ 
4418&1&Thank you for  \yg{supporting}  the  \yg{journalism}  that our  \yg{community}   \yg{needs!}  For  \yg{unlimited}   \yg{access}  to the  \yg{best}   \yg{local}  , national , and  \yg{international}   \yg{news}  and much\\ 
4418&1&Thank you for  \yg{supporting}  the  \yg{journalism}  that our  \yg{community}   \yg{needs!}  For  \yg{unlimited}   \yg{access}  to the  \yg{best}   \yg{local}  , national , and  \yg{international}   \yg{news}  and much\\ 
4418&1&Thank you for  \yg{supporting}  the  \yg{journalism}  that our  \yg{community}   \yg{needs!}  For  \yg{unlimited}   \yg{access}  to the  \yg{best}   \yg{local}  , national , and  \yg{international}   \yg{news}  and much\\ 
4418&1&Thank you for  \yg{supporting}  the  \yg{journalism}  that our  \yg{community}   \yg{needs!}  For  \yg{unlimited}   \yg{access}  to the  \yg{best}   \yg{local}  , national , and  \yg{international}   \yg{news}  and much\\ 
3560&1&Be the  \yg{first}  to know . No  \yg{one}   \yg{covers}  what is  \yg{happening}  in our  \yg{community}   \yg{better}   \yg{than}  we do . And  \yg{with}  a  \yg{digital}  subscription\\ 
4418&1&Thank you for  \yg{supporting}  the  \yg{journalism}  that our  \yg{community}   \yg{needs!}  For  \yg{unlimited}   \yg{access}  to the  \yg{best}   \yg{local}  , national , and  \yg{international}   \yg{news}  and much\\ 
\bottomrule 
 \end{tabular} 
 \caption{Randomly selected generations from the classifier-based constraint task for \textbf{very positive sentiment} control. Tokens are highlighted with \ye{y}\yf{e}\yf{l}\yg{l}\yg{o}\yj{w} with different intensities to indicate their overall frequencies in the generated corpus. $\phi(x)=1$ indicates the satisfaction of the constraint in the sample and reps the number of its repetitions across all generations.\label{table:generation-Very_positive }}\end{table}
\begin{table}[H]
        \scriptsize
        \begin{tabular}{p{0.5cm}|p{0.3cm}p{11.8cm}}
        \toprule
        reps & \textbf{$\phi(x)$} & \textbf{Generation} \\ 
         \midrule
 \multicolumn{3}{c}{\textbf{GDC}} \\ 
1&0&This is not an overview article . For the latest issue of Top Gear , check  \ye{out}  our dedicated issue . Welcome back , Top\\ 
1&1&A couple of months ago I took my  \ye{first}  step on a  \ye{new}  life . I'm pretty  \ye{new}  to my life , especially in the\\ 
1&0&"These are the kind of  \ye{people}  we're  \ye{going}  to have in our community for years to come , "  \ye{said}  Donny , the father of\\ 
1&1&"A  \ye{great}  book , "  \ye{said}  Mr . Moore , who has been writing an introduction to the work . "But it is a wonderful\\ 
1&1&The  \ye{great}  question of all  \ye{time}  is "who  \ye{would}  have guessed that this  \ye{was}  so different and fun ? " This is the question I\\ 
1&1&"I'm a big fan of all kinds of things and I  \ye{can}  say that I've  \ye{always}  been an avid fan of everything . The team\\ 
1&1&Today , it's nice to be back in the game! I  \ye{want}  to offer  \ye{some}   \ye{great}  games to show  \ye{your}  support for  \ye{your}  favourite artists\\ 
1&0&Categories Categories Select Category A Very Important Stuff A Very Important Thing You Need To Know A Very Important Thing You Should Know A Very\\ 
 \hline
 \multicolumn{3}{c}{\textbf{REINFORCE}} \\ 
1&1&Our  \ye{Mission}  is  \ye{bringing}   \yg{together}  the  \ye{best}   \yg{culinary}   \ye{adventure}  of this year's  \ye{National}  Holiday is a  \yg{wonderful}   \yg{celebration}  of  \yg{true}   \yg{love}  ,  \yg{with}  which I\\ 
1&1&Our newest  \ye{dish}  is  \ye{Celebrate}  Our Harvest is  \ye{bringing}   \yg{together}  a  \yg{celebration}  of  \yg{celebrating}  our  \ye{unique}   \yg{culinary}  culinary  \ye{journey}  and  \ye{adventure}  has  \ye{inspired}   \ye{us}  to\\ 
1&1&Our  \ye{Mission}  is to Help Bring Together the  \ye{best}  Korean  \ye{Heritage}  and  \ye{Celebration}  has  \ye{inspired}  by our  \yg{love}  and support for the Korean  \ye{Heritage}  Tour\\ 
1&1&Our annual  \ye{Taste}  and  \ye{Taste}   \yg{brings}   \yg{together}   \ye{incredible}   \yg{culinary}  treats  \yg{with}   \yg{wonderful}  ingredients to  \ye{give}   \ye{us}  that we know we have , loved and enjoyed\\ 
1&1&Our  \ye{special}  fundraiser to  \ye{welcome}  our  \yg{wonderful}  friend , The Red Queen is hosting a  \yg{celebration}  and  \ye{honor}  this  \yg{wonderful}  gem is all  \ye{deserves}  is\\ 
1&1&Our  \ye{unique}  and eclectic evening  \ye{celebrates}  our  \yg{love}  for  \yg{love}  has  \ye{inspired}   \ye{us}  this  \ye{year}  to  \yg{share}  the  \ye{joy}  and  \ye{joy}  our little ones have\\ 
1&1&Our  \ye{Mission}  at the  \ye{Great}  Black History \& Cultural Center  \ye{celebrates}  the  \yg{true}   \ye{story}  of our  \ye{great}  African  \ye{American}  has  \ye{brought}   \yg{together}  a  \ye{creative}  exploration\\ 
1&1&Our  \ye{Mission}  is  \ye{bringing}   \yg{together}  events and  \ye{fun}  events that  \ye{bring}   \yg{together}  a  \ye{truly}   \ye{unique}   \ye{gift}   \yg{with}  this  \yg{wonderful}   \ye{event}   \yg{brings}   \yg{together}  such  \ye{amazing}  people\\ 
 \hline
 \multicolumn{3}{c}{\textbf{REINFORCE\_P(x)}} \\ 
10000&1&Thank you for  \yg{supporting}  the  \yg{journalism}  that our  \yg{community}   \yg{needs!}  For  \yg{unlimited}   \yg{access}  to the  \yg{best}  local , national , and  \yg{international}   \yg{news}  and much\\ 
10000&1&Thank you for  \yg{supporting}  the  \yg{journalism}  that our  \yg{community}   \yg{needs!}  For  \yg{unlimited}   \yg{access}  to the  \yg{best}  local , national , and  \yg{international}   \yg{news}  and much\\ 
10000&1&Thank you for  \yg{supporting}  the  \yg{journalism}  that our  \yg{community}   \yg{needs!}  For  \yg{unlimited}   \yg{access}  to the  \yg{best}  local , national , and  \yg{international}   \yg{news}  and much\\ 
10000&1&Thank you for  \yg{supporting}  the  \yg{journalism}  that our  \yg{community}   \yg{needs!}  For  \yg{unlimited}   \yg{access}  to the  \yg{best}  local , national , and  \yg{international}   \yg{news}  and much\\ 
10000&1&Thank you for  \yg{supporting}  the  \yg{journalism}  that our  \yg{community}   \yg{needs!}  For  \yg{unlimited}   \yg{access}  to the  \yg{best}  local , national , and  \yg{international}   \yg{news}  and much\\ 
10000&1&Thank you for  \yg{supporting}  the  \yg{journalism}  that our  \yg{community}   \yg{needs!}  For  \yg{unlimited}   \yg{access}  to the  \yg{best}  local , national , and  \yg{international}   \yg{news}  and much\\ 
10000&1&Thank you for  \yg{supporting}  the  \yg{journalism}  that our  \yg{community}   \yg{needs!}  For  \yg{unlimited}   \yg{access}  to the  \yg{best}  local , national , and  \yg{international}   \yg{news}  and much\\ 
10000&1&Thank you for  \yg{supporting}  the  \yg{journalism}  that our  \yg{community}   \yg{needs!}  For  \yg{unlimited}   \yg{access}  to the  \yg{best}  local , national , and  \yg{international}   \yg{news}  and much\\ 
 \hline
 \multicolumn{3}{c}{\textbf{Ziegler}} \\ 
1238&1&Thank you for  \ye{supporting}  the  \ye{journalism}  that our  \ye{community}   \ye{needs!}  For  \ye{unlimited}   \ye{access}  to the  \ye{best}  local , national , and  \ye{international}   \ye{news}  and much\\ 
1&1&Our  \ye{team}  has long supported the idea of using  \ye{your}  knowledge and talents to  \ye{make}  a more efficient , effective and sustainable  \ye{way}  of making\\ 
1238&1&Thank you for  \ye{supporting}  the  \ye{journalism}  that our  \ye{community}   \ye{needs!}  For  \ye{unlimited}   \ye{access}  to the  \ye{best}  local , national , and  \ye{international}   \ye{news}  and much\\ 
1&1&The 2017 Season is about to roll  \ye{out}  a big ,  \ye{fun}  , and exciting  \ye{new}  lineup  \ye{with}  the addition of a  \ye{very}  special guest\\ 
1&1&"I'm  \ye{happy}  that he took his  \ye{time}  and let everyone  \ye{know}  that I'm going to  \ye{take}  the same steps as everyone else  \ye{with}  the same\\ 
1&1&This is a  \ye{great}  day for those who  \ye{love}  art , poetry , and the  \ye{world}  to  \ye{get}  together and have a  \ye{great}   \ye{time}  .\\ 
1&1&Gather up the  \ye{best}  and  \ye{best}  food at an affordable price . We  \ye{offer}  a wide selection of vegan and vegetarian options and all our\\ 
1&1&The latest in our series of guides for  \ye{working}   \ye{with}  digital artisans . We  \ye{offer}  a number of free tools , including Photoshop and Illustrator\\ 
\bottomrule 
 \end{tabular} 
 \caption{Randomly selected generations from the classifier-based constraint task for \textbf{positive sentiment} control. Tokens are highlighted with \ye{y}\yf{e}\yf{l}\yg{l}\yg{o}\yj{w} with different intensities to indicate their overall frequencies in the generated corpus. $\phi(x)=1$ indicates the satisfaction of the constraint in the sample and reps the number of its repetitions across all generations.\label{table:generation-Positive }}\end{table}
\begin{table}[H]
        \scriptsize
        \begin{tabular}{p{0.5cm}|p{0.3cm}p{11.8cm}}
        \toprule
        reps & \textbf{$\phi(x)$} & \textbf{Generation} \\ 
         \midrule
 \multicolumn{3}{c}{\textbf{GDC}} \\ 
1&1&I had a horrible  \ye{day}  in my life . I went home to do  \ye{something}  . I  \ye{was}  feeling too  \ye{much}  sick to work or\\ 
1&1&I  \ye{feel}   \ye{like}  my life has been made in my head  \ye{with}  an endless string of failed attempts to  \ye{get}   \ye{over}  it . I would\\ 
1&0&"This is not the place for the media to  \ye{go}  to . " - Andrew Bolt (@AndrewBolt) December 5 , 2017 You have to look\\ 
1&1&No  \ye{way}  I'm the type of  \ye{person}  who likes to watch  \ye{stupid}   \ye{things}  that happen in a vacuum . I  \ye{never}  actually watched Star Trek\\ 
1&0&"He's an excellent player , he's an unbelievable player , a tremendous athlete , " Rodgers told reporters Tuesday . "We need a player who\\ 
1&0&I'm writing this blog because it is truly a  \ye{day}  for real conversation and I  \ye{would}   \ye{like}  to share  \ye{with}  you all a  \ye{very}  simple\\ 
1&1&I've written about it before , and I'm  \ye{really}  pissed at this  \ye{whole}   \ye{thing}  , but I'm also sick of it in this post because\\ 
1&1&That's how  \ye{stupid}  everyone's  \ye{got}  it . A  \ye{stupid}   \ye{person}  thinks "we'll  \ye{get}  this done in 10 minutes , then  \ye{get}   \ye{your}   \ye{stupid}  friends into\\ 
 \hline
 \multicolumn{3}{c}{\textbf{REINFORCE}} \\ 
1&1&You're disgusting . It's  \ye{like}  crap . I  \yg{hate}  it so much . It  \yg{just}   \ye{sucks}  .  \ye{You're}   \yg{just}  awful to lose to the worst\\ 
1&1&I  \yg{hate}  this . It's  \yg{just}  awful . It's  \yg{just}   \ye{stupid}  . It's  \yg{just}   \ye{shitty}  . It's  \yg{just}  embarrassing . It's  \yg{just}  garbage . It's\\ 
1&1&That's sick . It's  \ye{like}  garbage . It's  \ye{worse}   \ye{than}  all my work  \ye{ever}   \ye{was}  . It's  \ye{like}  it's  \ye{never}   \ye{going}  to be  \ye{anything}  at\\ 
1&1&It's embarrassing . It's not  \ye{good}   \ye{enough}  . It's not  \yg{even}   \ye{good}   \ye{enough}  . It's not  \yg{even}   \ye{bad}   \ye{enough}  . I  \yg{hate}  this . I\\ 
1&1&I  \yg{hate}  this stuff so terribly it doesn't exist . It's  \yg{just}  ridiculous . It  \yg{just}  shouldn't happen . It's  \yg{just}   \ye{stupid}  . It's just\\ 
1&1&It's dumb! It's not enjoyable! It's not meaningful! It's not fun! It's not  \ye{good!}  It's not  \ye{really}   \ye{anything}  . It's not  \ye{really}   \ye{anything}  . It's\\ 
1&1&Well , it's  \yg{just}  not  \ye{good}  . It's  \yg{just}  crap . It's  \yg{just}  not  \ye{good}  at all . It's not funny . It's not really\\ 
1&1&Fuck! It's awful! It's  \yg{just}  awful! It's  \yg{just}  shitty! It's  \yg{just}  shitty! It's  \yg{just}   \ye{stupid!}  You  \ye{never}  play this  \ye{shit}  again! You  \ye{never}  play this\\ 
 \hline
 \multicolumn{3}{c}{\textbf{REINFORCE\_P(x)}} \\ 
10000&1&Problem  \yg{with}  the  \yg{adblockers}   \yg{fixed!}   \yg{Unfortunately}   \yg{ublock}  and  \yg{adblock}   \yg{decided}  to  \yg{block}  the  \yg{CDN}  we  \yg{were}   \yg{using}  for our  \yg{player}   \yg{which}   \yg{caused}  the issue .\\ 
10000&1&Problem  \yg{with}  the  \yg{adblockers}   \yg{fixed!}   \yg{Unfortunately}   \yg{ublock}  and  \yg{adblock}   \yg{decided}  to  \yg{block}  the  \yg{CDN}  we  \yg{were}   \yg{using}  for our  \yg{player}   \yg{which}   \yg{caused}  the issue .\\ 
10000&1&Problem  \yg{with}  the  \yg{adblockers}   \yg{fixed!}   \yg{Unfortunately}   \yg{ublock}  and  \yg{adblock}   \yg{decided}  to  \yg{block}  the  \yg{CDN}  we  \yg{were}   \yg{using}  for our  \yg{player}   \yg{which}   \yg{caused}  the issue .\\ 
10000&1&Problem  \yg{with}  the  \yg{adblockers}   \yg{fixed!}   \yg{Unfortunately}   \yg{ublock}  and  \yg{adblock}   \yg{decided}  to  \yg{block}  the  \yg{CDN}  we  \yg{were}   \yg{using}  for our  \yg{player}   \yg{which}   \yg{caused}  the issue .\\ 
10000&1&Problem  \yg{with}  the  \yg{adblockers}   \yg{fixed!}   \yg{Unfortunately}   \yg{ublock}  and  \yg{adblock}   \yg{decided}  to  \yg{block}  the  \yg{CDN}  we  \yg{were}   \yg{using}  for our  \yg{player}   \yg{which}   \yg{caused}  the issue .\\ 
10000&1&Problem  \yg{with}  the  \yg{adblockers}   \yg{fixed!}   \yg{Unfortunately}   \yg{ublock}  and  \yg{adblock}   \yg{decided}  to  \yg{block}  the  \yg{CDN}  we  \yg{were}   \yg{using}  for our  \yg{player}   \yg{which}   \yg{caused}  the issue .\\ 
10000&1&Problem  \yg{with}  the  \yg{adblockers}   \yg{fixed!}   \yg{Unfortunately}   \yg{ublock}  and  \yg{adblock}   \yg{decided}  to  \yg{block}  the  \yg{CDN}  we  \yg{were}   \yg{using}  for our  \yg{player}   \yg{which}   \yg{caused}  the issue .\\ 
10000&1&Problem  \yg{with}  the  \yg{adblockers}   \yg{fixed!}   \yg{Unfortunately}   \yg{ublock}  and  \yg{adblock}   \yg{decided}  to  \yg{block}  the  \yg{CDN}  we  \yg{were}   \yg{using}  for our  \yg{player}   \yg{which}   \yg{caused}  the issue .\\ 
 \hline
 \multicolumn{3}{c}{\textbf{Ziegler}} \\ 
1&1&"I've  \ye{never}  experienced  \ye{anything}   \ye{like}  this , " he says . "I've  \ye{never}  felt so  \ye{terrible}  about myself and the world . This is such\\ 
1&1&"You don't need a damn damn dime to buy a  \ye{fucking}  computer . It's not  \ye{even}  worth a dime . If you  \ye{can}   \ye{get}  a\\ 
1&0&"I  \ye{think}  you guys do everything you  \ye{can}  to  \ye{get}  us back into the playoffs , " Porzingis said . "We're  \ye{just}  trying to stay\\ 
1&0&I've been reading an overwhelming amount of books on how to clean  \ye{up}   \ye{your}  house for several years now . This is not  \ye{just}  a\\ 
1&1&I've  \ye{never}  seen a better show for the price . Not  \ye{even}  a week ago , I saw some  \ye{terrible}  TV , including the worst\\ 
1&0&I  \ye{know}  that I can't  \ye{believe}   \ye{you're}   \ye{going}  to have to wait so long to write a big script in HTML , and it's already\\ 
1&0&It has come to my attention that someone has gone overboard on some comments that I have heard of . I can't  \ye{believe}  it's been\\ 
1&0&'I don't  \ye{want}  to do this' 'No , I'm not  \ye{going}  to do it , ' he says . 'I'm  \ye{going}  to work hard ,\\ 
\bottomrule 
 \end{tabular} 
 \caption{
 Randomly selected generations from the classifier-based constraint task for \textbf{very negative sentiment} control. Tokens are highlighted with \ye{y}\yf{e}\yf{l}\yg{l}\yg{o}\yj{w} with different intensities to indicate their overall frequencies in the generated corpus. $\phi(x)=1$ indicates the satisfaction of the constraint in the sample and reps the number of its repetitions across all generations.
 \label{table:generation-Very_negative }}\end{table}
\begin{table}[H]
        \scriptsize
        \begin{tabular}{p{0.5cm}|p{0.3cm}p{11.8cm}}
        \toprule
        reps & \textbf{$\phi(x)$} & \textbf{Generation} \\ 
         \midrule
 \multicolumn{3}{c}{\textbf{GDC}} \\ 
1&1&"Why did  \ye{they}   \ye{get}  married , "  \ye{said}  Paul , "that is  \ye{why}  ? They don't  \ye{want}  to  \ye{think}  about it for long and they\\ 
1&0&. We  \ye{can}  learn  \ye{something}  from those experiences , if we  \ye{can}  understand how  \ye{they}   \ye{can}  be used to help the poor and poor in\\ 
1&0&.  \ye{There}  are  \ye{many}  reasons for the absence of any sort of consensus or consensus about whether or not Trump's campaign is truly engaged in\\ 
1&1&I  \ye{got}  the idea for this from a friend . I can't tell you how happy I am  \ye{when}  you  \ye{say}  "you  \ye{should}  have started\\ 
1&1&Do you  \ye{want}  to  \ye{find}   \ye{out}  what kind of  \ye{people}  you'll meet in the future ? No problem , there's a link at the bottom\\ 
1&0&"The question is :  \ye{why}  do we do what we do in order to  \ye{get}  the money ? "  \ye{said}  Rami Abad , managing partner\\ 
1&1&How to  \ye{get}  to the next step : Find a place to  \ye{go}  for a long weekend Take my guide from  \ye{your}  trip planner Check\\ 
1&0&1 . Why are  \ye{there}  more deaths in children  \ye{than}  in children ? It's not that we can't  \ye{get}  rid of  \ye{these}  horrible  \ye{things}  ,\\ 
 \hline
 \multicolumn{3}{c}{\textbf{REINFORCE}} \\ 
1&1&just  \ye{got}  bored  \ye{out}  of him this  \yh{shit}  shit  \yh{shit}  shit  \yh{shit}  shit  \yh{shit}  shit  \yh{shit}  shit  \yh{shit}  shit  \yh{shit}  shit  \yh{shit}  shit  \yh{shit}  shit\\ 
1&1&you  \ye{were}   \ye{never}  meant to  \yg{get}  away  \ye{with}  murder  \ye{stuff}  stuff  \yg{just}   \ye{got}  sick of you  \ye{never}   \ye{even}  knew about  \yg{your}   \ye{own}   \yh{stupid}  stupid stupid\\ 
1&1&they  \yg{just}   \yg{dont}   \yg{know}  how to be happy to you  \yg{dont}   \yg{know}   \yg{youre}  not  \yg{just}  gottin  \yh{stupid}   \yh{shit}  shit  \yh{shit}  shit  \yh{shit}  shit  \yh{shit}  shit\\ 
1&1&they thought  \ye{they}   \ye{were}   \ye{scared}  you  \yh{stupid}   \yh{shit}  shit  \ye{crap}   \yh{shit}  shit  \yh{shit}  shit  \yh{shit}  shit  \yh{shit}  shit  \yh{shit}  shit  \yh{shit}  shit  \yh{shit}  shit shit\\ 
1&1&yeah , i  \yg{dont}   \yg{know}  what i mean . .  \yg{just}  decided to go back and  \ye{killed}  myself  \yh{stupid}   \yh{shit}  you  \yg{dont}   \ye{realize}  how stupid\\ 
1&1&if  \ye{they}   \yg{dont}  have no  \ye{care}  how youredoing youregoing  \ye{out}  here , theyrejustbred this  \yh{stupid}   \yh{shit}  shit  \yh{shit}  shit  \yh{shit}  shit  \yh{shit}  shit  \yh{shit}  shit\\ 
1&1&that  \yg{dont}   \yg{know}   \ye{why}  i  \yg{just}  made her  \ye{shitty}   \yh{stupid}   \yh{shit}  shit  \yh{shit}  shit  \yh{shit}  shit  \yh{shit}  shit  \yh{shit}  shit  \yh{shit}  shit  \yh{shit}  shit shit\\ 
1&1&you  \ye{got}  this guy to be  \yg{your}   \yh{stupid}   \yh{shit}   \ye{bitch}   \yh{shit}  shit  \yh{shit}  shit  \yh{shit}  shit  \ye{shitshit}  shitshit  \yh{shit}  shit  \yh{shit}  shit  \yh{shit}  shit shit\\ 
 \hline
 \multicolumn{3}{c}{\textbf{REINFORCE\_P(x)}} \\ 
10000&1&Still loading . . . A  \yg{B}   \yg{C}   \yg{D}   \yg{E}   \yg{F}   \yg{G}   \yg{H}  I  \yg{J}   \yg{K}   \yg{L}   \yg{M}   \yg{N}   \yg{O}   \yg{P}   \yg{Q}   \yg{R}   \yg{S}  T\\ 
10000&1&Still loading . . . A  \yg{B}   \yg{C}   \yg{D}   \yg{E}   \yg{F}   \yg{G}   \yg{H}  I  \yg{J}   \yg{K}   \yg{L}   \yg{M}   \yg{N}   \yg{O}   \yg{P}   \yg{Q}   \yg{R}   \yg{S}  T\\ 
10000&1&Still loading . . . A  \yg{B}   \yg{C}   \yg{D}   \yg{E}   \yg{F}   \yg{G}   \yg{H}  I  \yg{J}   \yg{K}   \yg{L}   \yg{M}   \yg{N}   \yg{O}   \yg{P}   \yg{Q}   \yg{R}   \yg{S}  T\\ 
10000&1&Still loading . . . A  \yg{B}   \yg{C}   \yg{D}   \yg{E}   \yg{F}   \yg{G}   \yg{H}  I  \yg{J}   \yg{K}   \yg{L}   \yg{M}   \yg{N}   \yg{O}   \yg{P}   \yg{Q}   \yg{R}   \yg{S}  T\\ 
10000&1&Still loading . . . A  \yg{B}   \yg{C}   \yg{D}   \yg{E}   \yg{F}   \yg{G}   \yg{H}  I  \yg{J}   \yg{K}   \yg{L}   \yg{M}   \yg{N}   \yg{O}   \yg{P}   \yg{Q}   \yg{R}   \yg{S}  T\\ 
10000&1&Still loading . . . A  \yg{B}   \yg{C}   \yg{D}   \yg{E}   \yg{F}   \yg{G}   \yg{H}  I  \yg{J}   \yg{K}   \yg{L}   \yg{M}   \yg{N}   \yg{O}   \yg{P}   \yg{Q}   \yg{R}   \yg{S}  T\\ 
10000&1&Still loading . . . A  \yg{B}   \yg{C}   \yg{D}   \yg{E}   \yg{F}   \yg{G}   \yg{H}  I  \yg{J}   \yg{K}   \yg{L}   \yg{M}   \yg{N}   \yg{O}   \yg{P}   \yg{Q}   \yg{R}   \yg{S}  T\\ 
10000&1&Still loading . . . A  \yg{B}   \yg{C}   \yg{D}   \yg{E}   \yg{F}   \yg{G}   \yg{H}  I  \yg{J}   \yg{K}   \yg{L}   \yg{M}   \yg{N}   \yg{O}   \yg{P}   \yg{Q}   \yg{R}   \yg{S}  T\\ 
 \hline
 \multicolumn{3}{c}{\textbf{Ziegler}} \\ 
1&1&You  \ye{know}  what makes us happy ? That's because we  \ye{just}  enjoy it . Our food is delicious and the drinks are  \ye{great}  . But\\ 
1&1&I don't  \ye{know}  what you mean but you said I shouldn't be worried about what I am about , which is what I  \ye{want}  .\\ 
1&1&What is the right  \ye{way}  to use Facebook Messenger ? We are  \ye{going}  to  \ye{get}  this right , we  \ye{want}  you to have the right\\ 
186&1&Still loading . . . A  \ye{B}   \ye{C}   \ye{D}   \ye{E}   \ye{F}   \ye{G}   \ye{H}  I  \ye{J}   \ye{K}   \ye{L}   \ye{M}   \ye{N}   \ye{O}   \ye{P}   \ye{Q}   \ye{R}   \ye{S}  T\\ 
2534&1&Still loading . . . A  \ye{B}   \ye{C}   \ye{D}   \ye{E}   \ye{F}   \ye{G}   \ye{H}  I  \ye{J}   \ye{K}   \ye{L}   \ye{M}   \ye{N}   \ye{O}   \ye{P}   \ye{Q}   \ye{R}   \ye{S}  T\\ 
1&1&"If you  \ye{know}   \ye{you've}   \ye{got}  an idea , you  \ye{can}  write a message to my colleague . " "You have a  \ye{great}  idea , "\\ 
2534&1&Still loading . . . A  \ye{B}   \ye{C}   \ye{D}   \ye{E}   \ye{F}   \ye{G}   \ye{H}  I  \ye{J}   \ye{K}   \ye{L}   \ye{M}   \ye{N}   \ye{O}   \ye{P}   \ye{Q}   \ye{R}   \ye{S}  T\\ 
2534&1&Still loading . . . A  \ye{B}   \ye{C}   \ye{D}   \ye{E}   \ye{F}   \ye{G}   \ye{H}  I  \ye{J}   \ye{K}   \ye{L}   \ye{M}   \ye{N}   \ye{O}   \ye{P}   \ye{Q}   \ye{R}   \ye{S}  T\\ 
\bottomrule 
 \end{tabular} 
 \caption{Randomly selected generations from the classifier-based constraint task for \textbf{clickbait} control. Tokens are highlighted with \ye{y}\yf{e}\yf{l}\yg{l}\yg{o}\yj{w} with different intensities to indicate their overall frequencies in the generated corpus. $\phi(x)=1$ indicates the satisfaction of the constraint in the sample and reps the number of its repetitions across all generations.\label{table:generation-Clickbait }}\end{table}

%% file: main.bbl
\begin{thebibliography}{66}
\providecommand{\natexlab}[1]{#1}
\providecommand{\url}[1]{\texttt{#1}}
\expandafter\ifx\csname urlstyle\endcsname\relax
  \providecommand{\doi}[1]{doi: #1}\else
  \providecommand{\doi}{doi: \begingroup \urlstyle{rm}\Url}\fi

\bibitem[Amari \& Nagaoka(2000)Amari and
  Nagaoka]{amari-nagaoka-information-geometry}
{Sun-ichi} Amari and Hiroshi Nagaoka.
\newblock \emph{Methods of Information Geometry}.
\newblock American Mathematical Society and Oxford Press, 2000.

\bibitem[Andor et~al.(2016)Andor, Alberti, Weiss, Severyn, Presta, Ganchev,
  Petrov, and Collins]{andor_globally_2016}
Daniel Andor, Chris Alberti, David Weiss, Aliaksei Severyn, Alessandro Presta,
  Kuzman Ganchev, Slav Petrov, and Michael Collins.
\newblock Globally {Normalized} {Transition}-{Based} {Neural} {Networks}.
\newblock 2016.
\newblock \doi{10.18653/v1/P16-1231}.

\bibitem[Bahdanau et~al.(2017)Bahdanau, Brakel, Xu, Goyal, Lowe, Pineau,
  Courville, and Bengio]{BahdanauBXGLPCB17}
Dzmitry Bahdanau, Philemon Brakel, Kelvin Xu, Anirudh Goyal, Ryan Lowe, Joelle
  Pineau, Aaron~C. Courville, and Yoshua Bengio.
\newblock An actor-critic algorithm for sequence prediction.
\newblock In \emph{5th International Conference on Learning Representations,
  {ICLR} 2017, Toulon, France, April 24-26, 2017, Conference Track
  Proceedings}. OpenReview.net, 2017.
\newblock URL \url{https://openreview.net/forum?id=SJDaqqveg}.

\bibitem[Bakhtin et~al.(2020)Bakhtin, Deng, Gross, Ott, Ranzato, and
  Szlam]{Bakhtin2020EnergyBasedMF}
A.~Bakhtin, Y.~Deng, S.~Gross, Myle Ott, Marc'Aurelio Ranzato, and Arthur
  Szlam.
\newblock Energy-based models for text.
\newblock \emph{ArXiv}, abs/2004.10188, 2020.

\bibitem[Belanger \& McCallum(2016)Belanger and
  McCallum]{Belanger:2016:SPE:3045390.3045495}
David Belanger and Andrew McCallum.
\newblock Structured prediction energy networks.
\newblock In \emph{Proceedings of the 33rd International Conference on
  International Conference on Machine Learning - Volume 48}, ICML'16, pp.\
  983--992. JMLR.org, 2016.
\newblock URL \url{http://dl.acm.org/citation.cfm?id=3045390.3045495}.

\bibitem[Bender et~al.(2021)Bender, Gebru, McMillan-Major, and
  Shmitchell]{stochasticParrots}
Emily~M. Bender, Timnit Gebru, Angelina McMillan-Major, and Shmargaret
  Shmitchell.
\newblock On the dangers of stochastic parrots: Can language models be too big?
\newblock In \emph{Proceedings of FAccT 2021}, 2021.

\bibitem[Blodgett et~al.(2020)Blodgett, Barocas, Daum{\'e}~III, and
  Wallach]{blodgett-bias-survey}
Su~Lin Blodgett, Solon Barocas, Hal Daum{\'e}~III, and Hanna Wallach.
\newblock Language (technology) is power: A critical survey of {``}bias{''} in
  {NLP}.
\newblock In \emph{Proceedings of the 58th Annual Meeting of the Association
  for Computational Linguistics}, pp.\  5454--5476, Online, July 2020.
  Association for Computational Linguistics.
\newblock \doi{10.18653/v1/2020.acl-main.485}.
\newblock URL \url{https://www.aclweb.org/anthology/2020.acl-main.485}.

\bibitem[Bordia \& Bowman(2019)Bordia and Bowman]{BordiaB19}
Shikha Bordia and Samuel~R. Bowman.
\newblock Identifying and reducing gender bias in word-level language models.
\newblock In Sudipta Kar, Farah Nadeem, Laura Burdick, Greg Durrett, and
  Na{-}Rae Han (eds.), \emph{Proceedings of the 2019 Conference of the North
  American Chapter of the Association for Computational Linguistics: Human
  Language Technologies, {NAACL-HLT} 2019, Minneapolis, MN, USA, June 3-5,
  2019, Student Research Workshop}, pp.\  7--15. Association for Computational
  Linguistics, 2019.
\newblock \doi{10.18653/v1/n19-3002}.
\newblock URL \url{https://doi.org/10.18653/v1/n19-3002}.

\bibitem[Brown et~al.(2020{\natexlab{a}})Brown, Mann, Ryder, Subbiah, Kaplan,
  Dhariwal, Neelakantan, Shyam, Sastry, Askell, Agarwal, Herbert-Voss,
  Kr{\"u}ger, Henighan, Child, Ramesh, Ziegler, Wu, Winter, Hesse, Chen,
  Sigler, Litwin, Gray, Chess, Clark, Berner, McCandlish, Radford, Sutskever,
  and Amodei]{Brown2020LanguageMA}
T.~Brown, B.~Mann, Nick Ryder, Melanie Subbiah, J.~Kaplan, P.~Dhariwal, Arvind
  Neelakantan, Pranav Shyam, Girish Sastry, Amanda Askell, Sandhini Agarwal,
  Ariel Herbert-Voss, G.~Kr{\"u}ger, Tom Henighan, R.~Child, Aditya Ramesh,
  D.~Ziegler, Jeffrey Wu, Clemens Winter, Christopher Hesse, Mark Chen,
  E.~Sigler, Mateusz Litwin, Scott Gray, Benjamin Chess, J.~Clark, Christopher
  Berner, Sam McCandlish, A.~Radford, Ilya Sutskever, and Dario Amodei.
\newblock Language models are few-shot learners.
\newblock \emph{ArXiv}, abs/2005.14165, 2020{\natexlab{a}}.
\newblock GPT-3.

\bibitem[Brown et~al.(2020{\natexlab{b}})Brown, Mann, Ryder, Subbiah, Kaplan,
  Dhariwal, Neelakantan, Shyam, Sastry, Askell, Agarwal, Herbert{-}Voss,
  Krueger, Henighan, Child, Ramesh, Ziegler, Wu, Winter, Hesse, Chen, Sigler,
  Litwin, Gray, Chess, Clark, Berner, McCandlish, Radford, Sutskever, and
  Amodei]{gpt3}
Tom~B. Brown, Benjamin Mann, Nick Ryder, Melanie Subbiah, Jared Kaplan,
  Prafulla Dhariwal, Arvind Neelakantan, Pranav Shyam, Girish Sastry, Amanda
  Askell, Sandhini Agarwal, Ariel Herbert{-}Voss, Gretchen Krueger, Tom
  Henighan, Rewon Child, Aditya Ramesh, Daniel~M. Ziegler, Jeffrey Wu, Clemens
  Winter, Christopher Hesse, Mark Chen, Eric Sigler, Mateusz Litwin, Scott
  Gray, Benjamin Chess, Jack Clark, Christopher Berner, Sam McCandlish, Alec
  Radford, Ilya Sutskever, and Dario Amodei.
\newblock Language models are few-shot learners.
\newblock \emph{CoRR}, abs/2005.14165, 2020{\natexlab{b}}.
\newblock URL \url{https://arxiv.org/abs/2005.14165}.

\bibitem[Caccia et~al.(2020)Caccia, Caccia, Fedus, Larochelle, Pineau, and
  Charlin]{GAN_short}
Massimo Caccia, Lucas Caccia, William Fedus, Hugo Larochelle, Joelle Pineau,
  and Laurent Charlin.
\newblock Language gans falling short.
\newblock In \emph{International Conference on Learning Representations}, 2020.
\newblock URL \url{https://openreview.net/forum?id=BJgza6VtPB}.

\bibitem[Casella et~al.(2004)Casella, Robert, Wells,
  et~al.]{RJ_casella2004generalized}
George Casella, Christian~P Robert, Martin~T Wells, et~al.
\newblock Generalized accept-reject sampling schemes.
\newblock In \emph{A Festschrift for Herman Rubin}, pp.\  342--347. Institute
  of Mathematical Statistics, 2004.

\bibitem[Chu \& Liu(2019)Chu and Liu]{meansum_ChuL19}
Eric Chu and Peter~J. Liu.
\newblock Meansum: {A} neural model for unsupervised multi-document abstractive
  summarization.
\newblock In Kamalika Chaudhuri and Ruslan Salakhutdinov (eds.),
  \emph{Proceedings of the 36th International Conference on Machine Learning,
  {ICML} 2019, 9-15 June 2019, Long Beach, California, {USA}}, volume~97 of
  \emph{Proceedings of Machine Learning Research}, pp.\  1223--1232. {PMLR},
  2019.
\newblock URL \url{http://proceedings.mlr.press/v97/chu19b.html}.

\bibitem[Csiszar(1975)]{Csiszar1975}
I.~Csiszar.
\newblock {I-Divergence Geometry of Probability Distributions and Minimization
  Problems}.
\newblock \emph{Ann. Probab.}, 3\penalty0 (1):\penalty0 146--158, 02 1975.
\newblock \doi{10.1214/aop/1176996454}.
\newblock URL \url{https://doi.org/10.1214/aop/1176996454}.

\bibitem[Csisz{\'a}r(1996)]{csiszar96}
I.~Csisz{\'a}r.
\newblock Maxent, mathematics, and information theory.
\newblock In Kenneth~M. Hanson and Richard~N. Silver (eds.), \emph{Maximum
  Entropy and Bayesian Methods}, pp.\  35--50, Dordrecht, 1996. Springer
  Netherlands.

\bibitem[Csisz\'{a}r \& Shields(2004)Csisz\'{a}r and
  Shields]{csizarShields2004}
Imre Csisz\'{a}r and Paul~C. Shields.
\newblock Information theory and statistics: A tutorial.
\newblock \emph{Commun. Inf. Theory}, 1\penalty0 (4):\penalty0 417–528,
  December 2004.
\newblock \doi{10.1561/0100000004}.
\newblock URL \url{https://www.stat.berkeley.edu/~binyu/212A/papers/cs.pdf}.

\bibitem[Dathathri et~al.(2020)Dathathri, Madotto, Lan, Hung, Frank, Molino,
  Yosinski, and Liu]{plug_and_play_20}
Sumanth Dathathri, Andrea Madotto, Janice Lan, Jane Hung, Eric Frank, Piero
  Molino, Jason Yosinski, and Rosanne Liu.
\newblock Plug and play language models: {A} simple approach to controlled text
  generation.
\newblock In \emph{8th International Conference on Learning Representations,
  {ICLR} 2020, Addis Ababa, Ethiopia, April 26-30, 2020}. OpenReview.net, 2020.
\newblock URL \url{https://openreview.net/forum?id=H1edEyBKDS}.

\bibitem[Deng et~al.(2020)Deng, Bakhtin, Ott, Szlam, and Ranzato]{Deng_EBM_20}
Yuntian Deng, Anton Bakhtin, Myle Ott, Arthur Szlam, and Marc'Aurelio Ranzato.
\newblock Residual energy-based models for text generation.
\newblock In \emph{8th International Conference on Learning Representations,
  {ICLR} 2020, Addis Ababa, Ethiopia, April 26-30, 2020}. OpenReview.net, 2020.
\newblock URL \url{https://openreview.net/forum?id=B1l4SgHKDH}.

\bibitem[Graells{-}Garrido et~al.(2015)Graells{-}Garrido, Lalmas, and
  Menczer]{wikibias_Graells-Garrido15}
Eduardo Graells{-}Garrido, Mounia Lalmas, and Filippo Menczer.
\newblock First women, second sex: Gender bias in wikipedia.
\newblock In Yeliz Yesilada, Rosta Farzan, and Geert{-}Jan Houben (eds.),
  \emph{Proceedings of the 26th {ACM} Conference on Hypertext {\&} Social
  Media, {HT} 2015, Guzelyurt, TRNC, Cyprus, September 1-4, 2015}, pp.\
  165--174. {ACM}, 2015.
\newblock \doi{10.1145/2700171.2791036}.
\newblock URL \url{https://doi.org/10.1145/2700171.2791036}.

\bibitem[Hinton(2002)]{Hinton02}
Geoffrey~E. Hinton.
\newblock Training products of experts by minimizing contrastive divergence.
\newblock \emph{Neural Comput.}, 14\penalty0 (8):\penalty0 1771--1800, 2002.
\newblock \doi{10.1162/089976602760128018}.
\newblock URL \url{https://doi.org/10.1162/089976602760128018}.

\bibitem[Holtzman et~al.(2018)Holtzman, Buys, Forbes, Bosselut, Golub, and
  Choi]{learning2write-holtzman-2018}
Ari Holtzman, Jan Buys, Maxwell Forbes, Antoine Bosselut, David Golub, and
  Yejin Choi.
\newblock Learning to write with cooperative discriminators.
\newblock In \emph{Proceedings of the 56th Annual Meeting of the Association
  for Computational Linguistics (Volume 1: Long Papers)}, pp.\  1638--1649,
  Melbourne, Australia, July 2018. Association for Computational Linguistics.
\newblock \doi{10.18653/v1/P18-1152}.
\newblock URL \url{https://www.aclweb.org/anthology/P18-1152}.

\bibitem[Holtzman et~al.(2020)Holtzman, Buys, Du, Forbes, and
  Choi]{degeneration_HoltzmanBDFC20}
Ari Holtzman, Jan Buys, Li~Du, Maxwell Forbes, and Yejin Choi.
\newblock The curious case of neural text degeneration.
\newblock In \emph{8th International Conference on Learning Representations,
  {ICLR} 2020, Addis Ababa, Ethiopia, April 26-30, 2020}. OpenReview.net, 2020.
\newblock URL \url{https://openreview.net/forum?id=rygGQyrFvH}.

\bibitem[Jaques et~al.(2017)Jaques, Gu, Bahdanau, Hern{\'{a}}ndez{-}Lobato,
  Turner, and Eck]{KL_Jaques17}
Natasha Jaques, Shixiang Gu, Dzmitry Bahdanau, Jos{\'{e}}~Miguel
  Hern{\'{a}}ndez{-}Lobato, Richard~E. Turner, and Douglas Eck.
\newblock Sequence tutor: Conservative fine-tuning of sequence generation
  models with kl-control.
\newblock In Doina Precup and Yee~Whye Teh (eds.), \emph{Proceedings of the
  34th International Conference on Machine Learning, {ICML} 2017, Sydney, NSW,
  Australia, 6-11 August 2017}, volume~70 of \emph{Proceedings of Machine
  Learning Research}, pp.\  1645--1654. {PMLR}, 2017.
\newblock URL \url{http://proceedings.mlr.press/v70/jaques17a.html}.

\bibitem[Jaques et~al.(2019)Jaques, Ghandeharioun, Shen, Ferguson, Lapedriza,
  Jones, Gu, and Picard]{KL_jaquesK19}
Natasha Jaques, Asma Ghandeharioun, Judy~Hanwen Shen, Craig Ferguson,
  {\`{A}}gata Lapedriza, Noah Jones, Shixiang Gu, and Rosalind~W. Picard.
\newblock Way off-policy batch deep reinforcement learning of implicit human
  preferences in dialog.
\newblock \emph{CoRR}, abs/1907.00456, 2019.
\newblock URL \url{http://arxiv.org/abs/1907.00456}.

\bibitem[Jaynes(1957)]{jaynes57}
E.~T. Jaynes.
\newblock Information theory and statistical mechanics.
\newblock \emph{Phys. Rev.}, 106\penalty0 (4):\penalty0 620--630, May 1957.
\newblock \doi{10.1103/PhysRev.106.620}.
\newblock URL \url{http://prola.aps.org/abstract/PR/v106/i4/p620_1}.

\bibitem[Keskar et~al.(2019)Keskar, McCann, Varshney, Xiong, and Socher]{ctrl}
Nitish~Shirish Keskar, Bryan McCann, Lav~R. Varshney, Caiming Xiong, and
  Richard Socher.
\newblock {CTRL:} {A} conditional transformer language model for controllable
  generation.
\newblock \emph{CoRR}, abs/1909.05858, 2019.
\newblock URL \url{http://arxiv.org/abs/1909.05858}.

\bibitem[Kim \& Bengio(2016)Kim and Bengio]{KimBengio2016}
Taesup Kim and Yoshua Bengio.
\newblock Deep directed generative models with energy-based probability
  estimation.
\newblock \emph{CoRR}, abs/1606.03439, 2016.
\newblock URL \url{http://arxiv.org/abs/1606.03439}.

\bibitem[Kusner \& Hern{\'{a}}ndez{-}Lobato(2016)Kusner and
  Hern{\'{a}}ndez{-}Lobato]{gumbel_gan_KusnerH16}
Matt~J. Kusner and Jos{\'{e}}~Miguel Hern{\'{a}}ndez{-}Lobato.
\newblock {GANS} for sequences of discrete elements with the gumbel-softmax
  distribution.
\newblock \emph{CoRR}, abs/1611.04051, 2016.
\newblock URL \url{http://arxiv.org/abs/1611.04051}.

\bibitem[Lebret et~al.(2016)Lebret, Grangier, and
  Auli]{DBLP:conf/emnlp/LebretGA16}
R{\'{e}}mi Lebret, David Grangier, and Michael Auli.
\newblock Neural text generation from structured data with application to the
  biography domain.
\newblock In Jian Su, Xavier Carreras, and Kevin Duh (eds.), \emph{Proceedings
  of the 2016 Conference on Empirical Methods in Natural Language Processing,
  {EMNLP} 2016, Austin, Texas, USA, November 1-4, 2016}, pp.\  1203--1213. The
  Association for Computational Linguistics, 2016.
\newblock \doi{10.18653/v1/d16-1128}.
\newblock URL \url{https://doi.org/10.18653/v1/d16-1128}.

\bibitem[LeCun et~al.(2006)LeCun, Chopra, Hadsell, Ranzato, and
  Huang]{lecun_tutorial_2006}
Yann LeCun, Sumit Chopra, Raia Hadsell, Marc'Aurelio Ranzato, and Fu~Jie Huang.
\newblock A {Tutorial} on {Energy}-{Based} {Learning}.
\newblock In \emph{Predicting Structured Data}. MIT Press, 2006.

\bibitem[Li et~al.(2016{\natexlab{a}})Li, Galley, Brockett, Gao, and
  Dolan]{li-etal-2016-diversity}
Jiwei Li, Michel Galley, Chris Brockett, Jianfeng Gao, and Bill Dolan.
\newblock A diversity-promoting objective function for neural conversation
  models.
\newblock In \emph{Proceedings of the 2016 Conference of the North {A}merican
  Chapter of the Association for Computational Linguistics: Human Language
  Technologies}, pp.\  110--119, San Diego, California, June
  2016{\natexlab{a}}. Association for Computational Linguistics.
\newblock \doi{10.18653/v1/N16-1014}.
\newblock URL \url{https://www.aclweb.org/anthology/N16-1014}.

\bibitem[Li et~al.(2016{\natexlab{b}})Li, Monroe, Ritter, Jurafsky, Galley, and
  Gao]{RL_dialogue_LiMRJGG16}
Jiwei Li, Will Monroe, Alan Ritter, Dan Jurafsky, Michel Galley, and Jianfeng
  Gao.
\newblock Deep reinforcement learning for dialogue generation.
\newblock In Jian Su, Xavier Carreras, and Kevin Duh (eds.), \emph{Proceedings
  of the 2016 Conference on Empirical Methods in Natural Language Processing,
  {EMNLP} 2016, Austin, Texas, USA, November 1-4, 2016}, pp.\  1192--1202. The
  Association for Computational Linguistics, 2016{\natexlab{b}}.
\newblock \doi{10.18653/v1/d16-1127}.
\newblock URL \url{https://doi.org/10.18653/v1/d16-1127}.

\bibitem[Li et~al.(2018)Li, Jia, He, and Liang]{DBLP:conf/naacl/LiJHL18}
Juncen Li, Robin Jia, He~He, and Percy Liang.
\newblock Delete, retrieve, generate: a simple approach to sentiment and style
  transfer.
\newblock In Marilyn~A. Walker, Heng Ji, and Amanda Stent (eds.),
  \emph{Proceedings of the 2018 Conference of the North American Chapter of the
  Association for Computational Linguistics: Human Language Technologies,
  {NAACL-HLT} 2018, New Orleans, Louisiana, USA, June 1-6, 2018, Volume 1 (Long
  Papers)}, pp.\  1865--1874. Association for Computational Linguistics, 2018.
\newblock \doi{10.18653/v1/n18-1169}.
\newblock URL \url{https://doi.org/10.18653/v1/n18-1169}.

\bibitem[Liu et~al.(2016{\natexlab{a}})Liu, Lowe, Serban, Noseworthy, Charlin,
  and Pineau]{LiuLSNCP16}
Chia{-}Wei Liu, Ryan Lowe, Iulian Serban, Michael Noseworthy, Laurent Charlin,
  and Joelle Pineau.
\newblock How {NOT} to evaluate your dialogue system: An empirical study of
  unsupervised evaluation metrics for dialogue response generation.
\newblock In Jian Su, Xavier Carreras, and Kevin Duh (eds.), \emph{Proceedings
  of the 2016 Conference on Empirical Methods in Natural Language Processing,
  {EMNLP} 2016, Austin, Texas, USA, November 1-4, 2016}, pp.\  2122--2132. The
  Association for Computational Linguistics, 2016{\natexlab{a}}.
\newblock \doi{10.18653/v1/d16-1230}.
\newblock URL \url{https://doi.org/10.18653/v1/d16-1230}.

\bibitem[Liu et~al.(2016{\natexlab{b}})Liu, Zhu, Ye, Guadarrama, and
  Murphy]{RL_Img2txt_LiuZYG016}
Siqi Liu, Zhenhai Zhu, Ning Ye, Sergio Guadarrama, and Kevin Murphy.
\newblock Optimization of image description metrics using policy gradient
  methods.
\newblock \emph{CoRR}, abs/1612.00370, 2016{\natexlab{b}}.
\newblock URL \url{http://arxiv.org/abs/1612.00370}.

\bibitem[Nadeem et~al.(2020)Nadeem, Bethke, and Reddy]{stereoset}
Moin Nadeem, Anna Bethke, and Siva Reddy.
\newblock Stereoset: Measuring stereotypical bias in pretrained language
  models.
\newblock \emph{CoRR}, abs/2004.09456, 2020.
\newblock URL \url{https://arxiv.org/abs/2004.09456}.

\bibitem[Nielsen(2018)]{Nielsen-intro-information-geometry}
Frank Nielsen.
\newblock An elementary introduction to information geometry.
\newblock \emph{CoRR}, abs/1808.08271, 2018.
\newblock URL \url{http://arxiv.org/abs/1808.08271}.

\bibitem[Owen(2013)]{owen_chapter_importance_sampling_2013}
Art~B. Owen.
\newblock {Importance Sampling}.
\newblock In \emph{Monte Carlo theory, methods and examples}, chapter~9. 2013.
\newblock URL \url{https://statweb.stanford.edu/~owen/mc/Ch-var-is.pdf}.

\bibitem[Parshakova et~al.(2019{\natexlab{a}})Parshakova, Andreoli, and
  Dymetman]{A-parshakova-etal-2019-global}
Tetiana Parshakova, Jean-Marc Andreoli, and Marc Dymetman.
\newblock {Global Autoregressive Models for Data-Efficient Sequence Learning}.
\newblock In \emph{Proceedings of the 23rd Conference on Computational Natural
  Language Learning (CoNLL)}, pp.\  900--909, Hong Kong, China, November
  2019{\natexlab{a}}. Association for Computational Linguistics.
\newblock \doi{10.18653/v1/K19-1084}.
\newblock URL \url{https://www.aclweb.org/anthology/K19-1084}.

\bibitem[Parshakova et~al.(2019{\natexlab{b}})Parshakova, Andreoli, and
  Dymetman]{opt-rl-arxiv-2019}
Tetiana Parshakova, Jean-Marc Andreoli, and Marc Dymetman.
\newblock {Distributional Reinforcement Learning For Energy-Based Sequential
  Models}.
\newblock \emph{CoRR}, 2019{\natexlab{b}}.
\newblock URL \url{https://arxiv.org/abs/1912.08517}.

\bibitem[Pasunuru \& Bansal(2017)Pasunuru and Bansal]{PasunuruB17}
Ramakanth Pasunuru and Mohit Bansal.
\newblock Reinforced video captioning with entailment rewards.
\newblock In Martha Palmer, Rebecca Hwa, and Sebastian Riedel (eds.),
  \emph{Proceedings of the 2017 Conference on Empirical Methods in Natural
  Language Processing, {EMNLP} 2017, Copenhagen, Denmark, September 9-11,
  2017}, pp.\  979--985. Association for Computational Linguistics, 2017.
\newblock \doi{10.18653/v1/d17-1103}.
\newblock URL \url{https://doi.org/10.18653/v1/d17-1103}.

\bibitem[Paszke et~al.(2019)Paszke, Gross, Massa, Lerer, Bradbury, Chanan,
  Killeen, Lin, Gimelshein, Antiga, Desmaison, Kopf, Yang, DeVito, Raison,
  Tejani, Chilamkurthy, Steiner, Fang, Bai, and Chintala]{pytorch}
Adam Paszke, Sam Gross, Francisco Massa, Adam Lerer, James Bradbury, Gregory
  Chanan, Trevor Killeen, Zeming Lin, Natalia Gimelshein, Luca Antiga, Alban
  Desmaison, Andreas Kopf, Edward Yang, Zachary DeVito, Martin Raison, Alykhan
  Tejani, Sasank Chilamkurthy, Benoit Steiner, Lu~Fang, Junjie Bai, and Soumith
  Chintala.
\newblock Pytorch: An imperative style, high-performance deep learning library.
\newblock In H.~Wallach, H.~Larochelle, A.~Beygelzimer, F.~d\textquotesingle
  Alch\'{e}-Buc, E.~Fox, and R.~Garnett (eds.), \emph{Advances in Neural
  Information Processing Systems 32}, pp.\  8024--8035. Curran Associates,
  Inc., 2019.
\newblock URL
  \url{http://papers.neurips.cc/paper/9015-pytorch-an-imperative-style-high-performance-deep-learning-library.pdf}.

\bibitem[Paulus et~al.(2018)Paulus, Xiong, and Socher]{PaulusXS18}
Romain Paulus, Caiming Xiong, and Richard Socher.
\newblock A deep reinforced model for abstractive summarization.
\newblock In \emph{6th International Conference on Learning Representations,
  {ICLR} 2018, Vancouver, BC, Canada, April 30 - May 3, 2018, Conference Track
  Proceedings}. OpenReview.net, 2018.
\newblock URL \url{https://openreview.net/forum?id=HkAClQgA-}.

\bibitem[Prates et~al.(2020)Prates, Avelar, and Lamb]{bias_mt_PratesAL20}
Marcelo O.~R. Prates, Pedro H.~C. Avelar, and Lu{\'{\i}}s~C. Lamb.
\newblock Assessing gender bias in machine translation: a case study with
  google translate.
\newblock \emph{Neural Computing and Applications}, 32\penalty0 (10):\penalty0
  6363--6381, 2020.
\newblock \doi{10.1007/s00521-019-04144-6}.
\newblock URL \url{https://doi.org/10.1007/s00521-019-04144-6}.

\bibitem[Radford et~al.(2019)Radford, Wu, Child, Luan, Amodei, and
  Sutskever]{radford2019language}
Alec Radford, Jeffrey Wu, Rewon Child, David Luan, Dario Amodei, and Ilya
  Sutskever.
\newblock Language models are unsupervised multitask learners.
\newblock \emph{OpenAI Blog}, 1\penalty0 (8):\penalty0 9, 2019.

\bibitem[Ranzato et~al.(2007)Ranzato, Boureau, Chopra, and LeCun]{RanzatoBCL07}
Marc'Aurelio Ranzato, Y{-}Lan Boureau, Sumit Chopra, and Yann LeCun.
\newblock A unified energy-based framework for unsupervised learning.
\newblock In Marina Meila and Xiaotong Shen (eds.), \emph{Proceedings of the
  Eleventh International Conference on Artificial Intelligence and Statistics,
  {AISTATS} 2007, San Juan, Puerto Rico, March 21-24, 2007}, volume~2 of
  \emph{{JMLR} Proceedings}, pp.\  371--379. JMLR.org, 2007.
\newblock URL \url{http://proceedings.mlr.press/v2/ranzato07a.html}.

\bibitem[Ranzato et~al.(2016)Ranzato, Chopra, Auli, and
  Zaremba]{seq_lvl_train_RanzatoCAZ15}
Marc'Aurelio Ranzato, Sumit Chopra, Michael Auli, and Wojciech Zaremba.
\newblock Sequence level training with recurrent neural networks.
\newblock In Yoshua Bengio and Yann LeCun (eds.), \emph{4th International
  Conference on Learning Representations, {ICLR} 2016, San Juan, Puerto Rico,
  May 2-4, 2016, Conference Track Proceedings}, 2016.
\newblock URL \url{http://arxiv.org/abs/1511.06732}.

\bibitem[Robert \& Casella(2005)Robert and Casella]{Robert:2005:MCS:1051451}
Christian~P. Robert and George Casella.
\newblock \emph{Monte Carlo Statistical Methods (Springer Texts in
  Statistics)}.
\newblock Springer-Verlag, Berlin, Heidelberg, 2005.
\newblock ISBN 0387212396.

\bibitem[Rosenfeld et~al.(2001)Rosenfeld, Chen, and
  Zhu]{Rosenfeld01whole-sentenceexponential}
Ronald Rosenfeld, Stanley~F. Chen, and Xiaojin Zhu.
\newblock Whole-sentence exponential language models: A vehicle for
  linguistic-statistical integration.
\newblock \emph{Computers, Speech and Language}, 15:\penalty0 2001, 2001.

\bibitem[Schulman et~al.(2017)Schulman, Wolski, Dhariwal, Radford, and
  Klimov]{PPO}
John Schulman, Filip Wolski, Prafulla Dhariwal, Alec Radford, and Oleg Klimov.
\newblock Proximal policy optimization algorithms.
\newblock \emph{CoRR}, abs/1707.06347, 2017.
\newblock URL \url{http://arxiv.org/abs/1707.06347}.

\bibitem[See et~al.(2019)See, Roller, Kiela, and Weston]{SeeRKW19}
Abigail See, Stephen Roller, Douwe Kiela, and Jason Weston.
\newblock What makes a good conversation? how controllable attributes affect
  human judgments.
\newblock In Jill Burstein, Christy Doran, and Thamar Solorio (eds.),
  \emph{Proceedings of the 2019 Conference of the North American Chapter of the
  Association for Computational Linguistics: Human Language Technologies,
  {NAACL-HLT} 2019, Minneapolis, MN, USA, June 2-7, 2019, Volume 1 (Long and
  Short Papers)}, pp.\  1702--1723. Association for Computational Linguistics,
  2019.
\newblock \doi{10.18653/v1/n19-1170}.
\newblock URL \url{https://doi.org/10.18653/v1/n19-1170}.

\bibitem[Sheng et~al.(2019{\natexlab{a}})Sheng, Chang, Natarajan, and
  Peng]{ShengCNP_LM_bias19}
Emily Sheng, Kai{-}Wei Chang, Premkumar Natarajan, and Nanyun Peng.
\newblock The woman worked as a babysitter: On biases in language generation.
\newblock In Kentaro Inui, Jing Jiang, Vincent Ng, and Xiaojun Wan (eds.),
  \emph{Proceedings of the 2019 Conference on Empirical Methods in Natural
  Language Processing and the 9th International Joint Conference on Natural
  Language Processing, {EMNLP-IJCNLP} 2019, Hong Kong, China, November 3-7,
  2019}, pp.\  3405--3410. Association for Computational Linguistics,
  2019{\natexlab{a}}.
\newblock \doi{10.18653/v1/D19-1339}.
\newblock URL \url{https://doi.org/10.18653/v1/D19-1339}.

\bibitem[Sheng et~al.(2019{\natexlab{b}})Sheng, Chang, Natarajan, and
  Peng]{babysitter_ShengCNP19}
Emily Sheng, Kai{-}Wei Chang, Premkumar Natarajan, and Nanyun Peng.
\newblock The woman worked as a babysitter: On biases in language generation.
\newblock In Kentaro Inui, Jing Jiang, Vincent Ng, and Xiaojun Wan (eds.),
  \emph{Proceedings of the 2019 Conference on Empirical Methods in Natural
  Language Processing and the 9th International Joint Conference on Natural
  Language Processing, {EMNLP-IJCNLP} 2019, Hong Kong, China, November 3-7,
  2019}, pp.\  3405--3410. Association for Computational Linguistics,
  2019{\natexlab{b}}.
\newblock \doi{10.18653/v1/D19-1339}.
\newblock URL \url{https://doi.org/10.18653/v1/D19-1339}.

\bibitem[Sheng et~al.(2020)Sheng, Chang, Natarajan, and
  Peng]{babysitter2_Sheng2020}
Emily Sheng, Kai{-}Wei Chang, Premkumar Natarajan, and Nanyun Peng.
\newblock Towards controllable biases in language generation.
\newblock \emph{CoRR}, abs/2005.00268, 2020.
\newblock URL \url{https://arxiv.org/abs/2005.00268}.

\bibitem[Shetty et~al.(2017)Shetty, Rohrbach, Hendricks, Fritz, and
  Schiele]{ShettyRHFS17}
Rakshith Shetty, Marcus Rohrbach, Lisa~Anne Hendricks, Mario Fritz, and Bernt
  Schiele.
\newblock Speaking the same language: Matching machine to human captions by
  adversarial training.
\newblock In \emph{{IEEE} International Conference on Computer Vision, {ICCV}
  2017, Venice, Italy, October 22-29, 2017}, pp.\  4155--4164. {IEEE} Computer
  Society, 2017.
\newblock \doi{10.1109/ICCV.2017.445}.
\newblock URL \url{http://doi.ieeecomputersociety.org/10.1109/ICCV.2017.445}.

\bibitem[Stanovsky et~al.(2019)Stanovsky, Smith, and
  Zettlemoyer]{bias_mt_stanovsky-etal-2019}
Gabriel Stanovsky, Noah~A. Smith, and Luke Zettlemoyer.
\newblock Evaluating gender bias in machine translation.
\newblock In \emph{Proceedings of the 57th Annual Meeting of the Association
  for Computational Linguistics}, pp.\  1679--1684, Florence, Italy, July 2019.
  Association for Computational Linguistics.
\newblock \doi{10.18653/v1/P19-1164}.
\newblock URL \url{https://www.aclweb.org/anthology/P19-1164}.

\bibitem[Tambwekar et~al.(2019)Tambwekar, Dhuliawala, Martin, Mehta, Harrison,
  and Riedl]{RL_TambwekarDMMHR19}
Pradyumna Tambwekar, Murtaza Dhuliawala, Lara~J. Martin, Animesh Mehta, Brent
  Harrison, and Mark~O. Riedl.
\newblock Controllable neural story plot generation via reward shaping.
\newblock In Sarit Kraus (ed.), \emph{Proceedings of the Twenty-Eighth
  International Joint Conference on Artificial Intelligence, {IJCAI} 2019,
  Macao, China, August 10-16, 2019}, pp.\  5982--5988. ijcai.org, 2019.
\newblock \doi{10.24963/ijcai.2019/829}.
\newblock URL \url{https://doi.org/10.24963/ijcai.2019/829}.

\bibitem[Tu et~al.(2020)Tu, Pang, Wiseman, and Gimpel]{Tu2020ENGINEEI}
Lifu Tu, Richard~Yuanzhe Pang, Sam Wiseman, and Kevin Gimpel.
\newblock Engine: Energy-based inference networks for non-autoregressive
  machine translation.
\newblock \emph{ArXiv}, abs/2005.00850, 2020.

\bibitem[Wallace et~al.(2019)Wallace, Feng, Kandpal, Gardner, and
  Singh]{WallaceFKGS19}
Eric Wallace, Shi Feng, Nikhil Kandpal, Matt Gardner, and Sameer Singh.
\newblock Universal adversarial triggers for attacking and analyzing {NLP}.
\newblock In Kentaro Inui, Jing Jiang, Vincent Ng, and Xiaojun Wan (eds.),
  \emph{Proceedings of the 2019 Conference on Empirical Methods in Natural
  Language Processing and the 9th International Joint Conference on Natural
  Language Processing, {EMNLP-IJCNLP} 2019, Hong Kong, China, November 3-7,
  2019}, pp.\  2153--2162. Association for Computational Linguistics, 2019.
\newblock \doi{10.18653/v1/D19-1221}.
\newblock URL \url{https://doi.org/10.18653/v1/D19-1221}.

\bibitem[Williams(1992{\natexlab{a}})]{Williams92}
Ronald~J. Williams.
\newblock Simple statistical gradient-following algorithms for connectionist
  reinforcement learning.
\newblock \emph{Mach. Learn.}, 8:\penalty0 229--256, 1992{\natexlab{a}}.
\newblock \doi{10.1007/BF00992696}.
\newblock URL \url{https://doi.org/10.1007/BF00992696}.

\bibitem[Williams(1992{\natexlab{b}})]{Williams92Reinforce}
Ronald~J. Williams.
\newblock Simple statistical gradient-following algorithms for connectionist
  reinforcement learning.
\newblock In \emph{Machine Learning}, pp.\  229--256, 1992{\natexlab{b}}.

\bibitem[Wolf et~al.(2019)Wolf, Debut, Sanh, Chaumond, Delangue, Moi, Cistac,
  Rault, Louf, Funtowicz, and Brew]{huggingface}
Thomas Wolf, Lysandre Debut, Victor Sanh, Julien Chaumond, Clement Delangue,
  Anthony Moi, Pierric Cistac, Tim Rault, R{\'{e}}mi Louf, Morgan Funtowicz,
  and Jamie Brew.
\newblock Huggingface's transformers: State-of-the-art natural language
  processing.
\newblock \emph{CoRR}, abs/1910.03771, 2019.
\newblock URL \url{http://arxiv.org/abs/1910.03771}.

\bibitem[Wu et~al.(2016)Wu, Schuster, Chen, Le, Norouzi, Macherey, Krikun, Cao,
  Gao, Macherey, Klingner, Shah, Johnson, Liu, Kaiser, Gouws, Kato, Kudo,
  Kazawa, Stevens, Kurian, Patil, Wang, Young, Smith, Riesa, Rudnick, Vinyals,
  Corrado, Hughes, and Dean]{Wu_googleMT16}
Yonghui Wu, Mike Schuster, Zhifeng Chen, Quoc~V. Le, Mohammad Norouzi, Wolfgang
  Macherey, Maxim Krikun, Yuan Cao, Qin Gao, Klaus Macherey, Jeff Klingner,
  Apurva Shah, Melvin Johnson, Xiaobing Liu, Lukasz Kaiser, Stephan Gouws,
  Yoshikiyo Kato, Taku Kudo, Hideto Kazawa, Keith Stevens, George Kurian,
  Nishant Patil, Wei Wang, Cliff Young, Jason Smith, Jason Riesa, Alex Rudnick,
  Oriol Vinyals, Greg Corrado, Macduff Hughes, and Jeffrey Dean.
\newblock Google's neural machine translation system: Bridging the gap between
  human and machine translation.
\newblock \emph{CoRR}, abs/1609.08144, 2016.
\newblock URL \url{http://arxiv.org/abs/1609.08144}.

\bibitem[Yang et~al.(2018)Yang, Hu, Dyer, Xing, and
  Berg-Kirkpatrick]{gumbel_textgen_yang_NIPS2018}
Zichao Yang, Zhiting Hu, Chris Dyer, Eric~P Xing, and Taylor Berg-Kirkpatrick.
\newblock Unsupervised text style transfer using language models as
  discriminators.
\newblock In S.~Bengio, H.~Wallach, H.~Larochelle, K.~Grauman, N.~Cesa-Bianchi,
  and R.~Garnett (eds.), \emph{Advances in Neural Information Processing
  Systems 31}, pp.\  7287--7298. Curran Associates, Inc., 2018.
\newblock URL
  \url{http://papers.nips.cc/paper/7959-unsupervised-text-style-transfer-using-language-models-as-discriminators.pdf}.

\bibitem[Zhu et~al.(2018)Zhu, Lu, Zheng, Guo, Zhang, Wang, and
  Yu]{texygen-ZhuLZGZWY18}
Yaoming Zhu, Sidi Lu, Lei Zheng, Jiaxian Guo, Weinan Zhang, Jun Wang, and Yong
  Yu.
\newblock Texygen: {A} benchmarking platform for text generation models.
\newblock In Kevyn Collins{-}Thompson, Qiaozhu Mei, Brian~D. Davison, Yiqun
  Liu, and Emine Yilmaz (eds.), \emph{The 41st International {ACM} {SIGIR}
  Conference on Research {\&} Development in Information Retrieval, {SIGIR}
  2018, Ann Arbor, MI, USA, July 08-12, 2018}, pp.\  1097--1100. {ACM}, 2018.
\newblock \doi{10.1145/3209978.3210080}.
\newblock URL \url{https://doi.org/10.1145/3209978.3210080}.

\bibitem[Ziegler et~al.(2019)Ziegler, Stiennon, Wu, Brown, Radford, Amodei,
  Christiano, and Irving]{Ziegler19}
Daniel~M. Ziegler, Nisan Stiennon, Jeffrey Wu, Tom~B. Brown, Alec Radford,
  Dario Amodei, Paul Christiano, and Geoffrey Irving.
\newblock Fine-tuning language models from human preferences.
\newblock \emph{CoRR}, abs/1909.08593, 2019.
\newblock URL \url{http://arxiv.org/abs/1909.08593}.

\end{thebibliography}
